\documentclass[SIG,biber]{nowfnt} 

\usepackage[utf8]{inputenc}
\usepackage{float}
\usepackage{amsfonts}
\usepackage{graphics}
\usepackage{graphicx}
\usepackage{amsmath}
\usepackage{amssymb}
\usepackage{bm}
\usepackage{url}
\usepackage{algorithm,algorithmic}
\title{A Brief Introduction to Machine Learning for Engineers}

\maintitleauthorlist{
Osvaldo Simeone \\
Department of Informatics\\
King's College London\\
osvaldo.simeone@kcl.ac.uk
}

\usepackage{mwe}

\begin{document}

\author[1]{Simeone, Osvaldo}

\affil[1]{Department of Informatics, King's College London; osvaldo.simeone@kcl.ac.uk}


\makeabstracttitle

\begin{abstract}
This monograph aims at providing an introduction to key concepts, algorithms, and theoretical results in machine learning. The treatment concentrates on probabilistic models for supervised and unsupervised learning problems. It introduces fundamental concepts and algorithms by building on first principles, while also exposing the reader to more advanced topics with extensive pointers to the literature, within a unified notation and mathematical framework. The material is organized according to clearly defined categories, such as discriminative and generative models, frequentist and Bayesian approaches, exact and approximate inference, as well as directed and undirected models. This monograph is meant as an entry point for researchers with an engineering background in probability and linear algebra. 
\end{abstract}

\clearpage

\setcounter{page}{0}
\pagenumbering{arabic}

\chapter*{Notation}
\markboth{\sffamily\slshape Notation}
{\sffamily\slshape Notation}  
\vspace{-2cm}
\indent \indent $\bullet$ Random variables or random vectors -- both abbreviated as rvs -- are represented using roman typeface, while their
values and realizations are indicated by the corresponding standard
font. For instance, the equality $\mathrm{x}=x$ indicates that rv $\mathrm{x}$ takes value $x$.\\ 
\indent $\bullet$ Matrices are indicated using uppercase fonts, with roman typeface used for random matrices.\\
\indent $\bullet$ Vectors will be taken to be in column form.\\
\indent $\bullet$ $X^{T}$ and $X^{\dagger}$ are the transpose and
the pseudoinverse of matrix $X$, respectively.\\
\indent $\bullet$ The distribution of a rv $\mathrm{x}$, either probability
mass function (pmf) for a discrete rv or probability
density function (pdf) for continuous rvs, is denoted as $p_{\mathrm{x}}$, $p_{\mathrm{x}}(x)$,
or $p(x)$. \\
\indent $\bullet$ The notation $\mathrm{x}\sim p_{\mathrm{x}}$ indicates that rv
$\mathrm{x}$ is distributed according to $p_{\mathrm{x}}$.\\
\indent $\bullet$ For jointly distributed rvs $(\mathrm{x,y})\sim p_{\mathrm{xy}}$,
the conditional distribution of $\mathrm{x}$ given the observation
$\mathrm{y}=y$ is indicated as $p_{\mathrm{x|}\mathrm{y}=y}$, $p_{\mathrm{x|y}}(x|y)$
or $p(x|y)$. \\		\indent $\bullet$ The notation $\mathrm{x|y}=y\sim p_{\mathrm{x|y}=y}$ indicates that
rv $\mathrm{x}$ is drawn according to the conditional distribution
$p_{\mathrm{x|y}=y}$.\\
\indent $\bullet$ 	The notation $\textrm{E}_{\mathrm{x}\sim p_{\mathrm{x}}}[\cdot]$ indicates
the expectation of the argument with respect to the distribution of
the rv $\mathrm{x}\sim p_{\mathrm{x}}$. Accordingly,
we will also write $\textrm{E}_{\mathrm{x}\sim p_{\mathrm{x|y}}}[\cdot|y]$
for the conditional expectation with respect to the distribution $p_{\mathrm{x|}\mathrm{y}=y}$. When clear from the context, the distribution over which the expectation is computed may be omitted. \\
\indent $\bullet$ The notation $\textrm{Pr}_{\mathrm{x}\sim p_{\mathrm{x}}}[\cdot]$ indicates
the probability of the argument event with respect to the distribution of
the rv $\mathrm{x}\sim p_{\mathrm{x}}$. When clear from the context, the subscript is dropped.\\
\indent $\bullet$ 	The notation $\log$ represents the logarithm in base two, while $\ln$
represents the natural logarithm.\\
\indent $\bullet$ $\mathrm{x}\sim\mathcal{N}(\mu,\Sigma)$ indicates
that random vector $\mathrm{x}$ is distributed according
to a multivariate Gaussian pdf with mean vector $\mu$ and covariance
matrix $\Sigma$. The multivariate Gaussian pdf is denoted as $\mathcal{N}(x|\mu,\Sigma)$
as a function of $x$.\\
\indent $\bullet$	$\mathrm{x}\sim\mathcal{U}(a,b)$ indicates that rv $\mathrm{x}$
is distributed according to a uniform distribution in the interval
$[a,b]$. The corresponding uniform pdf is denoted as $\mathcal{U}(x|a,b)$.\\
\indent $\bullet$	$\delta(x)$ denotes the Dirac delta function or the Kronecker delta
function, as clear from the context.\\
\indent $\bullet$ $||a||^{2}=\sum_{i=1}^{N}a_{i}^{2}$ is the quadratic, or $l_2$, norm
of a vector $a=[a_{1},...,a_{N}]^{T}$. We similarly define the $l_1$ norm as $||a||_1=\sum_{i=1}^{N}|a_{i}|$, and the $l_0$ pseudo-norm $||a||_0$ as the number of non-zero entries of vector $a$.\\
\indent $\bullet$ $I$ denotes the identity matrix, whose dimensions will be clear from the context. Similarly, $1$ represents a vector of all ones.\\
\indent $\bullet$ $\mathbb{R}$ is the set of real numbers; $\mathbb{R}^{+}$ the set of non-negative real numbers; $\mathbb{R}^{-}$ the set of non-positive real numbers; and $\mathbb{R}^N$ is the set of all vectors of $N$ real numbers. \\
\indent $\bullet$ $1\left(\cdot\right)$ is the indicator function: $1\left(x\right)=1$ if $x$ is true, and $1\left(x\right)=0$ otherwise.\\
\indent $\bullet$ $|\mathcal{S}|$ represents the cardinality of a set $\mathcal{S}$.\\
\indent $\bullet$ $x_{\mathcal{S}}$ represents a set of rvs $x_k$ indexed by the integers $k\in \mathcal{S}$.\\

\chapter*{Acronyms}
\markboth{\sffamily\slshape Acronyms}
{\sffamily\slshape Acronyms}  
\vspace{-2cm}

\indent \indent AI: Artificial Intelligence\\
\indent AMP: Approximate Message Passing\\
\indent BN: Bayesian Network\\
\indent DAG: Directed Acyclic Graph\\
\indent ELBO: Evidence Lower BOund\\
\indent EM: Expectation Maximization\\
\indent ERM: Empirical Risk Minimization\\
\indent GAN: Generative Adversarial Network\\
\indent GLM: Generalized Linear Model\\
\indent HMM: Hidden Markov Model\\
\indent i.i.d.: independent identically distributed\\
\indent KL: Kullback-Leibler\\
\indent LASSO: Least Absolute Shrinkage and Selection Operator\\
\indent LBP: Loopy Belief Propagation \\
\indent LL: Log-Likelihood \\
\indent LLR: Log-Likelihood Ratio \\
\indent LS: Least Squares \\
\indent MC: Monte Carlo\\
\indent MCMC: Markov Chain Monte Carlo\\
\indent MDL: Minimum Description Length\\
\indent MFVI: Mean Field Variational Inference\\
\indent ML: Maximum Likelihood\\
\indent MRF: Markov Random Field\\
\indent NLL: Negative Log-Likelihood\\
\indent PAC: Probably Approximately Correct\\
\indent pdf: probability density function\\
\indent	pmf: probability mass function\\
\indent	PCA: Principal Component Analysis\\
\indent	PPCA: Probabilistic Principal Component Analysis\\
\indent QDA: Quadratic Discriminant Analysis\\
\indent RBM: Restricted Boltzmann Machine\\
\indent SGD: Stochastic Gradient Descent\\
\indent SVM: Support Vector Machine\\
\indent rv: random variable or random vector (depending on the context)\\
\indent s.t.: subject to\\
\indent VAE:  Variational AutoEncoder\\
\indent VC:  Vapnik–Chervonenkis\\
\indent VI: Variational Inference\\

\part{Basics}


\chapter{Introduction}

Having taught courses on machine learning, I am often asked by colleagues
and students with a background in engineering to suggest ``the best
place to start'' to get into this subject. I typically respond with
a list of books \textendash{} for a general, but slightly outdated
introduction, read this book; for a detailed survey of methods based
on probabilistic models, check this other reference; to learn about
statistical learning, I found this text useful; and so on. This answer
strikes me, and most likely also my interlocutors, as quite unsatisfactory.
This is especially so since the size of many of these books may be
discouraging for busy professionals and students working on other
projects. This monograph is an attempt to offer a basic and compact
reference that describes key ideas and principles in simple terms
and within a unified treatment, encompassing also more recent developments
and pointers to the literature for further study. 

\section{What is Machine Learning?}

A useful way to introduce the machine learning methodology is by means
of a comparison with the conventional engineering design flow. This
starts with a in-depth analysis of the problem domain, which culminates
with the definition of a mathematical model. The mathematical model
is meant to capture the key features of the problem under study, and
is typically the result of the work of a number of experts. The mathematical
model is finally leveraged to derive hand-crafted solutions to the
problem. 

For instance, consider the problem of defining a chemical process
to produce a given molecule. The conventional flow requires chemists
to leverage their knowledge of models that predict the outcome of
individual chemical reactions, in order to craft a sequence of suitable
steps that synthesize the desired molecule. Another example is the
design of speech translation or image/ video compression algorithms.
Both of these tasks involve the definition of models and algorithms
by teams of experts, such as linguists, psychologists, and signal
processing practitioners, not infrequently during the course of long
standardization meetings. 

The engineering design flow outlined above may be too costly and inefficient
for problems in which faster or less expensive solutions are desirable.
The machine learning alternative is to collect large data sets, e.g.,
of labelled speech, images or videos, and to use this information
to train general-purpose learning machines to carry out the desired
task. While the standard engineering flow relies on domain knowledge
and on design optimized for the problem at hand, machine learning
lets large amounts of data dictate algorithms and solutions. To this
end, rather than requiring a precise model of the set-up under study,
machine learning requires the specification of an objective, of a
model to be trained, and of an optimization technique.

Returning to the first example above, a machine learning approach
would proceed by training a general-purpose machine to predict the
outcome of known chemical reactions based on a large data set, and
by then using the trained algorithm to explore ways to produce more
complex molecules. In a similar manner, large data sets of images
or videos would be used to train a general-purpose algorithm with
the aim of obtaining compressed representations from which the original
input can be recovered with some distortion.

\section{When to Use Machine Learning?}

Based on the discussion above, machine learning can offer an efficient
alternative to the conventional engineering flow when development
cost and time are the main concerns, or when the problem appears to
be too complex to be studied in its full generality. On the flip side,
the approach has the key disadvantages of providing generally suboptimal
performance, or hindering interpretability of the solution, and to
apply only to a limited set of problems. 

In order to identify tasks for which machine learning methods may
be useful, reference \cite{mitchell17} suggests the following criteria: 
\begin{enumerate}
\item the task involves a function that maps well-defined inputs to well-defined
outputs; 
\item large data sets exist or can be created containing input-output pairs; 
\item the task provides clear feedback with clearly definable goals and
metrics; 
\item the task does not involve long chains of logic or reasoning that depend
on diverse background knowledge or common sense; 
\item the task does not require detailed explanations for how the decision
was made; 
\item the task has a tolerance for error and no need for provably correct
or optimal solutions; 
\item the phenomenon or function being learned should not change rapidly
over time; and 
\item no specialized dexterity, physical skills, or mobility is required. 
\end{enumerate}
These criteria are useful guidelines for the decision of whether machine
learning methods are suitable for a given task of interest. They also
offer a convenient demarcation line between machine learning as is
intended today, with its focus on training and computational statistics
tools, and more general notions of Artificial Intelligence (AI) based
on knowledge and common sense \cite{Levesque17} (see \cite{norvig2009}
for an overview on AI research).

\subsection{Learning Tasks}

We can distinguish among three different main types of machine learning
problems, which are briefly introduced below. The discussion reflects
the focus of this monograph on parametric probabilistic models, as
further elaborated on in the next section.

\indent \emph{ }\textbf{1. Supervised learning:} We have $N$ labelled
training examples $\text{\ensuremath{\mathcal{D}}=}\{(x_{n},t_{n})\}_{n=1}^{N},$
where $x_{n}$ represents a covariate, or explanatory variable, while
$t_{n}$ is the corresponding label, or response. For instance, variable
$x_{n}$ may represent the text of an email, while the label $t_{n}$
may be a binary variable indicating whether the email is spam or not.
The goal of supervised learning is to predict the value of the label
$t$ for an input $x$ that is not in the training set. In other words,
supervised learning aims at generalizing the observations in the data
set $\mathcal{D}$ to new inputs. For example, an algorithm trained
on a set of emails should be able to classify a new email not present
in the data set $\mathcal{D}$.

We can generally distinguish between \emph{classification} problems,
in which the label $t$ is discrete, as in the example above, and
\emph{regression} problems, in which variable $t$ is continuous.
An example of a regression task is the prediction of tomorrow's temperature
$t$ based on today's meteorological observations $x$. 

An effective way to learn a predictor is to identify from the data
set $\mathcal{D}$ a predictive distribution $p(t|x)$ from a set
of parametrized distributions. The conditional distribution $p(t|x)$
defines a profile of beliefs over all possible of the label $t$ given
the input $x$. For instance, for temperature prediction, one could
learn mean and variance of a Gaussian distribution $p(t|x)$ as a
function of the input $x$. As a special case, the output of a supervised
learning algorithm may be in the form of a deterministic predictive
function $t=\hat{t}(x)$. 

\indent \emph{ }\textbf{2. Unsupervised learning:} Suppose now that
we have an unlabelled set of training examples $\text{\ensuremath{\mathcal{D}}=}\{x_{n}\}_{n=1}^{N}.$
Less well defined than supervised learning, unsupervised learning
generally refers to the task of learning properties of the mechanism
that generates this data set. Specific tasks and applications include
clustering, which is the problem of grouping similar examples $x_{n}$;
dimensionality reduction, feature extraction, and representation learning,
all related to the problem of representing the data in a smaller or
more convenient space; and generative modelling, which is the problem
of learning a generating mechanism to produce artificial examples
that are similar to available data in the data set $\mathcal{D}$.

As a generalization of both supervised and unsupervised learning,
\emph{semi-supervised learning} refers to scenarios in which not all
examples are labelled, with the unlabelled examples providing information
about the distribution of the covariates $x$. 

\indent \emph{ }\textbf{3. Reinforcement learning:} Reinforcement
learning refers to the problem of inferring optimal sequential decisions
based on rewards or punishments received as a result of previous actions.
Under supervised learning, the ``label'' $t$ refers to an action
to be taken when the learner is in an informational state about the
environment given by a variable $x$. Upon taking an action $t$ in
a state $x$, the learner is provided with feedback on the immediate
reward accrued via this decision, and the environment moves on to
a different state. As an example, an agent can be trained to navigate
a given environment in the presence of obstacles by penalizing decisions
that result in collisions.

Reinforcement learning is hence neither supervised, since the learner
is not provided with the optimal actions $t$ to select in a given
state $x$; nor is it fully unsupervised, given the availability of
feedback on the quality of the chosen action. Reinforcement learning
is also distinguished from supervised and unsupervised learning due
to the influence of previous actions on future states and rewards.

This monograph focuses on supervised and unsupervised learning. These
general tasks can be further classified along the following dimensions.

\indent $\bullet$\emph{ Passive vs. active learning}: A passive learner
is given the training examples, while an active learner can affect
the choice of training examples on the basis of prior observations. 

\indent $\bullet$\emph{ Offline vs. online learning}: Offline learning
operates over a batch of training samples, while online learning processes
samples in a streaming fashion. Note that reinforcement learning operates
inherently in an online manner, while supervised and unsupervised
learning can be carried out by following either offline or online
formulations.

This monograph considers only passive and offline learning.

\section{Goals and Outline }

This monograph aims at providing an introduction to key concepts,
algorithms, and theoretical results in machine learning. The treatment
concentrates on probabilistic models for supervised and unsupervised
learning problems. It introduces fundamental concepts and algorithms
by building on first principles, while also exposing the reader to
more advanced topics with extensive pointers to the literature, within
a unified notation and mathematical framework. Unlike other texts
that are focused on one particular aspect of the field, an effort
has been made here to provide a broad but concise overview in which
the main ideas and techniques are systematically presented. Specifically,
the material is organized according to clearly defined categories,
such as discriminative and generative models, frequentist and Bayesian
approaches, exact and approximate inference, as well as directed and
undirected models. This monograph is meant as an entry point for researchers
with a background in probability and linear algebra. A prior exposure
to information theory is useful but not required. 

Detailed discussions are provided on basic concepts and ideas, including
overfitting and generalization, Maximum Likelihood and regularization,
and Bayesian inference. The text also endeavors to provide intuitive
explanations and pointers to advanced topics and research directions.
Sections and subsections containing more advanced material that may
be skipped at a first reading are marked with a star ($*$). 

The reader will find here neither discussions on computing platform
or programming frameworks, such as map-reduce, nor details on specific
applications involving large data sets. These can be easily found
in a vast and growing body of work. Furthermore, rather than providing
exhaustive details on the existing myriad solutions in each specific
category, techniques have been selected that are useful to illustrate
the most salient aspects. Historical notes have also been provided
only for a few selected milestone events.

Finally, the monograph attempts to strike a balance between the algorithmic
and theoretical viewpoints. In particular, all learning algorithms
are introduced on the basis of theoretical arguments, often based
on information-theoretic measures. Moreover, a chapter is devoted
to statistical learning theory, demonstrating how to set the field
of supervised learning on solid theoretical foundations. This chapter
is more theoretically involved than the others, and proofs of some
key results are included in order to illustrate the theoretical underpinnings
of learning. This contrasts with other chapters, in which proofs of
the few theoretical results are kept at a minimum in order to focus
on the main ideas. 

The rest of the monograph is organized into five parts. The first
part covers introductory material. Specifically, Chapter 2 introduces
the frequentist, Bayesian and Minimum Description Length (MDL) learning
frameworks; the discriminative and generative categories of probabilistic
models; as well as key concepts such as training loss, generalization,
and overfitting \textendash{} all in the context of a simple linear
regression problem. Information-theoretic metrics are also briefly
introduced, as well as the advanced topics of interpretation and causality.
Chapter 3 then provides an introduction to the exponential family
of probabilistic models, to Generalized Linear Models (GLMs), and
to energy-based models, emphasizing main properties that will be invoked
in later chapters. 

The second part concerns supervised learning. Chapter 4 covers linear
and non-linear classification methods via discriminative and generative
models, including Support Vector Machines (SVMs), kernel methods,
logistic regression, multi-layer neural networks and boosting. Chapter
5 is a brief introduction to the statistical learning framework of
the Probably Approximately Correct (PAC) theory, covering the Vapnik\textendash Chervonenkis
(VC) dimension and the fundamental theorem of PAC learning. 

The third part, consisting of a single chapter, introduced unsupervised
learning. In particular, in Chapter 6, unsupervised learning models
are described by distinguishing among directed models, for which Expectation
Maximization (EM) is derived as the iterative maximization of the
Evidence Lower BOund (ELBO); undirected models, for which Restricted
Boltzmann Machines (RBMs) are discussed as a representative example;
discriminative models trained using the InfoMax principle; and autoencoders.
Generative Adversarial Networks (GANs) are also introduced. 

The fourth part covers more advanced modelling and inference approaches.
Chapter 7 provides an introduction to probabilistic graphical models,
namely Bayesian Networks (BNs) and Markov Random Fields (MRFs), as
means to encode more complex probabilistic dependencies than the models
studied in previous chapters. Approximate inference and learning methods
are introduced in Chapter 8 by focusing on Monte Carlo (MC) and Variational
Inference (VI) techniques. The chapter briefly introduces in a unified
way techniques such as variational EM, Variational AutoEncoders (VAE),
and black-box inference. Some concluding remarks are provided in the
last part, consisting of Chapter 9. 

We conclude this chapter by emphasizing the importance of probability
as a common language for the definition of learning algorithms \cite{cheeseman1985defense}.
The centrality of the probabilistic viewpoint was not always recognized,
but has deep historical roots. This is demonstrated by the following
two quotes, the first from the first AI textbook published by P. H.
Winston in 1977, and the second from an unfinished manuscript by J.
von Neumann (see \cite{norvig2009,Hintoncourse} for more information). 

\begin{quote}``Many ancient Greeks supported Socrates opinion that
deep, inexplicable thoughts came from the gods. Today\textquoteright s
equivalent to those gods is the erratic, even probabilistic neuron.
It is more likely that increased randomness of neural behavior is
the problem of the epileptic and the drunk, not the advantage of the
brilliant.''\end{quote} from \emph{Artificial Intelligence}, 1977. 

\begin{quote}``All of this will lead to theories of computation
which are much less rigidly of an all-or-none nature than past and
present formal logic... There are numerous indications to make us
believe that this new system of formal logic will move closer to another
discipline which has been little linked in the past with logic. This
is thermodynamics primarily in the form it was received from Boltzmann.''\end{quote}
from \emph{The Computer and the Brain}, 1958.

\chapter{A Gentle Introduction through Linear Regression}

In this chapter, we introduce the frequentist, Bayesian and MDL learning
frameworks, as well as key concepts in supervised learning, such as
discriminative and generative models, training loss, generalization,
and overfitting. This is done by considering a simple linear regression
problem as a recurring example. We start by introducing the problem
of supervised learning and by presenting some background on inference.
We then present the frequentist, Bayesian and MDL learning approaches
in this order. The treatment of MDL is limited to an introductory
discussion, as the rest of monograph concentrates on frequentist and
Bayesian viewpoints. We conclude with an introduction to the important
topic of information-theoretic metrics, and with a brief introduction
to the advanced topics of causal inference and interpretation. 

\section{Supervised Learning\label{sec:Supervised-Learning}}

\begin{figure}[H] 
\centering\includegraphics[scale=0.45]{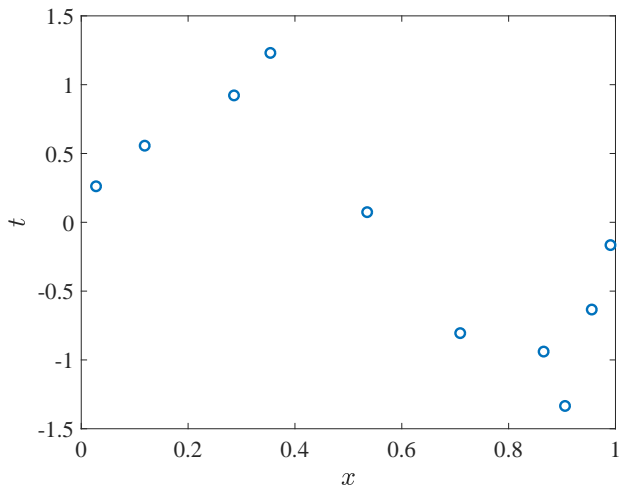} 
\caption{Example of a training set $\mathcal{D}$ with $N=10$ points ($x_n$,$t_n$), $n=1,...,N$. \label{fig:Fig1Ch2}} \end{figure}

In the standard formulation of a supervised learning problem, we are
given a training set $\mathcal{D}$ containing $N$ training points
$(x_{n},t_{n})$, $n=1,...,N$. The observations $x_{n}$ are considered
to be free variables, and known as \emph{covariates},\emph{ domain
points}, or\emph{ explanatory variables}; while the target variables
$t_{n}$ are assumed to be dependent on $x_{n}$ and are referred
to as \emph{dependent variables, labels}, or\emph{ responses}. An
example is illustrated in Fig. \ref{fig:Fig1Ch2}. We use the notation
$x_{\mathcal{D}}=(x_{1},....,x_{N})^{T}$ for the covariates and $t_{\mathcal{D}}=(t_{1},....,t_{N})^{T}$
for the labels in the training set $\mathcal{D}$. Based on this data,
the goal of supervised learning is to identify an algorithm to predict
the label $t$ for a new, that is, as of yet unobserved, domain point
$x$.

The outlined learning task is clearly impossible in the absence of
additional information on the mechanism relating variables $x$ and
$t$. With reference to Fig. \ref{fig:Fig1Ch2}, unless we assume,
say, that $x$ and $t$ are related by a function $t=f(x)$ with some
properties, such as smoothness, we have no way of predicting the label
$t$ for an unobserved domain point $x$. This observation is formalized
by the\emph{ no free lunch} \emph{theorem} to be reviewed in Chapter
5: one cannot learn rules that \emph{generalize} to unseen examples
without making assumptions about the mechanism generating the data.
The set of all assumptions made by the learning algorithm is known
as the \emph{inductive bias}.

This discussion points to a key difference between \emph{memorizing}
and learning. While the former amounts to mere retrieval of a value
$t_{n}$ corresponding to an already observed pair $(x_{n},t_{n})\in\mathcal{D},$
learning entails the capability to predict the value $t$ for an unseen
domain point $x$. Learning, in other words, converts experience \textendash{}
in the form of $\mathcal{D}$ \textendash{} into expertise or knowledge
\textendash{} in the form of a predictive algorithm. This is well
captured by the following quote by Jorge Luis Borges: ``\emph{To
think is to forget details, generalize, make abstractions.}'' \cite{sevenpilllars}.

By and large, the goal of supervised learning is that of identifying
a predictive algorithm that minimizes the \emph{generalization loss},
that is, the error in the prediction of a new label $t$ for an unobserved
explanatory variable $x$. How exactly to formulate this problem,
however, depends on one's viewpoint on the nature of the model that
is being learned. This leads to the distinction between the frequentist
and the Bayesian approaches, which is central to this chapter. As
it will be also discussed, the MDL philosophy deviates from the mentioned
focus on prediction as the goal of learning, by targeting instead
a parsimonious description of the data set $\mathcal{D}$.

\section{Inference\label{sec:Inference}}

Before we start our discussion of learning, it is useful to review
some basic concepts concerning statistical inference, as they will
be needed throughout this chapter and in the rest of this monograph.
We specifically consider the inference problem of predicting a rv
$\mathrm{t}$ given the observation of another rv $\mathrm{x}$ under
the assumption that their joint distribution $p(x,t)$ is \emph{known}.
As a matter of terminology, it is noted that here we will use the
term ``inference'' as it is typically intended in the literature
on probabilistic graphical models (see, e.g., \cite{Koller}), hence
diverging from its use in other branches of the machine learning literature
(see, e.g., \cite{Bishop}).

In order to define the problem of optimal inference, one starts by
defining a non-negative\emph{ loss function} $\ell(t,\hat{t})$. This
defines the cost, or loss or risk, incurred when the correct value
is $t$ while the estimate is $\hat{t}$. An important example is
the $\ell_{q}$ loss 
\begin{equation}
\ell_{q}(t,\hat{t})=|t-\hat{t}|^{q},
\end{equation}
which includes as a special case the \emph{quadratic loss} $\ell_{2}(t,\hat{t})=(t-\hat{t})^{2},$
and the \emph{0-1 loss, or detection error}, $\ell_{0}(t,\hat{t})=|t-\hat{t}|_{0},$
where $|a|_{0}=1$ if $a\neq0$ and $|a|_{0}=0$ otherwise. Once a
loss function is selected, the optimal prediction $\hat{t}(x)$ for
a given value of the observation $x$ is obtained by minimizing the
so-called \emph{generalization risk or generalization loss}\footnote{The term generalization error or population error are also often used,
but they will not be adopted in this monograph. } 
\begin{equation}
L_{p}(\hat{t})=\textrm{E}_{(\mathrm{x},\mathrm{t})\sim p_{\mathrm{xt}}}[\ell(\mathrm{t},\hat{t}(\mathrm{x}))].\label{eq:gen error inf}
\end{equation}
The notation $L_{p}$ emphasizes the dependence of the generalization
loss on the distribution $p(x,t)$. 

The solution of this problem is given by the optimal prediction or
decision rule\footnote{\noindent The optimal estimate (\ref{eq:opt decision}) is also known
as Bayes' prediction or Bayes' rule, but here we will not use this
terminology in order to avoid confusion with the Bayesian approach
discussed below. }

\noindent 
\begin{equation}
\hat{t}^{*}(x)=\textrm{arg}\min_{\hat{t}}\textrm{E}_{\mathrm{t}\sim p_{\mathrm{t}|\mathrm{x}}}[\ell(\mathrm{t},\hat{t})|x].\label{eq:opt decision}
\end{equation}
This can be seen by using the law of iterated expectations $\textrm{E}_{(\mathrm{x},\mathrm{t})\sim p_{\mathrm{xt}}}[\cdot]=\textrm{E}_{\mathrm{x}\sim p_{\mathrm{x}}}[\textrm{E}_{\mathrm{t}\sim p_{\mathrm{t}|\mathrm{x}}}[\cdot|\mathrm{x}]]$.
Equation (\ref{eq:opt decision}) shows that the optimal estimate,
or prediction, $\hat{t}^{*}(x)$ is a function of the \emph{posterior}
distribution $p(t|x)$ of the label given the domain point $x$ and
of the loss function $\ell$. Therefore, once the posterior $p(t|x)$
is known, one can evaluate the optimal prediction (\ref{eq:opt decision})
for any desired loss function, without the need to know the joint
distribution $p(x,t)$.

As a special case of (\ref{eq:opt decision}), with the quadratic
loss function $\ell_{2}$, the optimal prediction is the conditional
mean $\hat{t}^{*}(x)=\textrm{E}_{\mathrm{t}\sim p_{\mathrm{t}|\mathrm{x}}}[\mathrm{t}|x]$;
while for the 0-1 loss function $\ell_{0}$, the optimal decision
is the mode of the posterior distribution, i.e., $\hat{t}^{*}(x)=\textrm{arg}\max_{t}p(t|x)$. 

For example, assume that we have 
\begin{equation}
\mathrm{t}|\mathrm{x}=x\sim0.8\delta(t-x)+0.2\delta(t+x),
\end{equation}
so that, conditioned on the event $\mathrm{x}=x$, $\mathrm{t}$ equals
$x$ with probability 0.8 and $-x$ with probability 0.2. The optimal
prediction is $\hat{t}^{*}(x)=0.8x-0.2x=0.6x$ for the quadratic loss,
while it is $\hat{t}^{*}(x)=x$ for the 0-1 loss. 

The goal of supervised learning methods is broadly speaking that of
obtaining a predictor $\hat{t}(x)$ that performs close to the optimal
predictor $\hat{t}^{*}(x),$ based only on the training set $\mathcal{D}$,
and hence without knowledge of the joint distribution $p(x,t).$ The
closeness in performance is measured by the difference between the
generalization loss $L_{p}(\hat{t})$ achieved by the trained predictor
and the \emph{minimum} generalization loss $L_{p}(\hat{t}^{*})$ of
the optimal predictor, which depends on the true distribution $p(x,t)$.
Strictly speaking, this statement applies only for the frequentist
approach, which is discussed next. As it will be explained later in
the chapter, in fact, while the Bayesian approach still centers around
the goal of prediction, its modelling assumptions are different. Furthermore,
the MDL approach concentrates on the task of data compression rather
than prediction.

\section{Frequentist Approach}

According to the frequentist viewpoint, the training data points $(\mathrm{x}_{n},\mathrm{t}_{n})\in\mathcal{D}$
are independent identically distributed (i.i.d.) rvs drawn from a
\emph{true}, and unknown, distribution $p(x,t)$:
\begin{equation}
(\mathrm{x}_{n},\mathrm{t}_{n})\underset{\textrm{i.i.d.}}{\sim}p(x,t),\textrm{ \ensuremath{i=1,...,N.}}
\end{equation}
The new observation ($\mathrm{x},\mathrm{t}$) is also independently
generated from the same true distribution $p(x,t)$; the domain point
$\mathrm{x}$ is observed and the label $\mathrm{t}$ must be predicted.
Since the probabilistic model $p(x,t)$ is not known, one cannot solve
directly problem (\ref{eq:opt decision}) to find the optimal prediction
that minimizes the generalization loss $L_{p}$ in (\ref{eq:gen error inf}).

Before discussing the available solutions to this problem, it is worth
observing that the definition of the ``true'' distribution $p(x,t)$
depends in practice on the way data is collected. As in the example
of the ``beauty AI'' context, if the rankings $t_{n}$ assigned
to pictures $x_{n}$ of faces are affected by racial biases, the distribution
$p(x,t)$ will reflect these prejudices and produce skewed results
\citep{beautyai}. 

\textbf{Taxonomy of solutions.} There are two main ways to address
the problem of learning how to perform inference when not knowing
the distribution $p(x,t)$: 

\indent $\bullet$\emph{ Separate learning and (plug-in) inference}:
Learn first an approximation, say $p_{\mathcal{D}}(t|x)$, of the
conditional distribution $p(t|x)$ based on the data $\mathcal{D}$,
and then plug this approximation in (\ref{eq:opt decision}) to obtain
an approximation of the optimal decision as 
\begin{equation}
\hat{t}_{\mathcal{D}}(x)=\textrm{arg}\min_{\hat{t}}\textrm{E}_{\mathrm{t}\sim p_{\mathcal{D}}(t|x)}[\ell(\mathrm{t},\hat{t})|x].\label{eq:separate learning and inference}
\end{equation}

\indent $\bullet$\emph{ Direct inference via Empirical Risk Minimization
(ERM):} Learn directly an approximation $\hat{t}_{\mathcal{D}}(\cdot)$
of the optimal decision rule by minimizing an empirical estimate of
the generalization loss (\ref{eq:gen error inf}) obtained from the
data set as 
\begin{equation}
\hat{t}_{\mathcal{D}}(\cdot)=\textrm{arg}\min_{\hat{t}}L_{\mathcal{D}}(\hat{t}),\label{eq:ERM}
\end{equation}
where the empirical risk, or \emph{empirical loss}, is
\begin{equation}
L_{\mathcal{D}}(\hat{t})=\frac{1}{N}\sum_{n=1}^{N}\ell(t_{n},\hat{t}(x_{n})).
\end{equation}
The notation $L_{\mathcal{D}}(\hat{t})$ highlights the dependence
of the empirical loss on the predictor $\hat{t}(\cdot)$ and on the
training set $\mathcal{D}.$ 

In practice, as we will see, both approaches optimize a set of parameters
that define the probabilistic model or the predictor. Furthermore,
the first approach is generally more flexible, since having an estimate
$p_{\mathcal{D}}(t|x)$ of the posterior distribution $p(t|x)$ allows
the prediction (\ref{eq:separate learning and inference}) to be computed
for any loss function. In contrast, the ERM solution (\ref{eq:ERM})
is tied to a specific choice of the loss function $\ell$. In the
rest of this section, we will start by taking the first approach,
and discuss later how this relates to the ERM formulation.

\textbf{Linear regression example. }For concreteness, in the following,
we will consider the following running example inspired by \cite{Bishop}.
In the example, data is generated according to the true distribution
$p(x,t)=p(x)p(t|x)$, where $\mathrm{x}\sim\mathcal{U}(0,1)$ and
\begin{equation}
\mathrm{t}|\mathrm{x}=x\sim\mathcal{N}(\sin(2\pi x),0.1).\label{eq:true distributon Ch2}
\end{equation}
The training set in Fig. \ref{fig:Fig1Ch2} was generated from this
distribution. If this true distribution were known, the optimal predictor
under the $\ell_{2}$ loss would be equal to the conditional mean
\begin{equation}
\hat{t}^{*}(x)=\sin(2\pi x).\label{eq:opt predictor example}
\end{equation}
Hence, the minimum generalization loss is $L_{p}(\hat{t}^{*})=0.1$. 

It is emphasized that, while we consider this running example in order
to fix the ideas, all the definitions and ideas reported in this chapter
apply more generally to supervised learning problems. This will be
further discussed in Chapter 4 and Chapter 5. 

\subsection{Discriminative vs. Generative Probabilistic Models}

In order to learn an approximation $p_{\mathcal{D}}(t|x)$ of the
predictive distribution $p(t|x)$ based on the data $\mathcal{D}$,
we will proceed by first selecting a family of parametric probabilistic
models, also known as a \emph{hypothesis class}, and by then learning
the parameters of the model to fit (in a sense to be made precise
later) the data $\mathcal{D}.$ 

Consider as an example the linear regression problem introduced above.
We start by modelling the label $t$ as a polynomial function of the
domain point $x$ added to a Gaussian noise with variance $\beta^{-1}.$
Parameter $\beta$ is the precision, i.e., the inverse variance of
the additive noise. The polynomial function with degree $M$ can be
written as 
\begin{equation}
\mu(x,w)=\sum_{j=0}^{M}w_{j}x^{j}=w^{T}\phi(x),\label{eq:powers}
\end{equation}
where we have defined the weight vector $w=[w_{0}\textrm{ }w_{1}\cdots w_{M}]^{T}$
and the \emph{feature} vector $\phi(x)=[1\textrm{ \ensuremath{x\textrm{ \ensuremath{x^{2}}}}}\cdots x^{M}]^{T}$.
The vector $w$ defines the relative weight of the powers in the sum
(\ref{eq:powers}). This assumption corresponds to adopting a parametric
probabilistic model $p(t|x,\theta)$ defined as
\begin{equation}
\mathrm{t}|\mathrm{x}=x\sim\mathcal{N}(\mu(x,w),\beta^{-1}),\label{eq:discrmodellrCh2}
\end{equation}
with parameters $\theta=(w\textrm{,\ensuremath{\beta}}).$ Having
fixed this hypothesis class, the parameter vector $\theta$ can be
then learned from the data $\mathcal{D}$, as it will be discussed.

In the example above, we have parametrized the posterior distribution.
Alternatively, we can parametrize, and learn, the full joint distribution
$p(x,t)$. These two alternatives are introduced below.

\textbf{1. Discriminative probabilistic model.} With this first class
of models, the posterior, or predictive, distribution $p(t|x)$ is
assumed to belong to a hypothesis class $p(t|x,\theta)$ defined by
a parameter vector $\theta$. The parameter vector $\theta$ is learned
from the data set $\mathcal{D}.$ For a given parameter vector $\theta$,
the conditional distribution $p(t|x,\theta)$ allows the different
values of the label $t$ to be discriminated on the basis of their
posterior probability. In particular, once the model is learned, one
can directly compute the predictor (\ref{eq:separate learning and inference})
for any loss function. 

As an example, for the linear regression problem, once a vector of
parameters $\theta_{\mathcal{D}}=(w_{\mathcal{D}},\beta_{\mathcal{D}})$
is identified based on the data $\mathcal{D}$ during learning, the
optimal prediction under the $\ell_{2}$ loss is the conditional mean
$\hat{t}_{\mathcal{D}}(x)=\textrm{E}_{\mathrm{t}\sim p(t|x,\theta_{\mathcal{D}})}[\mathrm{t}|x]$,
that is, $\hat{t}_{\mathcal{D}}(x)=\mu(x,w_{\mathcal{D}})$. 

\textbf{2. Generative probabilistic model.} Instead of learning directly
the posterior $p(t|x)$, one can model the joint distribution $p(x,t)$
as being part of a parametric family $p(x,t|\theta$). Note that,
as opposed to discriminative models, the joint distribution $p(x,t|\theta$)
models also the distribution of the covariates $x$. Accordingly,
the term ``generative'' reflects the capacity of this type of models
to generate a realization of the covariates $\textrm{x}$ by using
the marginal $p(x|\theta).$

Once the joint distribution $p(x,t|\theta$) is learned from the data,
one can compute the posterior $p(t|x,\theta)$ using Bayes' theorem,
and, from it, the optimal predictor (\ref{eq:separate learning and inference})
can be evaluated for any loss function. Generative models make stronger
assumptions by modeling also the distribution of the explanatory variables.
As a result, an improper selection of the model may lead to more significant
bias issues. However, there are potential advantages, such as the
ability to deal with missing data or latent variables, such as in
semi-supervised learning. We refer to Chapter 6 for further discussion
(see also \cite{Bishop}). 

In the rest of this section, for concreteness, we consider discriminative
probabilistic models $p(t|x,\theta)$, although the main definitions
will apply also to generative models. 

\subsection{Model Order and Model Parameters}

In the linear regression example, the selection of the hypothesis
class (\ref{eq:discrmodellrCh2}) required the definition of the polynomial
degree $M$, while the determination of a specific model $p(t|x,\theta)$
in the class called for the selection of the parameter vector $\theta=(w,\beta)$.
As we will see, these two types of variables play a significantly
different role during learning and should be clearly distinguished,
as discussed next.

\indent\textbf{1. Model order $M$} (and\textbf{ hyperparameters}):
The model order defines the ``capacity'' of the hypothesis class,
that is, the number of the degrees of freedom in the model. The larger
$M$ is, the more capable a model is to fit the available data. For
instance, in the linear regression example, the model order determines
the size of the weight vector $w$. More generally, variables that
define the class of models to be learned are known as hyperparameters.
As we will see, determining the model order, and more broadly the
hyperparameters, requires a process known as \emph{validation}.

\indent\textbf{2. Model parameters $\theta$:} Assigning specific
values to the model parameters $\theta$ identifies a hypothesis within
the given hypothesis class. This can be done by using learning criteria
such as Maximum Likelihood (ML) and Maximum a Posteriori (MAP).

We postpone a discussion of validation to the next section, and we
start by introducing the ML and MAP learning criteria.

\subsection{{Maximum Likelihood (ML) Learning}\label{subsec:Maximum-Likelihood-(ML)}}

Assume now that the model order $M$ is fixed, and that we are interested
in learning the model parameters $\theta.$ The ML criterion selects
a value of $\theta$ under which the training set $\mathcal{D}$ has
the maximum probability of being observed. In other words, the value
$\theta$ selected by ML is the most likely to have generated the
observed training set. Note that there might be more than one such
value. 

To proceed, we need to write the probability (density) of the observed
labels $t_{\mathcal{D}}$ in the training set $\mathcal{D}$ given
the corresponding domain points $x$. Under the assumed discriminative
model, this is given as
\begin{align}
p(t_{\mathcal{D}}|x_{\mathcal{D}},w,\beta) & =\prod_{n=1}^{N}p(t_{n}|x_{n},w,\beta),\label{eq:prob data ML}\\
 & =\prod_{n=1}^{N}\mathcal{N}(t_{n}|\mu(x_{n},w),\beta^{-1})\nonumber 
\end{align}
where we have used the independence of different data points. Taking
the logarithm yields the \emph{Log-Likelihood (LL) function}
\begin{align}
\text{}\ln p(t_{\mathcal{D}}|x_{\mathcal{D}},w,\beta) & =\sum_{n=1}^{N}\ln p(t_{n}|x_{n},w,\beta)\textrm{ \textrm{ }}\nonumber \\
 & =\text{-}\frac{\beta}{2}\sum_{n=1}^{N}(\mu(x_{n},w)-t_{n})^{2}+\frac{N}{2}\ln\frac{\beta}{2\pi}.\label{eq:log likelihood linear}
\end{align}
The LL function should be considered as a function of the model parameters
$\theta=(w,\beta)$, since the data set $\mathcal{D}$ is fixed and
given. The ML learning problem is defined by the minimization of the
\emph{Negative LL} (NLL) function as
\begin{equation}
\min_{w,\beta}-\frac{1}{N}\sum_{n=1}^{N}\ln p(t_{n}|x_{n},w,\beta).\label{eq:ML problem}
\end{equation}
This criterion is also referred to as \emph{cross-entropy} or \emph{log-loss,}
as further discussed in Sec. \ref{sec:Information-Theoretic-Metrics}. 

If one is only interested in learning only the posterior mean, as
is the case when the loss function is $\ell_{2}$, then one can tackle
problem (\ref{eq:log likelihood linear}) only over the weights $w$,
yielding the optimization
\begin{equation}
\min_{w}L_{\mathcal{D}}(w)=\frac{1}{N}\sum_{n=1}^{N}(\mu(x_{n},w)-t_{n})^{2}.\label{eq:training loss}
\end{equation}
The quantity $L_{\mathcal{D}}(w)$ is known as the \emph{training
loss}. An interesting observation is that this criterion coincides
with the ERM problem (\ref{eq:ERM}) for the $\ell_{2}$ loss if one
parametrizes the predictor as $\hat{t}(x)=\mu(x,w)$. 

The ERM problem (\ref{eq:training loss}) can be solved in closed
form. To this end, we write the empirical loss as $L_{\mathcal{D}}(w)=N^{-1}||t_{\mathcal{D}}-X_{\mathcal{D}}w||^{2}$,
with the $N\times(M+1)$ matrix
\begin{equation}
X_{\mathcal{D}}=[\phi(x_{1})\textrm{ \ensuremath{\phi(x_{2})\cdots}}\phi(x_{N})]^{T}.\label{eq:matrixM}
\end{equation}
Its minimization hence amounts to a Least Squares (LS) problem, which
yields the solution
\begin{equation}
w_{ML}=(X_{\mathcal{D}}^{T}X_{\mathcal{D}})^{-1}X_{\mathcal{D}}^{T}t_{\mathcal{D}},\label{eq:wml}
\end{equation}
Note that, in (\ref{eq:wml}), we have assumed the typical overdetermined
case in which the inequality $N>(M+1)$ holds. More generally, one
has the ML solution $w_{ML}=X_{\mathcal{D}}^{\dagger}t_{\mathcal{D}}$.
Finally, differentiating the NLL with respect to $\beta$ yields instead
the ML estimate
\begin{equation}
\frac{1}{\beta_{ML}}=L_{\mathcal{D}}(w_{ML}).
\end{equation}

\begin{figure}[H] 
\centering\includegraphics[scale=0.45]{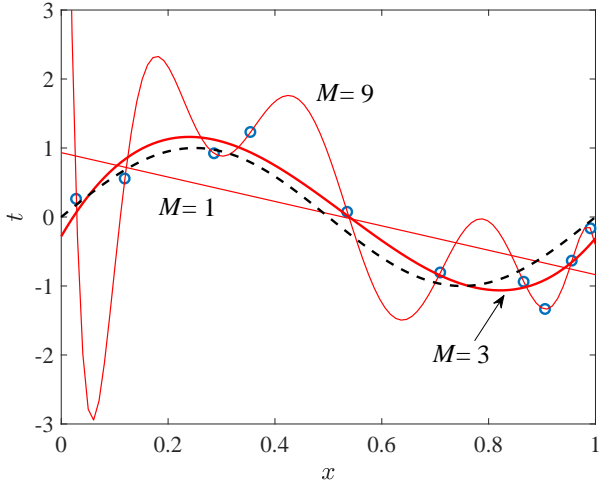}
\caption{Illustration of underfitting and overfitting in ML learning: The dashed line is the optimal predictor (\ref{eq:opt predictor example}), which depends on the unknown true distribution, while the other lines correspond to the predictor $\hat{t}_{ML}(x)=\mu(x,w_{ML})$ learned via ML with different model orders $M$. \label{fig:Fig2Ch2}} 
\end{figure}

\textbf{Overfitting and Underfitting. }Adopting the $\ell_{2}$ loss,
let us now compare the predictor $\hat{t}_{ML}(x)=\mu(x,w_{ML})$
learned via ML with the optimal, but unknown, predictor $\hat{t}^{*}(x)$
in (\ref{eq:opt predictor example}). To this end, Fig. \ref{fig:Fig2Ch2}
shows the optimal predictor $\hat{t}^{*}(x)$ as a dashed line and
the ML-based predictor $\hat{t}_{ML}(x)$ obtained with different
values of the model order $M$ for the training set $\mathcal{D}$
in Fig. \ref{fig:Fig1Ch2} (also shown in Fig. \ref{fig:Fig2Ch2}
for reference). 

We begin by observing that, with $M=1$, the ML predictor\emph{ underfits}
the data: the model is not rich enough to capture the variations present
in the data. As a result, the training loss $L_{\mathcal{D}}(w_{ML})$
in (\ref{eq:training loss}) is large. 

In contrast, with $M=9$, the ML predictor \emph{overfits} the data:
the model is too rich and, in order to account for the observations
in the training set, it yields inaccurate predictions outside it.
In this case, the training loss $L_{\mathcal{D}}(w)$ in (\ref{eq:training loss})
is small, but the generalization loss
\begin{equation}
L_{p}(w_{ML})=\textrm{E}_{(\mathrm{x},\mathrm{t})\sim p_{\mathrm{xt}}}[\ell(\mathrm{t},\mu(\mathrm{x},w_{ML}))]
\end{equation}
is large. With overfitting, the model is memorizing the training set,
rather than learning how to generalize to unseen examples. 

The choice $M=3$ appears to be the best by comparison with the optimal
predictor. Note that this observation is in practice precluded given
the impossibility to determine $\hat{t}^{*}(x)$ and hence the generalization
loss. We will discuss below how to estimate the generalization loss
using validation. \begin{figure}[H] 
\centering\includegraphics[scale=0.45]{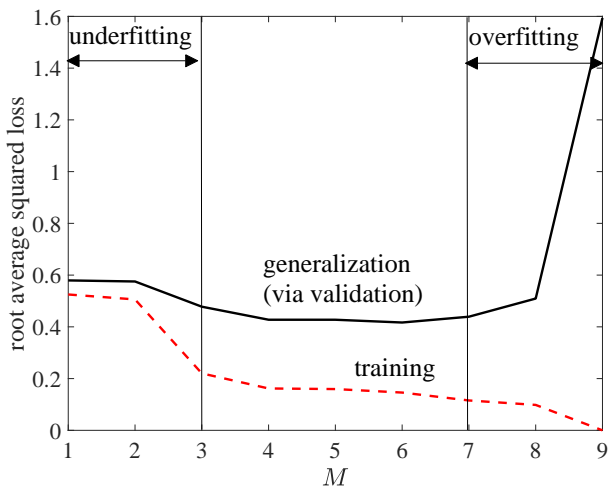}
\caption{Square root of the generalization loss $L_{p}(w_{ML})$ and of the training loss $L_{\mathcal{D}}(w_{ML})$ as a function of the model order $M$ for the training data set in Fig. \ref{fig:Fig1Ch2}.\label{fig:Fig3Ch2}} \end{figure}

The impact of the model order $M$ on training and generalization
losses is further elaborated on in Fig. \ref{fig:Fig3Ch2}, which
shows the squared root of the generalization loss $L_{p}(w_{ML})$
and of the training loss $L_{\mathcal{D}}(w_{ML})$ as a function
of $M$ for the same training data set. A first remark is that, as
expected, the training loss is smaller than the generalization loss,
since the latter accounts for all pairs ($\mathrm{x},\mathrm{t}$)$\sim p(x,t)$,
while the former includes only the training points used for learning.
More importantly, the key observation here is that increasing $\text{\ensuremath{M}}$
allows one to better fit \textendash{} and possibly overfit \textendash{}
the training set, hence reducing $L_{\mathcal{D}}(w_{ML})$. The generalization
loss $L_{p}(w_{ML})$ instead tends to decrease at first, as we move
away from the underfitting regime, but it eventually increases for
sufficiently large $\text{\ensuremath{M}}$. The widening of the gap
between training and generalization provides evidence that overfitting
is occurring. From Fig. \ref{fig:Fig3Ch2}, we can hence conclude
that, in this example, model orders larger than $M=7$ should be avoided
since they lead to overfitting, while model order less than $M=3$
should also not be considered in order to avoid underfitting.\begin{figure}[H] 
\centering\includegraphics[scale=0.45]{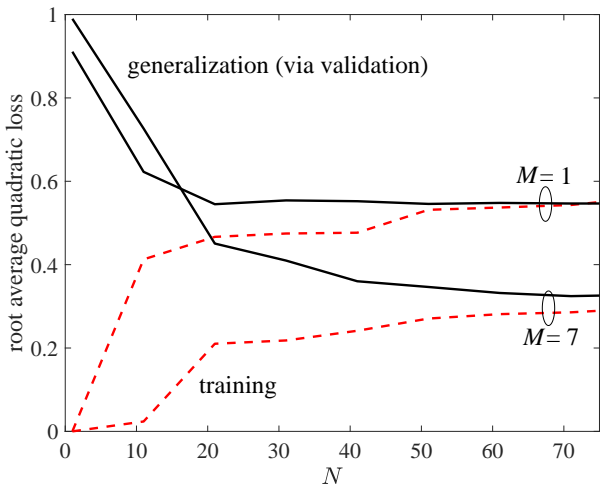}
\caption{Square root of the generalization loss $L_{p}(w_{ML})$ and of the training loss $L_{\mathcal{D}}(w_{ML})$ as a function of the training set size $N$. The asymptote of the generalization and training losses is given by the minimum generalization loss $L_{p}(w^{*})$ (cf. (\ref{eq:wstar})) achievable for the given model order (see Fig. \ref{fig:Fig5Ch2}). \label{fig:Fig4Ch2}} \end{figure}

\textbf{What if we had more data?} Extrapolating from the behavior
observed in Fig. \ref{fig:Fig2Ch2}, we can surmise that, as the number
$N$ of data points increases, overfitting is avoided even for large
values of $M$. In fact, when the training set is big as compared
to the number of parameters in $\theta$, we expect the training loss
$L_{\mathcal{D}}(w)$ to provide an accurate measure of the generalization
loss $L_{p}(w)$ \emph{for all possible values of} $w$. Informally,
we have the approximation $L_{\mathcal{D}}(w)\simeq L_{p}(w)$ simultaneously
for all values of $w$ as long as $N$ is large enough. Therefore,
the weight vector $w_{ML}$ that minimizes the training loss $L_{\mathcal{D}}(w)$
also (approximately) minimizes the generalization loss $L_{p}(w)$.
It follows that, for large $N$, the ML parameter vector $w_{ML}$
tends to the the optimal value $w^{*}$ (assuming for simplicity of
argument that it is unique) that minimizes the generalization loss
among all predictors in the model, i.e.,
\begin{equation}
w^{*}=\textrm{arg}\underset{w}{\textrm{min }}L_{p}(w).\label{eq:wstar}
\end{equation}
This discussion will be made precise in Chapter 5.

To offer numerical evidence for the point just made, Fig. \ref{fig:Fig4Ch2}
plots the (square root of the) generalization and training losses
versus $N$, where the training sets were generated at random from
the true distribution. From the figure, we can make the following
important observations. First, overfitting \textendash{} as measured
by the gap between training and generalization losses \textendash{}
vanishes as $N$ increases. This is a consequence of the discussed
approximate equalities $L_{\mathcal{D}}(w)\simeq L_{p}(w)$ and $w_{ML}\simeq w^{*}$,
which are valid as $N$ grows large, which imply the approximate equalities
$L_{\mathcal{D}}(w_{ML})\simeq L_{p}(w_{ML})\simeq L_{p}(w^{*})$.

Second, it is noted that the training loss $L_{\mathcal{D}}(w_{ML})$
tends to the minimum generalization loss $L_{p}(w^{*})$ for the given
$M$ from below, while the generalization loss $L_{p}(w_{ML})$ tends
to it from above. This is because, as $N$ increases, it becomes more
difficult to fit the data set $\mathcal{D}$, and hence $L_{\mathcal{D}}(w_{ML})$
increases. Conversely, as $N$ grows large, the ML estimate becomes
more accurate, because of the increasingly accurate approximation
$w_{ML}\simeq w^{*}$, and thus the generalization loss $L_{p}(w_{ML})$
decreases.

Third, selecting a smaller model order $M$ yields an improved generalization
loss when the training set is small, while a larger value of $M$
is desirable when the data set is bigger. In fact, as further discussed
below, when $N$ is small, the \emph{estimation error} caused by overfitting
dominates the \emph{bias} caused by the choice of a small hypothesis
class. In contrast, for sufficiently large training sets, the estimation
error vanishes and the performance is dominated by the bias induced
by the selection of the model.\begin{figure}[H] 
\centering\includegraphics[scale=0.5]{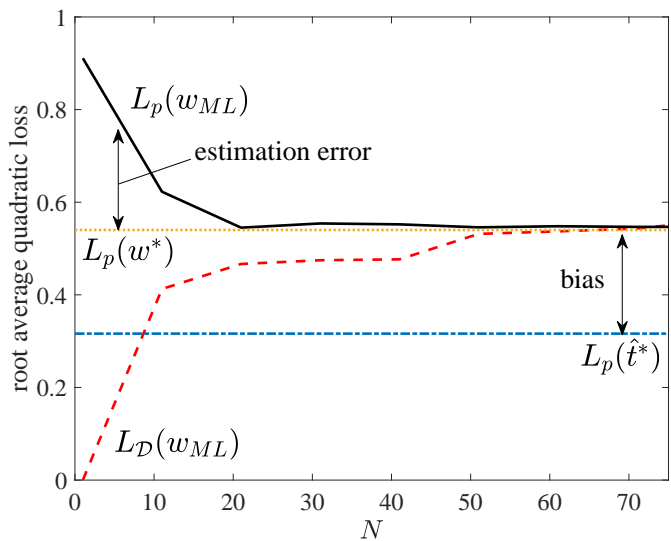}
\caption{Illustration of the bias and training error based on the decomposition (\ref{eq:decomposition}). \label{fig:Fig5Ch2}} \end{figure}

\textbf{Bias and generalization gap.} The previous paragraph introduced
the notions of \emph{estimation error} and \emph{bias} associated
with the selection of a given model order $M$. While Chapter 5 will
provide a more extensive discussion on these concepts, it is useful
to briefly review them here in the context of ML learning. Estimation
error and bias refer to the following decomposition of the generalization
loss achieved by the given solution $w_{ML}$ 
\begin{equation}
L_{p}(w_{ML})=L_{p}(\hat{t}^{*})+(L_{p}(w^{*})-L_{p}(\hat{t}^{*}))+(L_{p}(w_{ML})-L_{p}(w^{*})).\label{eq:decomposition}
\end{equation}
This decomposition is illustrated in Fig. \ref{fig:Fig5Ch2} for $M=1$.
In (\ref{eq:decomposition}), the term $L_{p}(\hat{t}^{*})=0.1$ (the
figure shows the square root) is, as seen, the minimum achievable
generalization loss without any constraint on the hypothesis class.
The term $(L_{p}(w^{*})-L_{p}(\hat{t}^{*}))$ represents the \emph{bias},
or approximation error, caused by the choice of the given hypothesis
class, and hence also by the choice of $M$. This is because, by (\ref{eq:wstar}),
$L_{p}(w^{*})$ is the best generalization loss for the given model.
Recall that the loss $L_{p}(w^{*})$ can be achieved when $N$ is
large enough. Finally, the term $(L_{p}(w_{ML})-L_{p}(w^{*}))$ is
the \emph{estimation error }or\emph{ generalization gap}\footnote{This is also defined as generalization error in some references. }
that is incurred due to the fact that $N$ is not large enough and
hence we have $w_{ML}\neq w^{*}$.

From the decomposition (\ref{eq:decomposition}), a large $N$ allows
us to reduce the estimation error, but it has no effect on the bias.
This is seen in Fig. \ref{fig:Fig4Ch2}, where the asymptote achieved
by the generalization loss as $N$ increases equals the minimum generalization
loss $L_{p}(w^{*})$ for the given model order. Choosing a small value
of $M$ in the regime of large data imposes a floor on the achievable
generalization loss that no amount of additional data can overcome.

\textbf{Validation and testing. }In the discussion above, it was assumed
that the generalization loss $L_{p}(w)$ can somehow be evaluated.
Since this depend on the true unknown distribution $p(x,t)$, this
evaluation is, strictly speaking, not possible. How then to estimate
the generalization loss in order to enable model order selection using
a plot as in Fig. \ref{fig:Fig3Ch2}? The standard solution is to
use validation.

The most basic form of validation prescribes the division of the available
data into two sets: a\emph{ hold-out, or validation, set} and the\emph{
training} set. The validation set is used to evaluate an approximation
of the generalization loss $L_{p}(w)$ via the empirical average
\begin{equation}
L_{p}(w)\simeq\frac{1}{N_{v}}\sum_{n=1}^{N_{v}}\ell(t_{n},\mu(x_{n},w)),
\end{equation}
where the sum is done over the $N_{v}$ elements of the validation
set. 

The just described hold-out approach to validation has a clear drawback,
as part of the available data needs to be set aside and not used for
training. This means that the number of examples that can be used
for training is smaller than the number of overall available data
points. To partially obviate this problem, a more sophisticated, and
commonly used, approach to validation is \emph{$k$-fold cross-validation}.
With this method, the available data points are partitioned, typically
at random, into $k$ equally sized subsets. The generalization loss
is then estimated by averaging $k$ different estimates. Each estimate
is obtained by retaining one of the $k$ subsets for validation and
the remaining $k-1$ subsets for training. When $k=N$, this approach
is also known as \emph{leave-one-out} cross-validation.

\textbf{Test set.} Once a model order $M$ and model parameters $\theta$
have been obtained via learning and validation, one typically needs
to produce an estimate of the generalization loss obtained with this
choice ($M,\theta$). The generalization loss estimated via validation
cannot be used for this purpose. In fact, the validation loss tends
to be smaller than the actual value of the generalization loss. After
all, we have selected the model order so as to yield the smallest
possible error on the validation set. The upshot is that the final
estimate of the generalization loss should be done on a separate set
of data points, referred to as the \emph{test set}, that are set aside
for this purpose and not used at any stage of learning. As an example,
in competitions among different machine learning algorithms, the test
set is kept by a judge to evaluate the submissions and is never shared
with the contestants.

\subsection{Maximum A Posteriori (MAP) Criterion \label{subsec:Maximum-A-Posteriori}}

We have seen that the decision regarding the model order $M$ in ML
learning concerns the tension between bias, whose reduction calls
for a larger $M$, and estimation error, whose decrease requires a
smaller $M$. ML provides a single integer parameter, $M$, as a gauge
to trade off bias and estimation error. As we will discuss here, the
MAP approach and, more generally, regularization, enable a finer control
of these two terms. The key idea is to leverage prior information
available on the behavior of the parameters in the absence, or presence,
of overfitting.

To elaborate, consider the following experiment. Evaluate the ML solution
$w_{ML}$ in (\ref{eq:wml}) for different values of $M$ and observe
how it changes as we move towards the overfitting regime by increasing
$M$ (see also \cite[Table 1.1]{Bishop}). For the experiment reported
in Fig. \ref{fig:Fig2Ch2}, we obtain the following values: for $M=1$,
$w_{ML}=[0.93\textrm{, }-1.76]^{T}$; for $M=3$, $w_{ML}=[-0.28,\textrm{ }13.32,\textrm{ }-35.92,\textrm{ }22.56]^{T}$;
and for $M=9$, $w_{ML}=[$$13.86$, $-780.33,$ $12.99$$\times10^{3},$
$-99.27\times10^{3},$ $416.49\times10^{3},$ $-1.03\times10^{6},$
$1.56\times10^{6},$ $1.40\times10^{6},$ $0.69\times10^{6},$ $-1.44\times10^{6}]$.
These results suggest that a manifestation of overfitting is the large
value of norm $\left\Vert w\right\Vert $ of the vector of weights.
We can use this observation as prior information, that is as part
of the inductive bias, in designing a learning algorithm.

To this end, we can impose a \emph{prior distribution} on the vector
of weights that gives lower probability to large values. A possible,
but not the only, way to do this is to is to assume a Gaussian prior
as
\begin{equation}
w\sim\mathcal{N}(0,\alpha^{-1}I),
\end{equation}
so that all weights are a priori i.i.d. zero-mean Gaussian variables
with variance $\alpha^{-1}$. Increasing $\alpha$ forces the weights
to be smaller as it reduces the probability associated with large
weights. The precision variable $\alpha$ is an example of a hyperparameter.
In a Bayesian framework, hyperparameters control the distribution
of the model parameters. As anticipated, hyperparameters are determined
via validation.

Rather than maximizing the LL, that is, probability density~~$p(t_{\mathcal{D}}|x_{\mathcal{D}},w,\beta)$
of the labels in the training set, as for ML, the MAP criterion prescribes
the maximization of the joint probability distribution of weights
and of labels given the prior $p(w)=\mathcal{N}(w|0,\alpha^{-1}I)$,
that is,
\begin{equation}
p(t_{\mathcal{D}},w|x_{\mathcal{D}},\beta)=p(w)\prod_{n=1}^{N}p(t_{n}|x_{n},w,\beta).
\end{equation}
Note that a prior probability can also be assumed for the parameter
$\beta$, but in this example we leave $\beta$ as a deterministic
parameter. The MAP learning criterion can hence be formulated as
\begin{equation}
\min_{w,\beta}-\sum_{n=1}^{N}\ln p(t_{n}|x_{n},w,\beta)-\ln p(w).\label{eq:MAP Ch2}
\end{equation}

The name ``Maximum a Posteriori'' is justified by the fact that
problem (\ref{eq:MAP Ch2}) is equivalent to maximizing the \emph{posterior
distribution} of the parameters $w$ given the available data, as
we will further discuss in the next section. This yields the following
problem for the weight vector
\begin{equation}
\min_{w}L_{\mathcal{D}}(w)+\frac{\lambda}{N}\Vert w\Vert^{2},\label{eq:training loss-map}
\end{equation}
where we have defined $\lambda=\alpha/\beta$ and we recall that the
training loss is $L_{\mathcal{D}}(w)=N^{-1}\left\Vert t_{\mathcal{D}}-X_{\mathcal{D}}w\right\Vert ^{2}.$

\textbf{ML vs. MAP.} Observing (\ref{eq:training loss-map}), it is
important to note the following \emph{general property} of the MAP
solution: As the number $N$ of data points grows large, the MAP estimate
tends to the ML estimate, given that the contribution of the prior
information term decreases as $1/N$. When $N$ is large enough, any
prior credence is hence superseded by the information obtained from
the data.

Problem (\ref{eq:training loss-map}), which is often referred to
as ridge regression, modifies the ML criterion by adding the quadratic
(or Tikhonov) \emph{regularization} function
\begin{equation}
R(w)=\Vert w\Vert^{2}
\end{equation}
multiplied by the term $\lambda/N.$ The regularization function forces
the norm of the solution to be small, particularly with larger values
of the hyperparameter $\lambda$, or equivalently $\alpha$. The solution
of problem (\ref{eq:training loss-map}) can be found by using standard
LS analysis, yielding
\begin{equation}
w_{MAP}=(\lambda I+X_{\mathcal{D}}^{T}X_{\mathcal{D}})^{-1}X_{\mathcal{D}}^{T}t_{\mathcal{D}}.\label{eq:wml-1}
\end{equation}
This expression confirms that, as $N$ grows large, the term $\lambda I$
becomes negligible and the solution tends to the ML estimate (\ref{eq:wml})
(see \cite{lehmann2006theory} for a formal treatment).\begin{figure}[H] 
\centering\includegraphics[width=0.8\textwidth]{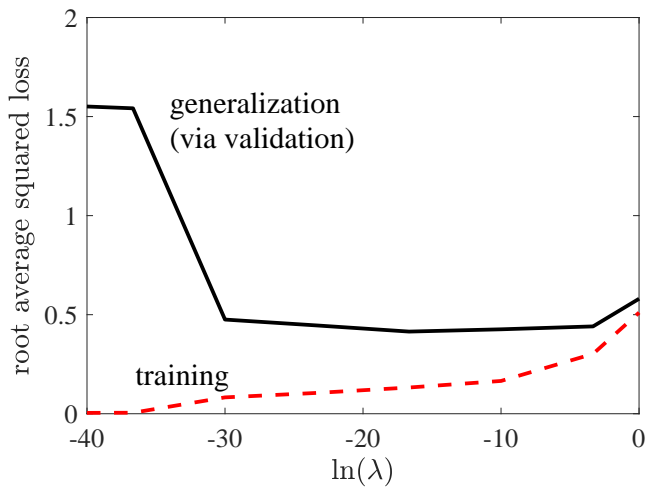}
\caption{Square root of the generalization loss $L_{p}(w_{MAP})$ and of the training loss $L_{\mathcal{D}}(w_{MAP})$ as a function of the regularization parameter $\lambda$ for the training data set in Fig. \ref{fig:Fig1Ch2} with $M=9$.\label{fig:Fig6Ch2}} \end{figure}

Fig. \ref{fig:Fig6Ch2} shows the squared root of the generalization
loss $L_{p}(w_{MAP})$ and of the training loss $L_{\mathcal{D}}(w_{MAP})$
as a function of $\lambda$ (in logarithmic scale) for the training
data set in Fig. \ref{fig:Fig1Ch2} with $M=9$. The generalization
loss is estimated using validation. We observe that increasing $\lambda$,
and hence the relevance of the regularization term, has a similar
impact as decreasing the model order $M$. A larger $\lambda$ reduces
the effective capacity of the model. Stated in different words, increasing
$\lambda$ reduces overfitting but may entail a larger bias.

Other standard examples for the prior distribution include the Laplace
pdf, which yields the $l_{1}$ norm regularization function $R(w)=\left\Vert w\right\Vert _{1}=\sum_{j=0}^{M}\left|w\right|_{1}$.
This term promotes the sparsity of the solution, which is useful in
many signal recovery algorithms \cite{baraniuk2007compressive} and
in non-parametric function estimation \cite{tsybakov2009introduction}.
The corresponding optimization problem
\begin{equation}
\min_{w}L_{\mathcal{D}}(w)+\frac{\lambda}{N}\Vert w\Vert_{1}
\end{equation}
 is known as LASSO (Least Absolute Shrinkage and Selection Operator).

\subsection{Regularization\label{subsec:RegularizationCh2}}

We have seen above that the MAP learning criterion amounts to the
addition of a regularization function $R(w)$ to the ML or ERM learning
losses. This function penalizes values of the weight vector $w$ that
are likely to occur in the presence of overfitting, or, generally,
that are improbable on the basis of the available prior information.
The net effect of this addition is to effectively decrease the capacity
of the model, as the set of values for the parameter vector $w$ that
the learning algorithm is likely to choose from is reduced. As a result,
as seen, regularization can control overfitting and its optimization
requires validation. 

Regularization generally refers to techniques that aim at reducing
overfitting during training. The discussion in the previous subsection
has focused on a specific form of regularization that is grounded
in a probabilistic interpretation in terms of MAP learning. We note
that the same techniques, such as ridge regression and LASSO, can
also be introduced independently of a probabilistic framework in an
ERM formulation. Furthermore, apart from the discussed addition of
regularization terms to the empirical risk, there are other ways to
perform regularization.

One approach is to modify the optimization scheme by using techniques
such as\emph{ early stopping} \cite{Goodfellowbook}. Another is to\emph{
augment} the training set by generating artificial examples and hence
effectively increasing the number $N$ of training examples. Related
to this idea is the technique known as \emph{bagging}. With bagging,
we first create a number $K$ of \emph{bootstrap} data sets. Bootstrap
data sets are obtained by selecting $N$ data points uniformly \emph{with
replacement} from $\mathcal{D}$ (so that the same data point generally
appears multiple times in the bootstrap data set). Then, we train
the model $K$ times, each time over a different bootstrap set. Finally,
we average the results obtained from the models using equal weights.
If the errors accrued by the different models were independent, bagging
would yield an estimation error that decreases with $K$. In practice,
significantly smaller gains are obtained, particularly for large $K$,
given that the bootstrap data sets are all drawn from $\mathcal{D}$
and hence the estimation errors are not independent \cite{Bishop}.

\section{Bayesian Approach}

The frequentist approaches discussed in the previous section assume
the existence of a true distribution, and aim at identifying a specific
value for the parameters $\theta$ of a probabilistic model to derive
a predictor (cf. (\ref{eq:opt decision})). ML chooses the value $\theta$
that maximizes the probability of the training data, while MAP includes
in this calculation also prior information about the parameter vector.
With the frequentist approach, there are hence two distributions on
the data: the true distribution, approximated by the empirical distribution
of the data and the model (see further discussion in Sec. \ref{sec:A-Tale-of-Ch2}). 

The Bayesian viewpoint is conceptually different: (\emph{i}) It assumes
that all data points are jointly distributed according to a distribution
that is known except for some hyperparameters; and (\emph{ii}) the
model parameters $\theta$ are jointly distributed with the data.
As a result, as it will be discussed, rather than committing to a
single value to explain the data, the Bayesian approach considers
the explanations provided by all possible values of $\theta$, each
weighted according to a generally different, and data-dependent, ``belief''.

More formally, the Bayesian viewpoint sees the vector of parameters
as rvs that are jointly distributed with the labels $\mathrm{t}_{\mathcal{D}}$
in the training data $\mathcal{D}$ and with the new example $\mathrm{t}$.
We hence have the joint distribution $p(t_{\mathcal{D}},w,t|x_{\mathcal{D}},x)$.
We recall that the conditioning on the domain points $x_{\mathcal{D}}$
and $x$ in the training set and in the new example, respectively,
are hallmarks of discriminative probabilistic models. The Bayesian
solution simply takes this modeling choice to its logical end point:
in order to predict the new label $\mathrm{t}$, it directly evaluates
the posterior distribution $p(t|x_{\mathcal{D}},t_{\mathcal{D}},x)=p(t|\mathcal{D},x)$
given the available information ($\mathcal{D},x$) by applying the
marginalization rules of probability to the joint distribution $p(t_{\mathcal{D}},w,t|x_{\mathcal{D}},x)$.

As seen, the posterior probability $p(t|\mathcal{D},x)$ can be used
as the predictive distribution in (\ref{eq:opt decision}) to evaluate
a predictor $\hat{t}(x)$. However, a fully Bayesian solution returns
the entire posterior $p(t|\mathcal{D},x)$, which provides significantly
more information about the unobserved label $\mathrm{t}$. As we will
discuss below, this knowledge, encoded in the posterior $p(t|\mathcal{D},x)$,
combines both the assumed prior information about the weight vector
$w$ and the information that is obtained from the data $\mathcal{D}$.

\begin{figure}[H] 
\centering\includegraphics[scale=0.55]{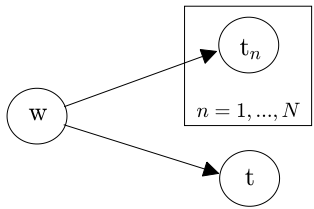}
\caption{Bayesian Network (BN) describing the joint distribution (\ref{eq:Bayesian}) of the weight vector $\mathrm{w}$, of the labels $\mathrm{t}_{\mathcal{D}}$ in the training data $\mathcal{D}$ and $\mathrm{t}$ in the new example, as used in the Bayesian approach.\label{fig:BNCh2}} \end{figure}

To elaborate, in the rest of this section, we assume that the precision
parameter $\beta$ is fixed and that the only learnable parameters
are the weights in vector $w$$.$ The joint distribution of the labels
in the training set, of the weight vector and of the new label, conditioned
on the domain points $x_{\mathcal{D}}$ in the training set and on
the new point $x$, is given as
\begin{equation}
p(t_{\mathcal{D}},w,t|x_{\mathcal{D}},x)=\underbrace{p(w)}_{\text{a priori distribution}}\underbrace{p(t_{\mathcal{D}}|x_{\mathcal{D}},w)}_{\text{ likelihood}}\underbrace{p(t|x,w)}_{\text{ distribution of new data}}.
\end{equation}
In the previous equation, we have identified the a priori distribution
of the data; the likelihood term $p(t_{\mathcal{D}}|x_{\mathcal{D}},w)=\prod_{n=1}^{N}\mathcal{N}(t_{n}|\mu(x_{n},w),\beta^{-1})$
in (\ref{eq:prob data ML})\footnote{The likelihood is also known as sampling distribution within the Bayesian
framework \cite{lunn2012bugs}. }; and the pdf of the new label $p(t|w,x)=\mathcal{N}(t|\mu(x,w),\beta^{-1})$.
It is often useful to drop the dependence on the domain points $x_{\,\mathcal{D}}$
and $x$ to write only the joint distribution of the random variables
in the model as
\begin{equation}
p(t_{\mathcal{D}},w,t)=\underbrace{p(w)}_{\text{a priori distribution}}\underbrace{p(t_{\mathcal{D}}|w)}_{\text{ likelihood}}\underbrace{p(t|w)}_{\text{ distribution of new data}}.\label{eq:Bayesian}
\end{equation}
This factorization can be represented graphically by the Bayesian
Network (BN) in Fig. \ref{fig:BNCh2}. The significance of the graph
should be clear by inspection, and it will be discussed in detail
in Chapter 7.

It is worth pointing out that, by treating all quantities in the model
\textendash{} except for the hyperparameter $\alpha$ \textendash{}
as rvs, the Bayesian viewpoint does away with the distinction between
learning and inference. In fact, since the joint distribution is assumed
to be known in a Bayesian model, the problem at hand becomes that
of inferring the unknown rv $\mathrm{t}$. To restate this important
point, the Bayesian approach subsumes all problems in the general
inference task of estimating a subset of rvs given other rvs in a
set of jointly distributed rvs with a known joint distribution.

As mentioned, we are interested in computing the posterior probability
$p(t|\mathcal{D},x)$ of the new label $\mathrm{t}$ given the training
data $\mathcal{D}$ and the new domain point $\mathrm{x}=x$. Dropping
again the domain variables to simplify the notation, we apply standard
rules of probability to obtain
\[
p(t|t_{\mathcal{D}})=\frac{p(t_{\mathcal{D}},t)}{p(t_{\mathcal{D}})}=\int\frac{p(w)p(t_{\mathcal{D}}|w)}{p(t_{\mathcal{D}})}p(t|w)dw
\]

\begin{equation}
=\int p(w|t_{\mathcal{D}})p(t|w)dw,
\end{equation}
where the second equality follows from the marginalization rule $p(t_{\mathcal{D}},t)=\int p(t_{\mathcal{D}},w,t)dw$
and the last equality from Bayes' theorem. Putting back the dependence
on the domain variables, we obtain the predictive distribution as
\begin{equation}
p(t|x,\mathcal{D})=\int\underbrace{p(w|\mathcal{D})}_{\text{posterior distribution of \ensuremath{w}}}p(t|x,w)dw.\label{eq:predictive}
\end{equation}

This is the key equation. Accordingly, the Bayesian approach considers
the predictive probability $p(t|x,w)$ associated with each value
of the weight vector $w$ weighted by the posterior belief
\begin{equation}
p(w|\mathcal{D})=\frac{p(w)p(t_{\mathcal{D}}|x_{\mathcal{D}},w)}{p(t_{\mathcal{D}}|x_{\mathcal{D}})}.\label{eq:belief Bayes}
\end{equation}
The posterior belief $p(w|\mathcal{D})$, which defines the weight
of the parameter vector $w,$ is hence proportional to the prior belief
$p(w)$ multiplied by the correction $p(t_{\mathcal{D}}|x_{\mathcal{D}},w)$
due to the observed data. 

Computing the posterior $p(w|\mathcal{D})$, and a fortiori also the
predictive distribution $p$$(t|x,\mathcal{D})$, is generally a difficult
task that requires the adoption of approximate inference techniques
covered in Chapter 8. For this example, however, we can directly compute
the predictive distribution as \cite{Bishop} 
\begin{equation}
t|x,\mathcal{D}\sim\mathcal{N}(\mu(x,w_{MAP}),s^{2}(x)),\label{eq:bayes}
\end{equation}
where $s^{2}(x)=\beta^{-1}(1+\phi(x)^{T}\left(\lambda I+X_{\mathcal{D}}^{T}X_{\mathcal{D}}\right)^{-1}\phi(x)).$
Therefore, in this particular example, the optimal predictor under
$\ell_{2}$ loss is MAP. This is consequence of the fact that mode
and mean of a Gaussian pdf coincide, and is not a general property.
Even so, as discussed next, the Bayesian viewpoint can provide significantly
more information on the label $t$ than the ML or MAP. 

\textbf{ML and MAP vs. Bayesian approach. }The Bayesian posterior
(\ref{eq:bayes}) provides a finer prediction of the labels $t$ given
the explanatory variables $x$ as compared to the predictive distribution
$p(t|x,\theta_{ML})=\mathcal{N}(\mu(x,w_{ML}),\beta^{-1})$ returned
by ML and similarly by MAP. To see this, note that the latter has
the same variance for all values of $x$, namely $\beta^{-1}.$ Instead,
the Bayesian approach reveals that, due to the uneven distribution
of the observed values of $x$, the accuracy of the prediction of
labels depends on the value of $x$: Values of $x$ closer to the
existing points in the training sets generally exhibit a smaller variance. 

\begin{figure}[H] 
\centering\includegraphics[scale=0.5]{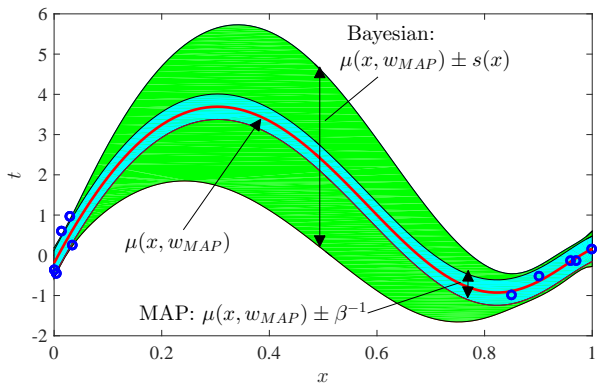}
\caption{Illustration of the predictive distribution $p(t|x,\mathcal{D})$ produced by the Bayesian method for the training set shown in the figure as compared to that obtained with the MAP criterion. The larger interval corresponds to $\mu(x,w_{MAP})\pm s(x)$ for the Bayesian method, while the smaller interval to $\mu(x,w_{MAP})\pm\beta^{-1/2}$ for MAP ($M=9$, $\beta^{-1}=0.1$ and $\alpha^{-1}=0.2\times10^{5}$).\label{fig:Fig7Ch2}} 
\end{figure}

This is shown in Fig. \ref{fig:Fig7Ch2}, which plots a training set,
along with the corresponding predictor $\mu(x,w_{MAP})$ and the high-probability
interval $\mu(x,w_{MAP})\pm s(x)$ produced by the Bayesian method.
We set $M=9$, $\beta^{-1}=0.1$ and $\alpha^{-1}=0.2\times10^{5}.$
For reference, we also show the interval $\mu(x,w_{MAP})\pm\beta^{-1/2}$
that would result from the MAP analysis. This intervals illustrate
the capability of the Bayesian approach to provide information about
the uncertainty associated with the estimate of $t$.

This advantage of the Bayesian approach reflects its conceptual difference
with respect to the frequentist approach: The frequentist predictive
distribution refers to a hypothetical new observation generated with
the same mechanism as the training data; instead, the Bayesian predictive
distribution quantifies the statistician's belief in the value of
$t$ given the assumed prior and the training data. 

From (\ref{eq:bayes}), we can make another important general observation
about the relationship with ML and MAP concerning the asymptotic behavior
when $N$ is large. In particular, when $N\rightarrow\infty$, we
have already seen that, informally, the limit $w_{MAP}\rightarrow w_{ML}$
holds. We now observe that it is also the case that the variance $s^{2}(x)$
of the Bayesian predictive distribution tends to $\beta^{-1}.$ As
a result, we can conclude that the Bayesian predictive distribution
approaches that returned by ML when $N$ is large. A way to think
about this conclusion is that, when $N$ is large, the posterior distribution
$p(w|\mathcal{D})$ of the weights tends to concentrate around the
ML estimate, hence limiting the average (\ref{eq:predictive}) to
the contribution of the ML solution.

\textbf{Marginal likelihood.} Another advantage of the Bayesian approach
is that, in principle, it allows model selection to be performed without
validation. Toward this end, compute the \emph{marginal likelihood
\begin{equation}
p(t_{\mathcal{D}}|x_{\mathcal{D}})=\int p(w)\prod_{n=1}^{N}p(t_{n}|x_{n},w)dw,\label{eq:marginal likelihood}
\end{equation}
}that is, the probability density of the training set when marginalizing
over the weight distribution. With the ML approach, the corresponding
quantity $p(t_{\mathcal{D}}|x_{\mathcal{D}},w_{ML})$ can only increase
by choosing a larger model order $M$. In fact, a larger $M$ entails
more degrees of freedom in the optimization (\ref{eq:training loss})
of the LL. A similar discussion applies also to MAP. However, this
is not the case for (\ref{eq:marginal likelihood}): a larger $M$
implies a more ``spread-out'' prior distribution of the weights,
which may result in a more diffuse distribution of the labels in (\ref{eq:marginal likelihood}).
Hence, increasing $M$ may yield a smaller marginal likelihood.\begin{figure}[H] 
\centering\includegraphics[width=0.8\textwidth]{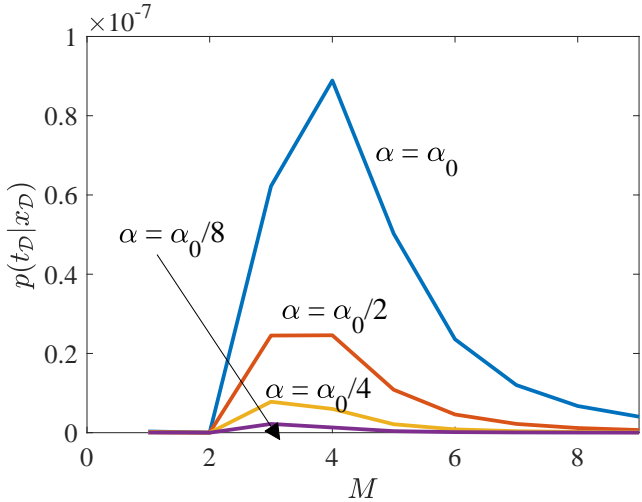}
\caption{Marginal likelihood versus the model order $M$ for the training set of Fig. \ref{fig:Fig1Ch2} ($\beta=10$, $\alpha_{0}=10^{-3}$).\label{fig:Fig9Ch2}} \end{figure}

To illustrate this point, Fig. \ref{fig:Fig9Ch2} plots the marginal
likelihood for the data set in Fig. \ref{fig:Fig1Ch2} for $\beta=10$
and three different values of $\alpha$ as a function of $M$. The
marginal likelihood in this example can be easily computed since we
have
\begin{equation}
\mathrm{t_{\mathcal{D}}}|\mathrm{x_{\mathcal{D}}}=x_{\mathcal{D}}\sim\mathcal{N}(0,\alpha^{-1}X_{\mathcal{D}}X_{\mathcal{D}}^{T}+\beta^{-1}I).
\end{equation}
It is observed that the marginal likelihood presents a peak at a given
value of $M$, while decreasing when moving away from the optimal
value. Therefore, we could take the value of $M$ at which the marginal
likelihood is maximized as the selected model order.

Does this mean that validation is really unnecessary when adopting
a Bayesian viewpoint? Unfortunately, this is not necessarily the case.
In fact, one still needs to set the hyperparameter $\alpha$. As seen
in Fig. \ref{fig:Fig9Ch2}, varying $\alpha$ can lead to different
conclusions on the optimal value of $M$. An alternative approach
would be to treat $\alpha$, and even $M$, as rvs with given priors
to be specified (see, e.g., \cite{MarcoScutari}). This would not
obviate the problem of selecting hyperparameters \textendash{} now
defining the prior distributions of $\alpha$ and $M$ \textendash{}
but it can lead to powerful hierarchical models. The necessary tools
will be discussed in Chapter 7.

As a final note, rather than using often impractical exhaustive search
methods, the optimization over the hyperparameters and the model order
$M$ for all criteria discussed so far, namely ML, MAP and Bayesian,
can be carried out using so-called Bayesian optimization tools \cite{Bayesianopt}.
A drawback of these techniques is that they have their own hyperparameters
that need to be selected.

\textbf{Empirical Bayes. }Straddling both frequentist and Bayesian
viewpoints is the so-called empirical Bayes method. This approach
assumes an a priori distribution for the parameters, but then estimates
the parameters of the prior \textendash{} say mean and variance of
a Gaussian prior \textendash{} from the data \cite{efron2016computer}.

\section{Minimum Description Length (MDL)$^{*}$\label{sec:Minimum-Description-Length}}

In this section, we briefly introduce a third, conceptually distinct,
learning philosophy \textendash{} the MDL criterion. The reader is
warned that the treatment here is rather superficial, and that a more
formal definition of the MDL criterion would would require a more
sophisticated discussion, which can be found in \cite{Grunwald}.
Furthermore, some background in information theory is preferable in
order to fully benefit from this discussion.

To start, we first recall from the treatment above that learning requires
the identification of a model, or hypothesis class \textendash{} here
the model order $M$ \textendash{} and of a specific hypothesis, defined
by parameters $\theta$ \textendash{} here $\theta=(w,\beta)$ \textendash{}
within the class. While MDL can be used for both tasks, we will focus
here only on the first. 

To build the necessary background, we now need to review the relationship
between probabilistic models and compression. To this end, consider
a signal $x$ taking values in a finite alphabet $\mathcal{X}$, e.g.,
a pixel in a gray scale image. Fix some probability mass function
(pmf) $p(x)$ on this alphabet. A key result in information theory
states that it is possible to design a lossless compression scheme
that uses $\left\lceil -\log p(x)\right\rceil $ bits to represent
value $x$\footnote{This is known as Kraft's inequality. More precisely, it states that
the lossless compression scheme at hand is prefix-free and hence decodable,
or invertible, without delay \cite{Cover}.}

By virtue of this result, the choice of a probability distribution
$p(x)$ is akin to the selection of a lossless compression scheme
that produces a description of around $-\log p(x)$ bits to represent
value $x$. Note that the description length $-\log p(x)$ decreases
with the probability assigned by $p(x)$ to value $x$: more likely
values under $p(x)$ are assigned a smaller description. Importantly,
a decoder would need to know $p(x)$ in order to recover $x$ from
the bit description.

At an informal level, the MDL criterion prescribes the selection of
a model that compresses the training data to the shortest possible
description. In other words, the model selected by MDL defines a compression
scheme that describes the data set $\mathcal{D}$ with the minimum
number of bits. As such, the MDL principle can be thought of as a
formalization of Occam's razor: choose the model that yields the simplest
explanation of the data. As we will see below, this criterion naturally
leads to a solution that penalizes overfitting.

What is the length of the description of a data set $\mathcal{D}$
that results from the selection of a specific value of $M$? The answer
is not straightforward, since, for a given value of $M$, there are
as many probability distributions as there are values for the corresponding
parameters $\theta$ to choose from. A formal calculation of the description
length would hence require the introduction of the concept of universal
compression for a given probabilistic model \cite{Grunwald}. Here,
we will limit ourselves to a particular class of universal codes known
as two-part codes.

Using two-part codes, we can compute the description length for the
data $\mathcal{D}$ that results from the choice of a model order
$M$ as follows. First, we obtain the ML solution ($w_{ML},\beta_{ML}$).
Then, we describe the data set by using a compression scheme defined
by the probability $p(t|x,w_{ML},\beta_{ML})=\mathcal{N}(t|\mu(x,w_{ML}),\beta_{ML}^{-1})$.
As discussed, this produces a description of approximately $-\sum_{n=1}^{N}\textrm{log}p(t_{n}|x_{n},w_{ML},\beta_{ML})$
bits\footnote{This neglects the technical issue that the labels are actually continuous
rvs, which could be accounted for by using quantization.}. This description is, however, not sufficient, since the decoder
of the description should also be informed about the parameters $(w_{ML},\beta_{ML}$).

Using quantization, the parameters can be described by means of a
number $C(M)$ of bits that is proportional to the number of parameters,
here $M+2.$ Concatenating these bits with the description produced
by the ML model yields the overall description length
\begin{equation}
-\sum_{n=1}^{N}\textrm{log}p(t_{n}|x_{n},w_{ML},\beta_{ML})+C(M).\label{eq:MDL}
\end{equation}
MDL \textendash{} in the simplified form discussed here \textendash{}
selects the model order $M$ that minimizes the description length
(\ref{eq:MDL}). Accordingly, the term $C(M)$ acts as a regularizer.
The optimal value of $M$ for the MDL criterion is hence the result
of the trade-off between the minimization of the overhead $C(M)$,
which calls for a small value of $M$, and the minimization of the
NLL, which decreases with $M$. 

Under some technical assumptions, the overhead term can be often evaluated
in the form $(K/2)\ln N+c$, where $K$ is the number of parameters
in the model and $c$ is a constant. This expression is not quite
useful in practice, but it provides intuition about the mechanism
used by MDL to combat overfitting.

\section{{Information-Theoretic Metrics\label{sec:Information-Theoretic-Metrics}}}

We now provide a brief introduction to information theoretic metrics
by leveraging the example studied in this chapter. As we will see
in the following chapters, information-theoretic metrics are used
extensively in the definition of learning algorithms. Appendix A provides
a detailed introduction to information-theoretic measures in terms
of inferential tasks. Here we introduce the key metrics of Kullback-Leibler
(KL) divergence and entropy by examining the asymptotic behavior of
ML in the regime of large $N$. The case with finite $N$ is covered
in Chapter 6 (see Sec. \ref{subsec:GAN}).

To start, we revisit the ML problem (\ref{eq:ML problem}), which
amounts to the minimization of the NLL $-N^{-1}\sum_{n=1}^{N}\ln p(t_{n}|x_{n},w,\beta)$,
also known as log-loss. According to the frequentist viewpoint, the
training set variables are drawn i.i.d. according to the true distribution
$p(x,t)$, i.e., $(\mathrm{x}_{n},\mathrm{t}_{n})\sim\ensuremath{p_{\mathrm{xt}}}$
i.i.d. over $n=1,...,N$. By the strong law of large numbers, we then
have the following limit with probability one
\begin{equation}
-\frac{1}{N}\sum_{n=1}^{N}\ln p(\mathrm{t}_{n}|\mathrm{x}_{n},w,\beta)\rightarrow\textrm{E}_{(\mathrm{x},\mathrm{t})\sim p_{\mathrm{xt}}}\left[-\ln p(\mathrm{t}|\mathrm{x},w,\beta)\right].\label{eq:expected log-loss}
\end{equation}
As we will see next, this limit has a useful interpretation in terms
of the KL divergence.

The KL divergence between two distributions $p$ and $q$ is defined
as
\begin{equation}
\textrm{KL}(p\Vert q)=\textrm{E}_{\mathrm{x}\sim p_{\mathrm{x}}}\left[\ln\frac{p(\mathrm{x})}{q(\mathrm{x})}\right].\label{eq:def of KL divergence Ch2}
\end{equation}
The KL divergence is hence the expectation of the Log-Likelihood Ratio
(LLR) $\ln(p(x)/q(x))$ between the two distributions, where the expectation
is taken with respect to the distribution at the numerator. The LLR
tends to be larger, on average, if the two distributions differ more
significantly, while being uniformly zero only when the two distributions
are equal. Therefore, the KL divergence measures the ``distance''
between two distributions. As an example, with $p(x)=\mathcal{N}(x|\mu_{1},\sigma_{1}^{2})$
and $q(x)=\mathcal{N}(x|\mu_{2},\sigma_{2}^{2})$, we have
\begin{equation}
\textrm{KL}(p\Vert q)=\frac{1}{2}\left(\frac{\sigma_{1}^{2}}{\sigma_{2}^{2}}+\frac{(\mu_{2}-\mu_{1})^{2}}{\sigma_{2}^{2}}-1+\ln\frac{\sigma_{2}^{2}}{\sigma_{1}^{2}}\right),
\end{equation}
and, in the special case $\sigma_{1}^{2}=\sigma_{2}^{2}=\sigma^{2}$,
we can write
\begin{equation}
\textrm{KL}(p\Vert q)=\frac{1}{2}\frac{(\mu_{2}-\mu_{1})^{2}}{\sigma^{2}},
\end{equation}
which indeed increase as the two distributions become more different.

The KL divergence is measured in nats when the natural logarithm is
used as in (\ref{eq:def of KL divergence Ch2}), while it is measured
in bits if a logarithm in base 2 is used. In general, the KL divergence
has several desirable properties as a measure of the distance of two
distributions \cite[pp. 55-58]{Bishop}. The most notable is \emph{Gibbs'
inequality
\begin{equation}
\textrm{KL}(p\Vert q)\geq0,\label{eq:GibbsineqCh2}
\end{equation}
}where equality holds if and only if the two distributions $p$ and
$q$ are identical. Nevertheless, the KL divergence has also some
seemingly unfavorable features, such as its non-symmetry, that is,
the inequality $\textrm{KL}(p\Vert q)\neq\textrm{KL}(q\Vert p)$.
We will see in Chapter 8 that the absence of symmetry can be leveraged
to define different types of approximate inference and learning techniques. 

Importantly, the KL divergence can be written as
\begin{align}
\textrm{KL}(p||q) & =\underbrace{\textrm{E}_{\mathrm{x}\sim p_{\mathrm{x}}}[-\ln q(\mathrm{x})]}_{H(p||q)}-\underbrace{\textrm{E}_{\mathrm{x}\sim p_{\mathrm{x}}}[-\ln p(\mathrm{x})]}_{H(p)},\label{eq:KL entropy Ch2}
\end{align}
where the first term, $H(p||q)$, is known as \emph{cross-entropy
}between $p(x)$ and $q(x)$ and plays an important role as a learning
criterion as discussed below; while the second term, $H(p)$, is the\emph{
entropy} of distribution $p(x)$, which is a measure of randomness.
We refer to Appendix A for further discussion on the entropy. 

Based on the decomposition (\ref{eq:KL entropy Ch2}), we observe
that the cross-entropy $H(p||q)$ can also be taken as a measure of
divergence between two distributions when one is interested in optimizing
over the distribution $q(x)$, since the latter does not appear in
the entropy term. Note that the cross-entropy, unlike the KL divergence,
can be negative.

Using the definition (\ref{eq:def of KL divergence Ch2}), the expected
log-loss on the right-hand side of (\ref{eq:expected log-loss}) can
be expressed as
\begin{equation}
\textrm{E}_{(\mathrm{x},\mathrm{t})\sim p_{\mathrm{xt}}}\left[-\ln p(\mathrm{t}|\mathrm{x},w,\beta)\right]=\textrm{E}_{\mathrm{x}\sim p_{\mathrm{x}}}[\textrm{\ensuremath{H}}(p(t|\textrm{x})||p(t|\mathrm{x},w,\beta))],\label{eq:KL div ML Ch2}
\end{equation}
which can be easily verified by using the law of iterated expectations.
Therefore, the average log-loss is the average over the domain point
$\mathrm{x}$ of the cross-entropy between the real predictive distribution
$p(t|x)$ and the predictive distribution $p(t|x,w,\beta)$ dictated
by the model. From (\ref{eq:KL div ML Ch2}), the ML problem (\ref{eq:ML problem})
can be interpreted as an attempt to make the model-based discriminative
distribution $p(t|x,w,\beta)$ as close as possible to the actual
posterior $p(t|x)$. This is done by minimizing the KL divergence,
or equivalently the cross-entropy, upon averaging over $p(x)$. 

As final remarks, in machine learning, it is common to use the notation
$\textrm{KL}(p\Vert q)$ even when $q$ is \emph{unnormalized}, that
is, when $q(x)$ is non-negative, but we may have the inequality $\int q(x)dx\neq1$.
We also observe that the entropy $H(p)=\textrm{E}_{\mathrm{x}\sim p_{\mathrm{x}}}[-\ln p(\mathrm{x})]$
is non-negative for discrete rvs, while it may be negative for continuous
rvs. Due to its different properties when evaluated for continuous
rvs, the quantity $H(p)$ should be more properly referred to as differential
entropy when the distribution $p$ is a pdf \cite{Cover}. In the
rest of this monograph, we will not always make this distinction.

\section{Interpretation and Causality$^{*}$\label{sec:Interpretation-and-Causality}}

Having learned a predictive model using any of the approaches discussed
above, an important, and often overlooked, issue is the interpretation
of the results returned by the learned algorithm. This has in fact
grown into a separate field within the active research area of deep
neural networks (see Chapter 4) \cite{montavon2017methods}. Here,
we describe a typical pitfall of interpretation that relates to the
assessment of causality relationships between the variables in the
model. We follow an example in \cite{Pearl}.

\begin{figure}[H] 
\centering\includegraphics[scale=0.4]{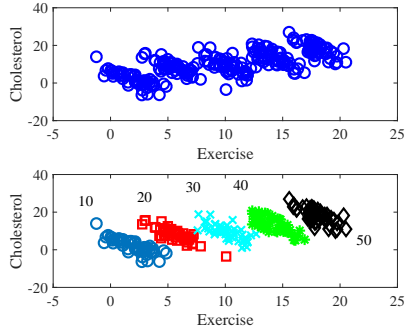}
\caption{Illustration of Simpson's paradox \cite{Pearl}.\label{fig:FigSimpson}} \end{figure}

Fig. \ref{fig:FigSimpson} (top) shows a possible distribution of
data points on the plane defined by coordinates $x=\textrm{exercise}$
and $t=\textrm{cholesterol}$ (the numerical values are arbitrary).
Learning a model that relates $t$ as the dependent variable to the
variable $x$ would clearly identify an upward trend \textendash{}
an individual that exercises more can be predicted to have a higher
cholesterol level. This prediction is legitimate and supported by
the available data, but can we also conclude that exercising less
would reduce one's cholesterol? In other words, can we conclude that
there exists a causal relationships between $x$ and $t$? We know
the answer to be no, but this cannot be ascertained from the data
in the figure.

The way out of this conundrum is to leverage prior information we
have about the problem domain.  In fact, we can explain away this
spurious correlation by including another measurable variable in the
model, namely age. To see this, consider the same data, now redrawn
by highlighting the age of the individual corresponding to each data
point. The resulting plot, seen in Fig. \ref{fig:FigSimpson} (bottom),
reveals that older people \textemdash{} within the observed bracket
\textemdash{} tend to have a higher cholesterol as well as to exercise
more. Therefore, age is a common cause of both exercise and cholesterol
levels. In order to capture the causality relationship between the
latter variables, we hence need to adjust for age. Doing this requires
to consider the trend within each age separately, recovering the expected
conclusion that exercising is useful to lower one\textquoteright s
cholesterol.\footnote{This example is an instance of the so called Simpson's paradox: patterns
visible in the data disappear, or are even reversed, on the segregated
data. }

We conclude that in this example the correlation between $x$ and
$t$, while useful for prediction, should not be acted upon for decision
making. When assessing the causal relationship between $x$ and $t$,
we should first understand which other variables may explain the observations
and then discount any spurious correlations. 

This discussion reveals an important limitation of most existing machine
learning algorithms when it comes to identifying causality relationships,
or, more generally, to answering counterfactual queries \cite{Pearl18}.
The study of causality can be carried out within the elegant framework
of \emph{interventions} on probabilistic graphical models developed
by Pearl \cite{Pearl,Koller,raginsky2011directed}. Other related
approaches are covered in \cite{Peters}. More discussion on probabilistic
graphical models can be found in Chapter 7.

\section{Summary\label{sec:A-Tale-of-Ch2}}

In this chapter, we have reviewed three key learning frameworks, namely
frequentist, Bayesian and MDL, within a parametric probabilistic set-up.
The frequentist viewpoint postulates the presence of a true unknown
distribution for the data, and aims at learning a predictor that generalizes
well on unseen data drawn from this distribution. This can be done
either by learning a probabilistic model to be plugged into the expression
of the optimal predictor or by directly solving the ERM problem over
the predictor. The Bayesian approach outputs a predictive distribution
that combines prior information with the data by solving the inference
problem of computing the posterior distribution over the unseen label.
Finally, the MDL method aims at selecting a model that allows the
data to be described with the smallest number of bits, hence doing
away with the need to define the task of generalizing over unobserved
examples. 

The chapter has also focused extensively on the key problem of overfitting,
demonstrating how the performance of a learning algorithm can be understood
in terms of bias and estimation error. In particular, while choosing
a hypothesis class is essential in order to enable learning, choosing
the ``wrong'' class constitutes an irrecoverable bias that can make
learning impossible. As a real-world example, as reported in \cite{Oneil},
including as independent variables in $x$ the ZIP code of an individual
seeking credit at a bank may discriminate against immigrants or minorities.
Another example of this phenomenon is the famous experiment by B.
F. Skinner on pigeons \cite{Shalev}. 

We conclude this chapter by emphasizing an important fact about the
probabilistic models that are used in modern machine learning applications.
In frequentist methods, typically at least two (possibly conditional)
distributions are involved: the empirical data distribution and the
model distribution. The former amounts to the histogram of the data
which, by the law of large numbers, tends to the real distribution
when the number of data points goes to infinity; while the latter
is parametrized and is subject to optimization. For this reason, divergence
metrics between the two distributions play an important role in the
development of learning algorithms. We will see in the rest of the
monograph that other frequentist methods may involve a single distribution
rather than two, as discussed in Sec. \ref{sec:Discriminative-Models-Ch6},
or an additional, so called variational, distribution, as covered
in Sec. \ref{sec:ML,-ELBO-and-Ch6} and Chapter 8. 

In contrast, Bayesian methods posit a single coherent distribution
over the data and the parameters, and frame the learning problem as
one of inference of unobserved variables. As we will discuss in Chapter
8, variational Bayesian methods also introduce an additional variational
distribution and are a building block for frequentist learning in
the presence of unobserved variables.

The running example in this chapter has been one of linear regression
for a Gaussian model. The next chapter provides the necessary tools
to construct and learn more general probabilistic models.

\chapter{Probabilistic Models for Learning}

In the previous chapter, we have introduced the frequentist, Bayesian,
and MDL learning frameworks. As we have seen, parametric probabilistic
models play a key role for all three of them. The linear regression
example considered in the previous chapter was limited to a simple
linear Gaussian model, which is insufficient to capture the range
of learning problems that are encountered in practice. For instance,
scenarios of interest may encompass discrete variables or non-negative
quantities. In this chapter, we introduce a family of probabilistic
models, known as the \emph{exponential family}, whose members are
used as components in many of the most common probabilistic models
and learning algorithms. The treatment here will be leveraged in the
rest of the monograph in order to provide the necessary mathematical
background. Throughout this chapter, we will specifically emphasize
the common properties of the models in the exponential family, which
will prove useful for deriving learning algorithms in the following
chapters.

\section{Preliminaries\label{sec:Preliminaries_Ch3}}

We start with a brief review of some definitions that will be used
throughout the chapter and elsewhere in the monograph (see \cite{Boyd}
for more details). Readers with a background in convex analysis and
calculus may just review the concept of sufficient statistic in the
last paragraph.

First, we define a \emph{convex set} as a subset of $\mathbb{R}^{D}$,
for some $D$, that contains all segments between any two points in
the set. Geometrically, convex sets hence cannot have ``indentations''.
Function $f(x)$ is convex if its domain is a convex set and if it
satisfies the inequality $f(\lambda x+(1-\lambda)y)\leq\lambda f(x)+(1-\lambda)f(y)$
for all $x$ and $y$ in its domain and for all $0\leq\lambda\leq1$.
Geometrically, this condition says that the function is ``$\cup$''-shaped:
the curve defining the function cannot lie above the segment obtained
by connecting any two points on the curve. A function is strictly
convex if the inequality above is strict except for $\lambda=0$ or
$\lambda=1$ when $x\neq y$. A concave, or strictly concave, function
is defined by reversing the inequality above \textendash{} it is hence
``$\cap$''-shaped.

The minimization of a convex (``$\cup$'') function over a convex
constraint set or the maximization of a concave (``$\cap$'') function
over a convex constraint set are known as \emph{convex optimization
problems}. For these problems, there exist powerful analytical and
algorithmic tools to obtain globally optimal solutions \cite{Boyd}.

We also introduce two useful concepts from calculus. The \emph{gradient}
of a differentiable function $f(x)$ with $x=[x_{1}\cdots x_{D}]^{T}\in\mathbb{R}^{D}$
is defined as the $D\times1$ vector $\nabla f(x)=[\partial f(x)/\partial x_{1}\cdots\partial f(x)/\partial x_{D}]^{T}$
containing all partial derivatives. At any point $x$ in the domain
of the function, the gradient is a vector that points to the direction
of locally maximal increase of the function. The \emph{Hessian} $\nabla^{2}f(x)$
is the $D\times D$ matrix with ($i,j$) element given by the second-order
derivative $\partial^{2}f(x)/\partial x_{i}\partial x_{j}$. It captures
the local curvature of the function. 

Finally, we define the concept of \emph{sufficient statistic}. Consider
a rv $\mathrm{x}\sim p(x|\theta)$, whose distribution depends on
some parameter $\theta$. A function $f(x)$ is a sufficient statistic\footnote{A statistic is a function of the data.}
for the estimate of $\theta$ if the likelihood $p(x|\theta)$ of
the parameters $\theta$ depends on $x$ only through the function
$f(x)$. As an example, for a rv $\mathrm{x}\sim\mathcal{N}(0,\sigma^{2})$,
the function $f(x)=x^{2}$ can be easily seen to be sufficient for
the estimate of the variance $\sigma^{2}.$

\section{The Exponential Family\label{sec:The-Exponential-Family}}

In this section, we introduce the exponential family of parametric
probabilistic models. As it will be discussed, this family includes
as special cases most of the distributions typically assumed when
solving machine learning problems. For example, it includes Gaussian,
Laplace, Gamma, Beta and Dirichlet pdfs, as well as Bernoulli, Categorical,
multinomial, and Poisson pmfs. An extensive list can be found in \cite{Wikiexp}. 

\subsection{Basic Definitions}

The exponential family contains probabilistic models of the form 
\begin{align}
p(x|\eta) & =\frac{1}{Z(\eta)}\exp\left(\sum_{k=1}^{K}\eta_{k}u_{k}(x)\right)m(x)\nonumber \\
 & =\frac{1}{Z(\eta)}\exp\left(\eta^{T}u(x)\right)m(x),\label{eq:exponential family}
\end{align}
where $x$ is a discrete or continuous-valued vector; $\eta=[\eta_{1}\cdots\eta_{K}]^{T}$
is the vector of \emph{natural parameters}; $u(x)=[u_{1}(x)\cdots u_{K}(x)]^{T}$
is the vector of \emph{sufficient statistics}, with each sufficient
statistic $u_{k}(x)$ being a function of $x$; $m(x)\geq0$ is the
\emph{base measure}, which is a function of $x$ that is\emph{ }independent
of the natural parameter vector $\eta;$ and $Z(\eta)$ is the \emph{partition
function} 
\begin{equation}
Z(\eta)=\int\exp\left(\eta^{T}u\left(x\right)\right)m\left(x\right)\mathrm{\textrm{d}}x
\end{equation}
for continuous rvs and $Z(\eta)=\sum_{x}\exp\left(\eta^{T}u\left(x\right)\right)m\left(x\right)$
for discrete rvs. The sufficient statistic vector $u(x)$ can be seen
to be a sufficient statistic for the estimation of the natural parameter
vector $\eta$ given $x$. 

The partition function normalizes the distribution so that it integrates,
or sums, to one. It is also often useful to use the \emph{unnormalized}
distribution
\begin{equation}
\tilde{p}(x|\eta)=\exp\left(\eta^{T}u(x)\right)m(x),
\end{equation}
since the latter is generally easier to evaluate. 

In short, distributions belonging to the exponential family are such
that the logarithm of the unnormalized distribution $\tilde{p}(x|\eta)$,
which is also known as the \emph{energy function}, is linear\footnote{Or, more precisely, affine given the presence of an additive constant.}
in the natural parameters, i.e.,
\begin{equation}
\ln\tilde{p}(x|\eta)=\eta^{T}u(x)+\ln m\left(x\right).
\end{equation}
For this reason, models of the form (\ref{eq:exponential family})
are also referred to as \emph{log-linear}.\footnote{There exists a more general version of the exponential family in which
the natural parameters are non-linear functions of the parameters
that identify the distribution. } Including the partition function, the LL of the natural parameters
can be written as
\begin{equation}
\ln p(x|\eta)=\eta^{T}u\left(x\right)-A(\eta)+\ln m(x),\label{eq:loglikelihood exp}
\end{equation}
where 
\begin{equation}
A(\eta)=\ln Z(\eta)\label{eq:logpartition}
\end{equation}
is the \emph{log-partition function. }

As per (\ref{eq:exponential family}), a probabilistic model belonging
to the exponential family is identified by the set of sufficient statistics
$\{u_{k}(x)\}_{k=1}^{K}$, whose order is irrelevant, and by the measure
$m(x)$. A specific hypothesis within the model is selected by determining
the natural parameter vector $\eta$. The set of feasible values for
the natural parameters contains all, and only, the vectors $\eta$
for which the unnormalized distribution $\tilde{p}(x|\eta)$ can be
normalized, that is, for which the inequality $A(\eta)<\infty$ holds.
We will see below that this set is convex. 

\medskip{}

\begin{example}\textbf{ (Gaussian pdf)} As a first example, consider
the Gaussian pdf
\begin{align*}
\mathcal{N}(x|\nu,\sigma^{2} & )=\frac{1}{(2\pi\sigma^{2})^{1/2}}\exp\left(\mathrm{-}\frac{x^{2}}{2\sigma^{2}}+\frac{\nu}{\sigma^{2}}x-\frac{\nu^{2}}{2\sigma^{^{2}}}\right).
\end{align*}
This can be written in the form (\ref{eq:exponential family}) upon
identification of the base measure $m(x)=1$ and of the sufficient
statistics $u(x)=[x\textrm{ \ensuremath{x^{2}}}]$. Note that, in
order to do this, we need to map the\emph{ }parameters $(\nu,\sigma^{2})$
to the natural parameters via the relationship $\eta=[\nu/\sigma^{2}\textrm{\ensuremath{\textrm{ }-1/}\ensuremath{(2}\ensuremath{\sigma^{2}}) }]$.
As a result, while the parameters $(\nu,\sigma^{2})$ take all possible
allowed values in $\mathbb{R}\times\mathbb{R}^{+}$, the natural parameter
vector $\eta$ takes values in the set $\mathbb{R}\times\mathbb{R}^{-}$.
Every value $\eta$ in this set corresponds to a valid pdf within
the hypothesis class of $\mathcal{N}(x|\nu,\sigma^{2})$ models. Finally,
we can compute the log-partition function as 
\begin{equation}
A(\eta)=\frac{\nu^{2}}{2\sigma^{2}}+\frac{1}{2}\text{ln}(2\pi\sigma^{2})=-\frac{\eta_{1}^{2}}{4\eta_{2}}-\frac{1}{2}\text{ln}\left(-\left(\frac{2\eta_{2}}{2\pi}\right)\right).
\end{equation}
\end{example}

\medskip{}

In order to ensure identifiability of the natural parameters, the
sufficient statistics $\{u_{k}(x)\}_{k=1}^{K}$ need to be linearly
independent. This means that no sufficient statistic $u_{k}(x)$ should
be computable, for all $x$, as a linear combination of other sufficient
statistics $u_{k'}(x)$ with $k'\neq k$. For example, this is the
case for the vector $u(x)=[x\textrm{ \ensuremath{x^{2}}}]$ of sufficient
statistics for the Gaussian distribution. This condition is referred
to as \emph{minimal representation} \cite{Wainwright}. Unless stated
otherwise, we will assume in the following that the exponential family
under study is minimal. Furthermore, we will assume the technical
condition that the set of feasible values for $\eta$ is also open
(that is, it excludes its boundary), which yields a \emph{regular}
exponential family.

\subsection{Natural Parameters, Mean Parameters and Convexity}

As suggested by the example above, once the sufficient statistics
and the base measure are fixed, a specific hypothesis \textendash{}
pdf or pmf \textendash{} can be identified by either specifying the
vector $\eta$ of natural parameters or the vector $\mu$ of \emph{mean
parameters. }The latter is defined as the expectation of the vector
of sufficient statistics 
\begin{equation}
\mu=\mathrm{E}_{\mathrm{x}\sim p(x|\eta)}[u(\mathrm{x})].
\end{equation}
For the preceding example, we have the mean parameter vector $\mu=[\textrm{E}[\mathrm{x}]=\nu,\textrm{ E\ensuremath{[\mathrm{x}^{2}]=\sigma^{2}+\nu^{2}}}]^{T}$.
We can therefore use the notation $p(x|\mu)$ as well as $p(x|\eta)$
to describe a model in the exponential family. As we will see below,
the availability of these two parametrizations implies that learning
can be done on either sets of variables. 

A key property of the exponential family is that the mapping between
natural parameters and mean parameters is given by the gradient of
the log-partition function. Specifically, it can be verified that
the partial derivative of the log-partition function equals the mean
of the corresponding sufficient statistic 
\begin{equation}
\frac{\partial A(\eta)}{\partial\eta_{k}}=\textrm{E}_{\mathrm{x}\sim p(x|\eta)}\left[u_{k}\left(\mathrm{x}\right)\right]=\mu_{k},\label{eq:gradient log part}
\end{equation}
or, in vector form, 
\begin{equation}
\nabla_{\eta}A(\eta)=\textrm{E}_{\mathrm{x}\sim p(x|\eta)}\left[u\left(\mathrm{x}\right)\right]=\mu.\label{eq:gradient log part-1}
\end{equation}
Although we will not be making use of this result here, the Hessian
$\nabla_{\eta}^{2}A(\eta)$ of the log-partition function can be similarly
seen to equal the covariance matrix of the sufficient statistics\footnote{More generally, the log-partition function $A(\eta)$ is the cumulant
function for rv $\mathrm{x}.$}. It is also equal to the Fisher information matrix for the natural
parameters \cite{Duchicourse}.

The log-partition function $A(\eta)$ in (\ref{eq:logpartition})
is strictly convex in $\eta$, as it follows from the fact that it
is a log-sum-exp function composed with an affine function \cite{Boyd}.
This property has the following important consequences. First, the
set of feasible values for the natural parameters is a convex set
\cite{Boyd}. Note that this is generally not the case for the corresponding
set of feasible values for the mean parameters. Second, the mapping
(\ref{eq:gradient log part-1}) between natural parameters $\eta$
and mean parameters $\mu$ is invertible (see, e.g., \cite{azoury2001relative})\footnote{The inverse mapping between mean parameters and natural parameters
is given by the gradient $\nabla_{\mu}A^{*}(\mu)$ of the convex conjugate
function $A^{*}(\mu)=\sup_{\eta}$$\eta^{T}\mu-A(\eta)$, where the
maximization is over the feasible set of natural parameters. }. 

Third, the LL function $\ln p(x|\eta)$ in (\ref{eq:loglikelihood exp})
is a concave function of $\eta$. As further discussed below, the
ML problem hence amounts to maximization of a convex optimization
problem. 

\subsection{Bernoulli Model}

Due to its importance for binary classification problems, we detail
here the Bernoulli model. We also introduce the important logistic
sigmoid function. 

The Bernoulli distribution for a binary rv $\mathrm{x}\in\{0,1\}$
is given as
\begin{equation}
\textrm{Bern}(x|\mu)=\mu^{x}\left(1-\mu\right)^{1-x},
\end{equation}
with $\mu=\textrm{E}_{\mathrm{x}\sim\textrm{Bern}(x|\mu)}\left[\mathrm{x}\right]=\textrm{Pr}[\mathrm{x}=1]$.
Since we can write the LL function as
\begin{equation}
\textrm{ln}\left(\textrm{Bern}(x|\mu)\right)=\textrm{ln\ensuremath{\left(\frac{\mu}{1-\mu}\right)x}}+\textrm{ln}\left(1-\mu\right),
\end{equation}
the sufficient statistic defining the Bernoulli model is $u(x)=x$
and the measure function is $m(x)=1$. The mapping between the natural
parameter $\eta$ and the mean parameter $\mu$ is given as 
\begin{equation}
\eta=\text{ln}\left(\frac{\mu}{1-\mu}\right),\label{eq:logit}
\end{equation}
that is, the natural parameter is the LLR $\eta=\textrm{ln}(\textrm{Bern}(1|\mu)/\textrm{Bern}(0|\mu))$.
Function (\ref{eq:logit}) is also known as the \emph{logit function}.
The corresponding set of feasible values is hence $\mathbb{R}$. 

The inverse mapping is instead given by the \emph{logistic sigmoid
function 
\begin{equation}
\mu=\sigma(\eta)=\frac{1}{1+e^{-\eta}}.
\end{equation}
}The sigmoid function converts a real number into the interval $[0,1]$
via an S-shape that associates values less than 0.5 to negative values
of the argument and larger values to positive numbers. Finally, the
log-partition function is given by the convex function of the natural
parameters
\begin{equation}
A(\eta)=-\text{ln}(1-\mu)=\text{ln}(1+e^{\eta}).
\end{equation}
Note that the relationship (\ref{eq:gradient log part-1}) is easily
verified.

\subsection{Categorical or Multinoulli Model}

For its role in multi-class classification, we introduce here in some
detail the Categorical or Multinoulli distribution, along with the
one-hot encoding of categorical variables and the soft-max function.

The Categorical model applies to discrete variables taking $C$ values,
here labeled without loss of generality as $\{0,1,...,C-1\}.$ Note
that setting $C=2$ recovers the Bernoulli distribution. Pmfs in this
model are given as
\begin{equation}
\textrm{Cat}(x|\mu)=\prod_{k=1}^{C-1}\mu_{k}^{1\left(x=k\right)}\times\mu_{0}^{1-\sum_{k=1}^{C-1}1(x=k)},
\end{equation}
where we have defined $\mu_{k}=\textrm{Pr}\left[\mathrm{x}=k\right]$
for $k=1,\ldots,C-1$ and $\mu_{0}=1-\sum_{k=1}^{C-1}\mu_{k}=\textrm{Pr}\left[\mathrm{x}=0\right]$.
The LL function is given as
\begin{equation}
\textrm{ln}\left(\textrm{Cat}(x|\mu)\right)=\sum_{k=1}^{C-1}1\left(x=k\right)\textrm{ln}\frac{\mu_{k}}{\mu_{0}}\textrm{+ln}\mu_{0}.
\end{equation}
This demonstrates that the categorical model is in the exponential
family, with sufficient statistics vector $u(x)=[1(x=1)\cdots1(x=C-1)]^{T}$
and measure function $m(x)=1$. Furthermore, the mean parameters $\mu=[\mu_{1}\cdots\mu_{C-1}]^{T}$
are related to the natural parameter vector $\eta=[\eta_{1}\cdots\eta_{C-1}]^{T}$
by the mapping 
\begin{equation}
\eta_{k}=\ln\left(\frac{\mu_{k}}{1-\sum_{j=1}^{C-1}\mu_{j}}\right),
\end{equation}
 which again takes the form of an LLR. The inverse mapping is given
by 
\begin{align}
 & \mu=\begin{bmatrix}\frac{e^{\eta_{1}}}{1+\sum_{k=1}^{C-1}e^{\eta_{k}}}\\
\vdots\\
\frac{e^{\eta_{C-1}}}{1+\sum_{k=1}^{C-1}e^{\eta_{k}}}
\end{bmatrix}.
\end{align}

The parametrization given here is minimal, since the sufficient statistics
$u(x)$ are linearly independent. An overcomplete representation would
instead include in the vector of sufficient statistics also the function
$1(x=0)$. In this case, the resulting vector of sufficient statistics
\begin{equation}
u(x)=\begin{bmatrix}1(x=0)\\
\vdots\\
1(x=C-1)
\end{bmatrix}
\end{equation}
is known as \emph{one-hot encoding} of the categorical variable, since
only one entry equals 1 while the others are zero. Furthermore, with
this encoding, the mapping between the natural parameters and the
mean parameters $\mu=[\mu_{0}\cdots\mu_{C-1}]^{T}$ can be expressed
in terms of the softmax function 
\begin{align}
 & \mu=\textrm{softmax(\ensuremath{\eta})=}\begin{bmatrix}\frac{e^{\eta_{0}}}{\sum_{k=0}^{C-1}e^{\eta_{k}}}\\
\vdots\\
\frac{e^{\eta_{C-1}}}{\sum_{k=0}^{C-1}e^{\eta_{k}}}
\end{bmatrix}.
\end{align}
The softmax function $\textrm{softmax}(\ensuremath{\eta})$ converts
a vector of ``scores'' $\eta$ into a probability vector. Furthermore,
the function has the property that, as $c$ grows to infinity, $\textrm{softmax(\ensuremath{c\eta})}$
tends to a vector with all zero entries except for the position corresponding
to the maximum value $\eta_{k}$ (assuming that it is unique). This
justifies its name.

\section{Frequentist Learning\label{sec:Frequentist-Learning Ch3}}

In this section, we provide general results concerning ML and MAP
learning when the probabilistic model belongs to the exponential family.
As seen in the previous chapter, with ML and MAP, one postulates that
the $N$ available data points $x_{\mathcal{D}}=\{x_{1},\ldots,x_{N}\}$
are i.i.d. realizations from the probabilistic model $p(x|\eta)$
as 
\begin{equation}
\mathrm{x}_{n}\underset{\textrm{i.i.d.}}{\sim}p(x|\mathbf{\eta}),\textrm{ \ensuremath{n=1,...,N}}.
\end{equation}
This data is used to estimate the natural parameters $\mathbf{\eta}$,
or the corresponding mean parameters $\mu$. 

Using (\ref{eq:loglikelihood exp}), the LL of the natural parameter
vector given the observation $x_{\mathcal{D}}$ can be written as
\begin{align}
\text{ln}p(x_{\mathcal{D}}|\eta) & =\sum_{n=1}^{N}\text{ln}p(x_{n}|\eta)\nonumber \\
 & =-NA(\eta)+\eta^{T}\sum_{n=1}^{N}u(x_{n})+\sum_{n=1}^{N}\textrm{ln\ensuremath{m}}(x_{n}).
\end{align}
Therefore, neglecting terms independent of $\eta$, we can write
\begin{align}
\text{ln}p(x_{\mathcal{D}}|\eta) & =-NA(\eta)+\eta^{T}u(x_{\mathcal{D}}),\label{eq:loglikelihood exp data}
\end{align}
where we have defined the cumulative sufficient statistics
\begin{equation}
u_{k}(x_{\mathcal{D}})=\sum_{n=1}^{N}u_{k}(x_{n}),\textrm{ }k=1,...,K,
\end{equation}
and the vector $u(x_{\mathcal{D}})=[u_{1}(x_{\mathcal{D}})\cdots u_{K}(x_{\mathcal{D}})]^{T}$.

A first important observation is that the LL function only depends
on the $K$ statistics $u_{k}(x_{\mathcal{D}})$, $k=1,...,K$. Therefore,
the vector $u(x_{\mathcal{D}})$ is a sufficient statistic for the
estimate of $\eta$ given the observation $x_{\mathcal{D}}$. Importantly,
vector $u(x_{\mathcal{D}})$ is of size $K$, and hence it does not
grow with the size $N$ of the data set. In fact, the exponential
family turns out to be unique in this respect: Informally, among all
distributions whose support does not depend on the parameters, only
distributions in the exponential family have sufficient statistics
whose number does not grow with the number $N$ of observations (Koopman-Pitman-Darmois
theorem) \cite{angelino2016patterns}.

\textbf{Gradient of the LL function.} A key result that proves very
useful in deriving learning algorithms is the expression of the gradient
of the LL (\ref{eq:loglikelihood exp data}) with respect to the natural
parameters. To start, the partial derivative with respect to $\eta_{k}$
can be written as
\begin{equation}
\frac{\partial\textrm{ln}p(x_{\mathcal{D}}|\eta)}{\partial\eta_{k}}=u_{k}(x_{\mathcal{D}})-N\frac{\partial A(\eta)}{\partial\eta_{k}}.
\end{equation}
Using (\ref{eq:gradient log part}) and (\ref{eq:gradient log part-1}),
this implies that we have
\begin{equation}
\frac{1}{N}\frac{\partial\textrm{ln}p(x_{\mathcal{D}}|\eta)}{\partial\eta_{k}}=\frac{1}{N}u_{k}(x_{\mathcal{D}})-\mu_{k},
\end{equation}
and for the gradient 
\begin{align}
\frac{1}{N}\nabla_{\eta}\textrm{ln}p(x_{\mathcal{D}}|\eta) & =\frac{1}{N}u(x_{\mathcal{D}})-\mu\nonumber \\
 & =\frac{1}{N}\sum_{n=1}^{N}u(x_{n})-\mu.\label{eq:gradient exp}
\end{align}
The gradient (\ref{eq:gradient exp}) is hence given by the difference
between the empirical average $N^{-1}\sum_{n=1}^{N}u(x_{n})$ of the
sufficient statistics given the data $x_{\,\mathcal{D}}$ and the
ensemble average $\mu$.

The following observation is instrumental in interpreting algorithms
based on gradient ascent or descent for exponential families. The
normalized gradient of the LL (\ref{eq:gradient exp}) has two components:
(\emph{i}) the \emph{``positive'' component} $u(x_{\mathcal{D}})/N$
points in a direction of the natural parameter space that maximizes
the unnormalized distribution $\ln\tilde{p}(x_{\mathcal{D}}|\eta)=\eta^{T}u(x_{\mathcal{D}})+\sum_{n=1}^{N}\ln m(x_{n})$,
hence maximizing the ``fitness'' of the model to the observed data
$x_{\mathcal{D}}$; while (\emph{ii}) the \emph{``negative'' component}
$-\mu=-\nabla_{\eta}A(\eta)$ points in a direction that minimizes
the partition function, thus minimizing the ``fitness'' of the model
to the unobserved data. The tension between these two components is
resolved when the empirical expectation of the sufficient statistics
equals that under the model, as discussed below.

\textbf{ML Learning.} Due to concavity of the LL function, or equivalently
convexity of the NLL, and assuming the regularity of the distribution,
the ML estimate $\text{\ensuremath{\eta}\ensuremath{\ensuremath{_{ML}}}}$
is obtained by imposing the optimality condition 
\begin{equation}
\nabla_{\eta}\text{ln}p(x_{\mathcal{D}}|\eta)=0,\label{eq:opt condition}
\end{equation}
which gives
\begin{equation}
\mu_{\text{\ensuremath{ML}}}=\frac{1}{N}\sum_{n=1}^{N}u(x_{n}).\label{eq:ML mean parameters exp}
\end{equation}
In words, the ML estimate of the mean parameters is obtained by matching
the ensemble averages obtained under the model to the empirical averages
observed from the data. This procedure is known as \emph{moment matching.
}Note that,\emph{ }from (\ref{eq:ML mean parameters exp}), if needed,
we can also compute the ML estimate $\eta_{\text{\ensuremath{ML}}}$
using the mapping between the two sets of parameters.

From (\ref{eq:ML mean parameters exp}), we can infer that the ML
estimate is \emph{consistent}: if the data is generated from a distribution
$p(x|\eta^{*})$ within the assumed family, the ML estimate will tend
to it with probability one as $N$ grows to infinity by the strong
law of large numbers \cite{lehmann2006theory}. However, for finite
$N$, ML may suffer from overfitting, as we will see next.

\begin{example}\textbf{ (Gaussian pdf).} The ML estimates of the
parameters $(\mu,\,\sigma^{2}$) for the Gaussian model are given
as 
\begin{align}
\mu_{\text{\ensuremath{ML}}} & =\frac{1}{N}\sum_{n=1}^{N}x_{n}\\
\sigma_{\text{\ensuremath{ML}}}^{2} & =\frac{1}{N}\sum_{n=1}^{N}x_{n}^{2}-\mu_{\text{\ensuremath{ML}}}^{2}.
\end{align}
For the Bernoulli model, we have the ML estimate of the mean parameter
$\mu=\textrm{Pr}\left[\mathrm{x}=1\right]$ 
\begin{equation}
\mu_{\text{\ensuremath{ML}}}=\frac{1}{N}\sum_{n=1}^{N}x_{n}=\frac{N[1]}{N},
\end{equation}
where $N\left[k\right]$ measures the number of observations equal
to $k$, i.e., 
\[
N[k]=|\{n:\;x_{n}=k\}|.
\]
Note that $N[1]$ has a binomial distribution. For the Categorical
model, we can similarly write the ML estimate of the mean parameters
$\mu_{k}=\textrm{Pr}\left[\mathrm{x}=k\right]$ as
\begin{align}
\mu_{k,\text{\ensuremath{ML}}} & =\frac{1}{N}\sum_{n=1}^{N}1(x_{n}=k)=\frac{N[k]}{N}.\label{eq:categorical ML}
\end{align}
The vector $[N[0],...,N[C-1]]^{T}$ has a multinomial distribution.\end{example}

To illustrate the problem of overfitting, consider the categorical
model. As per (\ref{eq:categorical ML}), if no instance of the data
equal to some value $k$ is observed, i.e., if $N[k]=0$, ML assigns
a zero probability to the event $\mathrm{x}=k$. In mathematical terms,
if $N[k]=0,$ the ML estimate of the probability of the event $\mathrm{x}=k$
is zero, that is, $\mu_{k,\text{\ensuremath{ML}}}=0$. So, ML gives
zero probability to any previously unobserved event. The problem,
which is an instance of overfitting, is known as the \emph{black swan
paradox }or\emph{ zero-count problem}: For the European explorers
of the 17th century \textendash{} or at least for those of them adhering
to the ML principle \textendash{} the black swans in the Americas
could not exist! \cite{Murphy}

\textbf{MAP Learning.} A MAP solution can be in principle derived
by including in the optimality condition (\ref{eq:opt condition})
the gradient of the prior distribution. We will instead solve the
MAP problem in the next section by computing the mode of the posterior
distribution of the parameters.

\section{Bayesian Learning}

As we discussed in the previous chapter, the Bayesian viewpoint is
to treat all variables as jointly distributed, including the model
parameters $\mu$ and the new, unobserved value $\mathrm{x}.$ The
joint distribution is given as
\begin{equation}
p(x_{\mathcal{D}},\mu,x|\alpha)=\underbrace{p(\mu|\alpha)}_{\text{a priori distribution}}\underbrace{p(x_{\mathcal{D}}|\mu)}_{\text{ likelihood}}\underbrace{p(x|\mu)}_{\text{ distribution of new data}},
\end{equation}
where $\alpha$ represents the vector of \emph{hyperparameters} defining
the prior distribution. The problem of inferring the unobserved value
$\mathrm{x}$ is solved by evaluating the predictive distribution
\begin{equation}
p(x|x_{\mathcal{D}},\alpha)=\int p(\mu|x_{\mathcal{D}},\alpha)p(x|\mu)\text{\ensuremath{d}}\mu.\label{eq:computing predictive}
\end{equation}
This distribution accounts for the weighted average of the contributions
from all values of the parameter vector $\mu$ according to the posterior
distribution $p(\mu|x_{\mathcal{D}},\alpha).$ Note that, for clarity,
we left indicated the dependence on the hyperparameters $\alpha.$
Using Bayes' theorem, the posterior of the parameter vector can be
written as
\begin{equation}
p(\mu|x_{\mathcal{D}},\alpha)=\frac{p(\mu|\alpha)p(x_{\mathcal{D}}|\mu)}{p(x_{\mathcal{D}}|\alpha)}\propto p(\mu|\alpha)p(x_{\mathcal{D}}|\mu).\label{eq:computing posterior}
\end{equation}
As discussed in Chapter 2, this relationship highlights the dependence
of the posterior on both the prior distribution and the likelihood
$p(x_{\mathcal{D}}|\mu)$. We also note the the denominator in (\ref{eq:computing posterior})
is the marginal likelihood.

\textbf{Prior distribution.} The first issue we should address is
the choice of the prior distribution. There are two main approaches:
1) \emph{Conjugate prior}: choose the prior $p(\mu|\alpha)$, so that
posterior $p(\mu|x_{\mathcal{D}},\alpha)$ has the same distribution
as the prior $p(\mu|\alpha)$ but with generally different parameters;
2) \emph{Non-informative prior}: choose the prior that is the least
informative given the observed data \cite[pp. 117-120]{Bishop}. Here,
we will work with conjugate priors, which are more commonly adopted
in applications. In fact, a key advantage of models in the exponential
family is that they all admit conjugate priors, and the conjugate
priors are also members of the exponential family.

Rather than providing a general discussion, which may be of limited
practical use, we proceed by means of representative examples. A table
of models with corresponding prior distributions can be found in \cite{Wikiconj}.

\subsection{Beta-Bernoulli Model{\label{subsec:Beta-Bernoulli-Model}}}

\begin{figure}[H] 
\centering\includegraphics[scale=0.45]{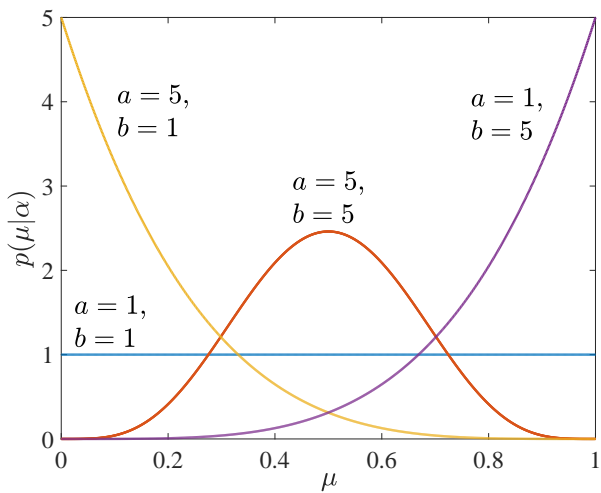}
\caption{Beta distribution with different values of the hyperparameters $a,b\geq1.$ \label{fig:Fig1Ch3}} \end{figure}

The Beta-Bernoulli model is suitable to study binary data. Conditioned
on the parameter $\mu=\textrm{Pr}\left[\mathrm{x}=1\right]$, the
pmf of the $N$ i.i.d. available observations $x_{\mathcal{D}}$ with
$x_{n}\in\{0,1\}$ is given as
\begin{equation}
p(x_{\mathcal{D}}|\mu)=\prod_{n=1}^{N}\textrm{Bern}(x_{n}|\mu)=\mu^{N[1]}(1-\mu)^{N[0]}.\label{eq:likelihood Bernoulli}
\end{equation}
As seen, a conjugate prior should be such that the posterior (\ref{eq:computing posterior})
has the same distribution of the prior but with different parameters.
For the likelihood (\ref{eq:likelihood Bernoulli}), the conjugate
prior is the \emph{Beta distribution}, which is defined as
\begin{equation}
\begin{split}p(\mu|a,b) & =\text{Beta}(\mu|a,b)\propto\mu^{a-1}(1-\mu)^{b-1},\end{split}
\label{eq:beta}
\end{equation}
where $a$ and $b$ are hyperparameters and the normalization constant
is not made explicit in order to simplify the notation. It is worth
emphasizing that (\ref{eq:beta}) is a probability distribution on
a probability $\mu$. Plots of the beta pdf for different values of
$a,b\geq1$ can be found in Fig. \ref{fig:Fig1Ch3}. 

Average and mode of the Beta pdf can be evaluated as
\begin{align}
\text{E}_{\mu\sim\text{Beta}(\mu|a,b)}[\mu] & =\frac{a}{a+b}\label{eq:mean beta}\\
\text{mode}_{\text{\ensuremath{\mu\sim}Beta}(\mu|a,b)} & [\mu]=\frac{a-1}{a+b-2}\quad,\label{eq:mode beta}
\end{align}
where the mode expression is only valid when $a,b>1$ (when this condition
is not met, the distribution is multi-modal, see Fig. \ref{fig:Fig1Ch3}).
The mean (\ref{eq:beta}) suggests that the hyperparameters $a$ and
$b$ can be interpreted as the number of observations that are expected
to equal ``1'' and ``0'', respectively, out of a total number
of $a+b$ measurements, based on prior information alone. More fittingly,
as we shall see next, we can think of these a priori observations
as ``virtual'' measurements, also known as \emph{pseudo-counts},
that should be used alongside the actual measurements $x_{\mathcal{D}}$
during learning.

We can now compute the posterior distribution of the parameter vector
using (\ref{eq:computing posterior}) as
\begin{align}
p(\mu|x_{\mathcal{D}},a,b) & \propto\text{Beta}\left(\mu\left|a+N[1],b+N[0]\right.\right)\nonumber \\
 & =\mu^{N[1]+a-1}(1-\mu)^{N[0]+b-1}.\label{eq:posterior beta}
\end{align}
This confirms that the posterior distribution is indeed a Beta pdf,
as dictated by the choice of a Beta conjugate prior. Furthermore,
the posterior Beta distribution has parameters $a+N[1]$ and $b+N[0]$.
This is consistent with the interpretation given above: the total
number of \emph{effective} observations equal to ``1'' and ``0''
are $a+N[1]$ and $b+N[0]$, respectively. As anticipated, the MAP
estimate of $\mu$ can be obtained by taking the mode of the posterior
(\ref{eq:computing posterior}), yielding
\begin{equation}
\mu_{\text{\ensuremath{MAP}}}=\frac{a+N[1]-1}{a+b+N-2}.
\end{equation}
As we discussed in the previous chapter, we have the limit $\mu_{\text{\ensuremath{MAP}}}\rightarrow\mu_{ML}$
for $N\rightarrow\infty$.

Back to the Bayesian viewpoint, the predictive distribution (\ref{eq:computing predictive})
is given as
\begin{align}
p(x=1|x_{\mathcal{D}},a,b) & =\int p(\mu|x_{\mathcal{D}},a,b)p(x=1|\mu)\text{\ensuremath{d}}\mu\nonumber \\
 & =\text{E}_{\mu\sim p(\mu|x_{\mathcal{D}},a,b)}[\mu]=\frac{N[1]+a}{N+a+b}\label{eq:predictive beta bernoulli}
\end{align}
where we have used the expression (\ref{eq:mean beta}) for the mean
of a Beta rv. We observe that, if $N$ is small, the predictive probability
is approximately equal to the mean of the prior, i.e., $p(x=1|x_{\mathcal{D}},a,b)\approx a/(a+b)$;
while, if $N$ is large, as we have seen in the previous chapter,
the predictive probability tends to the ML solution, i.e., $p(x=1|x_{\mathcal{D}},a,b)\approx N[1]/N.$ 

As an example of a non-conjugate prior, one could choose the distribution
of the natural parameter $\eta$, or equivalently of the logit function
(\ref{eq:logit}) of $\mu$, as Gaussian \cite{lunn2012bugs}. 

The following example illustrates the potential advantages, already
discussed in Chapter 2, of the Bayesian approach in avoiding overfitting.
Note that MAP would also yield similar results in this example.

\begin{figure} 
\centering 
\includegraphics[scale=0.4]{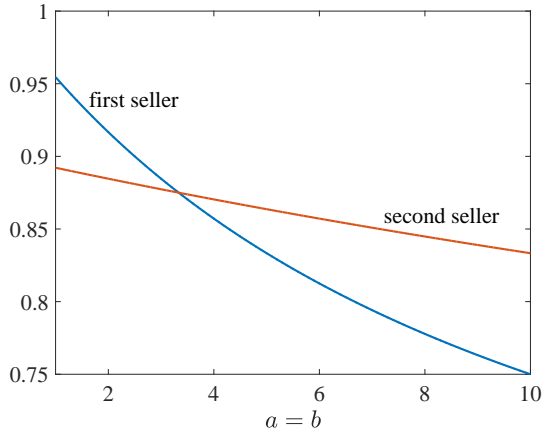} 
\caption{Probability that the next review is positive using the predictive distribution (\ref{eq:predictive beta bernoulli}) for Example 3.3.\label{fig:Fig2Ch3}} \end{figure}

\medskip{}
\begin{example}\textbf{ (Predicting online reviews)} On an online
shopping platform, there are two sellers offering a product at the
same price. The first has 30 positive reviews and 0 negative reviews,
while the second has 90 positive reviews and 10 negative reviews.
Which one to choose? To tackle this problem, we can learn a Beta-Bernoulli
model to predict whether the next review will be positive or not.
We can compute the probability that the next review is positive via
the predictive distribution (\ref{eq:predictive beta bernoulli}).
The result is shown in Fig. \ref{fig:Fig2Ch3}: While the first seller
has a 100\% positive rate, as opposed to the 90\% rate of the first
seller, we prefer to choose the first seller unless the prior distribution
is weak, which translates here into the condition $a=b\lesssim3$.
\end{example}

\subsection{Dirichlet-Categorical Model}

The Dirichlet-Categorical model generalizes the Beta-Bernoulli model
to the case of discrete observations that can take any number $C$
of values. The treatment follows along the same lines as for the Beta-Bernoulli
model, and hence we provide only a brief discussion. The likelihood
function can be written as
\begin{equation}
p(x_{\mathcal{D}}|\mu)=\prod_{k=0}^{C-1}\mu_{k}^{N[k]}.
\end{equation}
The conjugate prior is the Dirichlet distribution, a generalization
of the Beta distribution: 
\begin{equation}
\begin{split}p(\mu|\alpha) & =\text{Dir}(\mu|\alpha)\propto\prod_{k=0}^{C-1}\mu_{k}^{\alpha_{k}-1},\end{split}
\end{equation}
where $\alpha_{k}$ is the hyperparameter representing the number
of ``prior'' observations equal to $k$. Note that the Dirichlet
distribution is a joint pdf for the entries of the mean vector $\mu$.
Mean and mode vectors for the Dirichlet distribution are given as
\begin{align}
\text{E}_{\mu\sim\text{Dir}(\mu|\alpha)}[\mu] & =\frac{\alpha}{\sum_{j=0}^{C-1}\alpha_{j}}\\
\text{mode\ensuremath{_{\text{\ensuremath{\mu\sim}Dir}(\mu|\alpha)}}} & [\mu]=\frac{\alpha-1}{\sum_{j=0}^{C-1}\alpha_{j}-C}.
\end{align}
The posterior of the parameters is the Dirichlet distribution 
\begin{equation}
p(\mu|x_{\mathcal{D}},\alpha)\propto\prod_{k=0}^{C-1}\mu_{k}^{N[k]+\alpha_{k}-1}=\text{Dir}(\mu|\alpha+N),
\end{equation}
in which we can again interpret $\alpha_{k}+N[k]$ as the effective
number of observations equal to $k$. From this distribution, we can
obtain the MAP estimate as the mode. Finally, the Bayesian predictive
distribution is
\begin{equation}
p(x=k|x_{\mathcal{D}},\alpha)=\frac{N[k]+\alpha_{k}}{N+\sum_{j=0}^{C-1}\alpha_{j}}.
\end{equation}
One can check that the behavior in the two regimes of small and large
$N$ is consistent with the discussion on the Beta-Bernoulli model.

\subsection{Gaussian-Gaussian Model}

As a last example, we consider continuous observations that are assumed
to have a Gaussian distribution $\mathcal{N}(\mu,\sigma^{2})$ with
unknown mean $\mu$ but known variance $\sigma^{2}$. The likelihood
function is hence $p(x_{\mathcal{D}}|\mu)=\prod_{n=1}^{N}\mathcal{N}(x_{n}|\mu,\sigma^{2})$.
The conjugate prior is also Gaussian, namely $p(\mu|\mu_{0},\sigma_{0}^{2})=\mathcal{N}(\mu|\mu_{0},\sigma_{0}^{2}),$
with hyperparameters ($\mu_{0},\sigma_{0}^{2}$). The posterior $p(\mu|x_{\mathcal{D}\,},\mu_{0},\sigma_{0}^{2})=\mathcal{N}(\mu|\mu_{N},\sigma_{N}^{2})$
is hence Gaussian, with mean and variance satisfying
\begin{align}
\mu_{N} & =\frac{\sigma^{2}/N}{\sigma_{0}^{2}+\sigma^{2}/N}\mu_{0}+\frac{\sigma_{0}^{2}}{\sigma_{0}^{2}+\sigma^{2}/N}\mu_{\text{\ensuremath{ML}}}\\
\frac{1}{\sigma_{N}^{2}} & =\frac{1}{\sigma_{0}^{2}}+\frac{N}{\sigma^{2}},
\end{align}
where we recall that the ML estimate is $\mu_{ML}=\sum_{n=1}^{N}x_{n}/N.$
Note that, since mean and mode are equal for the Gaussian distribution,
the mean $\mu_{N}$ is also the MAP estimate $\mu_{MAP}$ of $\mu$.
Finally, the predictive distribution is also Gaussian and given as
$p(x|x_{\mathcal{D}},\mu_{0},\sigma_{0}^{2})=\mathcal{N}(x|\mu_{\text{\ensuremath{N}}},\sigma^{2}+\sigma_{N}^{2}$).
Once again, as $N$ grows large, the predictive distribution tends
to that returned by the ML approach, namely $\mathcal{N}(x|\mu_{ML},\sigma^{2})$.

\begin{figure} 
\centering 
\includegraphics[scale=0.6]{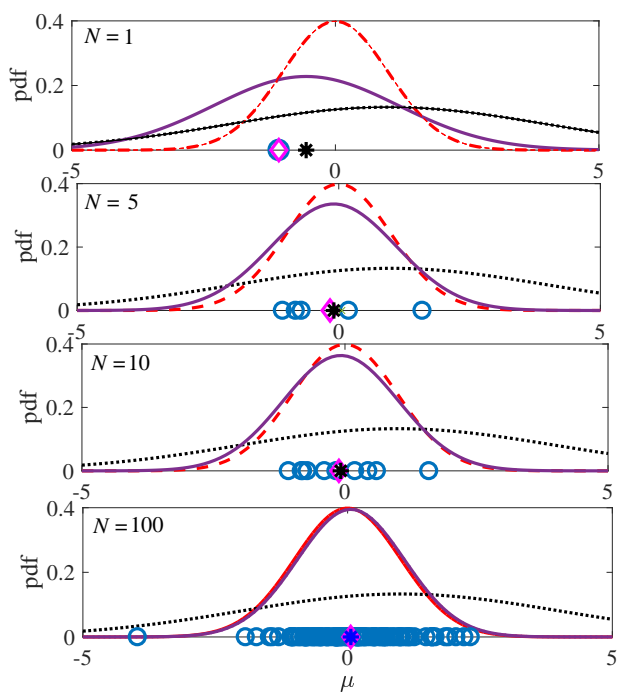} \caption{Gaussian-Gaussian model: prior distribution $\mathcal{N}$($x|\mu_{0}=1$,$\sigma_{0}^{2}=3$) (dotted), true distribution $\mathcal{N}$($x|\mu=0,$$\sigma^{2}=1$) (dashed), $N$ observations (circles), ML solution (diamond), the MAP estimate (star), and Bayesian predictive distribution (solid line).\label{fig:Fig3Ch3}} \end{figure}

Fig. \ref{fig:Fig3Ch3} illustrates the relationship among ML, MAP
and Bayesian solutions for the Gaussian-Gaussian model. In all the
panels, the dotted line represents the prior distribution, which is
characterized by the parameters ($\mu_{0}=1$, $\sigma_{0}^{2}=3$),
and the dashed line is the true distribution of the data, which is
assumed to be Gaussian with parameters ($\mu=0,$ $\sigma^{2}=1$),
hence belonging to the assumed model. Each subfigure plots a realization
of $N$ observations (circles), along with the ML solution (diamond)
and the MAP estimate (star). The solid lines represent the Bayesian
predictive distribution. 

As $N$ increases, we observe the following, already discussed, phenomena:
(\emph{i}) the ML estimate $\mu_{ML}$ consistently estimates the
true value $\mu=0$; (\emph{ii}) the MAP estimate $\mu_{MAP}$ tends
to the ML estimate $\mu_{ML}$; and (\emph{iii}) the Bayesian predictive
distribution tends to the ML predictive distribution, which in turn
coincides with the true distribution due to (\emph{i}).

\section{Supervised Learning via Generalized Linear Models (GLM)\label{sec:Supervised-Learning-using}}

Distributions in the exponential family are not directly suitable
to serve as discriminative probabilistic models to be used in supervised
learning tasks. In this section, we introduce Generalized Linear Models
(GLMs), which are popular probabilistic discriminative models that
build on members of the exponential family. 

To elaborate, let us denote as $\textrm{exponential(\ensuremath{\cdot|\eta})}$
a probabilistic model in the exponential family, that is, a model
of the form $(\ref{eq:exponential family}$) with natural parameters
$\eta$. We also write $\textrm{exponential(\ensuremath{\cdot|\mu})}$
for a probabilistic model in the exponential family with mean parameters
$\mu$.

Using the notation adopted in the previous chapter, in its most common
form, a GLM defines the probability of a target variable $\mathrm{t}$
as
\begin{equation}
p(t|x,W)=\textrm{exponential(}t|\eta=Wx),\label{eq:GLM}
\end{equation}
where we recall that $x$ is the vector of explanatory variables,
and $W$ here denotes a matrix of learnable weights of suitable dimensions.
According to (\ref{eq:GLM}), GLMs posit that the response variable
$\mathrm{t}$ has a conditional distribution from the exponential
family, with natural parameter vector $\eta=Wx$ given by a linear
function of the given explanatory variables $x$ with weights $W$.
More generally, we may have the parametrization
\begin{equation}
p(t|x,W)=\textrm{exponential(}t|\eta=W\phi(x))\label{eq:GLM-1}
\end{equation}
for some feature vector $\phi(\cdot)$ obtained as a function of the
input variables $x$ (see next chapter). 

While being the most common, the definition (\ref{eq:GLM-1}) is still
not the most general for GLMs. More broadly, GLMs can be interpreted
as a generalization of the linear model considered in the previous
chapter, whereby the mean parameters are defined as a linear function
of a feature vector. This viewpoint, described next, may also provide
a more intuitive understanding of the modelling assumptions made by
GLMs. 

Recall that, in the recurring example of Chapter 2, the target variable
was modelled as Gaussian distributed with mean given by a linear function
of the covariates $x$. Extending the example, GLMs posit the conditional
distribution
\begin{equation}
p(t|x,W)=\textrm{exponential(}t|\mu=g(W\phi(x))),\label{eq:GLM-2}
\end{equation}
where the mean parameter vector is parametrized as a function of the
feature vector $\phi(x)$ through a generally non-linear vector function
$g(\cdot)$ of suitable dimensions. In words, GLM assume that the
target variable is a ``noisy'' measure of the mean $\mu=g(W\phi(x)).$

When the function $g(\cdot)$ is selected as the gradient of the partition
function of the selected model, e.g., $g(\cdot)=\nabla_{\eta}A(\cdot),$
then, by (\ref{eq:gradient log part-1}), we obtain the GLM (\ref{eq:GLM-1}).
This choice for $g(\cdot)$ is typical, and is referred to as the
(inverse of the) canonical link function. For instance, the linear
regression model $p(t|x,w)=\mathcal{N}(t|w^{T}\phi(x),\sigma^{2})$
used in Chapter 2 corresponds to a GLM with canonical link function.
Throughout this monograph, when referring to GLMs, we will consider
models of the form (\ref{eq:GLM-1}), or, equivalently (\ref{eq:GLM-2}),
with canonical link function. For further generalizations, we refer
to \cite{Murphy,Barber}.

GLMs, especially in the form (\ref{eq:GLM-1}), are widely used. As
we will also discuss in the next chapter, learning the parameters
of GLMs can be done by means of gradient ascent on the LL using the
identity (\ref{eq:gradient exp}) and the chain rule of differentiation.

\section{Maximum Entropy Property$^{*}$}

In this more technical section, we review the \emph{maximum entropy}
property of the exponential family. Beside providing a compelling
motivation for adopting models in this class, this property also illuminates
the relationship between natural and mean parameters.

The key result is the following: The distribution $p(x|\mathbf{\eta})$
in (\ref{eq:exponential family}) obtains the maximum entropy over
all distributions $p(x)$ that satisfy the constraints $\textrm{E}\ensuremath{_{\mathrm{x}\sim p(x)}[u_{k}(\mathrm{x})]}=\ensuremath{\mu_{k}}$
for all $k=1,...,K$. Recall that, as mentioned in Chapter 2 and discussed
in more details in Appendix A, the entropy is a measure of randomness
of a random variable. Mathematically, the distribution $p(x|\mathbf{\eta})$
solves the optimization problem
\begin{equation}
\underset{p\left(x\right)}{\textrm{max }}H(p)\textrm{ \textrm{s.t.}\;E\ensuremath{_{\mathrm{x}\sim p(x)}\left[u_{k}(\mathrm{x})\right]}=\ensuremath{\mu_{k}}\;\textrm{for}\;\ensuremath{k=1,...,K.}}\label{eq:maxentropyCh3}
\end{equation}
Each natural parameter $\eta_{k}$ turns out to be the optimal Lagrange
multiplier associated with the $k$th constraint (see \cite[Ch. 6-7]{Duchicourse}). 

To see the practical relevance of this result, suppose that the only
information available about some data $x$ is given by the means of
given functions $u_{k}(x),$ $k=1,...,K$. The probabilistic model
(\ref{eq:exponential family}) can then be interpreted as encoding
the least additional information about the data, in the sense that
it is the ``most random'' distribution under the given constraints.
This observation justifies the adoption of this model by the maximum
entropy principle. 

Furthermore, the fact that the exponential distribution solves the
maximum entropy problem (\ref{eq:maxentropyCh3}) illuminates the
relationship between mean parameters $\{\mu_{k}\}$ and natural parameters
$\{\eta_{k}\}$, as the natural parameter $\eta_{k}$ is the optimal
Lagrange multiplier associated with the constraint $E\ensuremath{_{\mathrm{x}\sim p(x)}\left[u_{k}(\mathrm{x})\right]}=\ensuremath{\mu_{k}}.$ 

As another note on the exponential family and information-theoretic
metrics, in Appendix B we provide discussion about the computation
of the KL divergence between two distributions in the same exponential
family but with different parameters.

\section{Energy-based Models$^{*}$\label{sec:Energy-based-Models}}

A generalization of the exponential family is given by probabilistic
models of the form
\begin{equation}
p(x|\eta)=\frac{1}{Z(\eta)}\exp\left(-\sum_{c}E_{c}(x_{c}|\eta)\right),\label{eq:energy based models}
\end{equation}
where functions $E_{c}(x_{c}|\eta)$ are referred to as \emph{energy
functions} and $Z(\eta)$ is the partition function. Each energy function
$E_{c}(x_{c}|\eta)$ generally depends on a subset $x_{c}$ of the
variables in vector $x$. If each energy function depends linearly
on the parameter vector $\eta$, we recover the exponential family
discussed above. However, the energy functions may have a more general
non-linear form. An example is the function $E_{c}(x_{c}|\eta)=\ln\left(1+(\eta_{c}^{T}x_{c})^{2}\right)$
corresponding to a Student's t-distribution model\footnote{A Student\textquoteright s t-distribution can be interpreted as an
infinite mixture of Gaussians. As a result, it has longer tails than
a Gaussian pdf \cite[Chapter 2]{Bishop}.} \cite{lecun2006tutorial}. 

Models in the form (\ref{eq:energy based models}) encode information
about the plausibility of different configurations of subsets of rvs
$\mathrm{x}_{c}$ using the associated energy value: a large energy
entails an implausible configuration, while a small energy identifies
likely configurations. For example, a subset of rvs $\mathrm{x}_{c}$
may tend to be equal with high probability, implying that configurations
in which this condition is not satisfied should have high energy.
Energy-based models are typically represented via the graphical formalism
of Markov networks, as it will be discussed in Chapter 7.

With energy-based models, the key formula (\ref{eq:gradient exp})
of the gradient of the LL with respect to the model's parameters generalizes
as 
\begin{equation}
\frac{1}{N}\nabla_{\eta}\textrm{ln}p(x_{\mathcal{D}}|\eta)=-\frac{1}{N}\sum_{n=1}^{N}\sum_{c}\nabla_{\eta}E_{c}(x_{n}|\eta)+\sum_{c}\textrm{E}_{\mathrm{x}\sim p(x|\eta)}[\nabla_{\eta}E_{c}(\mathrm{x}|\eta)].\label{eq:gradient_exp-1}
\end{equation}
Generalizing the discussion around (\ref{eq:gradient exp}), the first
term in (\ref{eq:gradient_exp-1}) is the ``positive'' component
that points in a direction that minimizes the energy of the observations
$x_{\mathcal{D}}$; while the second term is the ``negative'' component
that pushes up the energy of the unobserved configurations. In gradient-ascent
methods, the application of the first term is typically referred to
as the positive phase, while the second is referred to as the negative
phase. (The negative phase is even taken by some authors to model
the working of the brain while dreaming! \cite{Goodfellowbook}) While
for the exponential family the expectation in the negative phase readily
yields the mean parameters, for more general models, the evaluation
of this term is generally prohibitive and typically requires Monte
Carlo approximations, which are discussed in Chapter 8.

\section{Some Advanced Topics$^{*}$}

The previous sections have focused on the important class of parametric
probabilistic models in the exponential family. Here we briefly put
the content of this chapter in the broader context of probabilistic
models for machine learning. First, it is often useful to encode additional
information about the relationships among the model variables by means
of a graphical formalism that will be discussed in Chapter 7. Second,
the problem of learning the distribution of given observations, which
has been studied here using parametric models, can also be tackled
using a non-parametric approach. Accordingly, the distribution is
inferred making only assumptions regarding its local smoothness. Typical
techniques in this family include Kernel Density Estimation and Nearest
Neighbor Density Estimation (see, e.g., \cite{sugiyama2012density}). 

Furthermore, rather than learning individual probability densities,
in some applications, it is more useful to directly estimate ratios
of densities. This is the case, for instance, when one wishes to estimate
the mutual information between two observed variables, or in two-sample
tests, whereby one needs to decide whether two sets of observations
have the same distribution or not. We refer to \cite{sugiyama2012density}
for details. Finally, there exist important scenarios in which it
is not possible to assign an explicit probabilistic model to given
observations, but only to specify a generative mechanism. The resulting
likelihood-free inference problems are covered in \cite{mohamed2016learning},
and will be further discussed in Chapter 6.

\section{Summary}

In this chapter, we have reviewed an important class of probabilistic
models that are widely used as components in learning algorithms for
both supervised and unsupervised learning tasks. Among the key properties
of members of this class, known as the exponential family, are the
simple form taken by the gradient of the LL, as well as the availability
of conjugate priors in the same family for Bayesian inference. An
extensive list of distributions in the exponential family along with
corresponding sufficient statistics, measure functions, log-partition
functions and mappings between natural and mean parameters can be
found in \cite{Wikiexp}. More complex examples include the Restricted
Boltzmann Machines (RBMs) to be discussed in Chapter 6 and Chapter
8. It is worth mentioning that there are also distribution not in
the exponential family, such as the uniform distribution parametrized
by its support. The chapter also covered the important idea of applying
exponential models to supervised learning via GLMs. Energy-based models
were finally discussed as an advanced topic. 

The next chapter will present various applications of models in the
exponential family to classification problems.

\part{Supervised Learning}

\chapter{Classification}

The previous chapters have covered important background material on
learning and probabilistic models. In this chapter, we use the principles
and ideas covered so far to study the supervised learning problem
of classification. Classification is arguably the quintessential machine
learning problem, with the most advanced state of the art and the
most extensive application to problems as varied as email spam detection
and medical diagnosis. Due to space limitations, this chapter cannot
provide an exhaustive review of all existing techniques and latest
developments, particularly in the active field of neural network research.
For instance, we do not cover decision trees here (see, e.g., \cite{witten2016data}).
Rather, we will provide a principled taxonomy of approaches, and offer
a few representative techniques for each category within a unified
framework. We will specifically proceed by first introducing as preliminary
material the Stochastic Gradient Descent optimization method. Then,
we will discuss deterministic and probabilistic discriminative models,
and finally we will cover probabilistic generative models. 

\section{Preliminaries: Stochastic Gradient Descent\label{sec:Stochastic-Gradient-Descent}}

In this section, we review a technique that is extensively used in
the solution of optimization problems that define learning problems
such as ML and MAP (see Chapter 2). The technique is known as Stochastic
Gradient Descent (SGD). SGD is introduced here and applied throughout
this monograph to other learning problems, including unsupervised
learning and reinforcement learning. Discussions about convergence
and about more advanced optimization techniques, which may be skipped
at a first reading, can be found in the Appendix A of this chapter.

SGD addresses optimization problems of the form
\begin{equation}
\underset{\theta}{\textrm{min}}\textrm{ \ensuremath{\sum_{n=1}^{N}f_{n}(\theta)}},\label{eq:cost function ERM SGD}
\end{equation}
where $\theta$ is the vector of variables to be optimized. The cost
function $f_{n}(\theta)$ typically depends on the $n$th example
in the training set $\mathcal{D}$. Following the notation set in
Chapter 2, for example, in the case of discriminative deterministic
models, the conventional form for the cost functions is
\begin{equation}
f_{n}(\theta)=\ell(t_{n},\hat{t}(x_{n},\theta)),
\end{equation}
where $\ell$ is a loss function; ($x_{n},t_{n}$) is the $n$th training
example; and $\hat{t}(x,\theta)$ is a predictor parametrized by vector
$\theta$.

SGD requires the differentiability of cost functions $f_{n}(\cdot)$.
The idea is to move at each iteration in the direction of maximum
descent for the cost function in (\ref{eq:cost function ERM SGD}),
when the latter is evaluated as $\sum_{n\in\mathcal{S}}f_{n}(\theta)$
over a subset, or\emph{ mini-batch}, $\mathcal{S}$ of samples from
the training set.\footnote{Strictly speaking, when the functions $f_{n}(\theta$) are fixed and
they are processed following a deterministic order, the approach should
be referred to as incremental gradient method \cite{bertsekas2011incremental}.
However, the term SGD is used in machine learning, capturing the fact
that the choice of the mini-batches may be randomized, and that the
sum (\ref{eq:cost function ERM SGD}) is considered to be an empirical
average of a target ensemble mean (see also \cite{2017arXiv170900599M}).} Given a learning rate schedule $\gamma^{(i)}$ and an initialization
$\theta$$^{(0)}$ of the parameters, SGD repeats in each iteration
until convergence the following two steps:

\indent $\bullet$ Pick a mini-batch $\mathcal{S}$ of $S$ indices
from the set $\{1,...,N\}$ according to some predetermined order
or randomly;

\indent $\bullet$ Update the weights in the direction of steepest
local descent as
\begin{equation}
\theta^{(i)}\leftarrow\theta^{(i-1)}-\frac{\gamma^{(i)}}{S}\sum_{n\in\mathcal{S}}\nabla_{\theta}f_{n}(\theta)|_{\theta=\theta^{(i-1)}}.
\end{equation}
The learning rate $\gamma^{(i)}$ as a function of the iteration $i$
is generally considered to be part of the hyperparameters to be optimized
via validation. More discussion on this can be found in Appendix A
of this chapter.

\section{Classification as a Supervised Learning Problem}

Classification is a supervised learning problem in which the label
$t$ can take a discrete finite number of values. We refer to Sec.
\ref{sec:Supervised-Learning} for an introduction to supervised learning.
In binary classification, each domain point $x$ is assigned to either
one of two classes, which are denoted as $\mathcal{C}_{0}$ and $\mathcal{C}_{1}$
and identified by the value of the label $t$ as follows\begin{subequations}\label{eq:mapppingtc}
\begin{align}
x\in\mathcal{C}_{0} & \textrm{ if }t=0\,\mbox{or}\,t=-1\\
x\in\mathcal{C}_{1} & \textrm{ if }t=1.
\end{align}
\end{subequations}Note that we will find it convenient to use either
the label $t=0$ or $t=-1$ to identify class $\mathcal{C}_{0}$.
In the more general case of $K$ classes $\mathcal{C}_{0},\mathcal{\textrm{ }C}_{1},...,\mathcal{\textrm{ }C}_{K-1}$,
we will instead prefer to use one-hot encoding (Sec. \ref{sec:The-Exponential-Family})
by labelling a point $x\in\mathcal{C}_{k}$ with a $K\times1$ label
$t$ that contains all zeros except for a ``1'' entry at position
$k+1$.

\medskip{}

\begin{example} Examples of binary classification include email spam
detection and creditworthiness assessment\footnote{While less useful, the ``hot dog/ not hot dog'' classifier designed
in the ``Silicon Valley'' HBO show (Season 4) is also a valid example. }. In the former case, the domain point $x$ may be encoded using the
\emph{bag-of-words} model, so that each entry represents the count
of the number of times that each term in a given set appears in the
email. In the latter application, the domain vector $x$ generally
includes valuable information to decide on whether a customer should
be granted credit, such as credit score and salary (see, e.g., \cite{abu2012learning}).
Examples of multi-class classification include classification of text
documents into categories such as sports, politics or technology,
and labelling of images depending on the type of depicted item. \end{example}

\begin{figure}[H] 
\centering\includegraphics[scale=0.6]{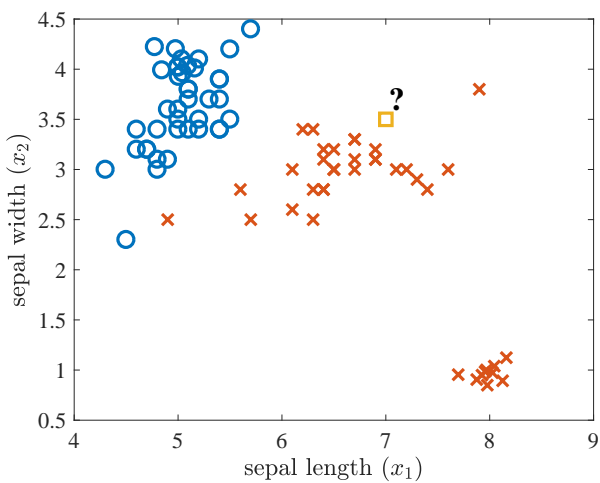}
\caption{Illustration of the binary ($K=2$ classes) classification problem with a domain space of dimension $D=2$: to which class should the new example $x$ be assigned?} \label{fig:Fig1Ch4} \end{figure}

The binary classification problem is illustrated in Fig. \ref{fig:Fig1Ch4}.
Given a training set $\mathcal{D}$ of labeled examples $x_{n}$,
$n=1,...,N$, the problem is to assign a new example $x$ to either
class $\mathcal{C}_{0}$ or $\mathcal{C}_{1}$. In this particular
standard data set, the two variables in each vector $x_{n}$ measure
the sepal length and sepal width of an iris flower. The latter may
belong to either the setosa or virginica family, as encoded by the
label $t_{n}$ and represented in the figure with different markers.
Throughout, we denote as $D$ the dimension of the domain point $x$
($D=2$ in Fig. \ref{fig:Fig1Ch4}). 

Following the taxonomy introduced in Chapter 2, we can distinguish
the following modeling approaches, which will be reviewed in the given
order throughout the rest of this chapter.

\indent $\bullet$\emph{ Discriminative deterministic models}: Model
directly the deterministic mapping between domain point and label
via a parametrized function $t=\hat{t}\left(x\right)$.

\indent $\bullet$\emph{ Discriminative probabilistic models}: Model
the probability of a point $x$ belonging to class $\mathcal{C}_{k}$
via a parametrized conditional pmf $p(t|x),$ with the relationship
between $t$ and $\mathcal{C}_{k}$ defined in (\ref{eq:mapppingtc}).
We will also write $p(\mathcal{C}_{k}|x)$ for the discriminative
probability when more convenient.

\indent $\bullet$\emph{ Generative probabilistic model}: Model the
joint distribution of domain point and class label by specifying the
prior distribution $p(t)$, or $p(\mathcal{C}_{k})$, and the class-dependent
probability distribution $p(x|t)$, or $p(x|\mathcal{C}_{k})$, of
the domain points within each class.

Discriminative models are arguably to be considered as setting the
current state of the art on classification, including popular methods
such as Support Vector Machine (SVM) and deep neural networks. Generative
models are potentially more flexible and powerful as they allow to
capture distinct class-dependent properties of the covariates $x$.

\section{Discriminative Deterministic Models}

In this section, we discuss binary classification using discriminative
deterministic models. Owing to their practical importance and to their
intuitive geometric properties, we focus on \emph{linear} models,
whereby the binary prediction $\hat{t}(x)$ is obtained by applying
a threshold rule on a decision variable $a(x,\tilde{w})$ obtained
as a linear function of the learnable weights $\tilde{w}$ (the notation
will be introduced below). Note that the decision variable $a(x,\tilde{w})$
may not be a linear function of the covariates $x$. As we will discuss,
this class of models underlie important algorithms that are extensively
used in practical applications such as SVM. A brief discussion on
multi-class classification using deterministic models is provided
at the end of this section. In the next two sections, we cover discriminative
probabilistic models, including GLMs and more general models.

\subsection{Model}

In their simplest form, linear discriminative deterministic classification
models are of the form
\begin{equation}
\hat{t}(x,\tilde{w})=\textrm{sign}\left(a(x,\tilde{w})\right),\label{eq:linclassbasicmodel}
\end{equation}
where the \emph{activation, }or decision variable, is given as
\begin{align}
a(x,\tilde{w}) & =\sum_{d=1}^{D}w_{d}x_{d}+w_{0}\nonumber \\
 & =w^{T}x+w_{0}=\tilde{w}^{T}\tilde{x},\label{eq:activation linear}
\end{align}
and we have defined the weight vectors $w=[w_{1}\cdots w_{D}]^{T}$
and $\tilde{w}=[w_{0}\textrm{ }w_{1}\cdots w_{D}]^{T}$, as well as
the extended domain point $\tilde{x}=[1\textrm{ }x^{T}]^{T},$ with
$x=[x_{1}\cdots x_{D}]^{T}$. The sign function in decision rule (\ref{eq:linclassbasicmodel})
outputs 1 if its argument is positive, and 0 or $-1$ if the argument
is negative depending on the assumed association rule in (\ref{eq:mapppingtc}).

\begin{figure}[H] 
\centering\includegraphics[scale=0.7]{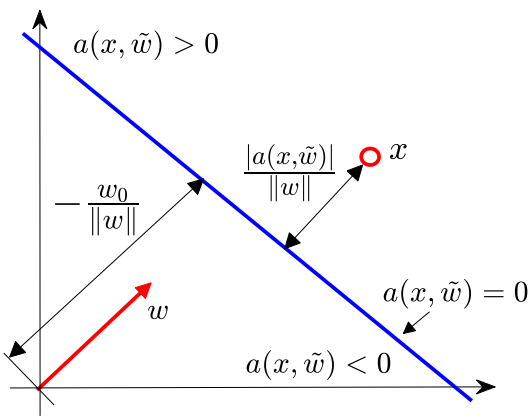}
\caption{Key definitions for a binary linear classifier.} \label{fig:FighypCh4} \end{figure}

\textbf{Geometric interpretation: classification, geometric and functional
margins. }The decision rule (\ref{eq:linclassbasicmodel}) defines
a hyperplane that separates the domain points classified as belonging
to either of the two classes. A hyperplane is a line when $D=2$;
a plane when $D=3$; and, more generally, a $D-1$-dimensional affine
subspace \cite{Boyd} in the domain space. The hyperplane is defined
by the equation $a(x,\tilde{w})=0$, with points on either side characterized
by either positive or negative values of the activation $a(x,\tilde{w}).$
The decision hyperplane can be identified as described in Fig. \ref{fig:FighypCh4}:
the vector $w$ defines the direction perpendicular to the hyperplane
and $-w_{0}/\left|\left|w\right|\right|$ is the bias of the decision
surface in the direction $w$.//

Given a point $x$, it is useful to measure the confidence level at
which the classifier assigns $x$ to the class identified through
rule (\ref{eq:linclassbasicmodel}). This can be done by quantifying
the Euclidean distance between $x$ and the decision hyperplane. As
illustrated in Fig. \ref{fig:FighypCh4}, this distance, also known
as \emph{classification margin}, can be computed as $|a\left(x,\tilde{w}\right)|/\left\Vert w\right\Vert $.

A point $x$ has a true label $t$, which may or may not coincide
with the one assigned by rule (\ref{eq:linclassbasicmodel}). To account
for this, we augment the definition of margin by giving a positive
sign to correctly classified points and a negative sign to incorrectly
classified points. Assuming that $t$ takes values in $\{-1,1\}$,
this yields the definition of \emph{geometric margin} as
\begin{equation}
\textrm{}\frac{t\cdot a\left(x,\tilde{w}\right)}{\left\Vert w\right\Vert },\label{eq:geometric margin}
\end{equation}
whose absolute value equals the classification margin. For future
reference, we also define the \emph{functional margin} as $\textrm{\ensuremath{t\cdot a\left(x,\tilde{w}\right)}}.$

\textbf{Feature-based model.} The model described above, in which
the activation is a linear function of the input variables $x$, has
the following drawbacks.

\begin{figure}[H] 
\centering\includegraphics[scale=0.6]{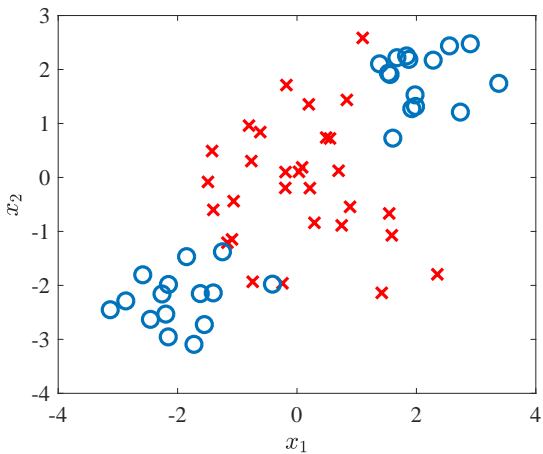}
\caption{A non-linearly separable training set.} \label{fig:Fig2Ch4} \end{figure}

1) \emph{Bias}: As suggested by the example in Fig. \ref{fig:Fig2Ch4},
dividing the domain of the covariates $x$ by means of a hyperplane
may fail to capture the geometric structure of the data. In particular,
in the example, the two classes are not \emph{linearly separable}
in the space of the covariates \textendash{} no hyperplane separates
exactly the domain points in the two classes. In such cases, classifiers
of the form (\ref{eq:linclassbasicmodel}) may yield large average
losses due to the bias induced by the choice of the model (see Sec.
\ref{subsec:Maximum-Likelihood-(ML)}).

2) \emph{Overfitting}: When $D$ is large and the data points $N$
are insufficient, learning the $D+1$ weights of the classifier may
cause overfitting. 

3) \emph{Data-dependent domain size}: In some applications, the dimension
$D$ may even change from data point to data point, that is, it may
vary with the index $n$. For example, a text $x_{n}$, e.g., represented
in ASCII format, will have a different dimension $D_{n}$ depending
on the number of words in the text.

To address these problems, a powerful approach is that of working
with feature vectors $\phi_{k}(x),\quad k=1,\ldots,D',$ rather than
directly with the covariates $x$, as the input to the classifier.
A feature $\phi_{k}(x)$ is a, generally non-linear, function of the
vector $x$. It is important to emphasize that these functions are
fixed and not learned. 

Choosing a number of features $D'>D$, which yields an overcomplete
representation of the data point $x$, may help against bias; while
opting for an undercomplete representation with $D'<D$ may help solve
the problem of overfitting. Furthermore, the same number of features
$D'$, e.g., word counts in a bag-of-words model, may be selected
irrespective of the size of the data point, addressing also the last
problem listed above.

The feature-based model can be expressed as (\ref{eq:linclassbasicmodel})
with activation
\begin{equation}
a\left(x,\tilde{w}\right)=\sum_{k=1}^{D'}w_{k}\phi_{k}(x)=\tilde{w}^{T}\mathbf{\phi}(x),\label{eq:featureBasedModelGeneral}
\end{equation}
where we have defined the feature vector $\mathbf{\phi}(x)=[\phi_{1}(x)\cdots\phi_{D'}(x)]^{T}$.
Note that model (\ref{eq:linclassbasicmodel}) is a special case of
(\ref{eq:featureBasedModelGeneral}) with the choice $\phi(x)=[1\textrm{ \ensuremath{x^{T}]^{T}}}$. 

\subsection{Learning}

As seen in Sec. \ref{subsec:Maximum-Likelihood-(ML)}, learning of
deterministic discriminative models can be carried out by means of
ERM for a given loss function $\ell$. Furthermore, as discussed in
Sec. \ref{subsec:RegularizationCh2}, overfitting can be controlled
by introducing a regularization function $R(\tilde{w})$ on the weight
vector $\tilde{w}$. Accordingly, a deterministic predictor $\hat{t}(x,\tilde{w})$
as defined in (\ref{eq:linclassbasicmodel}) can be learned by solving
the regularized ERM problem
\begin{equation}
\underset{\tilde{w}}{\mbox{min}}\quad L_{\mathcal{D}}(\tilde{w})+\frac{\lambda}{N}R(\tilde{w}),\label{eq:ERM classification}
\end{equation}
with the empirical risk
\begin{equation}
L_{\mathcal{D}}(\tilde{w})=\frac{1}{N}\sum_{n=1}^{N}\ell\left(t_{n},\hat{t}(x_{n},\tilde{w})\right).
\end{equation}
In (\ref{eq:ERM classification}), the hyperparameter $\lambda$ should
be selected via validation as explained in Sec. \ref{subsec:RegularizationCh2}. 

Extending the examples discussed in Sec. \ref{subsec:RegularizationCh2},
the regularization term is typically convex but possibly not differentiable,
e.g., $R\left(\tilde{w}\right)=\left\Vert \tilde{w}\right\Vert {}_{1}$.
 Furthermore, a natural choice for the loss function is the 0-1 loss,
which implies that the generalization loss $L_{p}$ in (\ref{eq:gen error inf})
is the probability of classification error. 

\begin{figure}[H] 
\centering\includegraphics[width=0.8\textwidth]{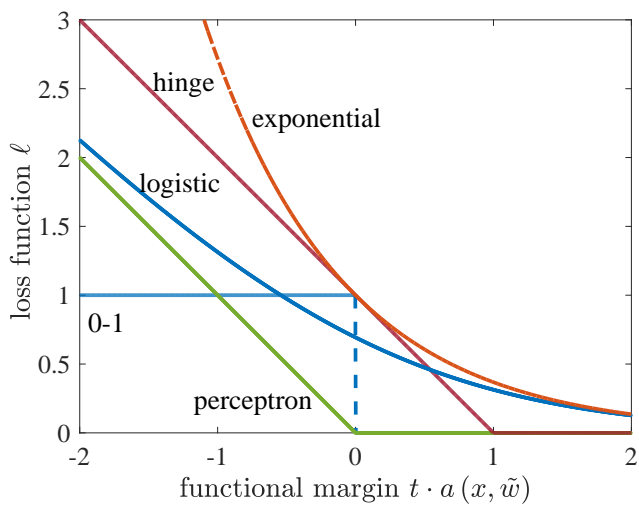}
\caption{Some notable surrogate loss functions for binary classification along with the 0-1 loss.} \label{fig:Figlossfun} \end{figure}

In the special case of linearly separable data sets, the resulting
ERM problem can be converted to a Linear Program (LP) \cite{Shalev}.
Since it is in practice impossible to guarantee the separability condition
a priori, one needs generally to solve directly the ERM problem (\ref{eq:ERM classification}).
The function sign$(\cdot)$ has zero derivative almost everywhere,
and is not differentiable when the argument is zero. For this reason,
it is difficult to tackle problem (\ref{eq:ERM classification}) via
standard gradient-based optimization algorithms such as SGD. It is
instead often useful to consider \emph{surrogate loss functions $\ell(t,a)$
}that depend directly on the differentiable (affine) activation $a(x,\tilde{w})$.
The surrogate loss function should preferably be convex in $a$, and
hence in $\tilde{w}$, ensuring that the resulting regularized ERM
problem
\begin{equation}
\underset{\tilde{w}}{\mbox{min}}\quad\sum_{n=1}^{N}\ell(t_{n},a(x_{n},\tilde{w}))+\frac{\lambda}{N}R(\tilde{w})\label{eq:surrogate ERM classification}
\end{equation}
is convex. This facilitates optimization \cite{Boyd}, and, under
suitable additional conditions, guarantees generalization \cite{Shalev}
(see also next chapter). The surrogate loss function should also ideally
be an upper bound on the original loss function. In this way, the
actual average loss is guaranteed to be smaller than the value attained
under the surrogate loss function. Examples of surrogate functions
that will be considered in the following can be found in Fig. \ref{fig:Figlossfun}.

\subsubsection{Perceptron Algorithm}

The perceptron algorithm is one of the very first machine learning
and AI algorithms. It was introduced by Frank Rosenblatt at the Cornell
Aeronautical Laboratory in 1957 to much fanfare in the popular press
\textendash{} it is ``the embryo of an electronic computer that {[}the
Navy{]} expects will be able to walk, talk, see, write, reproduce
itself and be conscious of its existence.'' reported The New York
Times \cite{NYT1}. The algorithm was implemented using analog electronics
and demonstrated impressive \textendash{} for the time \textendash{}
classification performance on images \cite{Bishop}. 

Using the feature-based model for generality, the perceptron algorithm
attempts to solve problem (\ref{eq:surrogate ERM classification})
with the surrogate \emph{perceptron loss function} defined as
\begin{align}
\ell(t,a(x,\tilde{w})) & =\max\left(0,-\textrm{\ensuremath{t\cdot a\left(x,\tilde{w}\right)}}\right).\label{eq:perceptron loss}
\end{align}
The perceptron loss assigns zero cost to a correctly classified example
$x$, whose functional margin $\ensuremath{t\cdot a\left(x,\tilde{w}\right)}$
is positive, and a cost equal to the absolute value of the functional
margin for a misclassified example, whose functional margin is negative.
A comparison with the 0-1 loss is shown in Fig. \ref{fig:Figlossfun}.
The perceptron algorithm tackles problem (\ref{eq:surrogate ERM classification})
with $\lambda=0$ via SGD with mini-batch size $S=1$. The resulting
algorithm works as follows. First, the weights $\tilde{w}^{(0)}$
are initialized. Then, for each iteration $i=1,2,...$

\indent $\bullet$ Pick a training example ($x_{n},t_{n}$) uniformly
with replacement from $\mathcal{D}$;

\indent $\bullet$ If the example is correctly classified, i.e., if
$\textrm{\ensuremath{t_{n}a(x_{n},\tilde{w}})}\geq0$, do not update
the weights: $\tilde{w}^{(i)}\leftarrow\tilde{w}^{(i-1)}$;

\indent $\bullet$ If the example is not correctly classified, i.e.,
if $\ensuremath{t_{n}a(x_{n},\tilde{w}})<0$, update the weights as:
\begin{align}
\tilde{w}^{(i)} & \leftarrow\tilde{w}^{(i-1)}-\nabla_{\tilde{w}}\ell(t_{n},a(x_{n},\tilde{w}))|_{\tilde{w}=\tilde{w}^{(i-1)}}=\tilde{w}^{(i-1)}+\phi(x_{n})t_{n}.
\end{align}

It can be proved that, at each step, the algorithm reduces the term
$\ell(t_{n},a(x_{n},\tilde{w}))$ in the perceptron loss related to
the selected training example $n$ if the latter is misclassified.
It can also be shown that, if the training set is linearly separable,
the perceptron algorithm finds a weight vector $\tilde{w}$ that separates
the two classes exactly in a finite number of steps \cite{Bishop}.
However, convergence can be slow. More importantly, the perceptron
fails on training sets that are not linearly separable, such as the
``XOR'' training set $\mathcal{D}$=$\{([0,0]^{T},0),([0,1]^{T},1),([1,0]^{T},1),([1,1]^{T},0)\}$
\cite{minsky1969perceptrons}. This realization came as a disappointment
and contributed to the first so-called AI winter period characterized
by a reduced funding for AI and machine learning research \cite{WikiAI}.

\subsubsection{Support Vector Machine (SVM)}

SVM, introduced in its modern form by Cortes and Vapnik \cite{cortes1995support}
in 1995, was among the main causes for a renewal of interest in machine
learning and AI. For this section, we will write explicitly (and with
a slight abuse of notation) the activation as
\begin{equation}
a\left(x,\tilde{w}\right)=w_{0}+w^{T}\phi(x),
\end{equation}
in order to emphasize the offset $w_{0}$. SVM solves the regularized
ERM problem (\ref{eq:surrogate ERM classification}) with the surrogate
\emph{hinge} \emph{loss function}
\begin{align}
\ell(t,a(x,\tilde{w})) & =\max(0,1-\textrm{\ensuremath{t\cdot a\left(x,\tilde{w}\right)}}),\label{eq:hinge loss}
\end{align}
and with the regularization function $R(\tilde{w})=\left\Vert w\right\Vert ^{2}$.
Note that the latter involves only the vector $w$ and not the bias
weight $w_{0}$ \textendash{} we will see below why this is a sensible
choice. The hinge loss function is also shown in Fig. \ref{fig:Figlossfun}. 

Therefore, unlike the perceptron algorithm, SVM includes a regularization
term, which was shown to ensure strong theoretical guarantees in terms
of generalization error \cite{cristianini2000introduction}. Furthermore,
rather than relying on SGD, SVM attempts to directly solve the regularized
ERM problem using powerful convex optimization techniques \cite{Boyd}.

To start, we need to deal with the non-differentiability of the hinge
loss (\ref{eq:hinge loss}). This can be done by introducing auxiliary
variables $z_{n}$, one for each training example $n$. In fact, imposing
the inequality $z_{n}\geq\ell(t_{n},a(x_{n},\tilde{w}))$ yields the
following equivalent problem \begin{subequations}\label{eq:SVM problem}
\begin{align}
\underset{\tilde{w},z}{\mbox{min}} & \sum_{n=1}^{N}z_{n}+\frac{\lambda}{N}\left\Vert w\right\Vert ^{2}\\
\textrm{s.t. } & \textrm{\ensuremath{\text{\ensuremath{t_{n}\cdot a(x_{n},\tilde{w})\geq1-z_{n}}}}}\label{eq:SVM constraint}\\
 & z_{n}\ge0\textrm{ for \ensuremath{n}=\ensuremath{1,...,N}},\label{eq:SVM constraint 1}
\end{align}
\end{subequations}where $z=[z_{1}\cdots z_{N}]^{T}$. The equivalence
between the original regularized ERM problem and problem (\ref{eq:SVM problem})
follows from the fact that any optimal value of the variables ($\tilde{w},z$)
must satisfy either constraint (\ref{eq:SVM constraint}) or (\ref{eq:SVM constraint 1})
with equality. This can be seen by contradiction: a solution for which
both constraints are loose for some $n$ could always be improved
by decreasing the value of the corresponding variables $z_{n}$ until
the most stringent of the two constraints in (\ref{eq:SVM problem})
is met. As a consequence, at an optimal solution, we have the equality
$z_{n}=\ell(t_{n},a(x_{n},\tilde{w}))$. 

The advantage of formulation (\ref{eq:SVM problem}) is that the problem
is convex, and can hence be solved using powerful convex optimization
techniques \cite{Boyd}. In fact, the cost function is strictly convex,
and thus the optimal solution is unique \cite{Boyd}. Furthermore,
the optimal solution has an interesting interpretation in the special
case in which the training data set is linearly separable. As we will
see, this interpretation justifies the name of this technique.

\textbf{Linearly separable sets and support vectors.} When the data
set is linearly separable, it is possible to find a vector $\tilde{w}$
such that all points are correctly classified, and hence all the functional
margins are positive, i.e., $t_{n}\cdot a(x_{n},\tilde{w})>0$ for
$n=1,...,N$. Moreover, by scaling the vector $\tilde{w}$, it is
always possible to ensure that the minimum functional margin equals
1 (or any other positive value). This means that we can impose without
loss of optimality the inequalities $t_{n}\cdot a(x_{n},\tilde{w})\geq1$
for $n=1,...,N$ and hence set $z=0$ in problem (\ref{eq:SVM problem}).
This yields the optimization problem \begin{subequations}\label{eq:SVM problem 1}
\begin{align}
\underset{\tilde{w}}{\mbox{min}} & \left\Vert w\right\Vert ^{2}\\
\textrm{s.t. } & \textrm{\ensuremath{\text{\ensuremath{t_{n}\cdot a(x_{n},\tilde{w})\geq1}}}}\textrm{ for \ensuremath{n}=\ensuremath{1,...,N}}.\label{eq:SVM lin con}
\end{align}
\end{subequations}

The problem above can be interpreted as the \emph{maximization of
the minimum geometric margin }across all training points\emph{. }To
see this, note that, under the constraint (\ref{eq:SVM lin con}),
the minimum geometric margin can be computed as
\begin{align}
 & \underset{n=1,...,N}{\textrm{min}}\frac{t_{n}a(x_{n},\tilde{w})}{\left\Vert w\right\Vert }=\frac{1}{\left\Vert w\right\Vert }.\label{eq:minmargin}
\end{align}
Furthermore, we call the vectors that satisfy the constraints (\ref{eq:SVM lin con})
with equality, i.e., $t_{n}\cdot a(x_{n},\tilde{w})=1$, as \emph{support
vectors, }since they support the hyperplanes parallel to the decision
hyperplane at the minimum geometric margin. At an optimum value $\tilde{w}$,
there are at least two support vectors, one on either side of the
separating hyperplane (see \cite[Fig. 7.1]{Bishop}). 

Using Lagrange duality, the support vectors can be easily identified
by observing the optimal values $\{\alpha_{n}\}$ of the multipliers
associated with the constraints (\ref{eq:SVM constraint}). Support
vectors $x_{n}$ correspond to positive Lagrange multipliers $\alpha_{n}>0$
(see, e.g., \cite{Bishop}), while all other points have zero Lagrange
multipliers. Note that the Lagrange multipliers are returned by standard
solvers such as the ones implemented by the CVX toolbox in MATLAB
\cite{cvx}.

\begin{figure}[H] 
\centering\includegraphics[width=0.8\textwidth]{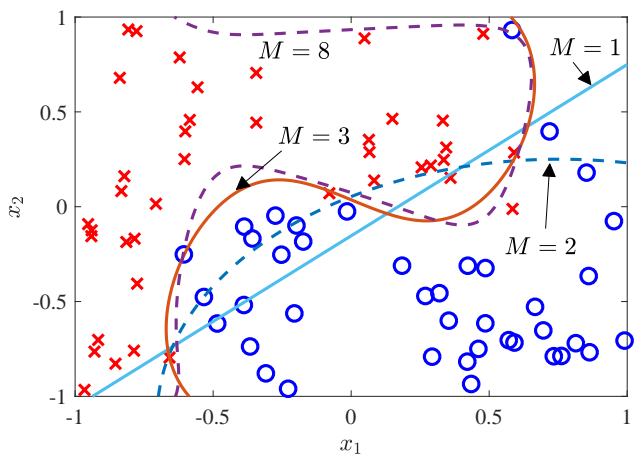}
\caption{Example of binary classification with SVM using polynomial features up to degree $M$ ($\lambda/N=0.2$).} \label{fig:SVMCh4} \end{figure}

\begin{example} In the example in Fig. \ref{fig:SVMCh4}, the illustrated
$N=80$ training samples are fed to a SVM using the monomial feature
vector $\phi(x)=[1\textrm{ \ensuremath{x_{1}\textrm{ }}\ensuremath{x_{2}\cdots x_{1}^{M}\textrm{ \ensuremath{x_{2}^{M}}}]}}$
and $\lambda/N=0.2$ for given model orders $M$. The decision boundary
is shown using dashed and solid lines. It is seen that, using a sufficiently
large order (here $M=3$), SVM is able to effectively partition the
two samples in the two classes. Furthermore, even with larger values
of $M$ (here $M=8$), SVM appears not to suffer from significant
overfitting thanks to the quadratic regularization term. \end{example}

The optimization problem (\ref{eq:SVM problem}) may be conveniently
tackled by using Lagrange duality techniques. This approach also allows
one to naturally introduce the powerful tool of kernel methods. The
interested reader can find this discussion in Appendix B of this chapter.

\subsection{Multi-Class Classification$^{*}$}

Here we briefly describe classification scenarios with $K>2$ classes.
As a first observation, it is possible to build multi-class classifiers
based solely on multiple binary classifiers, such as SVM. This can
be done by following one of two general strategies, namely \textit{one-versus-the-rest}
\textit{\emph{and }}\textit{one-versus-one} \cite[Chapter 7]{Bishop}.
The one-versus-the-rest approach trains $K$ separate binary classifiers,
say $k=1,...,K$, with the $k$th classifier operating on the examples
relative to class $\mathcal{C}_{k}$ against the examples from all
other classes. The one-versus-one method, instead, trains $K(K-1)/2$
binary classifiers, one for each pair of classes. Both approaches
can yield ambiguities in classification \cite[Chapter 7]{Bishop}. 

\section{Discriminative Probabilistic Models: Generalized Linear Models}

Discriminative probabilistic models are potentially more powerful
than deterministic ones since they allow to model sources of uncertainty
in the label assignment to the input variables. This randomness may
model noise, labelling errors, e.g., for crowdsourced labels, and/or
the residual uncertainty in the classification rule due to the availability
of limited data. Probabilistic models can also more naturally accommodate
the presence of more than two classes by producing a probability distribution
over the possible label values. 

In this section, we study GLMs, which were introduced in Sec. \ref{sec:Supervised-Learning-using}.
We recall that a GLM (\ref{eq:GLM-1}) posits that the conditional
pmf $p(t|x),$ or $p(\mathcal{C}_{k}|x)$, is a member of the exponential
family in which the natural parameter vector $\eta$ is given as a
linear function of a feature vector $\phi(x)$, i.e., $\eta=W\phi(x)$
for weight matrix $W$.\footnote{See Sec. \ref{sec:Supervised-Learning-using} for a more general definition.}
It is noted that GLMs are not linear: only the unnormalized log-likelihood
is linear in the weight matrix $W$ (see Sec. \ref{sec:The-Exponential-Family}).
We start by discussing binary classification and then cover the multi-class
case in Sec. \ref{subsec:Multi-Class-Classification_Ch4}.

\subsection{Model}

For classification, the label $t$ can take a finite number of values,
and it can hence be described by a Bernoulli variable in the binary
case or, more generally, by a Categorial variable (see Chapter 3).
The GLM (\ref{eq:GLM-1}) for binary classification is known as \emph{logistic
regression, }and it assumes the predictive distribution
\begin{equation}
p(t=1\vert x)=\sigma(\tilde{w}^{T}\phi(x)).\label{eq:logistic regression}
\end{equation}
We recall that $\sigma(a)=(1+\exp(-a))^{-1}$ is the sigmoid function
(see Chapter 2). We also observe that
\begin{equation}
\sigma(-a)=1-\sigma(a),
\end{equation}
which implies that we can write $p(t=0\vert x)=1-\sigma(\tilde{w}^{T}\phi(x))=\sigma(-\tilde{w}^{T}\phi(x))$.
Intuitively, the sigmoid function in (\ref{eq:logistic regression})
can be thought of as a ``soft'' version of the threshold function
$\textrm{sign}(a)$ used by the deterministic models studied in the
previous section.

We emphasize that the logistic regression model (\ref{eq:logistic regression})
is a GLM since it amounts to a Bernoulli distribution, which is in
the exponential family, with natural parameter vector $\eta=\tilde{w}^{T}\phi(x)$
as in 
\begin{equation}
t\vert x,w\sim\text{Bern}(t\vert\eta=\tilde{w}^{T}\phi(x)).\label{eq:log reg GLM}
\end{equation}

\textbf{Inference.} Before discussing learning, we observe that inference
is straightforward. In fact, once the discriminative model (\ref{eq:logistic regression})
is known, the average 0-1 loss, that is, the probability of error,
is minimized by choosing the label according to the following rule
\begin{equation}
p(\mathcal{C}_{1}\vert x)=p(t=1\vert x)\underset{\mathcal{C}_{0}}{\overset{\mathcal{C}_{1}}{\gtrless}}\frac{1}{2},
\end{equation}
or equivalently $\tilde{w}^{T}\phi(x)\underset{\mathcal{C}_{0}}{\overset{\mathcal{C}_{1}}{\gtrless}}0$.

\subsection{Learning}

Consider first ML. The NLL function can be written as
\begin{align}
-\ln p(t_{\mathcal{D}}\vert x_{\mathcal{D}},\tilde{w}) & =-\sum_{n=1}^{N}\ln p(t_{n}\vert x_{n},\tilde{w})\label{eq:crossentropy}\\
 & =-\sum_{n=1}^{N}\{t_{n}\ln(y_{n})+(1-t_{n})\ln(1-y_{n})\},
\end{align}
where we have defined $y_{n}=\sigma(\tilde{w}^{T}\phi(x_{n}))$. The
NLL (\ref{eq:crossentropy}) is also referred to as the \emph{cross
entropy }loss criterion, since the term $-t\ln(y)-(1-t)\ln(1-y)$
is the cross-entropy $H((t,1-t)||(y,1-y))$ (see Sec. \ref{sec:Information-Theoretic-Metrics}).
We note that the cross-entropy can be used to obtain upper bounds
on the probability of error (see, e.g., \cite{2018arXiv180209386F}).
The ML problem of minimizing the NLL is convex (see Sec. \ref{sec:Preliminaries_Ch3}),
and hence it can be solved either directly using convex optimization
tools, or by using iterative methods such as SGD or Newton (the latter
yields the iterative reweighed least square algorithm \cite[p. 207]{Bishop}).

The development of these methods leverages the expression of the gradient
of the LL function (\ref{eq:gradient exp}) (used with $N=1$) for
the exponential family. To elaborate, using the chain rule for differentiation,
we can write the gradient
\begin{equation}
\nabla_{\tilde{w}}\ln p(t|x,\tilde{w})=\nabla_{\eta}\ln\textrm{Bern}(t|\eta)|_{\eta=\tilde{w}\phi(x)}\times\nabla_{\tilde{w}}(\tilde{w}^{T}\phi(x)),
\end{equation}
which, recalling that $\text{\ensuremath{\nabla_{\eta}}ln\ensuremath{(}Bern}(t\vert\eta))=(t-\sigma(\eta))$
(cf. (\ref{eq:gradient exp})), yields
\begin{equation}
\nabla_{\tilde{w}}\ln p(t\vert x,\tilde{w})=(t-y)\phi(x).\label{eq:gradient GLM Bernoullui}
\end{equation}

Evaluating the exact posterior distribution for the Bayesian approach
turns out to be generally intractable due to the difficulty in normalizing
the posterior
\begin{equation}
p(w\vert\mathcal{\mathcal{D}})\propto p(w)\prod_{n=1}^{N}p(t_{n}\vert x_{n},\tilde{w}).\label{eq:posterior logistic}
\end{equation}
We refer to \cite[p. 217-220]{Bishop} for an approximate solution
based on Laplace approximation. Other useful approximate methods will
be discussed in Chapter 8.

As a final remark, with bipolar labels, i.e., $t$$\in\{-1,+1\}$,
the cross-entropy loss function can be written as
\begin{equation}
-\ln p(t_{\mathcal{D}}\vert x_{\mathcal{D}},\tilde{w})=\sum_{n=1}^{N}\ln(1+\exp(-t_{n}a(x_{n},\tilde{w})).
\end{equation}
This formulation shows that logistic regression can be thought of
as an ERM method with loss function $\ell(t,a(x,\tilde{w}))=\ln(1+\exp(-ta(\mathrm{x},\tilde{w})),$
which is seen in Fig. \ref{fig:Figlossfun} to be a convex surrogate
loss of the 0-1 loss.

\textbf{Mixture models.$^{*}$ }As seen, the Bayesian approach obtains
the predictive distribution by averaging over multiple models $p(t|x,w)$
with respect to the parameters' posterior $p(w|\mathcal{D})$ (cf.
(\ref{eq:predictive})). The resulting model hence \emph{mixes} the
predictions returned by multiple discriminative models $p(t|x,w)$
to obtain the predictive distribution $p(t|x)=\int p(w|\mathcal{D})p(t|x,w)dw$.
As we briefly discuss below, it is also possible to learn mixture
models within a frequentist framework. 

Consider $K$ probabilistic discriminative models $p(t|x,w_{k})$,
$k=1,...,K$, such as logistic regression. The mixture model is defined
as
\begin{equation}
p(t|x,\theta)=\sum_{k=1}^{K}\pi_{k}p(t|x,w_{k}).\label{eq:mixture model Ch4}
\end{equation}
In this model, the vector $\theta$ of learnable parameters includes
the probability vector $\pi$, which defines the relative weight of
the $K$ models, and the vectors $w_{1},...,w_{K}$ for the $K$ constituent
models. As discussed, in the Bayesian approach, the weights $\pi_{k}$
are directly obtained by using the rules of probability, as done in
(\ref{eq:posterior logistic}). Within a frequentist approach, instead,
ML training is typically performed via a specialized algorithm, which
will be described in Chapter 6, known as Expectation Maximization
(EM). 

Mixture models increase the capacity of discriminative models and
hence allow to learn more complex relationships between covariates
and labels. In particular, a mixture model, such as (\ref{eq:mixture model Ch4}),
has a number of parameters that increases proportionally with the
number $K$ of constituent models. Therefore, the capacity of a mixture
model grows larger with $K$. As an example, this increased capacity
may be leveraged by specializing each constituent model $p(t|x,w_{k})$
to a different area of the covariate domain. 

Given their larger capacity, mixture models may be prone to overfitting.
A way to control overfitting will be discussed in Sec. \ref{sec:Boosting}.

\subsection{Multi-Class Classification\label{subsec:Multi-Class-Classification_Ch4}}

In the case of $K$ classes, the relevant exponential family distribution
is Categorical with natural parameters depending linearly on the feature
vector. This yields the following discriminative model as a generalization
of logistic regression 
\begin{equation}
\mathrm{t}\vert x,W\sim\text{Cat}(t\vert\eta=W\phi(x)),\label{eq:softmax}
\end{equation}
where the label vector $\mathrm{t}$ is defined using one-hot encoding
(Chapter 3) and $W$ is a matrix of weights. We can also equivalently
write the vector of probabilities for the $K$ classes as
\begin{align}
y=\textrm{softmax(\ensuremath{W\phi(x)})=}\begin{bmatrix}\frac{e^{\eta_{0}}}{\sum_{k=0}^{K-1}e^{\eta_{k}}}\\
\vdots\\
\frac{e^{\eta_{K-1}}}{\sum_{k=0}^{K-1}e^{\eta_{k}}}
\end{bmatrix},\label{eq:GLM categorical}
\end{align}
where $y=[y_{1}\cdots y_{K}]^{T}$ with $y_{k}=p(\mathcal{C}_{k}|x)$;
and $\eta_{k}=w_{k+1}^{T}\phi(x)$ with $w_{k}^{T}$ being the $k$th
row of the weight matrix $W$. 

Learning follows as for logistic regression. To briefly elaborate
on this point, the NLL can be written as the cross-entropy function
\begin{align}
-\ln p(t_{\mathcal{D}}\vert x_{\,\mathcal{D}},W) & =-\sum_{n=1}^{N}\ln p(t_{n}\vert x_{n},W)\nonumber \\
 & =-\sum_{n=1}^{N}t_{n}^{T}\ln(y_{n}),\label{eq:crossentropy-1}
\end{align}
where the logarithm is applied element by element, and we have $y_{n}=\textrm{softmax}(W\phi(x_{n}))$.
Note that each term in (\ref{eq:crossentropy-1}) can be expressed
as the cross-entropy $-t_{n}^{T}\ln(y_{n})=H(t_{n}||y_{n}).$ The
ML problem is again convex and hence efficiently solvable. The gradient
of the NLL can be again found using the general formula (\ref{eq:gradient exp})
for exponential models and the chain rule for derivatives. We can
write
\begin{equation}
\nabla_{W}\ln p(t\text{\ensuremath{|x,W})=\ensuremath{\nabla}\ensuremath{\ensuremath{_{\eta}}}}\ln\textrm{Cat}(t|\eta)|_{\eta=W\phi(x)}\times\nabla_{W}(W\phi(x)),
\end{equation}
which yields
\begin{equation}
\nabla_{W}\ln p(\ensuremath{t}\vert x\ensuremath{,}W)=(t-y)\phi(x)^{T}.\label{eq:gradient GLM Cat}
\end{equation}

\subsection{Relation to Neural Networks}

\begin{figure}[H] 
\centering\includegraphics[scale=0.5]{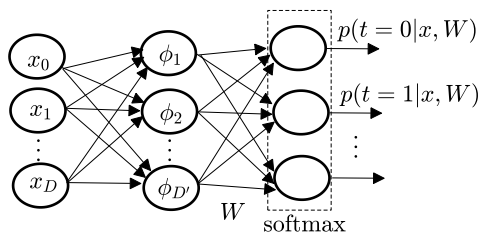}
\caption{GLM as a three-layer neural network with learnable weights only between hidden and output layers.} \label{fig:FigNNCh4sm} \end{figure}

GLM models of the form (\ref{eq:softmax}), or (\ref{eq:logistic regression})
in the special case of binary classification, can be interpreted in
terms of the neural network shown in Fig. \ref{fig:FigNNCh4sm}. A
neural network consists of a directed graph of computing elements,
known as neurons. Each neuron applies a deterministic transformation
of its inputs, as defined by the incoming edges, to produce a scalar
output on its outgoing edge.

In Fig. \ref{fig:FigNNCh4sm}, the input vector $x$ is first processed
by a hidden layer of neurons, in which each $k$th neuron computes
the feature $\phi_{k}(x)$. Then, each $k$th neuron in the output
layer applies the $k$th element of the softmax non-linearity (\ref{eq:GLM categorical})
to the numbers produced by the hidden neurons in order to compute
the probability $y_{k}=p(\mathcal{C}_{k}|x)$. Note that, in the case
of binary classification, only one output neuron is sufficient in
order to compute the probability (\ref{eq:logistic regression}).
It is also important to emphasize that only the weights between hidden
and output layers are learned, while the operation of the neurons
in the hidden layer is fixed.\\

We remark that one should not confuse a graph such as the one in Fig.
\ref{fig:FigNNCh4sm} with the BN representation previously seen in
Fig. \ref{fig:BNCh2}, which will be further discussed in Chapter
7. In fact, while BNs represent probability distributions, diagrams
of neural networks such as in Fig. \ref{fig:FigNNCh4sm} describe
deterministic functional relations among variables. This is despite
the fact that the output layer of the network in Fig. \ref{fig:FigNNCh4sm}
computes a vector of probabilities. In other words, while the nodes
of a BN are rvs, those in the graph of a neural network are computational
nodes. 

\textbf{``Extreme machine learning''.$^{*}$} An architecture in
which the features $\mathbf{\phi}(x)$ are selected via random linear
combinations of the input vector $x$ is sometimes studied under the
rubric of ``extreme machine learning'' \cite{HUANG2006489}. The
advantage of this architecture as compared to deep neural networks
with more hidden layers and full learning of the weights (Sec. \ref{subsec:Non-linear-Classification})
is its low complexity.

\section{Discriminative Probabilistic Models: Beyond GLM\label{subsec:Non-linear-Classification}}

\begin{figure}[H] 
\centering\includegraphics[scale=0.5]{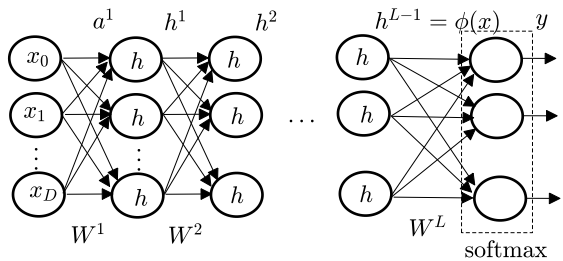}
\caption{A multi-layer neural network. } \label{fig:FigMultiLCh4} \end{figure}

As depicted in Fig. \ref{fig:FigNNCh4sm}, GLMs can be interpreted
as three-layer neural networks in which the only hidden layer computes
fixed features. The fixed features are then processed by the output
classification layer. In various applications, determining suitable
features is a complex and time-consuming task that requires significant
domain knowledge. Moving beyond GLMs allows us to work with models
that learn not only the weights used by the output layer for classification,
but also the vector of features $\phi(x)$ on which the output layer
operates. This approach yields a much richer set of models, which,
along with suitable learning algorithms, has led to widely publicized
breakthroughs in applications ranging from speech translation to medical
diagnosis. 

As a prominent example of beyond-GLM classification models, we describe
here feed-forward multi-layer neural networks, or deep neural networks
when the hidden layers are in large number. We note that, beside yielding
state-of-the-art classification algorithms, multi-layer feedforward
networks are also key components of the computational theory of mind
\cite{Pinker}.

\subsection{Model}

As illustrated in Fig. \ref{fig:FigMultiLCh4}, feed-forward multi-layer
networks consist of multiple layers with learnable weights. Focusing
on multi-class classification, we have the chain of vectors $x\rightarrow h^{1}\rightarrow\cdots\rightarrow h^{L}\rightarrow y$,
where $x$ is the $D\times1$ input (observed) vector; $y$ is the
$K\times$1 vector of output probabilities for the $K$ classes; and
$h^{l}$ represents the vector of outputs at the $l$th hidden layer.
The number of neurons in each hidden layer is a hyperparameter to
be optimized via validation or Bayesian methods (Chapter 2). Note
that the adopted numbering of the layers is not universally accepted,
and the reverse ordering is also used.

Referring to Fig. \ref{fig:FigMultiLCh4}, we can write the operation
of the neural network through the functional relations\begin{subequations}\label{eq:DNN}
\begin{align}
h^{1} & =h(a^{1})\textrm{ with \ensuremath{a^{1}=W^{1}}}x\\
h^{l} & =h(a^{l})\textrm{ }\textrm{with \ensuremath{a^{l}=W^{l}}\ensuremath{h}\ensuremath{^{l-1}}}\\
 & \textrm{ for \ensuremath{l=2,...,L}}\\
y & =\textrm{softmax}(a^{L+1})\textrm{ with \ensuremath{a^{L+1}}=\ensuremath{W^{L+1}h^{L}}}.
\end{align}
\end{subequations}The non-linear function $h(\cdot)$ is applied
element-wise and is typically selected as a sigmoid, such as the logistic
sigmoid or the hyperbolic tangent, or else, as has become increasingly
common, the Rectified Linear Unit (ReLU) $h(a)=\max(0,a).$ In (\ref{eq:DNN}),
we have defined the activation vectors $a^{l}$ for the hidden layers
$\ensuremath{l=1,...,L}$, and the matrices of weights $W^{l}$, $l=1,...,L+1$,
whose dimensions depend on the size of the hidden layers. We denote
the tensor\footnote{A tensor is a generalization of a matrix in that it can have more
than two dimensions, see, e.g., \cite{cichocki2015tensor}.} of all weights as $W$. 

The learnable weights of the hidden layers encode the feature vector
$\phi(x)=h_{L}$ used by the last layer for classification. With multi-layer
networks, we have then moved from having fixed features defined a
priori by vector $\phi(x)$ in linear models to designing optimal
features that maximize the classifier's performance in non-linear
models. Furthermore, in multi-layer networks, the learned features
$h^{1},...,h^{L}$ tend to progress from low-level features in the
lower layers, such as edges in an image, to higher-level concepts
and categories, such as ``cats'' or ``dogs'', in the higher layers
\cite{Goodfellowbook}.

\subsection{Learning}

Training deep neural networks is an art \cite{Goodfellowbook}. The
basic underlying algorithm is \emph{backpropagation}, first proposed
in 1969 and then reinvented in the mid-1980s \cite{norvig2009,Rumelhart}.
However, in practice, a number of tricks are required in order to
obtain state-of-the-art performance, including methods to combat overfitting,
such as dropout. Covering these solutions would require a separate
treatise, and here we refer to \cite{Hintoncourse,Goodfellowbook}
for an extensive discussion. 

Backpropagation \textendash{} or \emph{backprop} for short \textendash{}
extends the derivation done in (\ref{eq:gradient GLM Cat}) to evaluate
the gradient of the LL to be used within an SGD-based algorithm. Again,
the main ingredients are the general formula (\ref{eq:gradient exp})
for exponential models and the chain rule for derivatives. To elaborate,
select a given training example $(x_{n},t_{n})=(x,t)$ to be used
in an iteration of SGD. Backprop computes the derivative of the NLL,
or cross-entropy loss function, $L(W)=-\ln p(\ensuremath{t}\vert x\ensuremath{,}W)=-t^{T}\ln y$
(cf. (\ref{eq:crossentropy-1})), where the output $y$ is obtained
via the relations (\ref{eq:DNN}). It is important to note that, unlike
the linear models studied above, the cross-entropy for multi-layer
is generally a non-convex function of the weights. 

Backprop computes the gradients with respect to the weight matrices
by carrying out the following phases.

\indent $\bullet$\emph{ Forward pass}: Given $x$, apply formulas
(\ref{eq:DNN}) to evaluate $a^{1},h^{1},a{}^{2},$ $h^{2},...,a^{L},h^{L},\textrm{ and }y.$

\indent $\bullet$\emph{ Backward pass}: Given $a^{1},h^{1},a{}^{2},$$h^{2},...,a^{L},h^{L},a^{L+1},y$
and $t$, compute\begin{subequations}\label{eq:backprop_all_Ch4}
\begin{align}
\delta^{L+1} & =(y-t)\\
\delta^{l} & =(W^{l+1})^{T}\delta^{l+1}\cdot h'(a^{l})\textrm{ \textrm{ for \ensuremath{l=L,L-1,...,1}}}\label{eq:backwardmain_Ch4}\\
\nabla_{W^{l}}L(W) & =\delta^{l}(h^{l-1})^{T}\textrm{ for \ensuremath{l=1,2,...,L+1}},\label{eq:backwardman1_Ch4}
\end{align}
\end{subequations}where $h'(\cdot)$ denotes the first derivative
of the function $h(\cdot)$; the product $\cdot$ is taken element-wise;
and we set $h^{0}=x$.

Backprop requires a forward pass and a backward pass for every considered
training example. The forward pass uses the neural network as defined
by equations (\ref{eq:DNN}). This entails multiplications by the
weight matrices $W^{l}$ in order to compute the activation vectors,
as well as applications of the non-linear function $h(\cdot)$. In
contrast, the backward pass requires only linear operations, which,
by (\ref{eq:backwardmain_Ch4}), are based on the transpose $(W^{l})^{T}$
of the weight matrix $W^{l}$ used in the forward pass. 

The derivatives (\ref{eq:backwardman1_Ch4}) computed during the backward
pass are of the general form
\begin{equation}
\nabla_{w_{ij}^{l}}L(W)=h_{i}^{l-1}\times\delta_{j}^{l},\label{eq:backprop_simplified_Ch4}
\end{equation}
where $w_{ij}^{l}$ is the $(i,j)$th element of matrix $W^{l}$ corresponding
to the weight between the pre-synaptic neuron $j$ in layer $l-1$
and the post-synaptic neuron $i$ in layer $l$; $h_{i}^{l-1}$ is
the output of the pre-synaptic neuron $i$; and $\delta_{j}^{l}$
is the back-propagated error. The back-propagated error assigns ``responsibility''
for the error $y-t$ measured at the last layer (layer $L+1$) to
each synaptic weight $w_{ij}^{l}$ between neuron $j$ in layer $l-1$
and neuron $i$ in layer $l$. The back-propagated error is obtained
via the linear operations in (\ref{eq:backwardmain_Ch4}). 

\begin{figure}[H] 
\centering\includegraphics[width=0.8\textwidth]{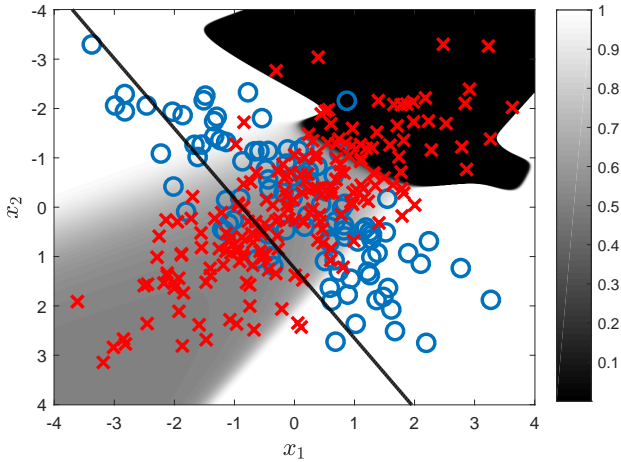}
\caption{Probability that the class label is the same as for the examples marked with circles according to the output of a feed-forward multi-layer network with one hidden layer ($L=1$) and six hidden neurons (with sigmoid non-linearity). The probability is represented by the color map illustrated by the bar on the right of the figure. For reference, the solid line represent the decision line for logistic regression.} \label{fig:discrCh4} \end{figure}

\begin{example} In the example in Fig. \ref{fig:discrCh4}, the illustrated
$N=300$ training examples are used to train a logistic regression,
i.e., GLM, model and a feed-forward multi-layer network with one hidden
layer ($L=1$) and six hidden neurons with sigmoid non-linearity $h(x)=\sigma(x)$.
The logistic model uses linear features $\phi(x)=[1\textrm{ \ensuremath{x}}]^{T}$.
Both networks are trained using SGD. For logistic regression, the
decision line is illustrated as a solid line, while for the multi-layer
network we plot the probability that the class label is the same as
for the examples marked with circles as color map. The GLM model with
linear features is seen to be unable to capture the structure of the
data, while the multi-layer network can learn suitable features that
improve the effectiveness of classification. \end{example}

\subsection{Some Advanced Topics$^{*}$}

We conclude this section by noting a few important aspects of the
ongoing research on deep neural networks. 

A first issue concerns the theoretical understanding of the generalization
properties of deep neural networks. On the face of it, the success
of deep neural networks appears to defy one of the the principles
laid out in Chapter 2, which will be formalized in the next chapter:
Highly over-parametrized models trained via ML suffer from overfitting
and hence do not generalize well. Recent results suggest that the
use of SGD, based on the gradient (\ref{eq:backprop_simplified_Ch4}),
may act a regularizer following an MDL perspective. In fact, SGD favors
the attainment of flat local maxima of the likelihood function. Flat
maxima require a smaller number of bits to be specified with respect
to sharp maxima, since, in flat maxima, the parameter vector can be
described with limited precision as long as it remain within the flat
region of the likelihood function \cite{hochreiter1997flat,keskar2016large,Ferenc}
(see also \cite{kawaguchi} for a different perspective). 

Another important aspect concerns the hardware implementation of backprop.
This is becoming extremely relevant given the practical applications
of deep neural networks for consumer's devices. In fact, a key aspect
of backprop is the need to propagate the error, which is measured
at the last layer, to each synapse via the backward pass. This is
needed in order to evaluate the gradient (\ref{eq:backprop_simplified_Ch4}).
While a software implementation of this rule does not present any
conceptual difficulty, realizing the computations in (\ref{eq:backprop_all_Ch4})
in hardware, or even on a biological neural system, is faced with
a number of issues. 

A first problem is the non-locality of the update (\ref{eq:backprop_simplified_Ch4}).
An update rule is local if it only uses information available at each
neuron. In contrast, as seen, rule (\ref{eq:backprop_simplified_Ch4})
requires back-propagation through the neural network. Another issue
is the need for the backward pass to use a different neural path with
respect to the forward pass, given that, unlike the backward pass,
the forward pass includes also non-linearities. A useful discussion
of hardware implementation aspects can be found in \cite{baldi2016learning}
(see also references therein). 

To obviate at least some of these practical issues, a number of variations
of the rule (\ref{eq:backprop_simplified_Ch4}) have been proposed
\cite{baldi2016learning}. For instance, the feedback alignment rule
modifies (\ref{eq:backwardmain_Ch4}) by using fixed random matrices
in lieu of the current weight matrices $W^{l}$; while the broadcast
alignment rule writes the vectors $\delta^{l}$ as a linear function
with fixed random coefficients of the error ($y-t$), hence removing
the need for back-propagation \cite{Samadi17}.

Furthermore, beside ML, there exist Bayesian learning algorithms \cite{Gal2016Uncertainty},
including simplified approaches such as dropout \cite{Goodfellowbook,Hintoncourse}.
Significant progress has also been made on applications such as image
recognition by leveraging the underlying structure, or geometry, of
the data \cite{bronstein2017geometric}. As an important case in point,
\emph{convolutional neural networks} leverage the stationarity, locality
and spatial invariance of image features by limiting the receptive
field of the neurons (i.e., by setting to zero weights that connect
to ``distant'' pixels) and by tying the weights of neurons in the
same layer. 

Another recent development is the design of event-driven \emph{spiking}
neural networks that can be implemented on neuromorphic computing
platforms with extremely low energy consumption (see, e.g., \cite{neuromorphic,2017arXiv171010704B}).

\section{Generative Probabilistic Models\label{sec:Generative-Probabilistic-Models Ch4}}

As discussed in Chapter 2, discriminative models do not attempt to
model the distribution of the domain points $x$, learning only the
predictive distribution $p(t|x)$. In contrast, generative models
aim at modelling the joint distribution by specifying parametrized
versions of the prior distribution $p(t)$, or $p(\mathcal{C}_{k})$,
and of the class-conditional probability distribution $p(x|t)$, or
$p(x|\mathcal{C}_{k})$. As a result, generative models make more
assumptions about the data by considering also the distribution of
the covariates $\mathrm{x}$. As such, generative models may suffer
from bias when the model is incorrectly selected. However, the capability
to capture the properties of the distribution of the explanatory variables
$\mathrm{x}$ can improve learning if the class-conditional distribution
$p(x|t)$ has significant structure. 

\subsection{Model}

Generative models for binary classification are typically defined
as follows\begin{subequations}\label{eq:Bayes GLM}
\begin{align}
\mathrm{t} & \sim\text{Bern}(\pi)\\
\mathrm{x}\vert\mathrm{t}=t & \sim\text{exponential}(\eta_{t}),
\end{align}
\end{subequations}where $\text{exponential}(\eta)$ represents a
distribution from the exponential family with natural parameter vector
$\eta$ (see previous chapter). Accordingly, the parameters of the
model are $\theta=(\pi,\text{\ensuremath{\eta}}_{0},\eta_{1})$, where
vectors $\text{\ensuremath{\eta}}_{t}$ represent the natural parameters
of the class-dependent distributions. As we have seen in Chapter 2,
we can also equivalently use mean parameters to define the exponential
family distributions. As a result of this choice, the joint distribution
for rv $(\mathrm{\mathrm{x}},\mathrm{t})$ is given as $p(x,t\vert\pi,\text{\ensuremath{\eta}}_{0},\eta_{1})=p(t\vert\pi)p(x\vert\text{\ensuremath{\eta}}_{t})$. 

\textbf{Inference.} Given a new point $x$, in order to minimize the
probability of error, the optimal prediction of the class under 0-1
loss can be seen to satisfy the maximum a posteriori rule
\begin{equation}
p(\mathcal{C}_{1}|x)=\frac{\pi p(x\vert\eta_{1})}{\pi p(x\vert\eta_{1})+(1-\pi)p(x\vert\eta_{0})}\underset{\mathcal{C}_{0}}{\overset{\mathcal{C}_{1}}{\gtrless}}\frac{1}{2}.\label{eq:predictive Ch4 gen}
\end{equation}

\subsection{Learning}

We now focus on ML learning. The LL function can be written as
\begin{equation}
\ln p(\mathcal{D}\vert\pi,\text{\ensuremath{\eta}}_{0},\eta_{1})=\sum_{n=1}^{N}\ln p(t_{n}\vert\pi)+\sum_{{n=1:\atop t_{n}=0}}^{N}\ln p(x_{n}|\eta_{0})+\sum_{{n=1:\atop t_{n}=1}}^{N}\ln p(x_{n}\vert\eta_{1}).\label{eq:lldecomposition}
\end{equation}
Given the decomposition of the LL in (\ref{eq:lldecomposition}),
we can optimize over $\pi$, $\eta_{0}$ and $\eta_{1}$ separately,
obtaining the respective ML estimates. Note, however, that, while
for $\pi$ we can use the entire data set, the optimization over parameters
$\eta_{0}$ and $\eta_{1}$ can leverage smaller data sets that include
only the samples $x_{n}$ with labels $t_{n}=0$ or $t_{n}=1$, respectively.
As we discussed in Chapter 2, ML estimates of exponential families
merely require moment matching, making these estimates generally easy
to obtain. We illustrate this point below with two important examples.

\textbf{Quadratic Discriminant Analysis (QDA)}. In QDA, the class-dependent
distributions are Gaussian with class-dependent mean and covariance:\begin{subequations}\label{eq:QDA1}
\begin{align}
\mathrm{t} & \sim\textrm{Bern}(\pi)\\
\mathrm{x}|\mathrm{t} & =k\sim\mathcal{N}(\mu_{k},\Sigma_{k}).
\end{align}
\end{subequations}By the general rules derived in Chapter 2 for the
exponential family, ML selects the moment matching estimates\begin{subequations}\label{eq:QDA ML1}
\begin{align}
\pi_{ML} & =\frac{N[1]}{N}\\
\mu_{k,ML} & =\frac{1}{N[k]}\sum_{{n=1:\atop t_{n}=k}}^{N}x_{n}\\
\Sigma_{k,ML} & =\frac{1}{N[k]}\sum_{{n=1:\atop t_{n}=k}}^{N}(x_{n}-\mu_{k})(x_{n}-\mu_{k})^{T}.
\end{align}
\end{subequations}The resulting predictive distribution for the label
of a new sample is then given by (\ref{eq:predictive Ch4 gen}) by
plugging in the estimates above as
\begin{equation}
p(\mathcal{C}_{1}\vert x)=\frac{\pi_{ML}\mathcal{N}(x|\mu_{1,ML},\Sigma_{1,ML})}{\pi_{ML}\mathcal{N}(x\vert\mu_{1,ML},\Sigma{}_{1,ML})+(1-\pi_{ML})\mathcal{N}(x\vert\mu_{0,ML},\Sigma_{0,ML})}.\label{eq:QDAprobpost}
\end{equation}

\textbf{Linear Discriminant Analysis (LDA)}. Setting $\Sigma_{k}=\Sigma$
for both classes $k=1,2$ yields the Linear Discriminant Analysis
(LDA) model \cite{Murphy}. Imposing that two generally distinct parameters,
such as $\Sigma_{1}$ and $\Sigma_{2}$ are equal is an example of
\emph{parameter tying or sharing}. By reducing the number of parameters
to be learned, parameter sharing can reduce overfitting at the potential
cost of introducing bias (see Chapter 2).

Under the assumption of conjugate priors, and of a priori independence
of the parameters, MAP and Bayesian approaches can be directly obtained
by following the derivations discussed in Chapter 2. We refer to \cite{Bishop,Barber,Murphy}
for details.

\begin{figure}[H] 
\centering\includegraphics[width=0.8\textwidth]{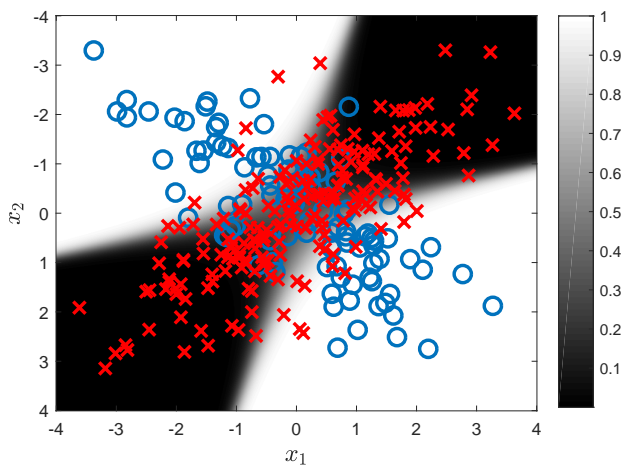}
\caption{Probability that the class label is the same as for the examples marked with circles according to the output of the generative model QDA. The probability is represented by the color map illustrated by the bar on the right of the figure. For this example, it can be seen that LDA fails to separate the two classes (not shown).} \label{fig:genCh4} \end{figure}

\begin{example} We continue the example in Sec. \ref{subsec:Non-linear-Classification}
by showing in Fig. \ref{fig:genCh4} the probability (\ref{eq:QDAprobpost})
that the class label is the same as for the examples marked with circles
according to the output of QDA. Given that the covariates have a structure
that is well modelled by a mixture of Gaussians with different covariance
matrices, QDA is seen to perform well, arguably better than the discriminative
models studied in Sec. \ref{subsec:Non-linear-Classification}. It
is important to note, however, that LDA would fail in this example.
This is because a model with equal class-dependent covariance matrices,
as assumed by LDA, would entail a significant bias for this example.
\end{example}

\subsection{Multi-Class Classification$^{*}$}

As an example of a generative probabilistic model with multiple classes,
we briefly consider the generalization of QDA to $K\geq2$ classes.
Extending (\ref{eq:QDA1}) to multiple classes, the model is described
as \begin{subequations}\label{eq:QDA}
\begin{align}
\mathrm{t} & \sim\textrm{Cat}(\pi)\\
\mathrm{x}|\mathrm{t} & =k\sim\mathcal{N}(\mu_{k},\Sigma_{k}),
\end{align}
\end{subequations}where $\mathrm{t}$ is encoded using one-hot encoding,
so that the label of each example is given by the vector $t_{n}=[t_{0n},...,t_{(K-1)n}]^{T}$.
Following the discussion above, moment matching yields the ML estimates\begin{subequations}\label{eq:QDA ML}
\begin{align}
\pi_{k,ML} & =\frac{N[k]}{N}=\frac{\sum_{n=1}^{N}t_{kn}}{N}\\
\mu_{k,ML} & =\frac{1}{N[k]}\sum_{n=1}^{N}t_{kn}x_{n}\\
\Sigma_{k,ML} & =\frac{1}{N[k]}\sum_{n=1}^{N}t_{kn}(x_{n}-\mu_{k})(x_{n}-\mu_{k})^{T}.
\end{align}
\end{subequations}

\section{Boosting$^{*}$\label{sec:Boosting} }

In this last section, we return to the mixture models of the form
(\ref{eq:mixture model Ch4}) and discuss a popular training approach
to reduce overfitting. We focus on deterministic discriminative models
with activations $a_{k}(x,\tilde{w}_{k})$, $k=1,...,K,$ in which
the mixture predictor is given as
\begin{equation}
a(x,\tilde{w})=\sum_{k=1}^{K}\pi_{k}a_{k}(x,\tilde{w}_{k})\label{eq:boosting Ch9}
\end{equation}
with learnable parameters $\{\pi_{k}\}$ and $\{\tilde{w}_{k}\}$.
The technique, known as boosting, trains one model $a_{k}(x,\tilde{w}_{k})$
at a time in a sequential fashion, from $k=1$ to $k=K$, hence adding
one predictor at each training step $k$. As a result, boosting increases
the capacity of the model in a sequential manner, extending the sum
in (\ref{eq:boosting Ch9}) over a growing number of predictors. In
this way, one starts by training a model with a large bias, or approximation
error; and progressively decreases the bias at the cost of a potentially
larger estimation error (see Chapter 2 and the next chapter for further
discussion on bias and estimation error). As we will discuss below,
each model is trained by solving an ERM problem in which the contribution
of a training example is weighted by the error rate of the previously
trained models. 

To elaborate, boosting \textendash{} more specifically the AdaBoost
scheme \textendash{} can be described as solving an ERM problem with
the exponential loss function $\ell(t,a(x,\tilde{w}))=\exp(-t\cdot a(x,\tilde{w}))$,
which is plotted in Fig. \ref{fig:Figlossfun}. When training the
$k$th model, the outputs $a_{1}(x,\tilde{w}_{1}),...,a_{k-1}(x,\tilde{w}_{k-1})$
of the previously trained models, as well as their weights $\pi_{1},...,\pi_{k-1}$,
are kept fixed. Excluding the models $k+1,...,K$, the training loss
can be written as 
\begin{equation}
\sum_{n=1}^{N}\alpha_{n}^{(k)}\exp(-\pi_{k}t_{n}\cdot a_{k}(x_{n},\tilde{w}_{k})),
\end{equation}
with the weights
\begin{equation}
\alpha_{n}^{(k)}=\exp\left(-t_{n}\cdot\sum_{j=1}^{k-1}\pi_{j}a_{j}(x_{n},\tilde{w}_{j})\right).\label{eq:weights boost Ch4}
\end{equation}
An important point is that the weights (\ref{eq:weights boost Ch4})
are larger for training samples $n$ with smaller functional margin
under the mixture model $\sum_{j=1}^{k-1}\pi_{j}a_{j}(x_{n},\tilde{w}_{j})$.
Therefore, when training the $k$th model, we give more importance
to examples that fare worse in terms of classification margins under
the current mixture model. Note that, at each training step $k$,
one trains a simple model, which has the added advantage of reducing
the computational complexity as compared to the direct learning of
the full training model. We refer to \cite[Ch. 14]{Bishop}\cite[Ch. 10]{Shalev}
for further details. 

\section{Summary}

This chapter has provided a brief review of the key supervised learning
problem of classification. Following the taxonomy put forth in Chapter
2, we have divided learning algorithms according to the type of models
used to relate explanatory variables and labels. Specifically, we
have described deterministic discriminative models, both linear and
non-linear, covering the perceptron algorithm, SVM, and backprop for
multi-layer neural networks; probabilistic discriminative models,
concentrating on GLM; and probabilistic generative models, including
QDA and LDA. We have also introduced the more advanced topics of mixture
models and boosting. We finally mention that supervised learning,
in the form of classification and regression, can also be used as
a building block for the task of sequential decision processing via
imitation learning (see, e.g., \cite{Levinerl}). 

While this chapter has focused on algorithmic aspects, the next chapter
discusses a well-established theoretical framework in which to study
the performance of learning for classification.

\section*{Appendix A: More on SGD\textmd{\normalsize{}$^{*}$}}

In this appendix, we provide some discussion on the convergence of
SGD and on more advanced optimization techniques.

\subsection*{Convergence}

To briefly discuss the convergence properties of SGD, consider first
for reference the conventional gradient descent algorithms, which
corresponds to choosing the entire training set, i.e., $\mathcal{S}=\{1,...,N\}$,
at each iteration. If the function to be optimized is strictly convex\footnote{If the function is twice differentiable, strict convexity is equivalent
to the requirement that all the eigenvalues of the Hessian matrix
are strictly positive.}, as for the quadratic loss, the algorithm is guaranteed to converge
to the (unique) minimum even with a fixed learning rate $\gamma^{(i)}=\gamma,$
as long as the latter is no larger than the inverse of the maximum
curvature of the loss function $L$, i.e., $\gamma\leq1/L$. For twice-differentiable
loss functions, the maximum curvature $L$ can be evaluated as the
maximum eigenvalue of the Hessian matrix. Functions with finite curvature
$L$ are known as Lipschitz smooth. For these functions, the convergence
is geometric, and hence the number of iterations needed to obtain
an error on the optimal solution equal to $\epsilon$ scales as $\textrm{ln}(1/\epsilon$)
(see, e.g., \cite[Chapter 8]{Aggelos}\cite{cevher2014convex,MAL-058}). 

We turn now to the proper SGD algorithm operating with a smaller mini-batch
size $S$. If the learning rate schedule is selected so as to satisfy
the Robbins\textendash Monro conditions
\begin{equation}
\sum_{i=1}^{\infty}\gamma^{(i)}=\infty\textrm{ and }\sum_{i=1}^{\infty}(\gamma^{(i)})^{2}<\infty,\label{eq:RobbinsMonro}
\end{equation}
the SGD algorithm is known to converge to the optimal solution of
problem (\ref{eq:ERM classification}) in the case of strictly convex
functions and to stationary points for non-convex functions with bounded
curvature (see \cite[Chapter 8]{Aggelos} for details). Learning rate
schedules that satisfy (\ref{eq:RobbinsMonro}) include $\gamma^{(i)}=1/i$.
The intuitive reason for the use of diminishing learning rates is
the need to limit the impact of the ``noise'' associated with the
finite-sample estimate of the gradient \cite{bertsekas2011incremental}.
The proof of convergence leverages the unbiasedness of the estimate
of the gradient obtained by SGD. 

In practice, a larger mini-batch size $S$ decreases the variance
of the estimate of the gradient, hence improving the accuracy when
close to a stationary point. However, choosing a smaller $S$ can
improve the speed of convergence when the current solution is far
from the optimum \cite[Chapter 8]{Aggelos}\cite{bertsekas2011incremental}.
A smaller mini-batch size $S$ is also known to improve the generalization
performance of learning algorithms by avoiding sharp extremal points
of the training loss function \cite{hochreiter1997flat,keskar2016large,Ferenc}
(see also Sec. \ref{subsec:Non-linear-Classification}). Furthermore,
as an alternative to decreasing the step size, one can also increase
the size of the mini-batch along the iterations of the SGD algorithm
\cite{Smithmini2017}. 

\subsection*{Variations and Generalizations}

Many variations of the discussed basic SGD algorithm have been proposed
and are routinely used. General principles motivating these schedule
variants include \cite[Chapter 8]{Goodfellowbook}: (\emph{i}) \emph{momentum,}
or \emph{heavy-ball,} \emph{memory}: correct the direction suggested
by the stochastic gradient by considering the ``momentum'' acquired
during the last update; (\emph{ii}) \emph{adaptivity}: use a different
learning rate for different parameters depending on an estimate of
the curvature of the loss function with respect to each parameter;
(\emph{iii}) \emph{control variates}: in order to reduce the variance
of the SGD updates, add control variates that do not affect the unbiasedness
of the stochastic gradient and are negatively correlated with the
stochastic gradient;  and (\emph{iv}) \emph{second-order updates}:
include information about the curvature of the cost or objective function
in the parameter update. 

As detailed in \cite[Chapter 8]{Goodfellowbook}\cite{johnson2013accelerating,de2015variance},
to which we refer for further discussions, schemes in the first category
include Nesterov momentum; in the second category, we find AdaGrad,
RMSprop and Adam; and the third encompasses SVRG and SAGA. Finally,
the fourth features Newton methods, which require calculation of the
Hessian to evaluate the local curvature of the objective, and approximated
Newton methods, which leverage an estimate of the Hessian. The practical
and theoretical implications of the use of these methods is still
under discussion \cite{Wilson}. 

Related to second-order methods is the \emph{natural gradient} approach,
which applies most naturally to probabilistic models in which the
function to be optimized is a LL \cite{amari1998natural}. The conventional
gradient method updates the parameters by moving in the direction
that minimizes the cost function under a constraint on the norm of
the update vector in the parameter space. A potentially problematic
aspect of this method is that the Euclidean distance $||\theta'-\theta''||^{2}$
between two parameter vectors $\theta'$ and $\theta'',$ e.g., two
mean parameters in a model within the exponential family, does not
provide a direct measure of the distance of the two corresponding
distributions in terms of relevant metrics such as the KL divergence.
The natural gradient method addresses this issue by measuring the
size of the update directly in terms of the KL divergence between
distributions. This modifies the update by pre-multiplying the gradient
with the inverse of the Fisher information matrix \cite{amari1998natural}.

The discussion above focuses on the common case of differentiable
cost functions. ERM problems typically include possibly non-differentiable
regularization terms. To tackle these problems, techniques such as
the subgradient method and proximal gradient can be used in lieu of
SGD \cite{bertsekas2011incremental}. Other important aspects of optimization
schemes include parallelism and non-convexity (see, e.g., \cite{Aldo,MM,EXTRA,Sayed}).
Alternatives to gradient methods that do not require differentiability
include evolutionary schemes \cite{evolutionary}.

\section*{Appendix B: Kernel Methods$^{*}$}

In this section, we provide a brief introduction to kernel methods.
This section requires some background in Lagrange duality.

We start by revisiting the problem (\ref{eq:SVM problem}) solved
by SVM. Using Lagrange duality, the optimization (\ref{eq:SVM problem})
can be solved in the dual domain, that is, by optimizing over the
dual variables or the Lagrange multipliers. Referring for details
to \cite[Chapter 7]{Bishop}, the resulting problem turns out to be
quadratic and convex. Importantly, the resulting optimal activation
can be expressed as
\begin{equation}
a(x,\tilde{w})=\sum_{n=1}^{N}\alpha_{n}t_{n}k(x,x_{n}),\label{eq:SVM dual}
\end{equation}
where $\alpha_{n}$ are the optimal dual variables, and we have defined
the \emph{kernel function}
\begin{equation}
k(x,y)=\phi(x)^{T}\phi(y),
\end{equation}
where $x$ and $y$ are two argument vectors. The kernel function
measures the correlation \textendash{} informally, the similarity
\textendash{} between the two input vectors $x$ and $y$. The activation
(\ref{eq:SVM dual}) has hence an intuitive interpretation: The decision
about the label of an example $x$ depends on the support vectors
$x_{n}$, which have $\alpha_{n}>0$, that are the most similar to
$x$. We note that equation (\ref{eq:SVM dual}) can also be justified
using the representer theorem in \cite[Chapter 16]{Shalev}, which
shows that the optimal weight vector must be a linear combination
of the feature vectors $\{\phi(x_{n})\}_{n=1}^{N}.$

Working in the dual domain can have computational advantages when
the number of the primal variables, here the size $D'$ of the weight
vector $\tilde{w}$, is larger than the number $N$ of dual variables.
While this seems a prior unlikely to happen in practice, it turns
out that this is not the case. The key idea is that one can use (\ref{eq:SVM dual})
with any other kernel function, not necessarily one explicitly defined
by a feature function $\phi(\cdot)$. A kernel function is any symmetric
function measuring the correlation of two data points, possibly in
an infinite-dimensional space. This is known as the \emph{kernel trick}.

As a first example, the polynomial kernel
\begin{equation}
k(x,y)=(\gamma x^{T}y+r)^{L},
\end{equation}
where $r>0$, corresponds to a correlation $\phi(x)^{T}\phi(y)$ in
a high-dimensional space $D'$. For instance, with $L=2$ and $D=1$,
we have $D'=6$ and the feature vector $\phi(x)=\left[1,\:\sqrt{2}x_{1},\:\sqrt{2}x_{2},\:x_{1}^{2}x_{2}^{2},\:\sqrt{2}x_{1}x_{2}\right]^{T}$
\cite{Murphy}. As another, more extreme, example, the conventional
Gaussian kernel
\begin{equation}
k(x,y)=e^{-r\left\Vert \mathrm{x}-y\right\Vert ^{2}}
\end{equation}
corresponds to an inner product in an infinite dimensional space \cite{Murphy}.
An extensive discussion on kernel methods can be found in \cite{Murphy}.

Before leaving the subject of kernel methods, it is worth noting that
an important class of methods including $k-$Nearest Neighbor ($k$-NN),
uses kernels that are data-dependent. $k$-NN is also an example of
\emph{non-parametric learning rules.} In contrast to the other schemes
studied here, it does not rely on a parametric model of the (probabilistic)
relationship between input and output. Instead, $k$-NN leverages
the assumption that the labels of nearby points $x$ should be similar
\cite{Koller}.

\chapter{Statistical Learning Theory$^{*}$}

Statistical learning theory provides a well-established theoretical
framework in which to study the trade-off between the number $N$
of available data points and the generalization performance of a trained
machine. The approach formalizes the notions of model capacity, estimation
error (or generalization gap), and bias that underlie many of the
design choices required by supervised learning, as we have seen in
the previous chapters. This chapter is of mathematical nature, and
it departs from the algorithmic focus of the text so far. While it
may be skipped at a first reading, the chapter sheds light on the
key empirical observations made in the previous chapters relative
to learning in a frequentist set-up. It does so by covering the theoretical
underpinnings of supervised learning within the classical framework
of statistical learning theory. To this end, the chapter contains
a number of formal statements with proofs. The proofs have been carefully
selected in order to highlight and clarify the key theoretical ideas.
This chapter follows mostly the treatment in \cite{Shalev}.

\section{A Formal Framework for Supervised Learning}

In this chapter, we concentrate on discriminative deterministic models
for binary classification, as it is typically done in statistical
learning theory. We also focus on the standard 0-1 loss $\ell(t,\hat{t})=1(\hat{t}\neq t)$,
for which the generalization loss is the probability of error. The
labels $t$ for the two classes take values in the set $\{0,1\}$
(cf. (\ref{eq:mapppingtc})).

The learning problem is formalized as follows. Assume that a model,
or hypothesis class, $\mathcal{H}$ has been selected. This set contains
a, possibly uncountable, number of predictors $\hat{t}$ that map
each point $x$ in the domain space to a label $\hat{t}(x)$ in $\{0,1\}$.
We would like to choose a specific hypothesis, or predictor, $\hat{t}\in\mathcal{H}$
that minimize the generalization error (cf. (\ref{eq:gen error inf}))
\begin{equation}
L_{p}(\hat{t})=E_{(\mathrm{x},\mathrm{t})\sim p_{\mathrm{xt}}}[\ell(\mathrm{t},\hat{t}(\mathrm{x}))].\label{eq:gen error SL}
\end{equation}
Solving this inference problem would yield an optimal model within
class $\mathcal{H}$ as 
\begin{equation}
\hat{t}_{\mathcal{H}}^{*}\in\textrm{arg}\underset{\hat{t}\in\mathcal{H}}{\min}\,L_{p}(\hat{t}).\label{eq:optimal predictor SL}
\end{equation}
The notation in (\ref{eq:optimal predictor SL}) emphasizes that there
may be multiple optimal hypotheses returned by the minimization of
the generalization error $L_{p}(\hat{t})$. Nevertheless, to fix the
ideas, it is useful to think of the case in which there is a unique
optimal hypothesis. This is, for instance, the case when the loss
function is strictly convex. Obtaining the optimal predictor (\ref{eq:optimal predictor SL})
requires knowledge of the true distribution $p(x,t)$, which is not
available. 

\begin{example}For the linear (deterministic) methods studied in
Chapter 4, the model is defined as 
\begin{equation}
\mathcal{H}=\{t_{\tilde{w}}(x)=\textrm{sign}(w^{T}x+w_{0})\}\label{eq:linear classifier SL}
\end{equation}
with $\tilde{w}=[w^{T}$ $w_{0}]^{T}$, and similarly for the feature-based
version. Identifying a hypothesis within this class requires the selection
of the weight vector $\tilde{w}\in\mathbb{R}^{D+1}$.\end{example}

In lieu of the true distribution $p(x,t)$, what is available is an
i.i.d. training set
\begin{equation}
\mathcal{D}=\{(\mathrm{x}_{n},\mathrm{t}_{n})\}_{n=1}^{N}\underset{\textrm{i.i.d.}}{\sim}p(x,t)
\end{equation}
distributed according to $p(x,t)$. A learning algorithm, such as
ERM, takes the training set $\mathcal{D}$ as input and returns a
predictor $\hat{t}_{\mathcal{D}}\in\mathcal{H}$ as output. We would
like the predictive model $\hat{t}_{\mathcal{D}}\in\mathcal{H}$ to
yield a generalization error $L_{p}(\hat{t}_{\mathcal{D}})$ that
is as close as possible to the minimum generalization loss $L_{p}(\hat{t}_{\mathcal{H}}^{*})$.
Note that the selected model $\hat{t}_{\mathcal{D}}$ is random due
to randomness of the data set $\mathcal{D}$. 

In this regard, we recall that the ERM learning rule chooses a hypothesis
$\hat{t}_{\mathcal{D}}^{ERM}\in\mathcal{H}$ by following the criterion
$\hat{t}_{\mathcal{D}}^{ERM}=\textrm{arg}\textrm{min}_{\hat{t}\in\mathcal{H}}L_{\mathcal{D}}(\hat{t})$,
where the empirical risk is
\begin{equation}
L_{\mathcal{D}}(\hat{t})=\frac{1}{N}\sum_{n=1}^{N}\ell(\mathrm{t}_{n},\hat{t}(\mathrm{x}_{n}))).\label{eq:empirical risk SL}
\end{equation}
The notation in (\ref{eq:empirical risk SL}) emphasizes the randomness
of the training set $\mathcal{D}=\{(\mathrm{x}_{n},\mathrm{t}_{n})\}_{n=1}^{N}$.

\begin{figure}[htbp]
\centering
\includegraphics[width=90mm]{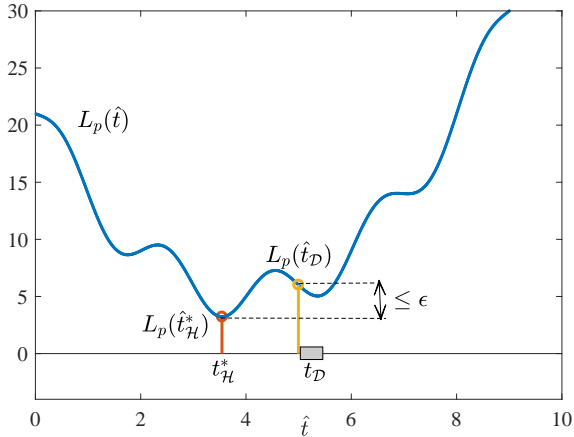}
\caption{A learning algorithm $\hat{t}_{\mathcal{D}}$ outputs a hypothesis that depends on the random training set $\mathcal{D}$. It hence takes values in a given interval (the box on the horizontal axis) with some large probability $1-\delta$. The accuracy level $\epsilon$ is measured by the difference with respect to optimal generalization loss $L_p(\hat{t}^{*}_{\mathcal{H}})$ for the worst-case $\hat{t}_{\mathcal{D}}$ in the high-probability interval. \label{fig:Fig1Ch5}}
\end{figure}

Since the distribution $p(x,t)$ is unknown, a learning rule $\hat{t}_{\mathcal{D}}$,
such as ERM, can only minimize the generalization loss $L_{p}(\hat{t})$
\emph{approximately} based on the observation of the data $\mathcal{D}$.
Furthermore, this approximation can only be guaranteed at some \emph{probability}
level due to the randomness of the data set $\mathcal{D}$. This is
illustrated in Fig. \ref{fig:Fig1Ch5}, in which we have represented
a high-probability interval for rv $L_{p}(\hat{t}_{\mathcal{D}})$
on the horizontal axis. We would like the approximation to be accurate
for all values of $L_{p}(\hat{t}_{\mathcal{D}})$ within this interval. 

But there is more: the probabilistic guarantee in terms of accuracy
cannot depend on the specific distribution $p(x,t)$, but it should
instead be \emph{universal} with respect to all distributions $p(x,t)$.
In summary, the best one can hope for is to have a learning rule $\hat{t}_{\mathcal{D}}$
that is \emph{Probably Approximately Correct} (PAC).

In order to formalize this notion, we introduce the following definition. 

\begin{definition}\textbf{ }A learning rule $\hat{t}_{\mathcal{D}}$
is ($N,\epsilon,\delta$) PAC if, when working on data sets $\mathcal{D}$
of $N$ examples, it satisfies the inequality
\begin{equation}
L_{p}(\hat{t}_{\mathcal{D}})\leq L_{p}(\hat{t}_{\mathcal{H}}^{*})+\epsilon\label{eq:PAC formal}
\end{equation}
with probability no smaller than $1-\delta,$ that is,
\begin{equation}
\textrm{P}\textrm{r}_{\mathcal{D\underset{\textrm{i.i.d.}}{\sim}}p_{\mathrm{xt}}}[L_{p}(\hat{t}_{\mathcal{D}})\leq L_{p}(\hat{t}_{\mathcal{H}}^{*})+\epsilon]\geq1-\delta,\label{eq:PAC formal 1}
\end{equation}
for any true distribution $p(x,t)$. \end{definition}\textbf{ }

In (\ref{eq:PAC formal}), we have defined $\epsilon$ as the \emph{accuracy}
parameter and $\delta$ as the \emph{confidence} parameter. The accuracy
is also known as \emph{estimation error }or\emph{ generalization gap}
according to the definition given in Sec. \ref{subsec:Maximum-Likelihood-(ML)}.
In words, the ($N,\epsilon,\delta$) PAC condition (\ref{eq:PAC formal})
requires that the learning rule $\hat{t}_{\mathcal{D}}$ operating
over $N$ data points is $\epsilon-$accurate with probability $1-\delta$
for any distribution $p(x,t)$.

\begin{figure}[htbp]
\centering
\includegraphics[width=95mm]{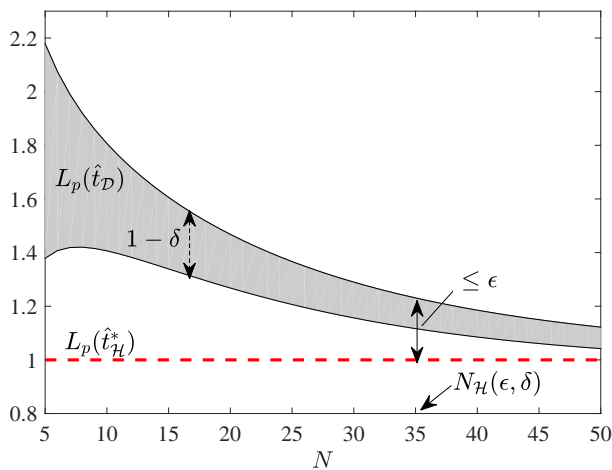}
\caption{High-probability interval (dashed arrow) for the generalization error $L_{p}(\hat{t}_{\mathcal{D}})$ versus the number $N$ of data points for a model $\mathcal{H}$.\label{fig:Fig2Ch5}}
\end{figure}

The key question is: \emph{Given a model $\mathcal{H},$ how large
should $N$ be in order to ensure the existence of an ($N,\epsilon$,$\delta$)
PAC learning scheme $\hat{t}_{\mathcal{D}}$ for given accuracy and
confidence levels $(\epsilon$,$\delta$)}? At a high level, we know
that a large model order implies the need for a larger $N$ in order
to avoid overfitting. More precisely, we expect to observe the behavior
illustrated in Fig. \ref{fig:Fig2Ch5}: As $N$ increases, (\emph{i})
the interval of values taken by the generalization loss $L_{p}(\hat{t}_{\mathcal{D}})$
with probability no smaller than $1-\delta$ shrinks; and (\emph{ii})
the generalization loss $\ensuremath{L_{p}(\hat{t}_{\mathcal{D}})}$
tends to the minimum generalization loss $\ensuremath{L_{p}(\hat{t}_{\mathcal{H}}^{*})},$
and hence the estimation error vanishes. As illustrated in the next
example, this expected behavior is a consequence of the law of large
numbers, since a larger $N$ allows an increasingly accurate estimation
of the true general loss. 

\begin{figure}[htbp]
\centering
\includegraphics[scale=0.65]{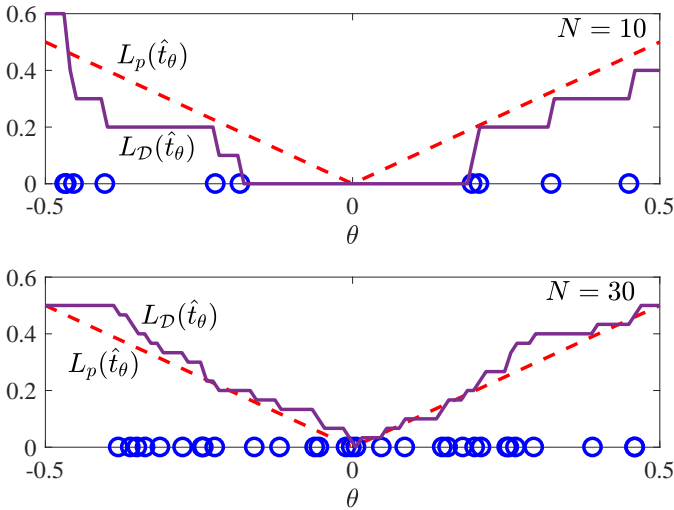}
\caption{Generalization and training losses for a scalar threshold classifier model.\label{fig:Fig3Ch5}}
\end{figure}

\begin{example}Consider the problem of binary classification using
the model of threshold functions, namely 
\begin{equation}
\mathcal{H}=\left\{ \hat{t}_{\theta}(x)=\begin{cases}
0, & \text{if }x<\theta\\
1, & \text{if }\,x\geq\theta
\end{cases}=1(x\geq\theta)\right\} ,\label{eq:threshold model}
\end{equation}
where $x$ is a real number ($D=1$). Note that the model is parametrized
by the threshold $\theta$. Make the \emph{realizability }assumption
that the true distribution is within the hypothesis allowed by the
model, i.e., $p(x,t)=p(x)1(t=\hat{t}_{0}(x))$ and hence the optimal
hypothesis is $\hat{t}_{\mathcal{H}}^{*}=\hat{t}_{0}$, or, equivalently,
the optimal threshold is $\theta^{*}=0$. Assuming a uniform distribution
$p(x)=\mathcal{U}(x|-0.5,0.5)$ in the interval $[-0.5,0.5]$ for
the domain points, Fig. \ref{fig:Fig3Ch5} shows the generalization
error $\textrm{Pr}[\hat{t}_{\theta}(\mathrm{x})\neq t_{0}(\mathrm{x})]=|\theta|$,
as well as the training loss $L_{\mathcal{D}}(\hat{t}_{\theta})$
for the training set shown on the horizontal axis, for two values
of $N$. Note that the training loss $L_{\mathcal{D}}(\hat{t}_{\theta})$
is simply the fraction of training examples that are correctly classified.
It is observed that, as $N$ increases, the training loss, or empirical
risk, becomes an increasingly reliable estimate of the generalization
loss uniformly for all hypotheses, parameterized by $\theta$, in
the model.\end{example}

As suggested by the example, if $N$ is large enough, the empirical
risk, or training loss, $L_{\mathcal{D}}(\hat{t})$ approximates increasingly
well (with high probability) the generalization loss $L_{p}(\hat{t})$
for \emph{any} fixed hypothesis in $\hat{t}\in\mathcal{H}$ by the
law of large numbers. It would then seem that the problem is solved:
Since $L_{\mathcal{D}}(\hat{t})\simeq L_{p}(\hat{t})$ for any $\hat{t}$,
the ERM solution $\hat{t}_{\mathcal{D}}^{ERM}$, which minimizes the
training loss $L_{\mathcal{D}}(\hat{t})$, should also approximately
minimize the generalization loss $L_{p}(\hat{t})$, and hence we have
$\hat{t}_{\mathcal{D}}^{ERM}\simeq\hat{t}_{\mathcal{H}}^{*}$. However,
this argument is incorrect. In fact, we need the training loss $L_{\mathcal{D}}(\hat{t})$
to be an accurate approximation of the generalization loss $L_{p}(\hat{t})$
\emph{uniformly} for \emph{all} hypotheses in $\hat{t}\in\mathcal{H}$
in order to ensure the condition $\hat{t}_{\mathcal{D}}^{ERM}\simeq\hat{t}_{\mathcal{H}}^{*}.$
As we will see in the rest of this chapter, guaranteeing this condition
requires to observe a number of samples $N$ that grows with the ``capacity''
of the model $\mathcal{H}$, that is, roughly, with the number of
parameters defining the hypotheses in $\mathcal{H}$. Moreover, some
models turn out to be impossible to learn \textendash{} in the sense
of PAC learnability formalized below \textendash{} no matter how large
$N$ is.

\section{PAC Learnability and Sample Complexity}

In order to formally address the key question posed above regarding
the learnability of a model $\mathcal{H}$, we make the following
definitions. As mentioned, for simplicity, we consider binary classification
under the 0-1 loss, although the analysis can be generalized under
suitable conditions \cite{Shalev}.

\begin{definition}A hypothesis class $\mathcal{H}$ is PAC learnable
if, for any $\epsilon,\delta\in(0,1)$, there exist an ($N,\epsilon,\delta$)
PAC learning rule as long as the inequality $N\geq N_{\mathcal{H}}(\epsilon,\delta)$
is satisfied for some function $N_{\mathcal{H}}(\epsilon,\delta)<\infty$.
\end{definition}

In words, a hypothesis class is PAC learnable if, as long as enough
data is collected, a learning algorithm can be found that obtains
any desired level of accuracy and confidence. An illustration of the
threshold $N_{\mathcal{H}}(\epsilon,\delta)$ can be found in Fig.
\ref{fig:Fig2Ch5}. A less strong definition of PAC learnability requires
(\ref{eq:PAC formal 1}) to hold only for all true distributions $p(x,t)$
that can be written as
\begin{equation}
p(x,t)=p(x)1(t=\hat{t}(x))\label{eq:realizability}
\end{equation}
for some marginal distribution $p(x)$ and for some hypothesis $\hat{t}(x)\in\mathcal{H}$.
The condition (\ref{eq:realizability}) is known as the \emph{realizability
}assumption, which implies that the data is generated from some mechanism
that is included in the hypothesis class. Note that realizability
implies the linear separability of any data set drawn from the true
distribution for the class of linear predictors (see Chapter 4). 

A first important, and perhaps surprising, observation is that not
all models are PAC learnable. As an extreme example of this phenomenon,
consider the class\textbf{ $\mathcal{H}$} of all functions from $\mathbb{R}^{D}$
to $\{0,1\}$. By the \emph{no free lunch theorem}, this class is
not PAC learnable. In fact, given any amount of data, we can always
find a distribution $p(x,t)$ under which the PAC condition is not
satisfied. Intuitively, even in the realizable case, knowing the correct
predictor $\hat{t}(x)$ in (\ref{eq:realizability}) for any number
of $x\in\mathbb{R}^{D}$ yields no information on the value of $\hat{t}(x)$
for other values of $x$. As another, less obvious, example the class
\begin{equation}
\mathcal{H}=\{h_{w}(x)=1(\sin(wx)>0)\}\label{eq:not PAC}
\end{equation}
 is not PAC learnable despite being parameterized by a single scalar
\cite{Shalev}.

\begin{definition}The sample complexity $N_{\mathcal{H}}^{*}(\epsilon,\delta)$
of model $\mathcal{H}$ is the minimal value of $N_{\mathcal{H}}(\epsilon,\delta)$
that satisfies the requirements of PAC learning for $\mathcal{H}$.\end{definition}

We will see next that the sample complexity depends on the capacity
of the model $\mathcal{H}$. Note that the sample complexity of the
two examples above is infinite since they are not PAC learnable. We
also remark that PAC learnability may be alternatively defined under
the additional conditions on the scaling of $N_{\mathcal{H}}^{*}(\epsilon,\delta)$
as a function of $\epsilon$ and $\delta$, as well as on the computational
complexity of the learning rule. We will not consider these more refined
definitions here, and we refer the reader to \cite{Hastie,Shalev}
for discussion.

\section{PAC Learnability for Finite Hypothesis Classes\label{sec:PAC-Learnability-for-Ch5}}

In this section, we consider models with a finite number of hypotheses.
The main result is summarized in the following theorem, which is proved
below in Sec. \ref{sec:Ch5proof}.

\begin{theorem}\label{th:Ch5thm1}A finite hypothesis class $\mathcal{H}$
is PAC learnable with sample complexity satisfying the inequality
\begin{equation}
N_{\mathcal{H}}^{*}(\epsilon,\delta)\leq\Biggl\lceil\frac{2\ln|\mathcal{H}|+2\ln(2/\delta)}{\epsilon^{2}}\Biggr\rceil\triangleq N_{\mathcal{H}}^{ERM}(\epsilon,\delta).\label{eq:sample complexity ERM}
\end{equation}
Moreover, the ERM algorithm achieves the upper bound $N_{\mathcal{H}}^{ERM}(\epsilon,\delta)$.
\end{theorem}

The previous theorem shows that all finite classes are PAC learnable.
Furthermore, for all finite classes, ERM is a PAC learning rule for
any desired levels of accuracy and confidence ($\epsilon,\delta$),
as long as $N$ is larger than the threshold $N_{\mathcal{H}}^{ERM}(\epsilon,\delta)$.
This threshold, which we will refer to as the \emph{ERM sample complexity
for class} $\mathcal{H}$, depends on the \emph{capacity} of the hypothesis
class, defined as $\ln|\mathcal{H}|$ (nats) or $\log_{2}|\mathcal{H}|$
(bits). This is the number of bits required to index the hypotheses
in $\mathcal{H}$. It is also interesting to note that increasing
the accuracy, i.e., decreasing $\epsilon$ is more demanding than
increasing the confidence, that is, decreasing $\delta$, in terms
of sample complexity.

Another way to understand the result (\ref{eq:sample complexity ERM})
is that, with $N$ data points, we can achieve the estimation error
\begin{equation}
\epsilon=\sqrt{\frac{2\ln(2|\mathcal{H}|/\delta)}{N}},\label{eq:epsilon Ch5}
\end{equation}
with probability $1-\delta$, by using ERM. As a result, with $N$
data points, we can upper bound the generalization loss of ERM as
\begin{equation}
L_{p}(\hat{t}_{\mathcal{D}}^{ERM})\leq L_{p}(\hat{t}_{\mathcal{H}}^{*})+\sqrt{\frac{2\ln|\mathcal{H}|/\delta}{N}}
\end{equation}
with probability $1-\delta$. In words, ERM achieves the optimal generalization
loss with an estimation error that scales with square root of the
model capacity and with the inverse square root of $N$.

As another important note, under the realizability assumption, the
theorem can be modified to yield the smaller upper bound \cite{Shalev}
\begin{equation}
N_{\mathcal{H}}^{*}(\epsilon,\delta)\leq\Biggl\lceil\frac{\ln|\mathcal{H}|+\textrm{ln(1/}\delta)}{\epsilon}\Biggr\rceil\triangleq N_{\mathcal{H}}^{ERM}(\epsilon,\delta),\label{eq:sample complexity real}
\end{equation}
which is also achievable by ERM.

What does the theorem say about infinite models such as the linear
classifier (\ref{eq:linear classifier SL})? One approach is to learn
a ``quantized'' version of $\mathcal{H},$ say $\mathcal{H}_{b}$,
in which each weight is represented by $b$ bits or, equivalently,
as one of $2^{b}$ pre-determined quantization levels. As a result,
the number of hypotheses in the hypothesis class $\mathcal{H}$ is
$|\mathcal{H}|=(2^{b})^{D+1}$, and the capacity of the hypothesis
class is $\log|\mathcal{H}|=(D+1)b$ (bits) or $\ln|\mathcal{H}|=b(D+1)\ln2$
(nats). It follows that, using (\ref{eq:linear classifier SL}), we
obtain the ERM sample complexity 
\begin{equation}
N_{\mathcal{H}}^{ERM}(\epsilon,\delta)=\Biggl\lceil\frac{2b(D+1)\ln2+2\ln(2/\delta)}{\epsilon^{2}}\Biggr\rceil.\label{eq:sample complexity ERM-1}
\end{equation}
It is observed that the ERM sample complexity scales proportionally
to the number of parameters $D+1$ and to the resolution $b$. Therefore,
obtaining an arbitrary precision by selecting larger values of $b$
yields a sample complexity that grows unbounded. We will see below
how to correct this result by introducing a more advanced theory of
generalization through the concept of Vapnik\textendash Chervonenkis
(VC) dimension.

\subsection{Proof of Theorem \ref{th:Ch5thm1}\label{sec:Ch5proof}}

The proof of Theorem \ref{th:Ch5thm1} reveals the role played by
the training loss $L_{\mathcal{D}}(\hat{t})$ in approximating the
generalization loss $L_{p}(\hat{t})$ uniformly for all hypotheses
$\hat{t}\in\mathcal{H}$. We start with the following key lemma.

\begin{lemma}\label{lemmaproofCh5}For any $N\geq N_{\mathcal{H}}^{ERM}(\epsilon,\delta)$,
we have 
\begin{equation}
\text{Pr}_{\mathcal{D}\underset{\textrm{i.i.d.}}{\sim}p(x,t)}\left[|L_{p}(\hat{t})-L_{\mathcal{D}}(\hat{t})|\leq\frac{\epsilon}{2}\;\textrm{for all }\hat{t}\in\mathcal{H}\right]\geq1-\delta.
\end{equation}
\end{lemma}

Reflecting the observation made above around Fig. \ref{fig:Fig3Ch5},
the lemma says that the training loss $L_{\mathcal{D}}(\hat{t}$)
is a uniformly accurate approximation, with accuracy level $\epsilon/2$,
of the generalization loss, as long as $N\geq N_{\mathcal{H}}^{ERM}(\epsilon,\delta)$. 

Assume now that the lemma is true \textendash{} a proof will be given
below. Using the lemma, Theorem \ref{th:Ch5thm1} follows immediately
from the inequalities
\begin{align}
L_{p}(\hat{t}_{\mathcal{D}}^{ERM}) & \leq L_{\mathcal{D}}(\hat{t}_{\mathcal{D}}^{ERM})+\frac{\epsilon}{2}\leq L_{\mathcal{D}}(\hat{t}^{*})+\frac{\epsilon}{2}\\
 & \leq L_{p}(\hat{t}^{*})+\frac{\epsilon}{2}+\frac{\epsilon}{2}=L_{p}(\hat{t}^{*})+\epsilon,
\end{align}
where the first inequality follows from the lemma; the second from
the definition of ERM; and the third by another application of the
lemma. 

We hence only need to prove the Lemma in order to conclude the proof.
To proceed, we will use Hoeffding's inequality, which says the following
(see, e.g., \cite{Shalev}). For i.i.d. rvs\textbf{ $\mathrm{u}_{1},\mathrm{u}_{2},\cdots,\mathrm{u}_{M}\sim p(u)$}
such that $\textrm{E}\left[\mathrm{u}_{i}\right]=\mu$ and $\textrm{Pr}[a\leq\mathrm{u}_{i}\leq b]=1$,
we have the large deviation inequality
\begin{equation}
\text{Pr}\left[\Biggl|\frac{1}{M}\sum_{m=1}^{M}\mathrm{u}_{m}-\mu\Biggr|>\epsilon\right]\leq2\exp\left(-\frac{2M\epsilon^{2}}{(b-a)^{2}}\right).
\end{equation}
We can now write the following sequence of equalities and inequalities,
which prove the lemma and hence conclude the proof:
\begin{align*}
 & \text{Pr}_{\mathcal{D}\underset{\textrm{i.i.d.}}{\sim}p(x,t)}\left[\exists\hat{t}\in\mathcal{H}:\textrm{}|L_{p}(\hat{t})-L_{\mathcal{D}}(\hat{t})|>\frac{\epsilon}{2}\right]\\
= & \text{Pr}_{\mathcal{D}\underset{\textrm{i.i.d.}}{\sim}p(x,t)}\left[\bigcup_{\hat{t}\in\mathcal{H}}\left\{ |L_{p}(\hat{t})-L_{\mathcal{D}}(\hat{t})|>\frac{\epsilon}{2}\right\} \right]\\
\leq & \sum_{\hat{t}\in\mathcal{H}}\text{Pr}_{\mathcal{D}\underset{\textrm{i.i.d.}}{\sim}p(x,t)}\left[|L_{p}(\hat{t})-L_{\mathcal{D}}(\hat{t})|>\frac{\epsilon}{2}\right]\\
\underset{}{\leq} & 2\sum_{\hat{t}\in\mathcal{H}}\exp\left(-\frac{N\epsilon^{2}}{2}\right)\\
= & 2|\mathcal{H}|\exp\left(-\frac{N\epsilon^{2}}{2}\right)\leq\delta,
\end{align*}
where the first inequality follows by the union bound; the second
by Hoeffding's inequality; and the third can be verified to be true
as long as the inequality $N\geq N_{\mathcal{H}}^{ERM}(\epsilon,\delta)$
is satisfied.

\subsection{Structural Risk Minimization$^{*}$}

The result proved above is useful also to introduce the Structural
Risk Minimization (SRM) learning approach. SRM is a method for joint
model selection and hypothesis learning that is based on the minimization
of an upper bound on the generalization loss. In principle, the approach
avoids the use of validation, and has deep theoretical properties
in terms of generalization \cite{Shalev}. In practical applications,
the approach is rarely used, and validation is often preferable. It
is nevertheless conceptually and theoretically a cornerstone of statistical
learning theory.

To elaborate, assume that we have a nested set of hypothesis classes
$\mathcal{H}_{1}\subseteq\mathcal{H}_{2}\subseteq...\subseteq\mathcal{H}_{M_{\max}}$.
For instance, the nested model may correspond to linear classifiers
with increasing orders $M\in\{1,2,...,M_{\max}\}$. From Lemma \ref{lemmaproofCh5},
we can obtain the following bound 
\begin{equation}
L_{p}(\hat{t})\leq L_{\mathcal{D}}(\hat{t})+\sqrt{\frac{\ln(2|\mathcal{H}_{M}|/\delta)}{2N}}\label{eq:SRM}
\end{equation}
for all $\hat{t}\in\mathcal{H}_{M},\text{ with probability }1-\delta.$
SRM minimizes this upper bound, which is a pessimistic estimate of
the generalization loss, over both the choice of the model $M$ and
the hypothesis $\hat{t}\in\mathcal{H}_{M}$. We note the similarity
of this approach with the simplified MDL criterion based on two-part
codes covered in Chapter 2.

\section{VC Dimension and Fundamental Theorem of PAC Learning}

We have seen that finite classes are PAC learnable with sample complexity
proportional to the model capacity $\ln|\mathcal{H}|$ by using ERM.
In this section, we address the following questions: Is $N_{\mathcal{H}}^{ERM}(\epsilon,\delta)$
the smallest sample complexity? How can we define the capacity of
infinite hypothesis classes? We have discussed at the end of Sec.
\ref{sec:PAC-Learnability-for-Ch5} that the answer to the latter
question cannot be found by extrapolating from results obtained when
considering finite hypothesis classes. In contrast, will see here
that the answers to both of these questions rely on the concept of
VC dimension, which serves as a more fundamental definition of capacity
of a model. The VC dimension is defined next. 

\begin{definition}A hypothesis class $\mathcal{H}$ is said to \emph{shatter
}a set of domain points $\mathcal{X}=\{x_{n}\}_{n=1}^{V}$ if, no
matter how the corresponding labels $\{t_{n}\in\{0,1\}\}_{n=1}^{V}$
are selected, there exists a hypothesis $\hat{t}\in\mathcal{H}$ that
ensures $\hat{t}(x_{n})=t_{n}$ for all $n=1,...,V$.\emph{ }\end{definition}

\begin{definition}The VC dimension $\text{VCdim}(\mathcal{H})$\textbf{
}of the model $\mathcal{H}$ is the size of the largest set $\mathcal{X}$
that is shattered by $\mathcal{H}.$\end{definition}

Based on the definitions above, to prove that a model has $\text{VCdim}(\mathcal{H})=V$,
we need to carry out the following two steps: 

\indent\emph{ Step 1}) Demonstrate the existence of a set $\mathcal{X}$
with $|\mathcal{X}|=V$ that is shattered by $\mathcal{H}$; and 

\indent\emph{ Step 2}) Prove that no set $\mathcal{X}$ of dimension
$V+1$ exists that is shattered by $\mathcal{H}$. 

The second step is typically seen to be more difficult, as illustrated
by the following examples.

\begin{example}The threshold function model (\ref{eq:threshold model})
has VCdim($\mathcal{H}$)$=1$, since there is clearly a set $\mathcal{X}$
of one sample ($V=1$) that can be shattered (Step 1); but there are
no sets of $V=2$ that can be shattered (Step 2). In fact, for any
set $\mathcal{X}=(x_{1},x_{2})$ of two points with $x_{1}\leq x_{2}$,
the label assignment $(t_{1},t_{2})=(1,0)$ cannot be realized by
any choice of the threshold $\theta$, which is the only parameter
in the model. \end{example}

\begin{example}The model $\mathcal{H}=\{\hat{t}_{a,b}(x)=1\left(a\leq x\leq b\right)\}$,
which assigns the label $t=1$ within an interval $[a,b]$ and the
label $t=0$ outside it, has VCdim($\mathcal{H}$)$=2$. In fact,
any set of $V=2$ points can be shattered \textendash{} and hence
there also exists one such set (Step 1); while there are no sets $\mathcal{X}$
of $V=3$ points that can be shattered (Step 2). For Step 2, note
that, for any set $\mathcal{X}=(x_{1},x_{2},x_{3})$ of three points
with $x_{1}\leq x_{2}\leq x_{3}$, the label assignment $(t_{1},t_{2},t_{3})=(1,0,1)$
cannot be realized by any choice of the two free parameters $(a,b)$.\end{example}

\begin{example}The model $\mathcal{H}=\{\hat{t}_{a_{1},a_{2},b_{1},b_{2}}(x)=1(a_{1}\leq x_{1}\leq a_{2}\;\text{and}\;b_{1}\leq x_{2}\leq b_{2})\}$,
which assigns the label $t=1$ within an axis-aligned rectangle defined
by parameters $a_{1},a_{2},b_{1}$ and $b_{2}$ has VCdim($\mathcal{H}$)$=4$,
as it can be proved by using arguments similar to the previous examples.
\end{example}

\begin{example}The linear classifier (\ref{eq:linear classifier SL})
has VCdim($\mathcal{H}$)$=D+1$ \cite{Shalev}. \end{example}

The example above suggests that the VC dimension of a set often coincides
with the number of degrees of freedom, or free parameters, in the
model. However, this is not necessarily the case.

\begin{example}Model (\ref{eq:not PAC}), while having a single parameter,
has an infinite VC dimension \cite{Shalev}. \end{example} 

We also note that, for finite classes, we have the inequality $\text{VCdim}(\mathcal{H})\leq\log|\mathcal{H}|$,
since $|\mathcal{H}|$ hypotheses can create at most $|\mathcal{H}|$
different label configurations. The next theorem, whose importance
is attested by its title of \emph{fundamental theorem of PAC learning,}
provides an answer to the two questions posed at the beginning of
this section.

\begin{theorem}A model $\mathcal{H}$ with finite VCdim($\mathcal{H}$)$=d<\infty$
is PAC learnable with sample complexity 
\begin{equation}
C_{1}\frac{d+\ln(1/\delta)}{\epsilon^{2}}\leq N_{\mathcal{H}}^{*}(\epsilon,\delta)\leq C_{2}\frac{d+\ln(1/\delta)}{\epsilon^{2}}
\end{equation}
for some constants $C_{1}$ and $C_{2}$. Moreover, the ERM learning
rule achieves the upper bound.\end{theorem}

The theorem shows that the sample complexity is proportional to $(\textrm{VCdim}(\mathcal{H})+\ln(1/\delta))/\epsilon^{2}$.
This reveals that VCdim($\mathcal{H}$) can be considered as the correct
definition of capacity for a hypothesis class $\mathcal{H}$, irrespective
of whether the class is finite or not: As VCdim($\mathcal{H}$) increases,
the number of required data points for PAC learning increases proportionally
to it. Furthermore, the theorem demonstrates that, if learning is
possible for a given model $\mathcal{H},$ then ERM allows us to learn
with close-to-optimal sample complexity.

For a proof of this result and for extensions, we refer to the extensive
treatment in \cite{Shalev}. We mention here the important extension
of the theory of generalization to convex learning problems \textendash{}
i.e., to problems with convex parameter set and a convex loss function.
Rather than depending on the model complexity as the theory developed
so far, generalization in this class of problems hinges on the stability
of the learning algorithms. Stability is the property that small changes
in the input do not affect much the output of the learning algorithm
\textendash{} a notion related to that of differential privacy \cite{raginsky2016information}.
We also point to the related notion of capacity of a perceptron introduced
in \cite{Mackay}.

\section{Summary}

This chapter has described the classical PAC framework for the analysis
of the generalization performance of supervised learning for classification.
We have seen that the concept of VC dimension defines the capacity
of the model, and, through it, the number of samples needed to learn
the model with a given accuracy and confidence, or sample complexity.
In the next chapter, we move from supervised learning to unsupervised
learning problems. 

\section*{Appendix: Minimax Redundancy and Model Capacity$^{*}$}

In this appendix, we describe an alternative definition of model capacity
that is directly related to the conventional notion of Shannon's capacity
of a noisy channel \cite{Cover}. As for Sec. \ref{sec:Minimum-Description-Length},
some background in information theory may be needed to fully appreciate
the content of this appendix.

To elaborate, consider a probabilistic model $\mathcal{H}$ defined
as the set of all pmfs $p(x|\theta)$ parametrized by $\theta$ in
a given set. With some abuse of notation, we take $\mathcal{H}$ to
be also the domain of parameter $\theta$. To fix the ideas, assume
that $x$ takes values over a finite alphabet. We know from Sec. \ref{sec:Minimum-Description-Length},
that a distribution $q(x)$ is associated with a lossless compression
scheme that requires around $-\log q(x)$ bits to describe a value
$x$. Furthermore, if we were informed about the true parameter $\theta$,
the minimum average coding length would be the entropy $H(p(x|\theta)),$
which requires setting $q(x)=p(x|\theta)$ (see Appendix A). 

Assume now that we only know that the parameter $\theta$ lies in
set $\mathcal{H}$, and hence the true distribution $p(x|\theta)$
is not known. In this case, we cannot select the true parameter distribution,
and we need instead to choose a generally different distribution $q(x)$
to define a compression scheme. With a given distribution $q(x)$,
the average coding length is given by $-\sum_{x}p(x|\theta)\log q(x)$.
Therefore, the choice of a generally incorrect distribution $q(x)$
entails a redundancy of 
\begin{equation}
\Delta R(q(x),\theta)=-\sum_{x}p(x|\theta)\log q(x)-H(p(x|\theta))\geq0\label{eq:redCh5app}
\end{equation}
bits. 

The redundancy $\Delta R(q(x),\theta)$ in (\ref{eq:redCh5app}) depends
on the true value of $\theta$. Since the latter is not known, this
quantity cannot be computed. We can instead obtain a computable metric
by maximizing over all values of $\theta\in\mathcal{H}$, which yields
the worst-case redundancy 
\begin{equation}
\Delta R(q(x),\mathcal{H})=\underset{\theta\in\mathcal{H}}{\textrm{ }\max}\Delta R(q(x),\theta).\label{eq:redCh5app-1}
\end{equation}
This quantity can be minimized over $q(x)$ yielding the so-called
\emph{minimax redundancy:}
\begin{align}
\Delta R(\mathcal{H}) & =\underset{q(x)}{\min}\Delta R(q(x),\mathcal{H})\\
 & =\underset{q(x)}{\min}\underset{\theta}{\textrm{ }\max}-\sum_{x}p(x|\theta)\log q(x)-H(p(x|\theta))\\
 & =\underset{q(x)}{\min}\underset{\theta}{\textrm{ }\max}\sum_{x}p(x|\theta)\log\frac{p(x|\theta)}{q(x)}.
\end{align}

The minimax redundancy can be taken as a measure of the capacity of
model $\mathcal{H},$ since a richer model tends to yield a larger
$\Delta R(\mathcal{H}).$ In fact, for a richer model, it is more
difficult to find a representative distribution $q(x)$ that yields
an average coding length close to the minimum $H(p(x|\theta))$ for
all values of $\theta$.

It turns out that the minimax redundancy equals the capacity $C(p(x|\theta))$
of the channel $p(x|\theta)$, which is defined as $C(p(x|\theta))=\max_{p(\theta)}I(\mathrm{x};\text{\ensuremath{\mathrm{\theta}}})$
\cite{Cover}. This is shown by the following sequence of equalities:\begin{subequations}
\begin{align}
\Delta R(\mathcal{H})= & \underset{q(x)}{\min}\underset{p(\theta)}{\textrm{ }\max}\sum_{x}\sum_{\theta}p(\theta)p(x|\theta)\log\frac{p(x|\theta)}{q(x)}\\
= & \underset{p(\theta)}{\max}\underset{q(x)}{\textrm{ }\min}\sum_{x}\sum_{\theta}p(\theta)p(x|\theta)\log\frac{p(x|\theta)}{q(x)}\\
= & \underset{p(\theta)}{\max}\sum_{x}\sum_{\theta}p(\theta)p(x|\theta)\log\frac{p(x|\theta)}{p(x)}\\
= & C(p(x|\theta)),
\end{align}
\end{subequations}where the first equality follows since the average
of a set of numbers is no larger than any of the numbers; the second
is a consequence of the minimax theorem since the term $\sum_{x}\sum_{\theta}p(\theta)p(x|\theta)\log(p(x|\theta)/q(x))$
is convex in $q(x)$ and concave in $p(\theta)$; and the third equality
follows by Gibbs' inequality (see Sec. \ref{sec:Information-Theoretic-Metrics}
and (\ref{eq:Shannon as log loss}) in Appendix A). As a final note,
the mutual information $I(\mathrm{x};\text{\ensuremath{\mathrm{\theta}}})$
between model parameter and data also plays a central role in obtaining
bounds on the performance of estimation \cite{loh2017lower}.

\part{Unsupervised Learning}

\chapter{Unsupervised Learning}

Unlike supervised learning, unsupervised learning tasks operate over
unlabelled data sets. Apart from this general statement, unsupervised
learning is more loosely defined than supervised learning, and it
also lacks a strong theoretical framework to mirror the PAC learning
theory covered in the previous chapter. Nevertheless, it is widely
expected that future breakthroughs in machine learning will come mainly
from advances in the theory and design of unsupervised learning algorithms.
This is due to the availability of huge repositories of unlabelled
data, as well as to the broader applicability of learning tasks whereby
the machine learns, as it were, without supervision or feedback. Unsupervised
learning is also considered by some as the key to the development
of general, as opposed to task, specific AI \cite{spectrumlc} (see
also \cite{Tegmark,Levesque17} for more on general AI). 

Generally speaking, unsupervised learning algorithms aim at learning
some properties of interest of the mechanism underlying the generation
of the data. In this sense, unsupervised learning concerns the study
of generative models, although, as we will see, this statement comes
with some caveats. A common aspect of many models used for unsupervised
learning is the presence of hidden, or latent, variables that help
explain the structure of the data. 

This chapter starts by discussing applications of unsupervised learning,
and by providing a description of the well-known $K$-means algorithm.
It then covers directed and undirected generative probabilistic models
for unsupervised learning. As we will detail, these models posit different
types of statistical dependence relations between hidden and measured
variables. Discriminative models, which capture the dependence of
hidden variables on observed variables, as well as autoencoders, which
combine discriminative and generative models, are also introduced.
The chapter is concluded with a discussion of a different type of
learning algorithm that may be considered as unsupervised, namely
PageRank, which is included due to its practical relevance. 

\section{Unsupervised Learning}

\textbf{Defining unsupervised learning.} A general, and rather imprecise,
definition of unsupervised learning tasks is the following. Taking
a frequentist viewpoint, we are given a data set $\mathcal{D}$ consisting
of $N$ i.i.d. unlabelled observations $\mathrm{x}_{n}\in\boldsymbol{\mathbb{R}}^{D}$.
These are assumed to be drawn i.i.d. from an unknown true distribution
as 
\begin{equation}
\mathcal{D}=\{\mathrm{x}_{n}\}_{n=1}^{N}\underset{\text{i.i.d.}}{\sim}p(x).\label{eq:iid data Ch6}
\end{equation}
The goal is to learn some useful properties of the distribution $p(x)$,
where the properties of interest depend on the specific application.
While this definition is general enough to include also the estimation
problems studied in Chapter 3, as mentioned, unsupervised learning
problems are typically characterized by the presence of \emph{hidden}
or\emph{ latent} variables. Notable examples include the following.

\indent $\bullet$\emph{ Density estimation}: Density estimation aims
at learning directly a good approximation of the distribution $p(x)$,
e.g., for use in plug-in estimators \cite{lehmann2006theory}, to
design compression algorithms (see Sec. \ref{sec:Minimum-Description-Length}),
or to detect outliers \cite{outliers}. 

\indent $\bullet$\emph{ Clustering}: Clustering assumes the presence
of an unobserved label $z_{n}$ associated to each data point $x_{n}$,
and the goal is that of recovering the labels $z_{n}$ for all points
in the data set $\mathcal{D}.$ For example, one may wish to cluster
a set $\mathcal{D}$ of text documents according to their topics,
by modelling the latter as an unobserved label $z_{n}$. Broadly speaking,
this requires to group together documents that are similar according
to some metric. It is important at this point to emphasize the distinction
between classification and clustering: While the former assumes the
availability of a labelled set of training examples and evaluates
its (generalization) performance on a separate set of unlabelled examples,
the latter works with a single, unlabelled, set of examples. The different
notation used for the labels \textendash{} $z_{n}$ in lieu of $t_{n}$
\textendash{} is meant to provide a reminder of this key difference.

\indent $\bullet$\emph{ Dimensionality reduction} \emph{and representation}:
Given the set $\mathcal{D}$, we would like to represent the data
points $x_{n}\in\mathcal{D}$ in a space of lower dimensionality.
This makes it possible to highlight independent explanatory factors,
and/or to ease visualization and interpretation \cite{visualization},
e.g., for text analysis via vector embedding (see, e.g., \cite{Rudolphemb}). 

\indent $\bullet$\emph{ Feature extraction}: Feature extraction is
the task of deriving functions of the data points $x_{n}$ that provide
useful lower-dimensional inputs for tasks such as supervised learning.
The extracted features are unobserved, and hence latent, variables.
As an example, the hidden layer of a deep neural network extract features
from the data for use by the output layer (see Sec. \ref{subsec:Non-linear-Classification}). 

\indent $\bullet$\emph{ Generation of new samples}: The goal here
is to learn a machine that is able to produce samples that are approximately
distributed according to the true distribution $p(x)$. For example,
in computer graphics for filmmaking or gaming, one may want to train
a software that is able to produce artificial scenes based on a given
description.

The variety of tasks and the difficulty in providing formal definitions,
e.g., on the realism of an artificially generated image, make unsupervised
learning, at least in its current state, a less formal field than
supervised learning. Often, loss criteria in unsupervised learning
measure the divergence between the learned model and the empirical
data distribution, but there are important exceptions, as we will
see.

\begin{figure}[H] 
\centering\includegraphics[scale=0.55]{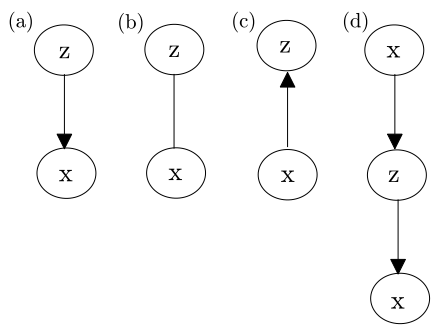}
\caption{(a) Directed generative models; (b) Undirected generative models; (c) Discriminative models;  (d) Autoencoders.} \label{fig:FigmodelsCh6} \end{figure}

\textbf{Models.} We now review the type of models that can be used
to tackle unsupervised learning problems. The models will be further
discussed in the subsequent sections of this chapter.

\indent $\bullet$\textit{ Directed generative models:} Directed generative
models are \emph{mixture} models in which the distribution $p(x|\theta)$
of the data is defined by a parametrized prior $p(z|\theta)$ of the
latent variables $\mathrm{z}$ and by a parametrized conditional distribution
$p(x|z,\theta)$ that defines the relationship between latent and
observed variables. Accordingly, the distribution can be expressed,
for discrete latent variables $\mathrm{\ensuremath{z}}$, as
\begin{equation}
p(x|\theta)=\sum_{z}p(z|\theta)p(x|z,\theta).\label{eq:directed UL}
\end{equation}
A similar expression applies to continuous hidden variables with an
integral in lieu of the sum. Directed models are suitable to capture
the cause-effect relationships between $\mathrm{z}$ and $\mathrm{x}$.
A BN describing directed generative models is shown in Fig. \ref{fig:FigmodelsCh6}(a).
Graphical models, including BNs, will be covered in detail in the
next chapter. Examples of directed generative models include the mixture
of Gaussians model, and the so-called \emph{likelihood-free models},
in which the conditional distribution $p(x|z,\theta)$ is implemented
by a deterministic transformation, most typically a multi-layer network. 

\indent $\bullet$\textit{ Undirected generative models:} Undirected
models parametrize directly the joint distribution of the observed
variables\textbf{ $\mathrm{x}$} and the hidden variables\textbf{
$\mathrm{z}$} as $p(x,z|\theta)$, and accordingly write the distribution
of the data as 
\begin{equation}
p(x|\theta)=\sum_{z}p(x,z|\theta).\label{eq:undirected UL}
\end{equation}
Unlike directed models, undirected models capture the affinity, or
compatibility, of given configurations of values for \textbf{$\mathrm{z}$}
and $\mathrm{x}$. An Markov Random Field (MRF) describing undirected
generative models is shown in Fig. \ref{fig:FigmodelsCh6}(b) (see
next chapter). A prominent example is given by RBMs.

\indent $\bullet$\emph{ Discriminative models}: Discriminative models
attempt to directly learn an \emph{encoding} probabilistic mapping
$p(z|x,\theta)$ between the data point $\mathrm{x}$ and a representation
$\mathrm{z}.$ This is represented by the BN in Fig. \ref{fig:FigmodelsCh6}(c).

\indent $\bullet$\emph{ Autoencoders}: As seen in Fig. \ref{fig:FigmodelsCh6}(d),
autoencoders compose a parametrized discriminative model $p(z|x,\theta)$,
which produces the hidden variables $\mathrm{z}$ from the data $\mathrm{x}$,
with a parametrized generative model $p(x|z,\theta).$ The former
is known as encoder, while the latter as decoder. Accordingly, the
latent variables are also referred to as the \emph{code}. The most
typical implementations use parameterized deterministic functions
$z=F_{\theta}(x)$ and $x=G_{\theta}(z)$ in lieu of the more general
probabilistic models $p(z|x,\theta)$ and $p(x|z,\theta)$, respectively.
As we will see, autoencoders are trained to reproduce the data $x$
at the output, turning the unsupervised problem into a supervised
one with ``labels'' given by the data point $x$ itself. 

\section{$K$-Means Clustering}

We start by reviewing the well-known $K$-means clustering algorithm.
The purpose here is to emphasize an algorithmic structure, namely
the Expectation Maximization (EM) algorithm, that is conceptually
at the core of many unsupervised learning algorithms. 

The problem is one of multi-cluster clustering: Given a data set $\mathcal{D}=\{x_{n}\}_{n=1}^{N}$,
we would like to assign every vector $x_{n}\in\mathbb{R}^{D}$ to
one of $K$ clusters. Cluster indices are encoded by categorical variables
$z_{n}$ via one-hot encoding (see Chapter 3). Accordingly, we write
the $k$th component of vector $z_{n}$ as $z_{kn}=1$ if $x_{n}$
is assigned to cluster $k$, while we write $z_{kn}=0$ otherwise.
It is emphasized that the labels are not given for any of the examples
in $\mathcal{D}$. Therefore, the algorithm should discern some regularity
in the data in order to divide the data set into $K$ classes. 

$K$-means is a heuristic method that attempts to cluster together
points that are mutually close in Euclidean distance. To this end,
$K$-means assigns all points in the same cluster a given ``prototype''
representative vector $\mu_{k}$. This vector can be thought of in
terms of quantization: all points within a given cluster can be quantized
to the prototype $\mu_{k}$ with minimal quadratic loss. This is formalized
by the following optimization problem over the cluster assignment
variables $z_{n}$ and the cluster representatives $\mu_{k}$:
\begin{equation}
\underset{\{z_{n}\},\{\mu_{k}\}}{\textrm{min}}{\displaystyle \sum_{n=1}^{N}}\sum_{k=1}^{K}z_{kn}d(x_{n},\mathbf{\mu}_{k}),\label{eq:K means problem}
\end{equation}
where $d(x,\mathbf{\mu})=\left\Vert x-\mathbf{\mu}\right\Vert ^{2}$
is the squared Euclidean distance. When using a general distance metric,
the approach to be described is instead known as the $K$-medoids
algorithm. In this regard, we note, for instance, that it is possible
to apply clustering also to discrete data as long as the distance
$d$ is properly defined \textendash{} typically by a matrix of pairwise
dissimilarities.

The $K$-means algorithm performs alternatively optimization of the
cluster assignment variables $z_{n}$ and of the cluster representatives
$\mu_{k}$ as follows:

\indent $\bullet$ Initialize cluster representatives $\{\mu_{k}^{\textrm{old}}\}.$

\indent $\bullet$\emph{ Expectation step, or E step}: For fixed vectors
$\{\mu_{k}^{\textrm{old}}\},$ solve problem (\ref{eq:K means problem})
over the cluster assignment $\{z_{n}\}$:
\begin{equation}
z_{kn}^{\textrm{new}}=\begin{cases}
1 & \textrm{for }k=\text{\ensuremath{\textrm{argmin}_{j}} }d(x_{n},\mu{}_{j}^{\textrm{old}})\\
0 & \text{otherwise}
\end{cases}.
\end{equation}
Accordingly, each training point is assigned to the cluster with the
closest prototype. Note that this step generally requires the computation
of $K$ distances for each data point $x_{n}$.

\indent $\bullet$\emph{ Maximization step, or M step}: For fixed
vectors $\{z_{n}\}$, solve problem (\ref{eq:K means problem}) over
the cluster representatives $\{\mu_{k}\}$. Imposing the optimality
condition that the gradient of the objective function in (\ref{eq:K means problem})
be zero, we obtain
\begin{equation}
\mu_{k}^{\text{new}}=\frac{\sum_{n=1}^{N}z_{kn}^{\text{new}}x_{n}}{\sum_{n=1}^{N}z_{kn}^{\text{new}}}.
\end{equation}
The new cluster representative $\mu_{k}^{\text{new}}$ for each cluster
$k$ is the mean of the data points assigned to cluster $k$.

\indent $\bullet$ If a convergence criterion is not satisfied, set
\{$\mu_{k}^{\text{old}}\}\leftarrow\{\mu_{k}^{\text{new}}\}$ and
return to the E step.

Since both E step and M step minimize the objective function in (\ref{eq:K means problem}),
respectively over the cluster assignment variables $z_{n}$ and the
cluster representatives $\mu_{k}$, the value of the objective function
is non-decreasing across the iterations. This ensures convergence.
Illustrations of convergence and examples can be found in \cite[Chapter 9]{Bishop}.
As a note, the algorithm is also known as Lloyd-Max quantization \cite{Gray}.

At a high level, $K$-means alternates between: (\emph{i}) making
inferences about the hidden variables \{$z_{n}$\} based on the current
model defined by the representatives \{$\mu_{k}$\} in the E step;
and (\emph{ii}) updating the model \{$\mu_{k}$\} to match the data
\{$x_{n}$\} and the inferred variables \{$z_{n}$\} in the M step.
We will see that a similar algorithmic structure is applied by many
unsupervised learning algorithms. 

Before we move on to discussing more general solutions, it is worth
spending a few words on the problem of selecting the number $K$ of
clusters. A first possibility is to add or remove clusters until certain
heuristic criteria are satisfied, such as the ``purity'' of clusters.
A second approach is hierarchical clustering, whereby one builds a
tree, known as dendrogram, that includes clustering solutions with
an increasing number of clusters as one moves away from the root (see,
e.g., \cite{Hastie}). Yet another solution is to let $K$ be selected
automatically by adopting a non-parametric Bayesian approach via a
Dirichlet process prior \cite{Murphy}.

\section{ML, ELBO and EM\label{sec:ML,-ELBO-and-Ch6}}

In this section, we discuss two key technical tools that are extensively
used in tackling unsupervised learning problems, namely the Evidence
Lower BOund (ELBO) and the EM algorithm. The starting point is the
fundamental problem of learning a probabilistic, directed or undirected,
model $p(x|\theta)$ from the data using ML. We will discuss later
how to learn discriminative models and autoencoders.

\subsection{ML Learning}

From Sec. \ref{sec:Information-Theoretic-Metrics}, we know that ML,
asymptotically in $N$, tends to minimize a KL divergence between
the true distribution $p(x)$ and the selected hypothesis in the model.
This is a criterion that is well suited for many of the unsupervised
learning tasks mentioned above, such as density estimation or generation
of new samples, and it is hence useful to start the discussion with
ML learning. 

Before we do that, a few remarks are in order. First, it is often
useful to choose divergences other than KL, which are tailored to
the specific application of interest \cite{Wasserstein}. We will
further discuss this aspect in Sec. \ref{subsec:GAN}. Second, when
the goal is representation learning, the machine aims at obtaining
useful features $z$. Hence, minimizing the KL divergence to match
the true distribution $p(x)$ does not directly address the objective
of representation learning, unless appropriate restrictions are imposed
on the model $p(x|z,\theta).$ In fact, if the generative model $p(x|z,\theta)$
is too powerful, then it can disregard the features $\mathrm{z}$
and still obtain a high likelihood for the data (see, e.g., \cite{blogMLuns}).
We will get back to this point in Sec. \ref{sec:Discriminative-Models-Ch6}.
Finally, obtaining ML solutions is often impractical, particularly
in large models. Nevertheless, understanding the ML problem allows
one to better gauge the impact of the approximations and simplifications
made to obtain efficient algorithms. 

To proceed, we consider probabilistic, directed or undirected, model,
and focus, for the purpose of simplifying the notation, on a data
set with a single data point ($N=1$). The extension to a data set
with an arbitrary number $N$ of points only requires adding an outer
sum over the sample index to the LL function. We also concentrate
on discrete hidden rvs, and the generalization to continuous rvs is
obtained by substituting sums with suitable integrals. The ML problem
can be written as the maximization of the LL function as
\begin{equation}
\underset{\theta}{\text{max }}\ln p(x|\theta)=\ln\left(\sum_{z}p(x,z|\theta)\right),\label{eq:ML unsupervised Ch6}
\end{equation}
where \textbf{$x$} denotes the data and \textbf{$z$} the hidden
or latent variables. Note the marginalization with respect to the
hidden variables in (\ref{eq:ML unsupervised Ch6}). This problem
should be contrasted with the supervised learning ML problem obtained
when both $x$ and $z$ are observed, namely
\begin{equation}
\underset{\theta}{\text{max }}\ln p(x,z|\theta).\label{eq:ML supervised Ch6}
\end{equation}

\begin{figure}[H] 
\centering\includegraphics[width=0.75\textwidth]{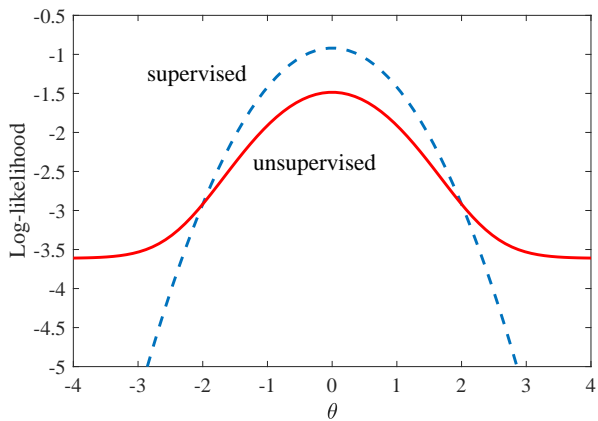}
\caption{Illustration of LL functions for supervised and unsupervised learning in a mixture of Gaussians model (Example \ref{ex:unsGaussCh5}).} \label{fig:FigMLCh6} \end{figure}

\begin{example}\label{ex:unsGaussCh5}Consider a directed generative
Bernoulli-Gaussian model characterized as\begin{subequations}\label{eq:BernGaussCh6}
\begin{align}
 & \mathrm{z}\sim\textrm{Bern}(0.5)\\
 & \mathrm{x}|\mathrm{z}=0\sim\mathcal{N}(2,1)\\
\textrm{} & \mathrm{x}|\mathrm{z}=1\sim\mathcal{N}(\theta,1).
\end{align}
\end{subequations}This corresponds to a mixture of Gaussians model,
in which the only parameter $\theta$ is the mean of one of the two
Gaussian components. Assume that we observe $x=0$. Consider first
the supervised learning case, where we assume that we also measure
$z=1$. In this case, the LL function in (\ref{eq:ML supervised Ch6})
is $\ln p(x=0,z=1|\theta)=\ln\mathcal{N}(x|\theta,1)+\textrm{\ensuremath{\ln(0.5)}}.$
In contrast, with unsupervised learning, the LL function in (\ref{eq:ML unsupervised Ch6})
is $\ln p(x=0|\theta)=\ln(0.5\mathcal{N}(0|2,1)+0.5\mathcal{N}(0|\theta,1))$.

The LL functions are shown in Fig. \ref{fig:FigMLCh6}. Unlike supervised
learning, the LL for unsupervised learning is seen to be \emph{non-concave}.
In this example, this is a consequence of the fact that, when $\theta$
is sufficiently large in absolute value, the probability of the data
$x$ resulting from the fixed Gaussian distribution centered at $x=2$
makes the contribution of the Gaussian centered $\theta$ increasingly
irrelevant. \end{example}

Solving problem (\ref{eq:ML unsupervised Ch6}) has two additional
complications as compared to the supervised learning counterpart (\ref{eq:ML supervised Ch6}).
The first issue was highlighted in the previous example: Even for
models in which the ML supervised learning problem is convex, the
LL is generally non-concave when the variables $\mathrm{z}$ are hidden,
which precludes the use of convex optimization algorithms. Barring
the use of often impractical global optimization algorithms, in general,
non-convex problems cannot be solved exactly. Rather, the best one
can hope for with standard local optimization schemes, such as SGD,
is obtaining stationary points or locally optimal points \cite{Boyd}.
In practice, this problem may not be critical since non-convexity
is not by itself a cause of poor learning performance. For instance,
as seen in Chapter 4, beyond-GLM supervised learning methods, including
deep neural networks, solve non-convex problems. 

The second complication is the need to sum \textendash{} or integrate
\textendash{} over the hidden variables in order to evaluate the LL.
This step is complicated by the fact that the distribution of the
hidden variables needs to be learned and is hence unknown. This is
a significant obstacle to the development of efficient learning algorithms,
and it generally needs to be addressed in order to make learning feasible.
In the rest of this section, we describe two technical tools that
are useful to tackle these problems. Chapter 8 will develop more complex
solutions for issues arising in the presence of large latent spaces
in which marginalization is not viable.

\subsection{ELBO}

Many methods to tackle the ML problem (\ref{eq:ML unsupervised Ch6})
are based on the maximization of the ELBO. These techniques include
the EM algorithm and some of the variational inference algorithms
to be discussed in Chapter 8. The key element in the development of
the ELBO is the introduction of an auxiliary distribution $q(z)$
on the latent variables. This is referred to as the \emph{variational
distribution} or the \emph{variational posterior }for reasons that\emph{
}will be made clear later. As we will see, computation of the ELBO
requires an average over distribution $q(z)$, which can be fixed
independently of the model parameters. This solves the key problem
identified above of averaging over the parameter-dependent marginal
of the hidden variables.

\begin{definition}For any fixed value $x$ and any distribution $q(z)$
on the latent variables $\mathrm{z}$ (possibly dependent on $x$),
the ELBO $\mathcal{L\mathrm{(\text{\ensuremath{q}},\theta)}}$ is
defined in one of the following equivalent forms
\begin{align}
\mathcal{L}(q,\theta)= & \textrm{E}_{\mathrm{z}\sim q(z)}[\underbrace{\ln p(x,\mathrm{z}|\theta)-\ln q(\mathrm{z})}_{\textrm{learning signal}}]\\
= & \underbrace{\textrm{E}_{\mathrm{z}\sim q(z)}[\ln p(x,\mathrm{z}|\theta)]}_{\text{negative energy }}+\underbrace{H(q)}_{\text{entropy }}\label{eq:ELBO1}\\
= & \underbrace{\textrm{E}_{\mathrm{z}\sim q(z)}[\ln p(x|\mathrm{z},\theta)]}_{\textrm{cross-entropy}}-\underbrace{\text{KL}\left(q(z)||p(z|\theta)\right)}_{\textrm{variational regularization}}\label{eq:ELBO2}\\
= & -\text{KL}\left(q(z)||p(x,z|\theta)\right)\label{eq:ELBO3}\\
= & \ln p(x|\theta)-\text{KL}\left(q(z)||p(z|x,\theta)\right),\label{eq:ELBO4}
\end{align}
where we have identified some terms using common terminology that
will be clarified in the following, and, in (\ref{eq:ELBO3}), we
have used the convention of defining $\text{KL}(p||q)$ even when
$q$ is not normalized (see Sec. \ref{sec:Information-Theoretic-Metrics}).
\end{definition}

The equivalence between the three forms of the ELBO can be easily
checked. The form (\ref{eq:ELBO1}) justifies the definition of the
negative of the ELBO as \emph{variational free energy or Gibbs free
energy}, which is the difference of energy and entropy. This form
is particularly useful for undirected models in which one specifies
directly the joint distribution $p(x,z|\theta)$, such as energy-based
models, while the form (\ref{eq:ELBO2}) is especially well suited
for directed models that account for the discriminative distribution
$p(x|z,\theta)$, such as for deep neural networks \cite{blundell2015weight}.
For both forms the first term can be interpreted as a cross-entropy
loss. The form (\ref{eq:ELBO3}) is more compact, and suggests that,
as we will formalize below, the ELBO is maximized when $q(z)$ is
selected to match the model distribution. The last form yields terms
that are generally not easily computable, but it illuminates the relationship
between the LL function and the ELBO, as we discuss next. 

The following theorem describes the defining property of the ELBO
as well as another important property. Taken together, these features
make the ELBO uniquely suited for the development of algorithmic solutions
for problem (\ref{eq:ML unsupervised Ch6}). 

\begin{theorem}\label{thm:ELBO}The ELBO is a global lower bound on
the LL function, that is,
\begin{equation}
\text{\ensuremath{\ln p(x|}\ensuremath{\theta}) }\geq\mathcal{L}(q,\theta),\label{eq:inequality ELBO}
\end{equation}
where equality holds at a value $\theta_{0}$ if and only if the distribution
$q(z)$ satisfies $q(z)=p(z|x,\theta_{0})$. Furthermore, the ELBO
is concave in $q(z)$ for a fixed $\theta$; and, if $\ln p(x,z|\theta)$
is concave in $\theta$, it is also concave in $\theta$ for a fixed
$q(z)$. \end{theorem}

\begin{proof}The first part of the theorem follows immediately from
the form (\ref{eq:ELBO4}), which we can rewrite as
\begin{equation}
\ln p(x|\theta)=\text{\ensuremath{\mathcal{L}}\ensuremath{(q,\ensuremath{\theta})+}KL}\left(q(z)||p(z|x,\theta)\right),\label{eq:equality ELBO}
\end{equation}
and from Gibbs' inequality. In fact, the latter implies that the KL
divergence $\text{KL}\left(q(z)||p(z|x,\theta)\right)$ is non-negative
and equal to zero if and only if the two distributions in the argument
are equal. The concavity of the ELBO can be easily checked using standard
properties of convex functions \cite{Boyd}. As a note, an alternative
proof of the first part of the theorem can be provided via the importance
sampling trick and Jensen's inequality. In fact, we can write \begin{subequations}\label{eq:jensenCh6}
\begin{align}
\ln p(x|\theta)= & \ln\left(\sum_{z}p(x,z|\theta)\right)\\
= & \ln\left(\sum_{z}q(z)\frac{p(x,z|\theta)}{q(z)}\right)\\
\underset{}{\geq} & \sum_{z}q(z)\ln\left(\frac{p(x,z|\theta)}{q(z)}\right)=\mathcal{L}(q,\theta),
\end{align}
\end{subequations}where the first equality is just obtained by multiplying
and dividing by $q(z)$ \textendash{} the importance sampling trick
\textendash{} and the last step is a consequence of Jensen's inequality.
We recall that Jensen's inequality says that for any concave function
$f(x)$ \textendash{} here $f(x)=\ln(x)$ \textendash{} we have the
inequality $\textrm{E}[f(\mathrm{x})]\leq f(\textrm{E}[\mathrm{x}])$.
\end{proof}

We illustrate the just described properties of the ELBO via the following
example.

\begin{figure}[H] 
\centering\includegraphics[scale=0.6]{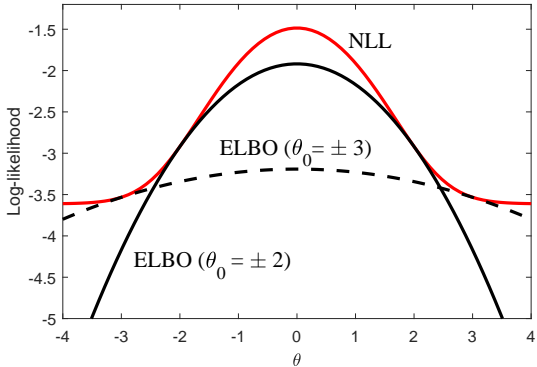}
\caption{Illustration of the ELBO for two different choices of the variational distribution that are tight at different values $\theta_0$ of the parameter.} \label{fig:FigELBOCh6} \end{figure}

\begin{example}Consider again the directed generative Bernoulli-Gaussian
model (\ref{eq:BernGaussCh6}). The posterior distribution of the
latent variable given an observation $x$ is given as
\begin{equation}
p(z=1|x=0,\theta)=\frac{\mathcal{N}(0|\theta,1)}{\mathcal{N}(0|2,1)+\mathcal{N}(0|\theta,1)}.
\end{equation}
Fix a parametrized variational distribution $q(z|\varphi)=\textrm{Bern\ensuremath{(z|\varphi)}}$.
Using (\ref{eq:ELBO1}), the ELBO is then given as 
\begin{equation}
\mathcal{L}(q,\theta)=\varphi(\ln(\mathcal{N}(0|\theta,1)+\ln(0.5))+(1-\varphi)(\ln(\mathcal{N}(0|2,1)+\ln(0.5))+H(q).
\end{equation}
The theorem above says that, given any value of $\varphi$, the ELBO
is a lower bound on the LL function, uniformly for all values of $\theta$.
Furthermore, this bound is tight. i.e., it equals the LL function,
at all values $\theta_{0}$ for which the selected variational distribution
$q(z|\varphi)$ equals the posterior of the hidden variables, that
is, for which we have $\varphi=p(z=1|x=0,\theta_{0})$. This is shown
in Fig. \ref{fig:FigELBOCh6}, where we plot the LL and the ELBO.
We see that indeed the ELBO is a uniform lower bound on the LL, which
is tight for specific values $\theta_{0}$ of the parameter $\theta$.
Reflecting the concavity property of the ELBO in the theorem, the
ELBO is also seen to be a concave function of the parameter $\theta$.
\end{example}

The lower bound property of the ELBO makes it useful not only for
the purpose of ML optimization but also as an estimate of LL, and
hence of how well the model fits the data, for the goal of model selection.
Furthermore, the ELBO can be computed analytically in special cases
for exponential models \cite{Wainwright,2017arXiv171105597Z}.

The ELBO can be generalized as the \emph{multi-sample ELBO} \cite{multisampleELBO}:
\begin{equation}
\ln p(x|\theta)\geq\textrm{E}_{\mathrm{z}_{1},...,\mathrm{z}_{K}\underset{\textrm{i.i.d.}}{\sim}q(z)}\left[\ln\left(\frac{1}{K}\sum_{k=1}^{K}\frac{p(x,\mathrm{z}_{k}|\theta)}{q(\mathrm{z}_{k})}\right)\right].
\end{equation}
The proof of this inequality follows in the same way as for the ELBO.
This bound has the advantage that, as $K$ grows large, it tends to
become increasingly accurate. In fact, as $K\rightarrow\infty$, by
the law of large numbers, we have the limit $K^{-1}\sum_{k=1}^{K}p(x,\mathrm{z}_{k}|\theta)/q(\mathrm{z}_{k})\rightarrow\sum_{z}p(x,z|\theta)$
with probability one. 

\textbf{ELBO and Bayesian inference.} In summary, for a given variational
distribution $q(z),$ the ELBO provides an upper bound on the LL function,
or equivalently a lower bound on the NLL function. This bound is tight
for values of the parameter vectors $\theta$ at which we have the
equality $q(z)=p(z|x,\theta)$. As such, given a certain value $\theta$
of the parameter vector, the variational distribution $q(z)$ that
provides the tightest bound is the posterior distribution $q(z)=p(z|x,\theta)$,
at which the KL divergence in (\ref{eq:equality ELBO}) vanishes.
That is, in order to obtain the tightest ELBO, one needs to solve
the Bayesian inference problem of computing the posterior $p(z|x,\theta)$
of the hidden variables for the given value $\theta$. This property
can be stated for reference as follows
\begin{equation}
\textrm{arg}\underset{q(z)}{\textrm{max }}\mathcal{L}(q,\theta)=p(z|x,\theta).\label{eq:EM property max}
\end{equation}

\textbf{Gradients of the LL and of the ELBO over the model parameters.$^{*}$}
Under suitable differentiability assumptions, the gradient of the
ELBO at the value $\theta_{0}$ in which the ELBO is tight coincides
with the gradient of the LL, i.e.,
\begin{align}
\nabla_{\theta}\ln p(x|\theta)|_{\theta=\theta_{0}} & =\nabla_{\theta}\mathcal{L}\left(p(z|x,\theta_{0}),\theta\right)|_{\theta=\theta_{0}}\nonumber \\
 & =\textrm{E}_{\mathrm{z}\sim p(z|x,\theta_{0})}\left[\nabla_{\theta}\ln p(x,\mathrm{z}|\theta)|_{\theta=\theta_{0}}\right].\label{eq:gradient unsupervised}
\end{align}
This is also suggested by the curves in Fig. \ref{fig:FigELBOCh6}.
We will see with an example in Sec. \ref{subsec:Restricted-Boltzmann-Machines}
that this formula is extremely useful when the gradient for the complete
log-likelihood $\nabla_{\theta}\ln p(x,\mathrm{z}|\theta)|_{\theta=\theta_{0}}$
can be easily computed, such as for exponential family models.

\textbf{Other global lower bounds on the likelihood.$^{*}$} Revisiting
the proof of Theorem \ref{thm:ELBO}, it can be concluded that the
following general family of lower bounds 

\begin{equation}
\text{\ensuremath{f(p(x|}\ensuremath{\theta})) }\geq\textrm{E}_{z\sim q(z)}\left[f\left(\frac{p(x,z|\theta)}{q(z)}\right)\right],\label{eq:inequality ELBO-1}
\end{equation}
for any concave function $f$. While the ELBO equals the negative
of a KL divergence between variational distribution and the true distribution,
as seen in (\ref{eq:ELBO3}), this representation yields more general
divergence measures, such as the $\alpha$-divergence to be discussed
in Chapter 8 \cite{LiTurner,2017arXiv170907433B,2017arXiv171105597Z}. 

\subsection{EM Algorithm}

As mentioned, a large number of practical schemes for unsupervised
learning using directed and undirected generative models are based
on the maximization of ELBO $\mathcal{L}(q,\theta)$ in lieu of the
LL function. As seen, this maximization has the key advantage that
the ELBO is a concave function of the parameters $\theta$. Furthermore,
the lower bound property (\ref{eq:inequality ELBO}) ensures that,
if the ELBO is tight at a value $\theta_{0}$, the result of the optimization
of the ELBO must necessarily yield a LL value that is no smaller than
$\textrm{ln}p(x|\theta_{0})$. This observation is leveraged by the
EM algorithm to obtain a procedure that is guaranteed to converge
to a stationary point of the original ML problem (\ref{eq:ML unsupervised Ch6}).

The EM algorithm is described in Table 6.1.

\begin{table} 

\indent $\bullet$ \textbf{Initialize} parameter vector $\theta^{\textrm{old}}.$

\indent $\bullet$\emph{ }\textbf{E step:} For fixed parameter vector
$\theta^{\textrm{old}}$, maximize the ELBO over the variational distribution
$q$, i.e., solve problem $\underset{q}{\text{max }}\mathcal{L}(q,\theta^{\textrm{old}})$,
which, by (\ref{eq:EM property max}), yields the new distribution
\begin{equation}
q^{\textrm{new}}(z)=p(z|x,\theta^{\textrm{old}}).\label{eq:EM E step}
\end{equation}

\indent $\bullet$\emph{ }\textbf{M step:} For fixed variational distribution
$q^{\textrm{new}}(z)$, maximize the ELBO over the parameter vector
$\theta$, i.e., solve problem $\underset{\theta}{\text{max }}\mathcal{L}(q^{\textrm{new}},\theta)$.
This convex problem can be equivalently written as the maximization
of the negative energy
\begin{align}
\underset{\theta}{\text{max }}Q(\theta,\theta^{\textrm{old}})= & \textrm{E}_{\mathrm{z}\sim p(z|x,\theta^{\textrm{old}})}\left[\ln p(x,\mathrm{z}|\theta)\right].\label{eq:EM M step}
\end{align}

\indent $\bullet$ If a convergence criterion is not satisfied, set
$\theta^{\textrm{new}}\leftarrow\theta^{\textrm{old}}$and return
to the E step.

 \caption{EM algorithm.}\end{table}

In many problems of interest, the model $p(x,\mathrm{z}|\theta)$
can be taken to be either the product of a prior and a likelihood
from the exponential family for directed models, or directly a member
of the exponential family for undirected models. In these cases, the
problem (\ref{eq:EM M step}) solved in the M step will be seen below
via an example to reduce to the corresponding supervised learning
problem, with the caveat that the sufficient statistics are averaged
over the posterior distribution $p(z|x,\theta^{\textrm{old}})$. 

\begin{figure}[H] 
\centering\includegraphics[scale=0.65]{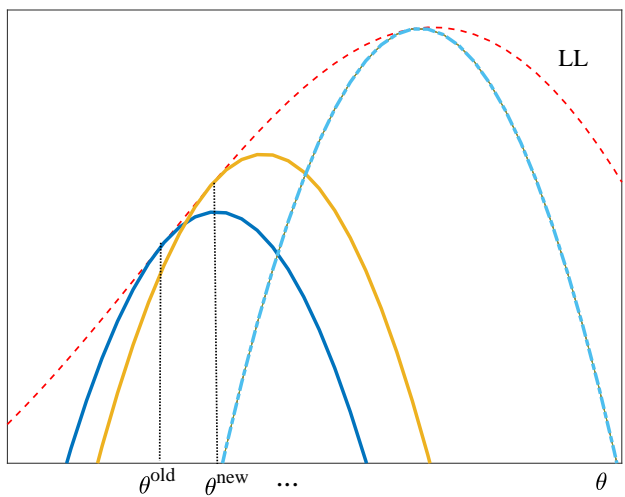}
\caption{Illustration of the EM algorithm. The dashed line is the LL function. The solid lines represent the ELBOs corresponding to the first two steps of the algorithm, while the dashed-dotted line is the ELBO after a number of iterations. At each step, EM maximizes the ELBO, which is a lower bound on the LL function. EM is an instance of the more general MM approach \cite{MM}.} \label{fig:FigEMCh6} \end{figure}

The EM algorithm is an instance of the more general Majorization Minimization
(MM) algorithm \cite{MM}. In this class of algorithms, at each iteration,
one constructs a tight lower bound of the objective function at the
current iterate $\theta^{\textrm{old}}.$ This bound, which should
be easy to maximize, is then optimized, yielding the new iterate $\theta^{\textrm{new}}$.
The process is illustrated in Fig. \ref{fig:FigEMCh6}. As it can
be seen, at each iteration one is guaranteed that the objective function
is not decreased, which ensures convergence to a local optimum of
the original problem. In EM, the tight lower bound is the ELBO $\mathcal{L}(q^{\textrm{new}},\theta)$,
which is obtained by computing the posterior distribution of the latent
variables $q^{\textrm{new}}(z)=p(z|x,\theta^{\textrm{old}})$ at the
current iterate $\theta^{\textrm{old}}$. 

Generalizing the $K$-means algorithm, the EM algorithm alternates
between: (\emph{i}) making inferences about the hidden variables $\mathrm{z}$
based on the model defined by the current iterate of the model parameter
$\theta$ in the E step; and (\emph{ii}) updating the model $\theta$
to match the data $x$ and the inferred variables $\mathrm{\ensuremath{z}}$
in the M step. It is useful to emphasize that unsupervised learning,
even in the assumed frequentist approach, entails the Bayesian inference
problem of evaluating the posterior of the hidden variables.

\textbf{Generalization to $N$ observations.} We conclude this section
by making explicit the generalization of the EM algorithm to the case
in which we have $N$ i.i.d. observations. To elaborate, assume that
we have pairs of observed/ unobserved i.i.d. variables $(\mathrm{x}_{n},\mathrm{z}_{n})$,
whose assumed joint distribution can be written as $p(x^{N},z^{N}|\theta)=\prod_{n=1}^{N}p(x_{n},z_{n}|\theta)$. 

\emph{E step}: The E step requires the computation of the posterior\emph{
$p(z^{N}|x^{N},\theta)$. }This can be seen to factorize across the
examples, since we have
\begin{align}
p(z^{N}|x^{N},\theta)=\frac{p(x^{N},z^{N}|\theta)}{p(x^{N}|\theta)}= & \prod_{n=1}^{n}\frac{p(x_{n},z_{n}|\theta)}{p(x_{n}|\theta)}\nonumber \\
= & \prod_{n=1}^{n}p(z_{n}|x_{n},\theta)\text{.\text{ }}\text{}\label{eq:E step N}
\end{align}
Therefore, in order to compute the posterior, we can operate separately
for each example in the data set by evaluating $p(z_{n}|x_{n},\theta)$
for all $n=1,...,N$. 

\emph{M step}: In a similar manner, the negative energy function $Q(\theta,\theta^{\textrm{old}})$
to be used in the \emph{M step} can be computed separately for each
example as
\begin{align}
Q(\theta,\theta^{\textrm{old}}) & =\textrm{E}_{\mathrm{z}^{N}\sim p(z^{N}|x^{N},\theta^{\textrm{old}})}\left[\ln p(x^{N},\mathrm{z}^{N}|\theta)\right]\nonumber \\
 & =\sum_{n=1}^{N}\textrm{E}_{\mathrm{z}_{n}\sim p(z_{n}|x_{n},\theta^{\textrm{old}})}\left[\ln p(x_{n},\mathrm{z}_{n}|\theta)\right],\label{eq:M step N}
\end{align}
and hence separately for each example. Note that the optimization
in the M step should instead be done jointly on (\ref{eq:M step N}).

\textbf{Extensions.$^{*}$} The EM algorithm solves the non-convexity
problem identified at the beginning of this section by optimizing
ELBOs iteratively according to the outlined MM mechanism. Nevertheless,
implementing the EM algorithm may be too computationally demanding
in practice. In fact, the E step requires to compute the posterior
distribution of the latent variables, which may be intractable when
the latent space is large; and the M step entails an average over
the posterior distribution, which can also be intractable. In Chapter
8, we will see approaches to overcome these problems via approximate
inference and learning techniques. For extensions of EM algorithm,
we refer to \cite[pp. 247-248]{Barber}. In this regard, it is observed
here that the EM algorithm applies also to scenarios in which different
data points have generally different subsets of unobserved variables.

\section{Directed Generative Models }

In this section, we discuss directed generative models, in which,
as seen in Fig. \ref{fig:FigmodelsCh6}(a), one posits a parametrized
prior $p(z|\theta)$ of the latent variables $\mathrm{z}$ and a parametrized
conditional decoding distribution $p(x|z,\theta)$. As discussed,
the goal can be that of learning the distribution $p(x)$ of the true
data or to extract useful features $\mathrm{z}$. In the latter case,
the use of directed generative model is referred to as performing
\emph{analysis by synthesis}. We first discuss two prototypical applications
of EM to perform ML learning. We then outline an alternative approach,
known as GAN, which can be thought of as a generalization of ML. Accordingly,
rather than selecting a priori the KL divergence as a performance
metric, as done implicitly by ML, GANs learn at the same time divergence
and generative model. 

Multi-layer extensions of generative directed models discussed here
fall in the category of Helmholtz machines. We refer to, e.g., \cite{dayan1995helmholtz},
for a discussion about the task of training these networks via an
approximation of the EM algorithm, which uses tools covered in Chapter
8.

\subsection{Mixture of Gaussians Model }

The mixture of Gaussians model can be described by the following directed
generative model\begin{subequations}
\begin{align}
\mathrm{z}_{n} & \sim\textrm{Cat}(\pi)\\
\mathrm{x}_{n}|\mathrm{z} & _{n}=k\sim\mathcal{N}(\mu_{k},\Sigma_{k}),
\end{align}
\end{subequations}with parameter vector $\theta=[\pi$,$\mu_{0},...,\mu_{K-1},\text{\ensuremath{\Sigma}}_{0},...,\Sigma_{K-1}].$
We use one-hot encoding for the categorical variables $z_{n}=[z_{0n},...,z_{(K-1)n}]^{T}.$
Note that this model can be thought as an unsupervised version of
the QDA model for the fully observed, or supervised, case studied
in Sec. \ref{sec:Generative-Probabilistic-Models Ch4} (cf. (\ref{eq:QDA})).
We will see below that the EM algorithm leverages this observation
in the M step. Furthermore, it will be observed that EM for mixture
of Gaussians generalizes the $K$-means algorithm.

\emph{E step}. In the E step, as per (\ref{eq:E step N}), we need
to solve the inference problem of computing the posterior $p(z_{kn}|x_{n},\theta^{\textrm{old}})$
for every example $n$ and cluster index $k=0,...,K-1$. In this case,
this can be done directly via Bayes' theorem since the normalization
requires to sum only over the $K$ possible values taken by $\mathrm{z}_{n}$,
yielding
\begin{equation}
p(z_{kn}=1|x_{n},\theta^{\textrm{old}})=\frac{\pi_{k}^{\textrm{old}}\mathcal{N}(x_{n}|\mu_{k}^{\textrm{old}},\Sigma_{k}^{\textrm{old}})}{\sum_{j=0}^{K-1}\pi_{j}^{\textrm{old}}\mathcal{N}(x_{n}|\mu_{j}^{\textrm{old}},\Sigma_{j}^{\textrm{old}})}\text{\text{.}}\label{eq:posterior EM ex1}
\end{equation}

\emph{M step}. In the M step, we need to maximize the negative energy
function $Q(\theta,\theta^{\textrm{old}})$ in (\ref{eq:M step N}).
Each term in the sum can be computed directly as
\begin{align}
\textrm{E}_{\mathrm{z}_{n}\sim p(z_{n}|x_{n},\theta^{\textrm{old}})}\left[\ln p(x_{n},\mathrm{z}_{n}|\theta)\right] & =\sum_{k=0}^{K-1}\bar{z}_{kn}\left\{ \ln\pi_{k}+\ln\mathcal{N}(x_{n}|\mu_{k},\Sigma_{k})\right\} 
\end{align}
with
\begin{equation}
\bar{z}_{kn}=\textrm{E}_{\mathrm{z}_{n}\sim p(z_{n}|x_{n},\theta^{\textrm{old}})}\left[\mathrm{z}_{kn}\right]=p(z_{kn}=1|x_{n},\theta^{\textrm{old}}).
\end{equation}
As it can be easily checked, the function $Q(\theta,\theta^{\textrm{old}})$
equals the LL function of the QDA supervised problem, in which the
variables $\mathrm{z}_{n}$ are observed, with the following caveat:
the sufficient statistics $z_{kn}$ are replaced by their posterior
averages $\bar{z}_{kn}$. As a result, as opposed to supervised learning,
EM describes each example $x_{n}$ as being part of all clusters,
with the\emph{ }``responsibility'' of cluster $k$ being given by
$\bar{z}_{kn}$. Having made this observation, we can now easily optimize
the $Q(\theta,\theta^{\textrm{old}})$ function by using the ML solutions
(\ref{eq:QDA ML}) for QDA by substituting $\bar{z}_{kn}$ for the
observed variable $t_{kn}$.

Setting $\Sigma_{k}=\epsilon I$ as a known parameter and letting
$\epsilon\rightarrow0$ recovers the $K$-means algorithm \cite[p. 443]{Bishop}.

\begin{figure}[H] 
\centering\includegraphics[scale=0.55]{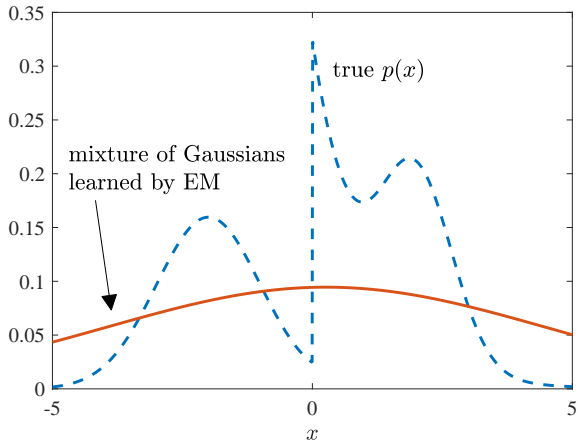}
\caption{True distribution (dashed line) and mixture of Gaussians model learned via EM (solid line).} \label{fig:figmixGaussCh6} \end{figure}

\begin{example}Consider data generated from the multi-modal distribution
shown in Fig. \ref{fig:figmixGaussCh6} as a dashed line, which obtained
as a mixture of two Gaussians and an exponential distribution. When
$M=1$, the mixture of Gaussians distribution learned via EM corresponds
to the conventional ML estimate, and is hence obtained via moment
matching (see Chapter 3). This is plotted in the figure for a given
data realization with $N=100$. Running EM with the larger values
$M=2$ and $M=3$ also returns the same distribution. This distribution
is seen to be inclusive of the entire support of the true distribution
and to smooth out the edges of the original distribution. As we will
further discuss in Sec. \ref{subsec:GAN}, this is a consequence of
the fact that ML minimizes the KL divergence over the second argument.
\end{example}

A similar model that applies to binary data, rather than continuous
data, such as black-and-white images, is the mixture of Bernoullis
model. The EM algorithm can be derived by following the same steps
detailed here \cite{Bishop}.

\subsection{Probabilistic Principal Component Analysis }

Probabilistic Principal Component Analysis (PPCA) is a popular generative
model that describes the data in terms of $M<D$ features that are
linearly related to the data vector. Specifically, PPCA uses a linear
factor generative model with $M<D$ features described as\begin{subequations}
\begin{align}
\mathrm{z} & _{n}\sim\mathcal{N}(0,I)\\
\mathrm{x}_{n}|\mathrm{z}_{n} & =z\sim\mathcal{N}(Wz+\mu,\sigma^{2}I),
\end{align}
\end{subequations} with parameter vector $\theta=(W\textrm{ }\mu\textrm{ }\sigma)$.
Equivalently, according to PPCA, the data vector can be written as
\begin{equation}
\mathrm{x}_{n}=W\mathrm{z}_{n}+\mu+\mathrm{q}_{n},\label{eq:lin features Ch6}
\end{equation}
with the latent variables $\mathrm{z}{}_{n}\sim\mathcal{N}(0,I)$
and the additive noise $\mathrm{q}_{n}\sim\mathcal{N}(0,\sigma^{2}I)$.
According to (\ref{eq:lin features Ch6}), the columns $\{w_{k}\}$
of the matrix $W=[w_{1}\textrm{ }w_{2}\cdots w_{M}]$ can be interpreted
as linear features of the data. This is in the sense that each data
point is written as a linear combination of the feature vectors as
\begin{equation}
\mathrm{x}_{n}=\sum_{m=1}^{M}w_{m}\mathrm{z}_{mn}+\mu+\mathrm{q}_{n},
\end{equation}
where $\mathrm{z}_{mn}$ is the $m$th component of the latent vector
$\mathrm{z}_{n}$. 

In the models studied above, the latent variable was a categorical
identifier of the class of the observation. As such, In PPCA, instead,
the representation of an observation $x_{n}$ in the hidden variable
space is \emph{distributed} across all the variables in vector $z_{n}$.
This yields a more efficient encoding of the hidden representation.
This is particularly clear in the case discrete hidden rv, which will
be further discussed in Sec. \ref{subsec:Restricted-Boltzmann-Machines}
(see also \cite{bengio2013representation}).

An EM algorithm can be devised to learn the parameter vector as described
in \cite[p. 577]{Bishop}. The E step leverages the known result that
the posterior of the latent variables can be written as
\begin{equation}
\mathrm{z}_{n}|\mathrm{x}_{n}=x_{n},\theta\sim\mathcal{N}(z_{n}|J^{-1}W^{T}(x_{n}-\mu),J^{-1}),
\end{equation}
with matrix $\ensuremath{J=\sigma^{-2}W^{T}W+I}$.

We note that, if we model the latent variables $\mathrm{z}_{kn}$,
$k=1,...,M$, are independent but not Gaussian, we obtain a type of
Independent Component Analysis (ICA) (see \cite[p. 591]{Bishop}).
It is also possible to impose structure on the latent vector by selecting
suitable marginal distributions $p(z)$. For instance, choosing Student's
t-distribution or a Laplace distribution tends to impose sparsity
on the learned model. A general discussion on linear factor models
can be found in \cite{Murphy} (see also \cite{collins2002generalization}
for a generalization of PPCA to any model in the exponential family).

\subsection{GAN\label{subsec:GAN}}

In Sec. \ref{sec:Information-Theoretic-Metrics}, we have seen that
ML can be interpreted, in the asymptotic regime of a large data set
$N$, as minimizing the KL divergence between the true distribution
of the data and the model. We start this section by revisiting this
argument for unsupervised learning in order to highlight a similar
interpretation that holds for finite values of $N$. This viewpoint
will lead us to generalize the ML learning approach to techniques
in which the choice of the divergence measure is adapted to the data.
As an important by-product, the resulting technique, known as GAN,
will be seen to accommodate easily likelihood-free models. GANs are
currently considered to yield state-of-the-art results for image generation
\cite{GoodfellowGANreview}.

To simplify the discussion, assume that variables $\mathrm{x}_{n}$
are categorical and take a finite number of values. To start, let
us observe that the NLL function (\ref{eq:iid data Ch6}) for i.i.d.
data can be written as 
\begin{equation}
-\frac{1}{N}\sum_{n=1}^{N}\textrm{ln\ensuremath{p(x_{n}|\theta)=-\sum_{x}\textrm{\ensuremath{\frac{N[x]}{N}}ln\ensuremath{p(x|\theta)}}},}\label{eq:NLL GAN}
\end{equation}
where we recall that $N[x]=|\{n:\textrm{ \ensuremath{x_{n}=x}}\}|$.
We now note that the ML problem of minimizing the NLL (\ref{eq:NLL GAN})
over $\theta$ is equivalent to the minimization of the KL divergence
\begin{align}
 & \underset{\theta}{\min}-\sum_{x}\textrm{\ensuremath{p_{\mathcal{D}}(x)}ln\ensuremath{p(x|\theta)}}+\sum_{x}\textrm{\ensuremath{p_{\mathcal{D}}(x)}ln\ensuremath{p_{\mathcal{D}}(x)}}\nonumber \\
 & \textrm{=\ensuremath{\underset{\theta}{\min}} }\textrm{KL}(p_{\mathcal{D}}(x)||p(x|\theta)),\label{eq:I proj GAN}
\end{align}
where we have defined the empirical distribution
\begin{equation}
p_{\mathcal{D}}(x)=\frac{N[x]}{N}.
\end{equation}
We hence conclude the ML attempts to minimize the KL divergence between
the empirical distribution $p_{\mathcal{D}}(x)$ of the data and the
model distribution $p(x|\theta)$. The ML problem min$_{\theta}$
$\textrm{KL}(p_{\mathcal{D}}(x)||p(x|\theta))$ is also known as the
\emph{M-projection }of the data distribution $p_{\mathcal{D}}(x)$
into the model $\{p(x|\theta)\}$ \cite{FnTCsi} (see Chapter 8 for
further discussion).

The KL divergence (\ref{eq:I proj GAN}) is a measure of the difference
between two distribution with specific properties that may not be
tailored to the particular application of interest. For instance,
distributions obtained by minimizing (\ref{eq:I proj GAN}) tend to
provide ``blurry'' estimates of the distribution of the data distribution,
as we have seen in Fig. \ref{fig:figmixGaussCh6}. As a result, learning
image distributions using M-projections is known to cause the learned
model to produce unfocused images \cite{GoodfellowGANreview}. 

The KL divergence is not, however, the only measure of the difference
between two distributions. As discussed in more detail in Appendix
A, the KL divergence is in fact part of the larger class of $f$-divergences
between two distributions $p(x)$ and $q(x)$. This class includes
divergence measures defined by the variational expression\footnote{The term variational refers to the fact that the definition involves
an optimization.}
\begin{equation}
D_{f}(p||q)=\max_{T(x)}\textrm{E}_{\mathrm{x}\sim p}[T(\mathrm{x})]-\textrm{E}_{\mathrm{x}\sim q}[g(T(\mathrm{x}))],\label{eq:f divergence Ch6}
\end{equation}
for some concave increasing function $g(\cdot)$ (the meaning of the
$f$ subscript is discussed in Appendix A). The key new element in
(\ref{eq:f divergence Ch6}) is the decision rule $T(x)$, which is
also known as \emph{discriminator} or critic. This function takes
as input a sample $x$ and ideally outputs a large value when $x$
is generated from distribution $p$, and a small value when it is
instead generated from $q$. Optimizing over $T(x)$ hence ensures
that the right-hand side of (\ref{eq:f divergence Ch6}) is large
when the two distributions are different and hence can be distinguished
based on the observation of $x$. When the domain over which the discriminator
$T(x)$ is optimized is left unconstrained, solving problem (\ref{eq:f divergence Ch6})
recovers the KL divergence and the Jensen-Shannon divergence, among
others, with specific choices of function $g$ (see Appendix A).\footnote{For a given function $T(x)$, the right-hand side of (\ref{eq:f divergence Ch6}),
excluding the maximization, provides a lower bound on the divergence
$D_{f}(p||q)$. Another useful related bound for the KL divergence
is the so-called Donsker-Varadhan representation (see, e.g., \cite{belghazi2018mine}).}

Generalizing the ML problem (\ref{eq:I proj GAN}), GANs attempt to
solve the problem
\begin{equation}
\textrm{\ensuremath{\underset{\theta}{\min}} }D_{f}(p_{\mathcal{D}}(x)||p(x|\theta)).\label{eq:ideal GAN}
\end{equation}
More precisely, GANs parametrize the discriminator $T(x)$ by choosing
a differentiable function $T_{\varphi}(x)$ of the parameter vector
$\varphi$. This effectively reduces the search space for the discriminator,
and defines a different type of divergence for each value of $\varphi$.
A typical choice for $T_{\varphi}(x)$ is the output of a multi-layer
neural network with weights $\varphi$ or a function thereof (see
Sec. \ref{subsec:Non-linear-Classification}). With this in mind,
using (\ref{eq:f divergence Ch6}) in (\ref{eq:ideal GAN}), we obtain
the minimax problem solved by GANs\footnote{The version of GAN given here is known as $f$-GANs \cite{fGAN},
which is a generalization of the original GAN.}
\begin{equation}
\textrm{\ensuremath{\underset{\theta}{\min}} }\max_{\varphi}\textrm{E}_{\mathrm{x}\sim p_{\mathcal{D}}}[T_{\varphi}(\mathrm{x})]-\textrm{E}_{\mathrm{x}\sim p(x|\theta)}[g(T_{\varphi}(\mathrm{x}))].\label{eq:ideal GAN-1}
\end{equation}
According to (\ref{eq:ideal GAN-1}), GANs select the divergence measure
adaptively through the optimization of the discriminator $T_{\varphi}(x)$.
Problem (\ref{eq:ideal GAN-1}) can also be interpreted in terms of
a strategic game played by generator and discriminator \cite{GoodfellowGANreview}.\footnote{It is noted that the GAN set-up is reminiscent of the adversarial
model used to define semantic security in cryptography.} 

The original GAN method \cite{GoodfellowGANreview} operates by setting
$T_{\varphi}(x)=\ln D_{\varphi}(x)$, where $D_{\varphi}(x)$ is the
output of a multi-layer perceptron, and $g(t)=\ensuremath{-\log(1-\exp(t))}$.
A variation that is shown to be more useful in practice is also discussed
in the first GAN paper (see \cite{Fedus17}). As another popular example,
Wasserstein GAN simply sets $T_{\varphi}(x)$ to be the output of
a multi-layer perceptron and chooses $g(t)=-(1-t)$. 

\textbf{Application to likelihood-free models.$^{*}$} GANs are typically
used with likelihood-free models. Accordingly, the model distribution
$p(x|\theta)$ is written as
\begin{equation}
p(x|\theta)=\int\delta(x-G_{\theta}(z))\mathcal{N}(z|0,I)dz.
\end{equation}
The samples $\mathrm{x}$ are hence modelled as the output of a \emph{generator}
function $G_{\theta}(z)$, whose input $\mathrm{z}$ is given by i.i.d.
Gaussian variables. This generative model can hence be interpreted
as a generalization of PPCA to non-linear encoders $G_{\theta}(z)$.
The latter is conventionally modelled again as a multi-layer neural
network.

Problem (\ref{eq:ideal GAN-1}) is typically tackled using SGD (Chapter
4) by iterating between the optimization of the generator parameters
$\theta$ and of the discriminator parameters $\varphi$ (see \cite[Algorithm 1]{Wasserstein}).
Learning requires empirical approximations for the evaluation of the
gradients that will be discussed in Chapter 8. 

\textbf{Likelihood ratio estimation viewpoint.$^{*}$} When no constraints
are imposed over the discriminator $T(x)$ in (\ref{eq:f divergence Ch6})
and the function $g$ is selected as $g(t)=\exp(t-1)$, the optimal
solution is given as $T(x)=1+\ln(p_{\mathcal{D}}(x)/p(x|\theta))$
and the corresponding divergence measure (\ref{eq:f divergence Ch6})
is the KL divergence KL$(p_{\mathcal{D}}(x)||p(x|\theta))$. Therefore,
solving problem (\ref{eq:f divergence Ch6}) over a sufficiently general
family of functions $T_{\varphi}(x)$ allows one to obtain an estimate
of the log-likelihood ratio $\ln(p_{\mathcal{D}}(x)/p(x|\theta))$.
As a result, GANs can be interpreted as carrying out the estimation
of likelihood ratios between the data and the model distributions
as a step in the learning process. The idea of estimating likelihood
ratios in order to estimate KL divergences is useful in other learning
problems based on variational inference, to be discussed in Chapter
8 (see \cite{huszar2017variational}).

\textbf{Some research topics.$^{*}$} Among topics of current research,
we mention here the problem of regularization of GANs \cite{roth2017stabilizing}.
We also note that GANs can also be extended to supervised learning
problems \cite{odena2016conditional}. The GAN approach of formulating
the divergence as an optimization over a discriminator function was
also recently found to be useful for ICA \cite{brakel17}. 

\section{Undirected Generative Models }

Unlike directed generative models, undirected generative models posit
a joint distribution for the hidden and the observed variables that
captures affinity, rather than causality, relationships between the
two sets of variables, as seen in Fig. \ref{fig:FigmodelsCh6}(b).
In this section, we discuss a representative example of undirected
generative models that is considered to be a powerful tool for a number
of applications, including feature selection, generation of new samples,
and even recommendation systems \cite{RusReccom}. The model, known
as RBM, also finds application as a components of larger multi-layer
structures, also for supervised learning \cite{Hintoncourse}. 

\subsection{Restricted Boltzmann Machines (RBM)\label{subsec:Restricted-Boltzmann-Machines} }

RBMs are typically characterized by an $M$-dimensional binary hidden
vector $z\in\{0,1\}^{M}$, while the observed variables may be discrete
or continuous. Here we consider the case of binary observed variables,
which is suitable to model, e.g., black-and-white images or positive/
negative recommendations. The RBM model is defined as (\ref{eq:undirected UL}),
where the joint distribution of visible and latent variables is a
log-linear model, and hence part of the exponential family. Mathematically,
we have
\begin{equation}
p(x,z|\theta)=\frac{1}{Z(\theta)}\exp(-E(x,z|\theta)),\label{eq:pdf RBM}
\end{equation}
with the energy function given as 
\begin{equation}
E(x,z|\theta)=-a^{T}z-b^{T}x-x^{T}Wz,\label{eq:energy RBM}
\end{equation}
and the partition function as $Z(\theta)=\sum_{x,z}\exp(-E(x,z|\theta)).$
The parameter vector $\theta$ includes the $M\times1$ vector $a$,
the $D\times1$ vector $b$ and the $D\times M$ matrix $W$.

\begin{figure}[H] 
\centering\includegraphics[scale=0.55]{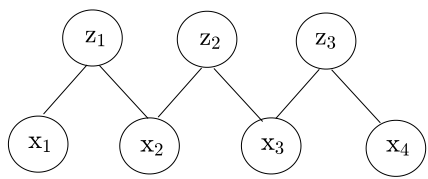}
\caption{An undirected graph (MRF) describing the joint distribution prescribed by the RBM model (\ref{eq:pdf RBM})-(\ref{eq:energy RBM}).} \label{fig:FigRBMCh6} \end{figure}

The qualifier ``restricted'' captures the fact that the energy function
(\ref{eq:energy RBM}) only features cross-terms that include one
variable from $\mathrm{x}$ and one from $\mathrm{\ensuremath{z}}$,
and not two variables from $\mathrm{x}$ or two from $\mathrm{z}$.
In other words, the model accounts for the affinities between variables
in $\mathrm{x}$ and $\mathrm{z}$, but not directly between variables
in either $\mathrm{x}$ or $\mathrm{z}$ \cite{bengio2013representation}.
For instance, if ($i,j$)th entry $w_{ij}$ of matrix $W$ is a large
positive number, variables $x_{i}$ and $z_{j}$ will tend to have
equal signs in order to minimize the energy and hence maximize the
probability; and the opposite is true when $w_{ij}$ is negative.
As we will seen in the next chapter, this type of probabilistic relationship
can be represented by the undirected graphical model known as MRF
shown in Fig. \ref{fig:FigRBMCh6}. 

From the model, it is easy to compute the distribution of each hidden
or observed variable when conditioning on all the observed or hidden
variables, respectively. In particular, we have that the variables
in $\mathrm{z}$ are mutually independent when conditioned on $\mathrm{x}$,
and, similarly, the variables in $\mathrm{x}$ are independent given
$\mathrm{z}$. Furthermore, the conditional distributions are given
as:\begin{subequations}\label{eq:RBM conditional}
\begin{align}
p(z_{j}=1|x,\theta)= & \sigma(w_{j}^{T}x+a_{j})\label{eq:RBM conditional 1}\\
p(x_{i}=1|z,\theta)= & \sigma(\tilde{w}_{i}^{T}z+b_{i}),\label{eq:RBM conditional 2}
\end{align}
\end{subequations}$\text{where \ensuremath{w_{j}} is the \ensuremath{j}th column of \ensuremath{W} and \ensuremath{\tilde{w}_{i}} is the \ensuremath{i}th row of matrix \ensuremath{W}}$
rearranged into a column vector via transposition. These relationships
reveal the significance of the binary hidden variables $z$ in terms
of features. 

In fact, as for PPCA, we can consider each column of matrix $W$ as
a feature vector that contributes in explaining the observations.
To see this, note that (\ref{eq:RBM conditional 1}) suggests that
the $j$th hidden variable $z_{j}$ equals 1 when the feature $w_{j}$
is well correlated with the data point $x$. Probability (\ref{eq:RBM conditional 2})
also confirms this interpretation, as each variable $z_{j}$ multiplies
the $i$th element of the feature vector $w_{j}$ in defining the
LLR $\tilde{w}_{i}^{T}z+b_{i}$ (see Sec. \ref{sec:The-Exponential-Family}).
Furthermore, the\emph{ distributed} representation of the data in
terms of the binary hidden vector $z$ is more efficient than that
provided by models with categorical variables such as the mixture
of Gaussian model. This is in the following sense. While categorical
models require to learn a number of parameters that increases linearly
with the number of classes, the distributed representation with a
binary vector $z$ can distinguish $2^{D}$ combinations of features
with a number of parameters that increases only linearly with $D$
\cite{bengio2013representation}.

Learning is typically done by means of an approximate SGD method that
leverages MC techniques. Recalling the general formula (\ref{eq:gradient exp})
for the exponential family, the gradient at the current iteration
$\theta=\theta^{\textrm{old}}$ can be computed using (\ref{eq:gradient unsupervised})
as
\begin{align}
\nabla_{w_{ij}}\ln p(x|\theta)|_{\theta=\theta^{\textrm{old}}} & =\textrm{E}_{\mathrm{\mathrm{z}}_{j}\sim p(z_{j}|x,\theta^{\textrm{old}})}[x_{i}\mathrm{z}_{j}]-\textrm{E}_{\textrm{\ensuremath{\mathrm{x}_{i},\mathrm{z}_{j}}}\sim p(\textrm{\ensuremath{x,z_{j}}}|\theta^{\textrm{old}})}[\ensuremath{\mathrm{x}_{i}\mathrm{z}_{j}]},
\end{align}
which can be further simplified using (\ref{eq:RBM conditional}).
The gradient presents the structure, noted in Chapter 3, given by
the difference between a positive component, which depends on the
data $x$, and a negative component, which instead requires an ensemble
average over all other possible samples $\textrm{\ensuremath{\mathrm{x}}}\sim p(\textrm{\ensuremath{x}}|\theta^{\textrm{old}})$.
Similar relations can be written for the gradients with respect to
$a$ and $b$, namely
\begin{align}
\nabla_{a_{j}}\ln p(x|\theta)|_{\theta=\theta^{\textrm{old}}} & =\textrm{E}_{\mathrm{\mathrm{z}}_{j}\sim p(z_{j}|x,\theta^{\textrm{old}})}[\mathrm{z}_{j}]-\textrm{E}_{\mathrm{\mathrm{z}}_{j}\sim p(z_{j}|\theta^{\textrm{old}})}[\mathrm{z}_{j}]
\end{align}
and
\begin{equation}
\nabla_{b_{i}}\ln p(x|\theta)|_{\theta=\theta^{\textrm{old}}}=x_{i}-\textrm{E}_{\textrm{\ensuremath{\mathrm{x}_{i}}}\sim p(x_{i}|\theta^{\textrm{old}})}[\mathrm{x}_{i}].
\end{equation}

In order to evaluate the expectations in the negative components of
the gradients, one typically uses a an MC technique known as Markov
Chain MC (MCMC) to be discussed in Chapter 8. More specifically, a
simplified approach known as Contrastive Divergence (CD) has been
found to be an effective approximation. Accordingly, one ``clamps''
the visible variables to the observed variables $\mathrm{x}=x$, and
generates the sequence $x\rightarrow z^{(0)}\rightarrow x^{(1)}\rightarrow z^{(1)}$,
using the conditional probability (\ref{eq:RBM conditional 1}) to
generate $\mathrm{z}$ from $\mathrm{x}$ and (\ref{eq:RBM conditional 2})
to generate $\mathrm{x}$ from $\mathrm{z}$. The resulting samples
are used to approximate the expectations as
\begin{align}
\nabla_{w_{ij}}\ln p(x|\theta)|_{\theta=\theta^{\textrm{old}}}\simeq x_{i} & z_{j}^{(0)}-x_{i}^{(1)}z_{j}^{(1)},
\end{align}
and similar expressions apply for the other gradients. The CD scheme
can also be generalized to CD-$k$ by increasing the Markov chain
sequence to $k$ steps, and the using the resulting samples $x^{(k)}$
and $z^{\ensuremath{(k)}}$ in lieu of $x^{(1)}$ and $z^{(1)}.$

Extensions and application of the RBM are discussed in \cite{bengio2013representation,welling2005exponential}.
Generalizations of RBMs with multiple layers of hidden variables are
referred to as Deep Boltzmann Machines. We refer to \cite{Hintoncourse}
for discussion about training and applications.

\section{Discriminative Models\label{sec:Discriminative-Models-Ch6}}

When the goal is learning a representation $\mathrm{z}$ of data $\mathrm{x}$,
one can try to directly learn a parametrized encoder, or discriminative
model, $p(z|x,\theta)$, as illustrated in Fig. \ref{fig:FigmodelsCh6}(c).
With an encoder $p(z|x,\theta)$, we can define the joint distribution
as $p(x,z|\theta)=p(x)p(z|x,\theta)$, where $p(x)$ is the true distribution
of the data. The latter is in practice approximated using the empirical
distribution $p_{\mathcal{D}}(x)$ of the data. This yields the joint
distribution 
\begin{equation}
p(x,z|\theta)=p_{\mathcal{D}}(x)p(z|x,\theta).\label{eq:joint distribution discr Ch6}
\end{equation}

From (\ref{eq:joint distribution discr Ch6}), it is observed that,
unlike the frequentist methods considered up to now, discriminative
models for unsupervised learning are based on the definition of a
single distribution over observed and unobserved variables. As such,
divergence measures, which involve two distributions, are not relevant
performance metrics. In contrast, a suitable metric is the mutual
information between the jointly distributed variables ($\mathrm{x}$,$\mathrm{z}$).
The mutual information is a measure of the statistical dependence
between the two rvs and is introduced in Appendix A.

To elaborate, a typical learning criterion is the maximization of
the mutual information $I_{p(x,z|\theta)}(\mathrm{x};\mathrm{z})$
between the representation $\mathrm{z}$ and the data $\mathrm{x}$
under the joint distribution $p(x,z|\theta)$. Note that, for clarity,
we explicitly indicated the joint distribution used to evaluate the
mutual information as a subscript. It can be related to the KL divergence
as $I_{p(x,z|\theta)}(\mathrm{x};\mathrm{z})=\textrm{KL}(p(x,z|\theta)||p(x)p(z))$.
This relationship indicates that the mutual information quantifies
the degree of statistical dependence, or the distance from independence,
for the two rvs. 

The resulting \emph{Information Maximization }problem is given as
\begin{equation}
\underset{\theta}{\max\textrm{ }}I_{p_{\mathcal{D}}(x)p(z|x,\theta)}(\mathrm{x};\mathrm{z}).\label{eq:infomaxCh6}
\end{equation}
As seen, in order to avoid learning the trivial identity mapping,
one needs to impose suitable constraints on the encoder $p(z|x,\theta)$
in order to restrict its capacity. In order to tackle problem (\ref{eq:infomaxCh6}),
a typical solution is to resort to an MM approach that is akin to
the EM algorithm for ML learning described above (see Fig. \ref{fig:FigEMCh6}). 

To this end, as for EM, we introduce a variational distribution $q(x|z)$,
and observe that we have the following lower bound on the mutual information\footnote{The bound is also used by the Blahut-Arimoto algorithm \cite{Cover}.}
\begin{equation}
I_{p_{\mathcal{D}}(x)p(z|x,\theta)}(\mathrm{x};\mathrm{z})\geq H(p_{\mathcal{D}}(x))+\textrm{E}_{\mathrm{x},\mathrm{z}\sim p_{\mathcal{D}}(x)p(z|x,\theta)}[\ln q(\mathrm{x}|\mathrm{z})].\label{eq:infomax}
\end{equation}
This result follows using the same steps as for the ELBO. The bound
is tight when the variational distribution $q(x|z)$ equals the exact
posterior $p(x|z,\theta)=p_{\mathcal{D}}(x)p(z|x,\theta)/(\sum_{x}p_{\mathcal{D}}(x)p(z|x,\theta))$.
Based on this inequality, one can design an iterative optimization
algorithm over the model parameters and the variational distribution
$q(x|z)$ (see, e.g., \cite{Akagov05,vincent2010stacked}). When the
latter is constrained to lie in a parametrized family, the optimization
over the decoder $q(x|z)$ is an example of variational inference,
which will be discussed in Chapter 8. The optimization over the model
parameters can be simplified when the model has additional structure,
such as in the InfoMax ICA method \cite{lee1999independent}.

\textbf{Inference-based interpretation.$^{*}$} The mutual information
can be related to the error probability of inferring $\mathrm{x}$
given the representation $\mathrm{z}$. This can be seen, e.g., by
using Fano's inequality \cite{Cover,Akagov05}. More concretely, criterion
(\ref{eq:infomax}) can be interpreted in terms of inference of $\mathrm{x}$
given $\mathrm{z}$ by noting that its right-hand side can be written
as the difference $H(p_{\mathcal{D}}(x))-(\textrm{E}_{\mathrm{x},\mathrm{z}\sim p_{\mathcal{D}}(x)p(z|x,\theta)}[-\ln q(\mathrm{x}|\mathrm{z})]).$
In fact, the second term is the cross-entropy measure for the prediction
of $\mathrm{x}$ given $\mathrm{z}$ by means of the variational distribution
$q(x|z)$. Therefore, by maximizing (\ref{eq:infomax}) over $\theta$,
the model $p(z|x,\theta)$ obtains a representation $z$ such that
the predictor $q(x|z)$ ensures, on average, a good reconstruction
of $x$ given $z$. This point is further elaborated on in \cite{Alemi17}.
We specifically point to \cite[Fig. 2]{Alemi17}, where a comparison
with methods based on ML is provided. 

\textbf{Information bottleneck method.$^{*}$} An important variation
of the InfoMax principle yields the \emph{information bottleneck}
method. In the latter, one assumes the presence of an additional variable
$\mathrm{y}$, jointly distributed with $\mathrm{x}$, which represents
the desired information, but is unobserved. The goal is, as above,
to learn an encoder $p(z|x,\theta)$ between the observed $\mathrm{x}$
and the representation $\mathrm{z}$. However, here, this is done
by maximizing the mutual information $I(\mathrm{y};\mathrm{z})$ between
the target unobserved variable $\mathrm{y}$ and the representation
$\mathrm{z}$, in the presence of a regularization term that aims
at minimizing the complexity of the representation. This penalty term
is given by the mutual information $I(\mathrm{x};\mathrm{z})$, which
finds justification in rate-distortion arguments \cite{bottleneck}.

\section{Autoencoders }

As seen in Fig. \ref{fig:FigmodelsCh6}, autoencoders include parametric
models for both encoder and decoder. We focus here for brevity on
deterministic autoencoders, in which the encoder is defined by a function
$z=F_{\theta}(x)$ and the decoder by a function $x=G_{\theta}(z)$.
Note that the parameters defining the two functions may be tied or
may instead be distinct, and that the notation is general enough to
capture both cases. We will mention at the end of this section probabilistic
autoencoders, in which encoder and decoder are defined by conditional
probability distributions. 

Autoencoders transform the unsupervised learning problem of obtaining
a representation $z=F_{\theta}(x)$ of the input $x$ based solely
on unlabelled examples $\{x_{n}\}_{n=1}^{N}$ into an instance of
supervised learning. They do so by concatenating the encoder $z=F_{\theta}(x)$
with a decoder $x=G_{\theta}(z)$, so as to obtain the input-output
relationship $t=G_{\theta}(F_{\theta}(x))$. The key idea is to train
this function by setting the target $t$ to be equal to the input
$x$. As such, the machine learns to obtain an intermediate representation
$z=F_{\theta}(x)$ that makes it possible to recover a suitable estimate
$t=G_{\theta}(z)\simeq x$ of $x$.

To formalize the approach, learning is typically formulated in terms
of the ERM problem
\begin{equation}
\underset{\theta}{\text{min}}\sum_{n=1}^{N}\ell(x_{n},G_{\theta}(F_{\theta}(x_{n}))),\label{eq:PCA problem}
\end{equation}
in which, as explained, the encoder-decoder mapping $G_{\theta}(F_{\theta}(\cdot))$
is trained to reproduce the input at its output. 

In the absence of constraints on encoder and decoder, the problem
above trivially returns an identify mapping, i.e., $G_{\theta}(F_{\theta}(x))=x.$
In order to potentially learn a useful model, one should hence impose
constraints, such as dimensionality or sparsity, on the latent vector
$z$. Some notable examples are discussed next. 

\textbf{PCA.} PCA assumes linear encoder and decoder, and ties their
weight matrices via a transposition. Specifically, PCA sets the encoder
as $F_{\theta}(x)=W^{T}x$ and the decoder as $G_{\theta}(z)=Wz$.
The parameters $\theta$ are hence given by the $D\times M$ matrix
$W$. By these definitions, the $M$ columns of $W$ have the interpretation
of linear features as for PPCA. 

With a quadratic loss function, the learning problem (\ref{eq:PCA problem})
is given as
\begin{equation}
\underset{W}{\text{min}}\sum_{n=1}^{N}\Vert x_{n}-WW^{T}x_{n}\Vert^{2}.
\end{equation}
This problem can be solved in closed form. The solution is in fact
given by the $M$ principal eigenvectors of the sample covariance
matrix $N^{-1}\sum_{n=1}^{N}x_{n}x_{n}^{T}.$ Extensions of PCA that
use the kernel trick (Chapter 4) can be also devised (see, e.g., \cite{Murphy}).
Furthermore, PCA can be seen to be a special case of PPCA by setting
in the latter $\mu=0$ and by taking the limit $\sigma^{2}\rightarrow0$.

\textbf{Dictionary learning.} In dictionary learning, the decoder
is linear, i.e., $G_{\theta}(z)=Wz$, as for PCA. However, the encoder
is limited only by constraints on the set of feasible latent variables
$z$. A typical such constraint is sparsity: the latent vector should
have a small number of non-zero elements. Defining as $\mathcal{C}$
the set of feasible values for $z$, the dictionary learning problem
can be formulated as
\begin{equation}
\underset{W,\{z_{n}\}\in\mathcal{C}}{\text{min}}\frac{1}{N}\sum_{n=1}^{N}\Vert x_{n}-Wz_{n}\Vert^{2}.
\end{equation}
The name of the method accounts for the fact that matrix $W$ can
be thought of as the dictionary of $M$ features \textendash{} its
columns \textendash{} that are used to describe the data. The problem
above is typically solved using alternate optimization. Accordingly,
one optimizes over $W$ for a fixed set of latent vectors $\{z_{n}\}$,
which is a standard least squares problem; and the optimizes over
each latent vector $z_{n}$ for a fixed $W$. The second problem can
be tackled by using standard sparsity-based methods, such as LASSO
(Chapter 2).

\textbf{Multi-layer autoencoders.$^{*}$} One can also represent both
encoder and decoder as multi-layer neural networks. In this case,
the weights of encoder and decoders are typically tied to be the transpose
of one another, in a manner similar to PCA. Training is often done
first layer-by-layer, e.g., using RBM training, and then via backpropagation
across the entire network \cite{Hintoncourse}. 

\textbf{Denoising autoencoders.$^{*}$} An alternative approach to
facilitate the learning of useful features is that taken by \emph{denoising
autoencoders} \cite{vincent2010stacked}. Denoising autoencoders add
noise to each input $x_{n}$, obtaining a noisy version $\tilde{x}_{n}$,
and to then train the machine with the aim of recovering the input
$x_{n}$ from its noisy version $\tilde{x}_{n}$. Formally, this can
be done by minimizing the empirical risk $\sum_{n=1}^{N}\ell(x_{n},G_{\theta}(F_{\theta}(\tilde{x}_{n})))$.

\textbf{Probabilistic autoencoders.$^{*}$} Instead of using deterministic
encoder and decoder, it is possible to work with probabilistic encoders
and decoders, namely $p(z|x,\theta)$ and $p(x|z,\theta)$, respectively.
Treating the decoder as a variational distribution, learning can then
be done by a variant of the EM algorithm. The resulting algorithm,
known as Variational AutoEncoder (VAE), will be mentioned in Chapter
8.

\section{Ranking$^{*}$}

We conclude this chapter by briefly discussing the problem of ranking.
When one has available ranked examples, ranking can be formulated
as a supervised learning problem \cite{Shalev}. Here, we focus instead
on the problem of ranking a set of webpages based only on the knowledge
of their underlying graph of mutual hyperlinks. This set-up may be
considered as a special instance of unsupervised learning. We specifically
describe a representative, popular, scheme known as PageRank \cite{pagerank},
which uses solely the web of hyperlinks as a form of supervision signal
to guide the ranking.

To elaborate, we define the connectivity graph by including a vertex
for each webpage and writing the adjacent matrix as
\begin{equation}
L_{ij}=\begin{cases}
1 & \text{if page \ensuremath{j} links to page \ensuremath{i}}\\
0 & \text{otherwise}
\end{cases}.
\end{equation}
The outgoing degree of a webpage is hence given as
\begin{equation}
C_{j}=\sum_{i}L_{ij}.
\end{equation}
PageRank computes the rank $p_{i}$ of a webpage $i$ as 

\begin{equation}
p_{i}=(1-d)+d\sum_{j\neq i}\frac{L_{ij}}{C_{j}}p_{j},\label{eq:pagerank}
\end{equation}
where $0<d<1$ is a parameter. Hence, the rank of a page is a weighted
sum of a generic rank of equal to 1, which enables the choice of new
pages, and of an aggregate ``vote'' from other pages. The latter
term is such that any other page $j$ votes for page $i$ if it links
to page $i$ and its vote equals its own rank divided by the total
number of outgoing links, that is, $p_{j}/C_{j}$. In essence, a page
$i$ is highly ranked if it is linked by pages with high rank. The
equation (\ref{eq:pagerank}) can be solved recursively to obtain
the ranks of all pages. A variation of PageRank that tailors ranking
to an individual's preferences can be obtained as described in \cite{haveliwala2003topic}.

\section{Summary}

In this chapter, we have reviewed the basics of unsupervised learning.
A common aspect of all considered approaches is the presence of hidden,
or latent, variables that help explain the structure of the data.
We have first reviewed ML learning via EM, and variations thereof,
for directed and undirected models. We have then introduced the GAN
method as a generalization of ML in which the KL divergence is replaced
by a divergence measure that is learned from the data. We have then
reviewed discriminative models, which may be trained via the InfoMax
principle, and autoencoders. 

In the next section, we broaden the expressive power of the probabilistic
models considered so far by discussing the powerful framework of probabilistic
graphical models.

\part{Advanced Modelling and Inference}

\chapter{Probabilistic Graphical Models}

As we have seen in the previous chapters, probabilistic models are
widely used in machine learning. Using Fig. \ref{fig:FigmodelsCh6}
as an example, we have encountered both directed and undirected models,
which have been used to carry out supervised and unsupervised learning
tasks. Graphical models encode structural information about the rvs
of interest, both observed and latent. They hence provide a principled
way to define parametric probabilistic models with desired features. 

The selection of a probabilistic graphical model hence follows the
same general rules that have been discussed so far: A more specialized,
or structured, model may help reduce overfitting, and hence the generalization
gap. This is done, as we will see, by reducing the number of parameters
to be learned. On the flip side, specialization may come at the cost
of an irrecoverable bias.

In this chapter, we provide an introduction to the vast field probabilistic
graphical models, which is a powerful framework that allows us to
represent and learn structured probabilistic models. The goal here
is to introduce the main concepts and tools, while referring to the
extensive treatments in \cite{Koller,Barber,Murphy,Wainwright} for
additional information.

\section{Introduction}

In this section, we start by discussing two examples that illustrate
the type of structural information that can be encoded by means of
probabilistic graphical models. We then provide an overview of this
chapter. 

As illustrated by the two examples below, structured probabilistic
models can be used to set up parametric models for both supervised
and unsupervised learning. In the former case, all variables are observed
in the training set, with some rvs being inputs, i.e., covariates
($x$), and others being considered as outputs, or targets ($t$).
In contrast, in the latter case, some variables are unobserved and
play the role of latent variables ($z$) that help explain or generate
the observed variables ($x$).\footnote{Strictly speaking, this distinction applies to the frequentist approach,
since in the Bayesian approach the model parameters are always treated
as unobserved rvs.}

\begin{example}Consider the tasks of text classification via supervised
learning or text clustering via unsupervised learning. In the supervised
learning case, the problem is to classify documents depending on their
topic, e.g., sport, politics or entertainment, based on set of labelled
documents. With unsupervised learning, the problem is to cluster documents
according to the similarity of their contents based on the sole observation
of the documents themselves. 

A minimal model for this problem should include a variable $\mathrm{\ensuremath{t}}$
representing the topic and a variable $\mathrm{x}$ for the document.
The topic can be represented by a categorical variable taking $T$
values, i.e.., $\mathrm{t}\in\left\{ 1,...,T\right\} ,$ which is
observed for supervised learning and latent for unsupervised learning.
As for the document, with ``bag-of-words'' encoding, a set of $W$
words of interest is selected, and a document is encoded as a $W\times1$
binary vector $x=[x_{1},...,x_{W}]^{T}$, in which $x_{w}=1$ if word
$w$ is contained in the document.\footnote{Note that $W$ here does not represent a matrix of weights!}

To start, we could try to use an unstructured directed model defined
as\begin{subequations}
\begin{align}
 & \mathrm{t\sim}\textrm{Cat}(\pi)\\
 & \mathrm{x}|\mathrm{t}=t\sim\textrm{Cat}(\pi_{t}),
\end{align}
\end{subequations}where the parameter vector includes the $T\times1$
probability vector $\pi$ and the $T$ probability vectors $\pi_{t},$
one for each class, each of dimension $2^{W}$. Note, in fact, that
the vector $\mathrm{x}$ can take $2^{W}$ possible values. As such,
this model would require to learn $(T-1)+T(2^{W}-1)$ parameters,
which quickly becomes impractical when the number $W$ of words of
interest is large enough. Furthermore, as we have seen, a learned
model with a large number of parameters is bound to suffer from overfitting
if the available data is relatively limited. 

Instead of using this unstructured model, we can adopt a model that
encodes some additional assumptions that can be reasonably made on
the data. Here is one possible such assumption: once the topic is
fixed, the presence of a word is independent on the presence of other
words. The resulting model is known as Bernoulli\emph{ naive Bayes},
and can be described as follows\begin{subequations}\label{eq:naiveBern}
\begin{align}
\mathrm{} & \mathrm{t}\sim\textrm{Cat}(\pi)\\
 & \mathrm{x}|\mathrm{t}=t\sim\prod_{w=1}^{W}\textrm{Bern}(x_{w}|\pi_{w|t}),
\end{align}
\end{subequations}with parameter vector including the $T\times1$
probability vector $\pi$ and $T$ sets of $W$ probabilities $\pi_{w|t}$,
$w=1,...,W$, for each topic $t$. Parameter $\pi_{w|t}$ represents
the probability of word $w$ occurring in a document of topic $t$.
The mentioned independence assumption hence allowed us to bring the
number of parameters down to $(T-1)+TW$, which corresponds to an
exponential reduction. \begin{figure}[H] 
\centering\includegraphics[scale=0.7]{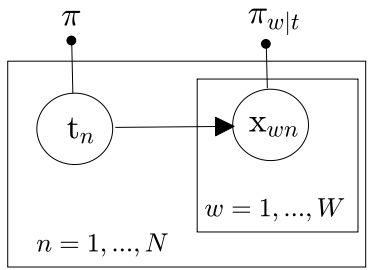}
\caption{BN for the naive Bayes model using the plate notation. Learnable parameters are represented as dots.} \label{fig:FigBN1Ch7} \end{figure}

The naive Bayes model can be represented graphically by the BN illustrated
in Fig. \ref{fig:FigBN1Ch7}, where we have considered $N$ i.i.d.
documents. Note that the graph is directed: in this problem, it is
sensible to model the document as being \emph{caused} by the topic,
entailing a directed causality relationship. Learnable parameters
are represented as dots. BNs are covered in Sec. \ref{sec:Bayesian-Networks}.
\end{example}

\begin{figure}[H] 
\centering\includegraphics[scale=0.7]{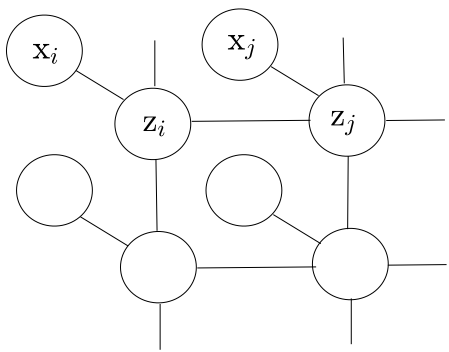}
\caption{MRF for the image denoising example. Only one image is shown and the learnable parameters are not indicated in order to simplify the illustration.} \label{fig:FigMRF1Ch7} \end{figure}

\begin{example}The second example concerns image denoising using
supervised learning. For this task, we wish to learn a joint distribution
$p(x,z|\theta)$ of the noisy image $\mathrm{x}$ and of the corresponding
desired noiseless image $\mathrm{z}$. We encode the images using
a matrix representing the numerical values of the pixels. A structured
model in this problem can account for the following reasonable assumptions:
(\emph{i}) neighboring pixels of the noiseless image are correlated,
while pixels further apart are not directly dependent on one another;
and (\emph{ii}) noise acts independently on each pixel of the noiseless
image to generate the noisy image. These assumptions are encoded by
the MRF shown in Fig. \ref{fig:FigMRF1Ch7}. Note that this is an
undirected model. This choice is justified by the need to capture
the mutual correlation among neighboring pixels, which cannot be described
as a directed causality relationship. We will study MRFs in Sec. \ref{sec:Markov-Random-Fields}.
\end{example}

As suggested by the examples above, structure in probabilistic models
can be conveniently represented in the form of graphs. At a fundamental
level, structural properties in a probabilistic model amount to \emph{conditional
independence} assumptions. For instance, in the naive Bayes model,
the word indicators are conditionally independent given the topic.
As we will see in the rest of this chapter, conditional independence
assumptions translate into factorizations of the joint distributions
of the variables under study. Factorizations, and associated conditional
independence properties, can be represented by three different graphical
frameworks, namely BNs, MRFs and factor graphs. For brevity, this
chapter will only focus on the first two. 

\section{Bayesian Networks\label{sec:Bayesian-Networks}}

This section provides a brief introduction to BNs by focusing on key
definitions and on the problem of ML learning with some note on MAP
and Bayesian viewpoints. 

\subsection{Definitions and Basics}

BNs encode a probability factorization or, equivalently, a set of
conditional independence relationships through a directed graph. The
starting point is the chain rule for probabilities for a generic set
of $K$ rvs $\{\mathrm{x}_{1},...,\mathrm{x}_{K}\}$:
\begin{align}
p(x_{1},...,x_{k}) & =p(x_{1})p(x_{2}|x_{1})...p(x_{K}|x_{1},...,x_{K-1}),\nonumber \\
 & =\prod_{k=1}^{K}p(x_{k}|x_{1},...,x_{k-1}),\label{eq:chain rule}
\end{align}
where the order of the variables is arbitrary. The factorization ($\ref{eq:chain rule}$)
applies for a generic joint distribution, and it does not encode any
additional structural information. Note that the notation here is
general and not meant to indicate that the variables are necessary
observed.

\begin{example}Consider again the naive Bayes model for text classification/
clustering. There, we imposed the structural constraint that word
indicator variables $\{\mathrm{x}_{w}\}_{w=1}^{W}$ be conditionally
independent given the topic $\mathrm{t}$. This conditional independence
assumption can be expressed using the ``perp'' notation
\begin{equation}
\mathrm{x}_{w}\perp\{\mathrm{x}_{w'}\}_{w'\neq w}|\mathrm{t},
\end{equation}
or the Markov chain notation
\begin{equation}
\mathrm{x}_{w}-\mathrm{t}-\{\mathrm{x}_{w'}\}_{w'\neq w}.
\end{equation}
Mathematically, this condition means that we have $p(x_{w}|t,\{x_{w'}\})=p(x_{w}|t)$,
where $\{x_{w'}\}$ is any subset of the variables $\{x_{w'}\}_{w'\neq w}$.
Applying the chain rule to the rvs $(\{\mathrm{x}_{w}\}_{w=1}^{W},\mathrm{t})$
using order $\mathrm{t},\mathrm{x}_{1},...,\mathrm{x}_{W}$ (or any
other order on the $\{\mathrm{x}_{w}\}_{w=1}^{W}$ variables), we
can hence write
\begin{equation}
p(x,t)=p(t)\prod_{w=1}^{W}p(x_{w}|t).\label{eq:naive bayes again}
\end{equation}

This factorization is represented by the BN in Fig. \ref{fig:FigBN1Ch7}.
In the directed graph shown in the figure, each vertex corresponds
to a rv, and a directed edge is included from $\mathrm{t}$ to each
variable $\mathrm{x}_{w}$. This edge captures the fact that the conditional
probability of the variable $\mathrm{x}_{w}$ in (\ref{eq:naive bayes again})
is conditioned on rv $\mathrm{t}$. Informally, $\mathrm{t}$ ``causes''
all variables in vector $\mathrm{x}.$ The graph accounts for multiple
i.i.d. realization $(x_{\mathcal{D}},t_{\mathcal{D}})=\{\mathrm{x}_{n},\mathrm{t}_{n}\}_{n=1}^{N}$,
where we denote the $n$th sample as $\mathrm{x}_{n}=[\mathrm{x}_{1n}\cdots\mathrm{x}_{Wn}]^{T}$.
The joint distribution factorizes as 
\begin{equation}
p(x_{\mathcal{D}},t_{\mathcal{D}})=\prod_{n=1}^{N}p(t_{n})\prod_{w=1}^{W}p(x_{wn}|t_{n}).
\end{equation}
The BN uses the tile notation that will be formalized below to indicate
multiple independent realizations of rvs with the same distribution.\end{example}

Generalizing the example above, we define BNs as follows.

\begin{definition}A BN is a directed acyclic graph (DAG)\footnote{In a DAG, there are no directed cycles, that is, no closed paths following
the direction of the arrows.}, whose vertices represent rvs $\{\mathrm{x}_{1},....,\mathrm{x}_{K}\}$
with an associated joint distribution that factorizes as
\begin{equation}
p(x_{1},...,x_{K})=\underset{k=1}{\prod^{K}}p(x_{k}|x_{\mathcal{P}(k)})\label{eq:definition BN}
\end{equation}
where $\mathcal{P}(k)$ denotes the set of parents of node $k$ in
the DAG. In a BN, rvs are represented by full circles, while learnable
parameters defining the conditional distributions are represented
by dots.\end{definition}

As per the definition, the parents $\mathrm{x}_{\mathcal{P}(k)}$
of a rv $\mathrm{x}_{k}$ in the DAG account for the statistical dependence
of $\mathrm{x}_{k}$ with all the preceding variables $\mathrm{x}_{1},...,\mathrm{x}_{k-1}$
according to the selected order. That is, the BN encodes the local
conditional independence relationships 
\begin{equation}
\mathrm{x}_{k}\perp\{\mathrm{x}_{1},...,\mathrm{x}_{k-1}\}|\mathrm{x}_{\mathcal{P}(k)}
\end{equation}
using the ``perp'' notation, or equivalently $\mathrm{x}_{k}-\mathrm{x}_{\mathcal{P}(k)}-\{x_{1},...,x_{k-1}\}$
using the Markov chain notation, for $k=1,...,K$. As seen in Fig.
\ref{fig:FigBN1Ch7}, \emph{plates} are used to represent independent
replicas of a part of the graph. 

\textbf{When to use BNs.} BNs are suitable models when one can identify
causality relationships among the variables. In such cases, there
exists a natural order on the variables, such that rvs that appear
later in the order are caused by a subset of the preceding variables.
The causing rvs for each rv $\mathrm{x}_{k}$ are included in the
parent set $\mathcal{P}(k)$, and are such that, when conditioning
on rvs $\mathrm{x}_{\mathcal{P}(k)}$, rv $\mathrm{x}_{k}$ is independent
on all other preceding rvs $\{\mathrm{x}_{1},...,\mathrm{x}_{k-1}\}.$
BNs also underlie the framework of interventions that allows the assessment
of causality, as opposed to mere correlation, among observed variables,
as briefly discussed in Sec. \ref{sec:Interpretation-and-Causality}
\cite{Pearl}. 

\textbf{Sampling from a BN.} The causality relationship among the
ordered variables $\{\mathrm{x}_{1},...,\mathrm{x}_{K}\}$ encoded
by a BN makes it easy, at least in principle, to draw samples from
a BN. This can be done by using \emph{ancestral sampling}: Generate
rv $\mathrm{x}_{1}\sim$$p(x_{1})$; then, $\mathrm{x}_{2}\sim p(x_{2}|x_{\mathcal{P}(2)})$;
and so on, with rv $\mathrm{x}_{k}$ being generated as $\mathrm{x}_{k}\sim p(x_{k}|x_{\mathcal{P}(k)})$.

\begin{example}Hidden Markov Models (HMMs) are used to study time
series, or more generally sequential data, measured through a memoryless
transformation, such as an additive noise channel. Mathematically,
HMMs can be represented by two sets of variables: the underlying sequence
$\mathrm{z}_{1},\mathrm{z}_{2},...,\mathrm{z}_{D},$ and the measured
``noisy'' data $\mathrm{x}_{1},\mathrm{x}_{2},...,\mathrm{x}_{D}$.
HMMs encode two assumptions: (\emph{i}) each sample $z_{i}$ depends
on the past samples only through the previous sample $\mathrm{z}_{i-1}$;
and (\emph{ii}) each measured data $\mathrm{x}_{i}$ depends only
on $\mathrm{z}_{i}$. Assumption (\emph{i}) makes process $\mathrm{z}_{1},\mathrm{z}_{2}$,...
a Markov chain. 

Using the order $\mathrm{z}_{1},\mathrm{x}_{1},\mathrm{z}_{2},\mathrm{x}_{2},...$,
we can write the joint distribution as
\begin{equation}
p(x,z)=p(z_{1})p(x_{1}|z_{1})\prod_{i=1}^{D}p(z_{i}|z_{i-1})p(x_{i}|z_{i})\label{eq:HMM}
\end{equation}
by enforcing the local independencies $\mathrm{z}_{i}-\mathrm{z}_{i-1}-\{\mathrm{z}_{1},...,\mathrm{z}_{i-2},\mathrm{x}_{1},...,\mathrm{x}_{i-2}\}$
and $\mathrm{x}_{i}-\mathrm{z}_{i}-\{\mathrm{z}_{1},...,\mathrm{z}_{i-1},\mathrm{x}_{1},...,\mathrm{x}_{i-1}\}.$
This factorization is represented by the BN in Fig. \ref{fig:FigBN2Ch7}.
While not indicated in the figure for clarity, an important aspect
of HMMs is that the learnable parameters defining the transitions
probabilities $p(z_{i}|z_{i-1})$ and the transformations $p(x_{i}|z_{i})$
are respectively imposed to be equal, or tied, for all $i$, hence
reducing the number of parameters to be leaned.

Among the many relevant examples of applications of HMMs (see \cite{Koller,Murphy,Barber,rabiner1986introduction}),
we mention here text autocorrection, in which the underlying sequential
data $\mathrm{z}_{1},\mathrm{z}_{2},...$ amount to the correct text
while the measured data $\mathrm{x}_{1},\mathrm{x}_{2},...$ to the
typed text; and speech recognition, in which the underlying time series
$\mathrm{z}_{1},\mathrm{z}_{2},...$ is a sequence of words and the
transformation to the measured recorded speech $\mathrm{x}_{1},\mathrm{x}_{2},...$
translates words into sounds. 

\begin{figure}[H] 
\centering\includegraphics[scale=0.7]{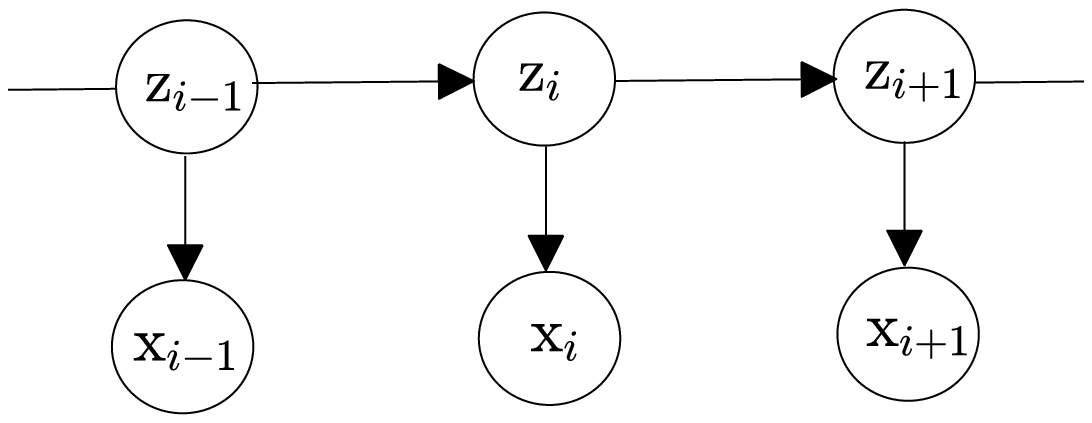}
\caption{BN representing an HMM. The lernable parameters are not explicitly indicated.} \label{fig:FigBN2Ch7} \end{figure}

In supervised learning applications, both sequences are observed in
the training data, with the goal of learning to predict the sequence
$\mathrm{z}_{1},\mathrm{z}_{2},...,$ for new measured data points
$\mathrm{x}_{1},\mathrm{x}_{2},...$ via to the trained model. This
task, in this context, is also known as \emph{denoising}, since\emph{
}$\mathrm{x}_{1},\mathrm{x}_{2},...$ can be thought of as a noisy
version of $\mathrm{z}_{1},\mathrm{z}_{2},...$ With unsupervised
learning, only the sequence $\mathrm{x}_{1},\mathrm{x}_{2},...$ is
observed. \end{example}

\begin{figure}[H] 
\centering\includegraphics[scale=0.7]{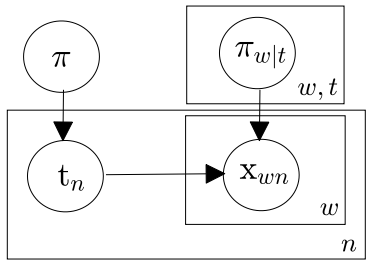}
\caption{BN for the Bayesian version of the naive Bayes model (\ref{eq:naive Bayes param}). Hyperparameters are not indicated and tiles indices are marked without their ranges for clarity.} \label{fig:FigBN3Ch7} \end{figure}

\begin{example}This example demonstrates how easily Bayesian modelling
can be incorporated in a BN. To this end, consider again the Bernoulli
naive Bayes model (\ref{eq:naiveBern}) for text classification. Taking
a Bayesian viewpoint, parameters $\pi$ and $\{\pi_{w|t}\}$ are to
be considered as rvs. We further assume them to be a priori independent,
which is known as the \emph{global independence assumption}. As a
result, the joint probability distribution for each document factorizes
as
\begin{align}
p(x,t,\pi,\pi_{w|t}|\alpha,a,b) & =\text{Dir}(\pi|\alpha)\textrm{\ensuremath{\prod_{t=1}^{T}\prod_{w=1}^{W}\textrm{Beta}(\pi_{w|t}|a,b)}}\nonumber \\
 & \times\textrm{Cat}(t|\pi)\prod_{w=1}^{W}\textrm{Bern}(x_{w}|\pi_{w|t}).\label{eq:naive Bayes param}
\end{align}
In the factorization above, we have made the standard assumption of
a Dirichlet prior for the probability vector $\pi$ and a Beta prior
for parameters $\{\pi_{w|t}\}$, as discussed in Chapter 3. The quantities
$\alpha,a,b$ are hyperparameters. Note that the hyperparameters $(a,b)$
are shared for all variables $\pi_{w|t}$ in this example. The corresponding
BN is shown in Fig. \ref{fig:FigBN3Ch7}, which can be compared to
Fig. \ref{fig:FigBN1Ch7} for the corresponding frequentist model.\end{example}

We invite the reader to consider also the Latent Dirichlet Allocation
(LDA) model and the other examples available in the mentioned textbooks.

\subsection{Global Conditional Independence}

As we have discussed, BNs are defined by local conditional independence
properties that are encoded in the factorization of a joint distribution
and in the supporting DAG. BNs can also be used to assess generic
global conditional independence queries of the type: Is a given subset
of variables $\mathcal{A}$ independent of another set $\mathcal{B}$
conditioned on a third subset $\mathcal{C}$ of variables? The $d$-separation
algorithm outputs either a positive response or a ``maybe'' answer
to this question. It is briefly described as follows:

\indent $\bullet$ Build a subgraph $\mathcal{G}'$ from the original
DAG by keeping all vertices in the subsets $\mathcal{A}$, $B$ and
$\mathcal{C}$, as well as all the edges and vertices encountered
by moving backwards on the DAG one or more edges from the vertices
in $\mathcal{A}$, $B$ and $\mathcal{C}$;

\indent $\bullet$ Build a subgraph $\mathcal{G}''$ from $\mathcal{G}'$
by deleting all edges coming out of the vertices in $\mathcal{C}$;

\indent $\bullet$ If there no path, neglecting the directionality
of the edges, between a node in $\mathcal{A}$ and a node in $\mathcal{B}$,
then the conditional independence relation $\mathcal{A}\perp\mathcal{B}|\mathcal{C}$
holds. Else, if one such path exists, then there is at least one joint
distribution that factorizes as for the given DAG for which the condition
$\mathcal{A}\perp\mathcal{B}|\mathcal{C}$ does not hold.

\begin{example}\label{ex:V}Consider the so-called V-structure $\mathrm{x}\rightarrow\mathrm{y}\leftarrow\mathrm{z}$.
Using $d$-separation, it can be seen that the conditional independence
$x-y-z$ does not hold in general.\end{example}

\subsection{Learning }

Assume that the DAG of a BN is given. Structure learning, that is,
the decision of which edges should be included in the graph based
on available training data, is also an important problem that will
not be considered here. Making explicit the dependence of the probability
factors on learnable parameters $\mu$, the joint distribution encoded
by a BN can be written as
\begin{equation}
p(x_{1},...,x_{K})=\underset{k=1}{\prod^{K}}p(x_{k}|x_{\mathcal{P}(k)},\mu_{k|x_{\mathcal{\mathcal{P}}(k)}}),
\end{equation}
where $\mu_{k|x_{\mathcal{\mathcal{P}}(k)}}$ are the parameters defining
the conditional distribution $p(x_{k}|x_{\mathcal{P}(k)})$. Note
that the parameters $\mu_{k|x_{\mathcal{\mathcal{P}}(k)}}$ are generally
different for different values of $k$ and of the parents' variables
$x_{\mathcal{\mathcal{P}}(k)}$ (see, e.g., (\ref{eq:naive Bayes param})).
In most cases of interest, the probability distribution $p(x_{k}|x_{\mathcal{P}(k)},\mu_{k|x_{\mathcal{P}(k)}})$
is in the exponential family or is a GLM (see Chapter 3).

As we have already seen in previous examples, the parameters $\mu_{k|x_{\mathcal{P}(k)}}$
can either be \emph{separate}, that is, distinct for each $k$ and
each value of $x_{\mathcal{P}(k)}$, or they can be \emph{tied}. In
the latter case, some of the parameters $\mu_{k|x_{\mathcal{P}(k)}}$
are constrained to be equal across different values of $x_{\mathcal{P}(k)}$
and/or across different values of $k$. As a special case of tied
parameters, the value of $\mu_{k|x_{\mathcal{P}(k)}}$ may also be
independent of $x_{\mathcal{P}(k)}$, such as in the case for GLMs. 

As for the data, we have seen that the rvs $\mathrm{x}_{1},...,\mathrm{x}_{K}$
can be either fully observed in the training set, as in supervised
learning, or they can be partially observed as in unsupervised learning.

For the sake of brevity, here we describe learning only for the case
of fully observed data with separate parameters, and we briefly mention
extensions to the other cases. 

\subsubsection{Fully Observed Data with Separate Parameters}

We are given a fully observed data set $\mathcal{D}=\{x_{n}\}_{n=1}^{N}$,
with each data point written as $x_{n}=[x_{1n},...,x_{Kn}]^{T}$.
For concreteness, assume that all variables are categorical. Denoting
as $\mathrm{x}_{\mathcal{P}(k)n}$ the parents of variable $\mathrm{x}_{kn}$,
the LL function can be factorized as:\begin{subequations}
\begin{align}
\ln p(\mathcal{D}|\mu) & =\sum_{n=1}^{N}\sum_{k=1}^{K}\ln p(x_{kn}|x_{\mathcal{P}(k)n},\mu_{k|x_{\mathcal{P}(k)n}})\\
 & =\sum_{k=1}^{K}\sum_{n=1}^{N}\ln p(x_{kn}|x_{\mathcal{P}(k)n},\mu_{k|x_{\mathcal{P}(k)n}})\\
 & =\sum_{k=1}^{K}\sum_{x_{\mathcal{P}(k)}}\sum_{n\in\mathcal{N}_{x_{\mathcal{P}(k)}}\textrm{ }}\ln p(x_{kn}|x_{\mathcal{P}(k)},\mu_{k|x_{\mathcal{P}(k)}}),\label{eq:LLlearningBNCh7}
\end{align}
\end{subequations}where we have defined the set of indices
\begin{equation}
\mathcal{N}_{x_{\mathcal{P}(k)}}=\{n:\textrm{ }x_{\mathcal{P}(k)n}=x_{\mathcal{P}(k)}\}.
\end{equation}
This set includes the indices $n$ for which the parents $\mathrm{x}_{\mathcal{P}(k)n}$
of node $k$ take a specific vector of values $x_{\mathcal{P}(k)}$.
In (\ref{eq:LLlearningBNCh7}), the inner sum depends only on the
parameters $\mu_{k|x_{\mathcal{P}(k)}}$corresponding to the given
value $x_{\mathcal{P}(k)}$ of the rvs $\mathrm{x}_{\mathcal{P}(k)n}$
for $n=1,...,N$. Therefore, in the case of separate parameters, the
ML estimate for each parameter $\mu_{k|x_{\mathcal{P}(k)}}$ can be
carried out independently. While this simplifies the problem, it may
also cause overfitting issues owing to the problem of data fragmentation,
as each parameter is estimated based only on a fraction of the data
set. 

As an even more concrete example, consider binary variables modelled
as $\mathrm{x}_{k}\sim\textrm{Bern}$$(\mu_{k|x_{\mathcal{P}(k)}}).$
Accordingly, the probability of each rv depends on the value $x_{\mathcal{P}(k)}$
of the parents. For this case, we can write the ML estimate as
\begin{equation}
\hat{\mu}_{k|x_{\mathcal{P}(k)},ML}=\frac{\sum_{n\in\mathcal{N}_{x_{\mathcal{P}(k)}}}x_{kn}}{\left|\mathcal{N}_{x_{\mathcal{P}(k)}}\right|}.
\end{equation}

\begin{figure}[H] 
\centering\includegraphics[scale=0.7]{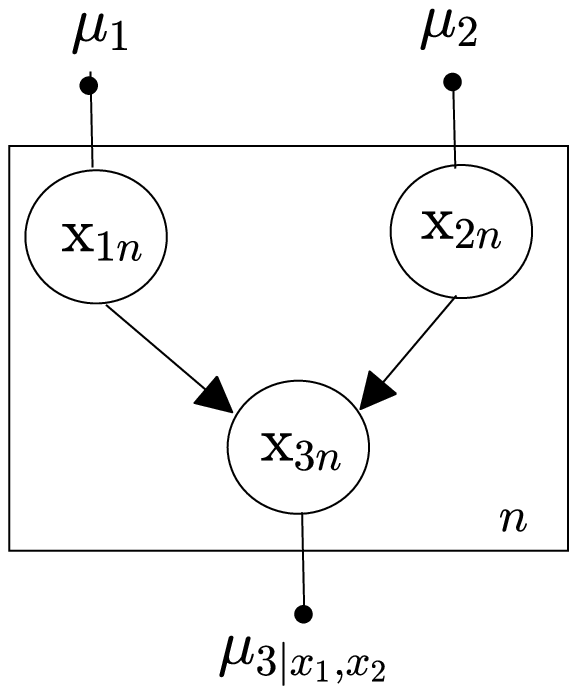}
\caption{BN for Example \ref{ex:ExampleBN}.} \label{fig:FigBN4Ch7} \end{figure}

\begin{example}\label{ex:ExampleBN}Consider the BN shown in Fig.
\ref{fig:FigBN4Ch7} with binary rvs in the alphabet $\{0,1\}$. The
observed data $\mathcal{D}$ is given as
\[
\left\{ \underset{10}{\underbrace{\left(\begin{array}{c}
1\\
0\\
1
\end{array}\right)}},\underset{14}{\underbrace{\left(\begin{array}{c}
0\\
1\\
1
\end{array}\right)}},\underset{8}{\underbrace{\left(\begin{array}{c}
1\\
1\\
0
\end{array}\right)}},\underset{12}{\underbrace{\left(\begin{array}{c}
0\\
0\\
0
\end{array}\right)}},\underset{1}{\underbrace{\left(\begin{array}{c}
1\\
0\\
0
\end{array}\right)}},\underset{2}{\underbrace{\left(\begin{array}{c}
0\\
1\\
0
\end{array}\right)}},\underset{1}{\underbrace{\left(\begin{array}{c}
1\\
1\\
1
\end{array}\right)}},\underset{2}{\underbrace{\left(\begin{array}{c}
0\\
0\\
1
\end{array}\right)}}\right\} ,
\]
where the bottom row indicates the number of observations equal to
the vector above it. The ML estimates are: $\mu_{1,ML}=\frac{10+8+1+1}{50}=\frac{20}{50}=\frac{2}{5}$,
$\mu_{2,ML}=\frac{14+8+2+1}{50}=\frac{1}{2}$, $\hat{\mu}_{3|00}=\frac{2}{12+2}=\frac{1}{7},$
$\hat{\mu}_{3|11}=\frac{1}{8+1}=\frac{1}{9}$, $\hat{\mu}_{3|01}=\frac{14}{14+2}=\frac{7}{8}$
and $\hat{\mu}_{3|10}=\frac{10}{10+1}=\frac{10}{11}$.\end{example}

MAP estimates can be seen to decompose in a similar way under global
independence assumptions on the parameters, and the same holds for
Bayesian approach \cite{Koller}.

\subsubsection{Notes on the General Case}

With shared parameters, obtaining ML and MAP estimates require aggregating
statistics across all variables that share the same parameters. The
Bayesian approach is more complex, and we refer to \cite{Koller}
for discussion. An alternative approach with ``soft sharing'' is
the hierarchical Bayes model (see \cite[Fig. 17.11]{Koller}). In
the presence of missing data, learning typically involves the EM algorithm
described in Chapter 6 or approximate learning counterparts to be
introduced in the next chapter.

\section{Markov Random Fields\label{sec:Markov-Random-Fields}}

In this section, we turn to MRF by following the same approach as
for BNs in the previous section.

\subsection{Definitions and Basics}

As BNs, MRFs encode a probability factorization or, equivalently,
a set of conditional independence relationships. They do so, however,
through an undirected graph. 

\begin{definition}A MRF is an undirected graph, whose vertices represent
rvs with associated joint distribution that factorizes as
\begin{equation}
p(x)=\frac{1}{Z}\underset{c}{\prod}\psi_{c}(x_{c}),\label{eq:MRF}
\end{equation}
where $c$ is the index of a clique\footnote{A clique is a fully connected subgraph.}
in the graph; $\mathrm{x}_{c}$ is the set of rvs associated with
the vertices in clique $c$; $\psi_{c}(x_{c})\geq0$ is the factor
or potential for clique $c$; and $Z=\underset{x}{\sum}\underset{c}{\prod}\psi_{c}(x_{c})$
is the partition function. With no loss of generality, the sum can
be limited to maximal cliques (i.e., cliques that are not fully included
in a larger clique). In a MRF, rvs are represented by full circles,
while learnable parameters defining the conditional distributions
are represented by dots. \end{definition}

Each factor $\psi_{c}(x_{c})$ in (\ref{eq:MRF}) encodes the compatibility
of the values $x_{c}$ in each clique, with larger values of $\psi_{c}(x_{c})$
corresponding to configurations $x_{c}$ that are more likely to occur.
Factors are generally not probability distributions, i.e., they are
not normalized to sum or integrate to one. Furthermore, unlike BNs,
each factor does not distinguish between conditioning and conditioned
rvs. Rather, all variables in the clique $x_{c}$ can play the same
role in defining the value of the potential $\psi_{c}(x_{c})$. 

\textbf{When to use MRFs.} This discussion points to the fact that
MRFs are especially well suited to model mutual relationships of compatibility,
or lack thereof, among variables, rather than causality effects. 

\textbf{Evaluating Probabilities and Sampling from an MRF}. This distinguishing
feature of MRF, which brings potential modelling advantages in some
applications, comes with the added difficulty of \emph{evaluating}
the joint probability distribution and\emph{ sampling} from the joint
distribution. In fact, the computation of the probability (\ref{eq:MRF})
requires the calculation of the partition function $Z$, which is
generally intractable when the alphabet of the vector $\mathrm{x}$
is large enough. This is typically not the case for BNs, in which
each conditional probability is conventionally selected from a known
(normalized) distribution. Furthermore, unlike BNs, MRFs do not allow
ancestral sampling, as all the conditional distributions of the individual
rvs in $\mathrm{x}$ are generally tied together via the partition
function.

\begin{example}\label{ex:Ising}The image denoising example described
at the beginning of this chapter leverages the MRF in Fig. \ref{fig:FigMRF1Ch7}.
In it, the maximal cliques are given by the pairs of rvs \{$\mathrm{z}_{i},\mathrm{z}_{j}$\}
and $\{\mathrm{x}_{i},\mathrm{z}_{i}\}$ that are connected by an
edge. Associating a factor or potential to each clique yields the
factorized joint distribution
\begin{equation}
p(x,z)=\frac{1}{Z}\underset{\{i,j\}}{\prod}\psi_{i,j}(z_{i},z_{j})\cdot\underset{i}{\prod}\psi_{i}(z_{i},x_{i}),
\end{equation}
where $\{i,j\}$ represents an edge of the undirected graph. As a
notable example, in the \emph{Ising model}, the variables are bipolar,
i.e., $\mathrm{z}_{i},\mathrm{x}_{i}\in\{-1,+1$\}, and the potentials
are defined as $\psi_{ij}(z_{i},z_{j}|\eta_{1})=\textrm{exp}(-E(z_{i},z_{j}|\eta_{1}))$
and $\psi_{i}(z_{i},x_{i}|\eta_{2})=\textrm{exp}(-E(z_{i},x_{i}|\eta_{2}))$,
with energy functions
\begin{equation}
E(z_{i},z_{j}|\eta_{1})=-\eta_{1}z_{i}z_{j}\textrm{ and }E(z_{i},x_{i}|\eta)=-\eta_{2}z_{i}x_{i}.
\end{equation}
From this definition, a large natural parameter $\eta_{1}>0$ yields
a large probability \textendash{} or a low energy \textendash{} when
$z_{i}$ and $z_{j}$ are equal; and, similarly, a large $\eta_{2}>0$
favors configurations in which $z_{i}=x_{i}$, that is, with low noise.\end{example}

\begin{example}Another related example is given by the RBMs studied
in Sec. \ref{subsec:Restricted-Boltzmann-Machines}, whose undirected
graph is shown in Fig. \ref{fig:FigRBMCh6}. \end{example}

As illustrated by the previous examples, potentials are typically
parameterized using the energy-based form
\begin{equation}
\psi_{c}(x_{c})=\textrm{exp}(-E_{c}(x_{c}|\eta_{c})),\label{eq:energybasedpotential}
\end{equation}
with parameter vector $\eta_{c}.$ This form ensures that the factors
are strictly positive as long as the energy is upper bounded. A special
class of such models is given by log-linear models, such as the Ising
model, in which, as seen in Chapter 3, the energy is a linear function
of the parameters.

\subsection{Global Conditional Independence}

MRF, in a manner similar to BNs, enable the assessment of conditional
independence properties globally in the graph. The procedure is, in
fact, easier with MRFs thanks to the Hammersley\textendash Clifford
theorem. The latter says that, if the potentials $\psi_{c}(x_{c})$
are strictly positive for all cliques $c$, as for the energy-based
potentials (\ref{eq:energybasedpotential}), the conditional independence
relationship $\mathcal{A}\perp\mathcal{B}|\mathcal{C}$ can be tested
via the following simple algorithm.

\indent $\bullet$ Eliminate all variables in $\mathcal{C}$ and all
the connected edges;

\indent $\bullet$ If there is no path between rvs in $\mathcal{A}$
and $\mathcal{B}$, then the relationship $\mathcal{A}\perp\mathcal{B}|\mathcal{C}$
holds; if, instead, there is a path, then there exists at least one
joint distribution that factorizes as for the undirected graph at
hand for which the relationship $\mathcal{A}\perp\mathcal{B}|\mathcal{C}$
does not hold.

\subsection{Learning }

Learning in MRFs is made complicated by the partition function. In
fact, the partition function couples together all the parameters.
This makes it impossible to carry out separately the estimate of the
parameters $\eta_{c}$ associated with each clique even in the fully
observed case with separate parameters. Nevertheless, MRFs with energy-based
potentials (\ref{eq:energybasedpotential}) fall either in the exponential
family, when the factors are log-normal, or in the more general class
of energy-based models. In both cases, gradient-based algorithms can
be devised by using methods similar to those used in Sec. \ref{subsec:Restricted-Boltzmann-Machines}
for RBMs.

\subsection{Converting BNs to MRFs\label{subsec:Convering-BNs-to} }

As suggested by the discussion above, BNs and MRFs are suitable to
encode different types of statistical dependencies, with the former
capturing causality while the latter accounting for mutual compatibility.
There are in fact conditional independence properties that can be
expressed by BN or MRF but not by both. An example is the V-structure
$\mathrm{x}\rightarrow\mathrm{y}\leftarrow\mathrm{z}$ discussed in
Example \ref{ex:V}, whose independencies cannot be captured by an
MRF.

That said, given a BN with factorization (\ref{eq:definition BN}),
we can define potential functions
\begin{equation}
\psi_{k}(x_{k},x_{\mathcal{P}(k)})=p(x_{k}|x_{\mathcal{P}(k)})
\end{equation}
to obtain the factorization
\begin{equation}
p(x)=\underset{k=1}{\prod^{K}}\psi_{k}(x_{k},x_{\mathcal{P}(k)})
\end{equation}
with partition function $Z=1$. This factorization defines an MRF
in which each maximal clique contains a rv $\mathrm{x}_{k}$ and its
parents $\mathrm{x}_{\mathcal{P}(k)}$. The corresponding undirected
graph can be directly obtained from the DAG that defines the BN via
the following two steps: 

\indent $\bullet$ Connect all pairs of parents by an undirected edge
\textendash{} this step is known as ``moralization'';

\indent $\bullet$ Make all edges undirected.

As per the discussion above, the resulting MRF may not account for
all the independencies encoded in the original graph. This can be
easily seen by applying the procedure to the V-structure.

\section{Bayesian Inference in Probabilistic Graphical Models\label{sec:Bayesian-Inference-in}}

Bayesian inference amounts to the computation of the posterior probability
of unobserved, or latent, variables given observed variables. In this
regard, it is useful to differentiate between \emph{intensive} and
\emph{extensive} latent variables. Intensive latent variables are
model parameters whose number does not increase with the number $N$
of data points, such as the probability vectors $(\pi$,$\{\pi_{w|t}\})$
in the Bernoulli naive Bayes model. Extensive latent variables are
instead rvs indexed by the example index $n$, whose number grows
with the sample size $N$. These correspond to the latent variables
$\mathrm{z}_{n}$ introduced in the previous chapter. 

Bayesian inference is a fundamental task that lies at the heart of
both inference and learning problems. As seen in Chapter 2, it underlies
supervised learning for generative probabilistic models, which requires
the computation of the predictive probability $p(t|x,\theta)$ for
the new sample ($\mathrm{x}$,$\mathrm{t}$) from the learned model
$p(x,t\text{|}\theta)$. It is also at the core of Bayesian supervised
learning, which evaluates the posterior $p(\theta|\mathcal{D})$ of
the parameter vector $\theta$ (an intensive variable) in order to
obtain the predictive posterior $p(t|x,\mathcal{D})=\int p(\theta|\mathcal{D})p(t|x,\theta)d\theta$
for the new sample ($\mathrm{x},\mathrm{t}$). As studied in Chapter
6, Bayesian inference is a key step in unsupervised learning even
under a frequentist viewpoint, since the EM algorithm requires the
evaluation of the posterior $p(z_{n}|x_{n},\theta$) of the latent
(extensive) variables $\{z_{n}\}$ given the current iterate $\theta.$

As discussed, when performing Bayesian inference, we can distinguish
between observed variables, say $\mathrm{x}$, and latent variables
$\mathrm{z}$. In general, only a subset of latent variables may be
of interest, say $\mathrm{z}_{i}$, with the rest of the rvs in $\mathrm{z}$
being denoted as $\mathrm{z}_{-i}.$ The quantity to be computed is
the posterior distribution
\begin{equation}
p(z_{i}|x)=\frac{p(x,z_{i})}{p(x)},\label{eq:posteriorBayesCh7}
\end{equation}
where 
\begin{align}
p(x,z_{i}) & =\sum_{z_{-i}}p(x,z)\label{eq:maginalization Bayes}
\end{align}
and
\begin{equation}
\textrm{ \ensuremath{p(x)=\sum_{z_{i}}p(x,z_{i})}},\label{eq:maginalization Bayes 1}
\end{equation}
with the sum being replaced by an integral for continuous variables.
The key complication in evaluating these expressions is the need to
sum over potentially large sets, namely the domains of variables $\mathrm{z}_{-i}$
and $\mathrm{z}_{i}$. Note that the sum in (\ref{eq:maginalization Bayes}),
which appears at the numerator of (\ref{eq:posteriorBayesCh7}), is
over all hidden variables that are of no interest. In contrast, the
sum in (\ref{eq:maginalization Bayes 1}), which is at the denominator
of (\ref{eq:posteriorBayesCh7}), is over the variables whose posterior
probability (\ref{eq:posteriorBayesCh7}) is the final objective of
the calculation.

The complexity of the steps (\ref{eq:maginalization Bayes}) and (\ref{eq:maginalization Bayes 1})
is exponential in the respective numbers of latent variables over
which the sums are computed, and hence it can be prohibitive.

\begin{example}Consider an HMM, whose BN is shown in Fig. \ref{fig:FigBN2Ch7}.
Having learned the probabilistic model, a typical problem is that
of inferring a given hidden variable $\mathrm{z}_{i}$ given the observed
variables $\mathrm{x=\{x}_{1},....,\mathrm{x}_{D}\}$. Computing the
posterior $p(z_{i}|x)$ requires the evaluation of the sums in (\ref{eq:maginalization Bayes})
and (\ref{eq:maginalization Bayes 1}). When the hidden variables
$\mathrm{z}_{1},...,\mathrm{z}_{D}$ are discrete with alphabet size
$\mathcal{Z}$, the complexity of step (\ref{eq:maginalization Bayes})
is of the order $\left|\mathcal{Z}\right|{}^{D-1}$, since one needs
to sum over the $\left|\mathcal{Z}\right|{}^{D-1}$ possible values
of the hidden variables. \end{example}

The structure encoded by probabilistic graphic models can help reduce
the discussed complexity of Bayesian inference. Therefore, probabilistic
graphical models can not only enhance learning by controlling the
capacity of the model, but also enable Bayesian inference. To elaborate,
consider a joint distributions defined by an MRF as in (\ref{eq:MRF}).
Note that, as we have seen in Sec. \ref{subsec:Convering-BNs-to},
one can easily convert BNs into MRFs, although possibly at the cost
of not preserving some linear independence relationship.  With this
factorization, the marginalizations (\ref{eq:maginalization Bayes})-(\ref{eq:maginalization Bayes 1})
require solving the so-called \emph{sum-product inference task}
\begin{equation}
\underset{z}{\sum}\underset{c}{\prod}\psi_{c}(x_{c}),\label{eq:sum-productCh7}
\end{equation}
where the variables $\mathrm{z}$ are a subset of the variables in
x. 

As an important observation, formulation (\ref{eq:sum-productCh7})
highlights the fact that the problem of computing the partition function
$Z$ for MRFs is a special case of the sum-product inference task.
In fact, in order to compute $Z$, the sum in (\ref{eq:sum-productCh7})
is carried out over all variables in $\mathrm{x}$.

When the undirected graph describing the joint distribution is a tree\footnote{In a tree, there is only one path between any pairs of nodes (no loops).},
the complexity of sum-product inference becomes exponential only in
the maximum number of variables in each factor, also known as \emph{treewidth}
of the graph. In this case, the sum-product inference problem can
be exactly solved via \emph{message passing} \emph{belief propagation}
over the factor graph associated to the MRF. We refer to the textbooks
\cite{Koller,Barber,Murphy} for details on factor graphs and belief
propagation.

\begin{example}The MRF associated with an HMM is obtained from the
BN in Fig. \ref{fig:FigBN2Ch7} by simply substituting directed for
undirected edges, and the distribution (\ref{eq:HMM}) factorizes
as
\begin{equation}
p(x,z)=\psi(z_{1})\psi(x_{1},z_{1})\prod_{i=1}^{D}\psi(z_{i},z_{i-1})\psi(x_{i},z_{i}).\label{eq:HMM-1}
\end{equation}
The undirected graph is a tree with treewidth equal to 2, since, as
per (\ref{eq:HMM-1}), there are at most two variables in each clique.
Therefore, belief propagation allows to evaluate the posteriors $p(z_{i}|x)$
with a complexity of the order $\left|\mathcal{Z}\right|{}^{2}$,
which does not scale exponentially with the number $D$ of time samples.\end{example}

When the undirected graph is not a tree, one can use the\emph{ junction
tree algorithm} for exact Bayesian inference. The idea is to group
subsets of variables together in cliques, in such a way that the resulting
graph is a tree. The complexity depends on the treewidth of the resulting
graph. When this complexity is too high for the given application,
approximate inference methods are necessary. This is the subject of
the next chapter.

\section{Summary}

Probabilistic graphical models encode a priori information about the
structure of the data in the form of causality relationships \textendash{}
via directed graphs and BNs \textendash{} or mutual affinities \textendash{}
via undirected graphs and MRFs. This structure translates into conditional
independence conditions. The structural properties encoded by probabilistic
graphical models have the potential advantage of controlling the capacity
of a model, hence contributing to the reduction of overfitting at
the expense of possible bias effects (see Chapter 5). They also facilitate
Bayesian inference (Chapters 2-4), at least in graphs with tree-like
structures. Probabilistic graphical models can be used as the underlying
probabilistic framework for supervised, unsupervised, and semi-supervised
learning problems, depending on which subsets of rvs are observed
or latent. 

While graphical models can reduce the complexity of Bayesian inference,
this generally remains computationally infeasible for most models
of interest. To address this problem, the next chapter discusses approximate
Bayesian inference, as well as associated learning problems (Chapter
6).

\chapter{Approximate Inference and Learning}

In Chapters 6 and 7, we have seen that learning and inference tasks
are often made difficult by the need to compute the posterior distribution
$p(z|x)$ of an unobserved variables $\mathrm{z}$ given an observed
variables $\mathrm{x}$. This task requires the computation of the
normalizing marginal
\begin{equation}
p(x)=\sum_{z}p(x,z),\label{eq:sum Ch8}
\end{equation}
where the sum is replaced by an integral for continuous variables.\footnote{Note that this task subsumes (\ref{eq:maginalization Bayes}) after
appropriate redefinitions of the variables. } This computation is intractable when the alphabet of the hidden variable
$z$ is large enough. Chapter 7 has shown that the complexity of computing
(\ref{eq:sum Ch8}) can be alleviated in the special case in which
the factorized joint distribution $p(x,z)$ is defined by specific
classes of probabilistic graphical models. 

What to do when the complexity of computing (\ref{eq:sum Ch8}) is
excessive? In this chapter, we provide a brief introduction to two
popular approximate inference approaches, namely MC methods and Variational
Inference (VI). We also discuss their application to learning. As
for the previous chapter, the reader is referred to \cite{Koller,Barber,Murphy,Wainwright}
for details and generalizations (see also \cite{Schniterreview}).

\section{Monte Carlo Methods}

To introduce MC methods, we start by observing that the expression
(\ref{eq:sum Ch8}) can be rewritten as the ensemble average
\begin{equation}
p(x)=\sum_{z}p(z)p(x|z)=\textrm{E}_{\mathrm{z}\sim p(z)}[p(x|\mathrm{z})]\label{eq:ens av Ch8}
\end{equation}
over the latent rvs $\mathrm{z}\text{\ensuremath{\sim}\ensuremath{p(z)}}$.
The general idea behind MC methods is replacing ensemble averages
with empirical averages over randomly generated samples. In the most
basic incarnation of MC, $M$ i.i.d. samples $\mathrm{z}_{m}\sim p(z)$,
$m=1,...,M,$ are generated from the marginal distribution $p(z)$
of the latent variables, and then the ensemble average (\ref{eq:ens av Ch8})
is approximated by the empirical average
\begin{equation}
p(x)\simeq\frac{1}{M}\sum_{m=1}^{M}p(x|\mathrm{z}_{m}).\label{eq:basic MC}
\end{equation}
By the law of large numbers, we know that this estimate is consistent,
in the sense that it tends with probability one to the ensemble average
(\ref{eq:ens av Ch8}) when $M$ is large. Furthermore, the error
of the approximation scales as $1/\sqrt{M}$. 

\begin{example}Consider the Ising model for image denoising introduced
in Example \ref{ex:Ising}. We recall that the joint distribution
can be factorized as
\begin{equation}
p(x,z)=\frac{1}{Z}\underset{\{i,j\}}{\prod}\psi_{i,j}(z_{i},z_{j})\cdot\underset{i}{\prod}\psi_{i}(z_{i},x_{i}),\label{eq:Ising model Ch8}
\end{equation}
where $\{i,j\}$ represents an edge of the undirected graph, with
energy-based potentials $\psi_{ij}(z_{i},z_{j})=\textrm{exp}(\eta_{1}z_{i}z_{j})$
and $\psi_{i}(z_{i},x_{i})=\textrm{exp}(\eta_{2}z_{i}x_{i}).$ We
have removed the dependence of the potentials on the (natural) parameters
$\eta_{1}$ and $\eta_{2}$ for simplicity of notation. In order to
compute the posterior $p(z|x),$ which may be used for image denoising,
the MC approach (\ref{eq:basic MC}) requires to sample from the marginal
$p(z)$. This is, however, not easy given the impossibility to perform
ancestral sampling over MRFs as discussed in Sec. \ref{sec:Markov-Random-Fields}.\end{example}

\textbf{Importance Sampling.} As seen in the previous example, the
MC procedure explained above is not always feasible in practice, since
drawing samples from the marginal $p(z)$ may not be tractable. For
instance, the marginal $p(z)$ may not be known or it may be difficult
to draw samples from it. In such common cases, one can instead resort
to a simplified distribution $q(z)$ from which sampling is easy.
This distribution typically has convenient factorization properties
that enable ancestral sampling. 

The starting observation is that the marginal distribution (\ref{eq:sum Ch8})
can be expressed as an ensemble average over a rv $\mathrm{z}\sim q(z)$
as
\begin{equation}
\begin{aligned}p(x) & =\sum_{z}p(z)p(x|z)\frac{q(z)}{q(z)}\\
 & =\sum_{z}q(z)\frac{p(z)}{q(z)}p(x|z)\\
 & =\textrm{E}_{\mathrm{z}\sim q(z)}\left[\frac{p(\mathrm{z})}{q(\mathrm{z})}p(x|\mathrm{z})\right],
\end{aligned}
\end{equation}
as long as the support of distribution $q(z)$ contains that of $p(z)$.
This expression suggests the following empirical estimate, which goes
by the name of Importance Sampling: Generate $M$ i.i.d. samples $\mathrm{z}_{m}\sim q(z)$,
$m=1,...,M,$ and then compute the empirical approximation
\begin{equation}
p(x)\simeq\frac{1}{M}\sum_{m=1}^{M}\frac{p(\mathrm{z}_{m})}{q(\mathrm{z}_{m})}p(x|\mathrm{z}_{m}).
\end{equation}
This estimate is again consistent, but its variance depends on how
well $q(z)$ approximates $p(z)$. Note this approach requires knowledge
of the marginal $p(z)$ but not the ability to sample from it.

\textbf{Markov Chain Monte Carlo (MCMC) via Gibbs Sampling.} Rather
than drawing samples from a distribution that mimics $p(z)$ in order
to compute an approximation of the posterior $p(z|x)=p(x,z)/p(x)$,
MCMC methods aim at obtaining samples $\{\mathrm{z}_{m}$\} directly
from the posterior $p(z|x)$. With such samples, one can compute empirical
approximations of any ensemble average with respect to $p(z|x)$.
This is sufficient to carry out most tasks of interest, including
the expectation needed to evaluate the predictive posterior in Bayesian
methods for supervised learning or the average energy in the EM algorithm. 

MCMC methods generate a sequence of correlated samples $\mathrm{z}_{1},\mathrm{z}_{2},...$
from an easy-to-sample Markov chain $\mathrm{z}_{1}-\mathrm{z}_{2}-...$
that has the key property of having the desired distribution $p(z|x)$
as the stationary distribution. Such a Markov chain can be designed
automatically once a BN or MRF factorization of the joint distribution
is available. To this end, Gibbs sampling samples sequentially subsets
of rvs. For each subset, the sampling distribution is obtained by
normalizing the product of all factors including the rv being sampled
(see, e.g., \cite{Koller}). 

The mechanical nature of this procedure makes it a \emph{universal},
or \emph{black-box}, inference method, in the sense that it can be
applied to any typical probabilistic graphical model in an automatic
fashion. This has led to the recent emergence of \emph{probabilistic
programming}, whereby Bayesian inference is automatically performed
by software libraries that are given as input a probabilistic graphical
model for the joint distribution (see, e.g., \cite{probprog,Pyroai}).

MC methods are often used in combination with VI, as discussed in
Sec. \ref{sec:Monte-Carlo-Based-Variational}.

\section{Variational Inference}

The general idea behind VI is to replace the ensemble average in (\ref{eq:ens av Ch8})
with a suitable optimization that returns an approximation of the
posterior distribution $p(z|x)$. Specifically, VI methods introduce
an additional distribution on the hidden variables $\mathrm{z}$ that
is optimized in order to approximate the desired posterior $p(z|x)$. 

\textbf{I-projection.} We start with the observation that the solution
to the optimization problem 
\begin{equation}
\textrm{\ensuremath{\underset{q(z)}{\textrm{min}}} KL}(q(z)||p(z|x))\label{eq:I projection-1}
\end{equation}
for a fixed value $x,$ yields the unique solution $q(z)=p(z|x)$
if no constraints are imposed on $q(z)$. This is due to Gibbs' inequality
(\ref{eq:GibbsineqCh2}). This result is, by itself, not useful, since
evaluating the KL divergence $\textrm{KL}(q(z)||p(z|x))$ requires
knowledge of $p(z|x)$, which is exactly what we are after. 

However, the equality between (\ref{eq:ELBO3}) and (\ref{eq:ELBO4}),
namely
\begin{equation}
\textrm{KL}(q(z)||p(x,z))=\textrm{KL}(q(z)||p(z|x))-\ln p(x),
\end{equation}
demonstrates that problem (\ref{eq:I projection-1}) is equivalent
to solving
\begin{equation}
\textrm{\ensuremath{\underset{q(z)}{\textrm{min}}}}\textrm{ KL}(q(z)||p(x,z)),\label{eq:I projection}
\end{equation}
where, by (\ref{eq:ELBO1}), we can write 
\begin{equation}
\textrm{KL}(q(z)||p(x,z))=-\textrm{E}_{\mathrm{z}\sim q(z)}[\ln p(x,\mathrm{z})]-H(q).\label{eq:var free energy}
\end{equation}
In words, solving problem (\ref{eq:I projection-1}) is equivalent
to minimizing the \emph{variational free energy or Gibbs free energy}
(\ref{eq:var free energy}) \textendash{} or the negative of the ELBO.
A key feature of this alternative formulation is that it does not
require knowledge of the unavailable posterior $p(z|x)$. 

From the derivation above, solving problem (\ref{eq:I projection})
exactly without imposing any constraint on $q(z)$ would yield the
desired posterior $p(z|x)$ as the output. The key idea of VI is to
choose a parametric form $q(z|\varphi)$ for the variational posterior
that enables the solution of problem
\begin{equation}
\begin{aligned}\underset{\varphi}{\text{min}}\text{ KL}(q(z|\varphi)||p(x,z)).\end{aligned}
\label{eq:I projection true}
\end{equation}
By the discussion above, this is equivalent to minimizing $\text{KL}(q(z|\varphi)||p(z|x))$,
despite the fact that $p(z|x)$ is not known. 

The solution $q(z|\varphi^{*})$ to problem (\ref{eq:I projection true})
is known as\emph{ I-projection} of the distribution $p(z|x)$ in the
set of distributions $\{q(z|\varphi)\}$ defined by the given parametrization.
The I-projection can be taken as an estimate of the posterior $p(z|x)$.
In fact, if the parametrized family $\{q(z|\varphi)\}$ is rich enough
to contain distributions close to the true posterior $p(z|x)$, the
minimization (\ref{eq:I projection true}) guarantees the approximate
equality $q(z|\varphi^{*})\simeq p(z|x)$. 

In order to ensure the feasibility of the optimization (\ref{eq:I projection true}),
the parametrized distribution $q(z|\varphi)$ is typically selected
to have a convenient factorization and to have factors with tractable
analytical forms, such as members of the exponential family or GLMs
\cite{beal2003variational}. 

\textbf{Amortized VI.$^{*}$} The variational posterior $q(z|\varphi)$
obtained from the I-projection (\ref{eq:I projection true}) depends
on the specific value of the observed variables $\mathrm{x}=x$. Problem
(\ref{eq:I projection true}) is, in fact, solved separately for each
value $\mathrm{x}=x$. A potentially more efficient solution is to
define an \emph{inference variational distribution }$q(z|x,\varphi)$,
which models the posterior distribution of $\mathrm{z}$ for any value
of $\mathrm{x}=x$. The inference distribution is parametrized by
a vector $\varphi$, and is typically implemented using a multi-layer
neural network. In this case, it is typically referred to as \emph{inference
network}. This approach is referred to as amortized VI. 

Amortized VI has the key advantage that, once the inference distribution
is learned, one does not need to carry out I-projections for previously
unobserved values of $\mathrm{x}=x$. Instead, one can directly apply\emph{
}$q(z|x,\varphi)$ for the learned values of parameter $\varphi$
\cite{VAE}.

The inference distribution $q(z|x,\varphi)$ can be obtained by solving
the \emph{amortized I-projection} problem
\begin{equation}
\textrm{\ensuremath{\underset{\varphi}{\textrm{min}}} E\ensuremath{_{\mathrm{x}\sim p(x)}[\textrm{KL}(q(z|\mathrm{x,\varphi})||p(\mathrm{x},z))]}},\label{eq:I projection-1-2}
\end{equation}
where the ensemble average is, in practice, replaced by an empirical
average over available data points $\{x_{n}\}$. The solution of the
VI problem (\ref{eq:I projection-1-2}) is hence ``amortized'' across
multiple values of $\mathrm{x}$.

\textbf{M-projection.$^{*}$ }By Theorem \ref{thm:ELBO}, the I-projection
maximizes a lower bound \textendash{} the ELBO \textendash{} on the
log-distribution of the observed data $\mathrm{x}$. This gives I-projections
a strong theoretical justification grounded in the ML learning principle.
Recalling that the KL divergence is not symmetric (see Sec. \ref{sec:Information-Theoretic-Metrics}
and Appendix A), one could also define the alternative problem
\begin{equation}
\textrm{\ensuremath{\underset{\varphi}{\textrm{min}}} KL}(p(z|x)||q(z|\varphi)).\label{eq:I projection-1-1}
\end{equation}
The solution $q(z|\varphi^{*})$ to this problem is known as\emph{
M-projection} of the distribution $p(z|x)$ in the set of distributions
$\{q(z|\varphi)\}$ defined by the given parametrization. 

As for the counterpart (\ref{eq:I projection-1}), this problem does
not appear to be solvable since it requires knowledge of the desired
posterior $p(z|x)$. However, the problem turns out to have an easy
solution if $q(z|\varphi)$ belongs to the exponential family. In
fact, the gradient with respect to the natural parameters $\varphi$
of the KL divergence in (\ref{eq:I projection-1}) can be computed
by following the same steps detailed for ML learning in Sec. \ref{sec:Frequentist-Learning Ch3}.
The upshot of this computation and of the enforcement of the optimality
condition is that the M-projection is obtained by \emph{moment matching.}
Specifically, one needs to find a value of parameter $\varphi$ such
that the expectations of the sufficient statistics of the model $q(z|\varphi)$
under distribution $q(z|\varphi)$ match with the same expectations
under the true distribution $p(z|x)$. 

In mathematical terms, the M-projection in the exponential family
model $q(z|\varphi)\propto\text{exp}(\varphi^{T}u(z))$, with sufficient
statistics $u(z)$, yields natural parameters $\varphi^{*}$ that
satisfy the moment matching condition
\begin{equation}
\textrm{E}_{\mathrm{z}\sim p(z|x)}[u(\mathrm{z})]=\textrm{E}_{\mathrm{z}\sim q_{(z|\varphi^{*})}}[u(\mathrm{z})].\label{eq:M proj optimality}
\end{equation}
This derivation is detailed in the Appendix of this chapter. Amortized
inference can be defined in a similar way as for I-projection. 

\begin{figure}[H] 
\centering\includegraphics[scale=0.65]{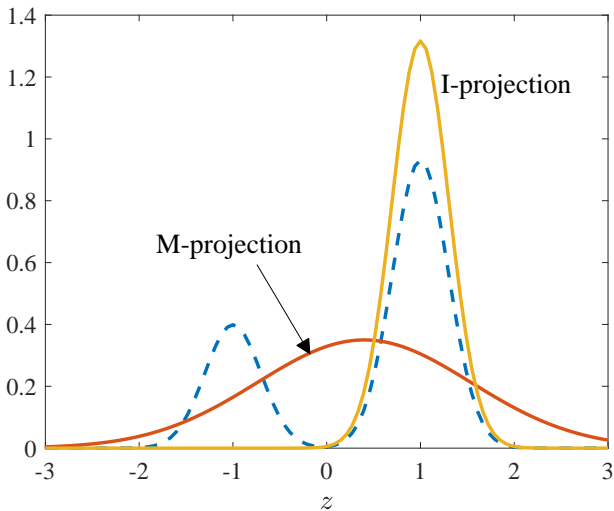}
\caption{Example of I- and M-projections of a mixture of Gaussians distribution (dashed line).\label{fig:FigIMCh8}} \end{figure}

\begin{example}This simple example is meant to provide an intuitive
comparison between the approximations produced by I- and M-projections.
To this end, consider a mixture of Gaussians distribution
\begin{equation}
\text{\ensuremath{p(z|x)=0.3\mathcal{N}}}(z|\mu_{1}=-1,\sigma_{1}^{2}=0.3)+0.7\mathcal{N}(z|\mu_{2}=1,\sigma_{2}^{2}=0.3),
\end{equation}
as shown in Fig. \ref{fig:FigIMCh8}. Note that this example is clearly
idealized, since in practice the conditional distribution $p(z|x)$
is not known. Assume the variational distribution $q(z|\varphi)=\mathcal{N}(z|m,\gamma^{2})$
with variational parameters $\varphi=(m,\gamma^{2})$. The M-projection
returns the moment matching estimates $m=\textrm{E}_{\mathrm{z\sim}p(z|x)}[\mathrm{z}]=0.4$
and $\gamma^{2}=\textrm{var}_{\mathrm{z\sim}p(z|x)}[\mathrm{z}]=0.3((\mu_{1}-m)^{2}+\sigma_{1}^{2})+0.7((\mu_{2}-m)^{2}+\sigma_{2}^{2})=1.93$
for $i=1,2.$ Instead, the I-projection can be computed numerically,
yielding $m=1$ and $\gamma^{2}=0.3$. The I- and M-projections are
also plotted in Fig. \ref{fig:FigIMCh8}. \end{example}

The previous example illustrates a few important facts about I- and
M-projections. First, the I-projection tends to be \emph{mode-seeking}
and \emph{exclusive}. Mathematically, this is because the variational
posterior $q(z|\varphi)$ determines the support over which the distributions
$p(z|x)$ and $q(z|\varphi)$ are compared by the KL divergence. Therefore,
I-projections tend to underestimate the variance of a distribution\footnote{See \cite{turner2011two} for an example that demonstrates the limitations of this general statement.}.
Furthermore, the I-projection is generally more accurate where $p(z|x)$
is larger. In contrast, the M-projection tends to be \emph{inclusive}
and to span the entire support of $p(z|x)$. This is because the M-projection
prefers to avoid zero values for $q(z|\varphi)$ at values of $z$
such that $p(z|x)\ne0$ in order to avoid an infinite KL divergence.
We refer to Fig. \ref{fig:figmixGaussCh6} for a related example.

\textbf{$\alpha-$Divergence.$^{*}$} As already discussed in Sec.
\ref{subsec:GAN}, the KL divergence is only one among many possible
ways to define a measure of distance between two distributions. A
metric that has found useful applications in the context of VI is
the $\alpha$-divergence introduced in \cite{minka2005divergence}.
The $\alpha$-divergence between two distributions $p(x)$ and $q(x)$
is defined as 
\begin{equation}
D_{\alpha}(p||q)=\frac{\sum_{x}\alpha p(x)+(1-\alpha)q(x)-p(x)^{\alpha}q(x)^{1-\alpha}}{\alpha(1-\alpha)},\label{eq:alpha divergence Ch8}
\end{equation}
where $p$ and $q$ need not be normalized, and $\alpha$ is a parameter.
It can be proved that, as $\alpha\rightarrow0$, we obtain $D_{\alpha}(p||q)=\textrm{KL}(q||p)$,
and, when $\alpha\rightarrow1$, we have $D_{\alpha}(p||q)=\textrm{KL}(p||q).$
Consistently with the discussion about I- and M-projections in the
previous example, performing projections of the type $\textrm{min}_{\varphi}D_{\alpha}(p(x|z)||q(z|\varphi))$
with $\alpha\leq0$ and decreasing values of $\alpha$ yields an increasingly
mode-seeking, or exclusive, solution; while increasing values of $\alpha\geq1$
yield progressively more inclusive and zero-avoiding solutions (see
\cite[Fig. 1]{minka2005divergence}). Finally, for all $\alpha\neq0$,
it can be proved that the stationary points of the projection $\textrm{min}_{\varphi}D_{\alpha}(p(x|z)||q(z|\varphi))$
coincide with those of the projection $\textrm{min}_{\varphi}\textrm{KL}(p(x|z)||p(x|z)^{\alpha}q(z|\varphi)^{1-\alpha})$.
The $\alpha$-divergence can be further generalized as discussed in
Appendix A.

\subsection{Mean Field Variational Inference}

Mean Field VI (MFVI) is a VI method that leads to a universal, or
black-box, Bayesian inference technique, such as Gibbs sampling for
MC techniques. MFVI assumes the factorization $q(z)=\prod_{j}q(z_{j})$,
and performs an I-projection iteratively for one hidden variable $\mathrm{z}_{i}$
at a time, while fixing the factors $q_{j}(z_{j})$ for the other
latent variables $\mathrm{z}_{j}$ with $j\neq i$. No constraints
are imposed on each factor $q_{j}(z_{j}).$ This corresponds to tackling
the I-projection problem using block coordinate descent within the
given factorized family of distributions.

For each factor $q_{i}(z_{i})$, the I-projection problem minimizes
the variational free energy
\begin{equation}
-\textrm{E}_{\mathrm{z}_{i}\sim q_{i}(z_{i})}\left[\textrm{E}_{j\neq i}[\text{ln}p(x,\mathrm{z})]\right]-\sum_{j}H(q_{j}(z_{j})),
\end{equation}
where we have defined for brevity the expectation 
\begin{equation}
\textrm{E}_{j\neq i}[\text{ln}p(x,\mathrm{z})]=\textrm{E}_{\{\mathrm{z}_{j}\}_{j\ne i}\sim\prod_{j\ne i}q_{j}(z_{j})}[\text{ln}p(x,z_{i},\{\mathrm{z}_{j}\}_{j\ne i})].
\end{equation}
Neglecting constants independent of $q_{i}(z_{i})$, this problem
is equivalent to the minimization
\begin{equation}
\underset{q_{i}}{\text{min}}\text{ }\text{KL}(q_{i}(z_{i})||\text{exp}(\textrm{E}_{j\neq i}[\text{ln}p(x,\mathrm{z})])).
\end{equation}
The solution to this problem can be easily seen to be obtained by
normalizing the right-hand argument of the divergence as
\begin{equation}
q_{i}(z_{i})=\frac{\text{exp}(\textrm{E}_{j\ne i}[\text{ln}p(x,\mathrm{z})])}{\sum_{z_{i}}\text{exp}(\textrm{E}_{j\ne i}[\text{ln}p(x,\mathrm{z})])}.\label{eq: 13}
\end{equation}

MFVI solves the system of equations \eqref{eq: 13} for all $i$ by
cycling through the factors $q_{i}(z_{i})$ or by choosing them randomly.
Note that, as discussed more generally above, this procedure does
not require knowledge of the desired posterior $p(z|x)$ but only
of the joint distribution $p(x,z)$. The MFVI iterations are guaranteed
to converge to a stationary point of the KL minimization problem.

It remains to discuss how to evaluate the expectations in (\ref{eq: 13}).
To this end, let us assume that the joint distribution $p(x,z)$ factorizes
as $p(x,z)=Z^{-1}\prod_{c}\psi_{c}(x_{c},z_{c})$ as for MRFs or BNs
\textendash{} recall that in the latter case, we have $Z=1$. The
MFVI equation (\ref{eq: 13}) can be written as
\begin{equation}
\begin{aligned}q_{i}(z_{i})\propto & \text{ exp}\left(\textrm{E}_{j\ne i}\left[\sum_{c}\text{ln}\psi_{c}(x_{c},\mathrm{z}_{c})\right]\right)\\
= & \exp\left(\textrm{E}_{j\ne i}\left[\sum_{c:\textrm{ \ensuremath{z_{i}\in z_{c}}}}\text{ln}\psi_{c}(x_{c},\mathrm{z}_{c})\right]\right).
\end{aligned}
\label{eq:MFVI MRF}
\end{equation}
In the second line we have used the fact that it is sufficient to
consider only the factors corresponding to cliques that include $z_{i}$.
This enables implementations by means of local message passing (see,
e.g., \cite{Koller}). Furthermore, additional simplifications are
possible when the factors $\psi_{c}(x_{c},z_{c})$ are log-linear,
as illustrated by the next example.

\begin{figure}[H] 
\centering\includegraphics[scale=0.65]{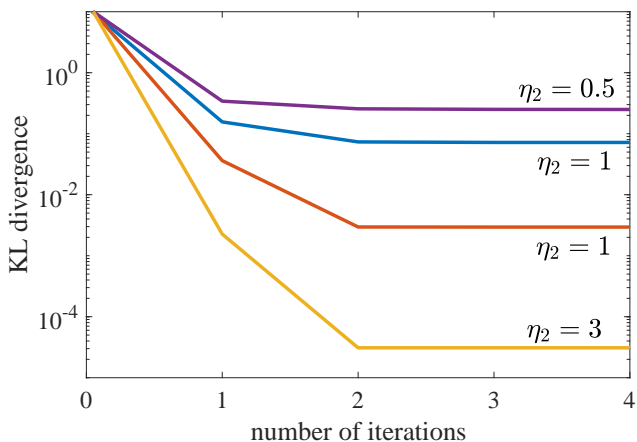}
\caption{KL divergence between actual posterior and mean-field approximation as a function of the number of iterations of MFVI ($\eta_1=0.15$).\label{fig:FigKLCh8}} \end{figure}

\begin{example}Consider again the Ising model (\ref{eq:Ising model Ch8}).
The MFVI equations (\ref{eq:MFVI MRF}) for approximating the posterior
$p(z|x)$ are given as
\begin{equation}
\begin{aligned}q_{i}(z_{i})\propto & \exp\Big(\eta_{1}\sum_{\{i,j\}}\textrm{E}_{\mathrm{z}_{j}\sim q(z_{j})}[z_{i}\mathrm{z}_{j}]+\eta_{2}x_{i}z_{i}\Big)\\
= & \exp\Big(z_{i}\Big(\eta_{1}\sum_{\{i,j\}}\textrm{E}_{\mathrm{z}_{j}\sim q(z_{j})}[\mathrm{z}_{j}]+\eta_{2}x_{i}\Big)\Big),
\end{aligned}
\end{equation}
 and hence, upon normalization, we have 
\begin{align}
q_{i}(z_{i} & =1)=\frac{1}{1+\exp(-2(\eta_{1}\sum_{\left\{ i,j\right\} }\mu_{j}+\eta_{2}x_{i}))},\label{eq:meanfieldIsingCh8}\\
 & \textrm{ }=\sigma\left(2\left(\eta_{1}\sum_{\left\{ i,j\right\} }\mu_{j}+\eta_{2}x_{i}\right)\right),\nonumber 
\end{align}
where $\mu_{i}=q_{i}(z_{i}=1)-q_{i}(z_{i}=-1)=2q_{i}(z_{i}=1)-1$.

For a numerical example, consider a $4\times4$ binary image $\mathrm{z}$
observed as matrix $\mathrm{x}$, where the joint distribution of
$\mathrm{x}$ and $\mathrm{z}$ is given by the Ising model. Note
that, according to this model, the observed matrix $\mathrm{x}$ is
such that each pixel of the original matrix $\mathrm{z}$ is flipped
independently with probability $\sigma(-2\eta_{2})$. In this small
example, it is easy to generate an image $x$ distributed according
to the model, as well as to compute the exact posterior $p(z|x)$
by enumeration of all possible images $z$. The KL divergence $\textrm{KL}(p(x|z)||\prod_{i}q_{i}(z_{i}))$
obtained at the end of each iteration of MFVI \textendash{} with one
iteration applying (\ref{eq:meanfieldIsingCh8}) one by one to all
variables \textendash{} is shown in Fig. \ref{fig:FigKLCh8} for $\eta_{1}=0.15$
and various values of $\eta_{2}.$ As $\eta_{2}$ increases, the posterior
distribution tends to become deterministic, since $x$ is an increasingly
accurate measurement of $z$. As a result, the final mean-field approximation
is more faithful to the real posterior, since a product distribution
can capture a deterministic pmf. For smaller values of $\eta_{2},$
however, the bias due to the mean-field assumption yields a significant
floor on the achievable KL divergence.\end{example}

\textbf{Beyond MFVI.$^{*}$} MFVI assumes a fully factorized variational
distribution $q(z)$. It is also possible to develop methods that
are based on the same factorization as the joint distribution $p(x,z)$.
This is known as the \emph{Bethe approach}, and yields Loopy Belief
Propagation (LBP) as a specific solution technique. LBP can also be
interpreted as the application of message passing belief propagation,
which was described in Sec. \ref{sec:Bayesian-Inference-in}, to factor
graphs with loops. Simplifications of loopy belief propagation include
Approximate Message Passing (AMP). When applied to M-projections with
factors restricted to lie in the exponential family, the approach
yields the \emph{Expectation Propagation} method. We refer to \cite{Koller,Schniterreview}
for details.

\section{Monte Carlo-Based Variational Inference\label{sec:Monte-Carlo-Based-Variational}$^{*}$}

The VI methods described in the previous section require the evaluation
of expectations with respect to the variational posterior (see, e.g.,
(\ref{eq: 13})). The feasibility of this operation relies on specific
assumptions about the variational posterior, such as its factorization
properties and membership of the exponential family. It is hence desirable
to devise methods that do not require the exact computation of the
ensemble averages with respect to the variational posterior $q(z|\varphi)$.
As we will see below, this is possible by combining VI methods with
MC approximations. The resulting methods can leverage SGD, scale to
large data sets, and have found a large number of recent applications
\cite{blei2017variational,bleivariational,angelino2016patterns}. 

The key idea of MC-based VI is to approximate the KL divergence (\ref{eq:var free energy})
via MC by drawing one or more samples $\mathrm{z}_{m}\sim$$q(z)$,
with $m=1,...,M$, and by computing the empirical average
\begin{equation}
\textrm{KL}(q(z)||p(x,z))\simeq\frac{1}{M}\sum_{m=1}^{M}(\textrm{ln}q(z_{m})-\ln p(x,z_{m})).\label{eq:var free energy-1}
\end{equation}
In order to optimize over the parameters $\varphi$ of the variational
posterior $q(z|\varphi)$, it is in fact more useful to approximate
the gradient of the KL divergence, as discussed next.

\textbf{REINFORCE approach.} To proceed, assume the following two
rather mild conditions on the parametrized variational posterior:
(\emph{i}) it is easy to draw samples $\mathrm{z}\sim$$q(z|\varphi)$;
and (\emph{ii}) it is possible to compute the gradient $\nabla_{\varphi}\textrm{ln}q(z|\varphi)$.
We can now develop an SGD-based scheme as follows. The main result
is that the gradient of the KL divergence in the I-projection problem
(\ref{eq:I projection true}) with respect to the variational parameters
$\varphi$ can be written as
\begin{equation}
\nabla_{\varphi}\textrm{KL}(q(z|\varphi)||p(x,z))=\textrm{E}_{\mathrm{z}\sim q(z|\varphi)}\left[\nabla_{\varphi}\textrm{ln}q(\mathrm{z}|\varphi)l_{\varphi}(x,\mathrm{z})\right],\label{eq:gradient free energy}
\end{equation}
where we have defined the \emph{learning signal}
\begin{equation}
l_{\varphi}(x,z)=\textrm{ln}q(z|\varphi)-\textrm{ln}p(x,z).
\end{equation}
To obtain (\ref{eq:gradient free energy}), we have used the identity
$\nabla_{\varphi}\textrm{ln}q(z|\varphi)=\nabla_{\varphi}q(z|\varphi)/q(z|\varphi)$,
which follows from the chain rule for differentiation, as well as
the equality $\textrm{E}_{\mathrm{z}\sim q(z|\varphi)}\left[\nabla_{\varphi}\textrm{ln}q(\mathrm{z}|\varphi)\right]=0$
(see \cite{bleivariational} for details).

MC-based VI methods evaluate an MC approximation of the gradient (\ref{eq:gradient free energy})
at a given value $\varphi$ by drawing one or more samples $\mathrm{z}_{m}\sim$$q(z|\varphi)$,
with $m=1,...,M$, and then computing
\begin{equation}
\nabla_{\varphi}\textrm{KL}(q(z|\varphi)||p(x,z))\simeq\frac{1}{M}\sum_{m=1}^{M}\left[\nabla_{\varphi}\textrm{ln}q(z_{m}|\varphi)l_{\varphi}(x,z_{m})\right].\label{eq:gradient free energy-1}
\end{equation}
This estimate is also known as likelihood ratio or REINFORCE gradient,
and it can be used to update the value of $\varphi$ using SGD (see
Sec. \ref{sec:Stochastic-Gradient-Descent}). The name reflects the
origin and the importance of the approach in the reinforcement learning
literature, as we briefly discuss in Chapter 9\footnote{In reinforcement learning, it is useful to compute the gradient of
the average reward $\textrm{E}_{\mathrm{t}\sim q(t|x,\varphi)}\left[R(\mathrm{t},x)\right]$
with respect to the parameters $\varphi$ defining the distribution
$q(t|x,\varphi)$ of the action $\mathrm{t}$ given the current state
$x$ of the environment. The reward $R(\mathrm{t},x)$ is a function,
possibly stochastic, of the action and of the state.}

In practice, these gradients have high variance. This is intuitively
due to the fact that this estimator does not use any information about
how a change in $z$ affects the learning signal $l_{\varphi}(x,z)$,
since it only depends on the value of this signal. Therefore, techniques
such as Rao-Blackwellization or control variates need to be introduced
in order to reduce its variance (see \cite{Yingzhen17} for a review).
Simplifications are possible when the variational posterior is assumed
to factorize, e.g., according to the mean-field full factorization.
This approach is used in the \emph{black-box inference} method of
\cite{blackbox}. 

\textbf{Reparametrization trick.} In order to mitigate the problem
of the high variance of the REINFORCE estimator, the reparametrization
trick leverage additional information about the dependence of the
variational distribution $q(z|\varphi),$ and hence of the learning
signal $l_{\varphi}(x,z)$, on the variable $z$. This approach is
applicable if: (\emph{i}) the latent variable $\mathrm{z}\sim q(z|\varphi)$
can be written as $\mathrm{z=}G_{\varphi}(\mathrm{w})$ for some differentiable
function $G_{\varphi}(\cdot)$ and for some rv $\mathrm{w}\sim s(z)$
whose distribution does not depend on $\varphi$; and (\emph{ii})
the \emph{variational regularization} term $\text{KL}(q(z|\varphi)||p(z)$)
in (\ref{eq:ELBO2}) can be computed and differentiated with respect
to $\varphi$. Assumption (\emph{ii}) is satisfied by members of the
exponential family (see Appendix B). A typical choice that satisfies
these conditions is to set $p(z)$ to be an i.i.d. Gaussian distribution;
$\mathrm{w}$ as i.i.d. Gaussian rvs; and function $G_{\varphi}(w)$
as $G_{\varphi}(w)=A_{\varphi}w+b_{\varphi}$, where matrix $A_{\varphi}$
and vector $b_{\varphi}$ are parametrized by a multi-layer neural
networks. Note that, with this choice, we have $q(z|\varphi)=\mathcal{N}(z|b_{\varphi},A_{\varphi}A_{\varphi}^{T})$.
We refer to \cite{jang2016categorical,maddison2016concrete} for other
examples.

To see why these assumptions are useful, let us first rewrite the
objective of the I-projection problem using (\ref{eq:ELBO2}) as
\begin{align}
\textrm{KL}(q(z|\varphi)||p(x,z)) & =-\textrm{E}_{\mathrm{w}\sim s(z)}\left[\textrm{ln}p(x|G_{\varphi}(\mathrm{w}))\right]+\text{KL}(q(z|\varphi)||p(z)).\label{eq:reparm}
\end{align}
We can now approximate the expectation in the first term via an empirical
average by generating i.i.d. samples $\mathrm{w}_{m}\sim s(w),$ $m=1,...,M$\footnote{This can be seen as an application of the Law of the Unconscious Statistician.}.
Note that this distribution does not depend on the current iterate
$\varphi$. We can then estimate the gradient by writing
\begin{equation}
\nabla_{\varphi}\textrm{KL}(q(z|\varphi)||p(x,z))\simeq-\frac{1}{M}\sum_{m=1}^{M}\left[\nabla_{\varphi}\textrm{ln}p(x|G_{\varphi}(w_{m}))\right]+\nabla_{\varphi}\text{KL}(q(z|\varphi)||p(z)).\label{eq:gradient free energy-1-1}
\end{equation}
This approach tends to provide estimate with lower variance than the
REINFORCE gradient, since it exploits the structure of the distribution
$q(z|\varphi)$.

Both REINFORCE and reparametrization trick can be naturally combined
with amortized inference. We also refer to \cite{2017arXiv171100123G}
for a proposed combination of both methods.

\section{Approximate Learning$^{*}$ }

As we have discussed, Bayesian inference plays a key role in learning
problems. In this section, we briefly discuss representative schemes
that incorporate approximate inference for learning. Since Bayesian
learning can directly benefit from approximate inference in evaluating
the posterior of the parameters and the variational posterior, we
focus here on the frequentist viewpoint. 

ML learning in the presence of hidden variables can be approximated
by maximizing the ELBO with respect to both the model parameters $\theta$
that define the forward model $p(x|z,\theta)$ and the parameters
$\varphi$ of the variational (amortized) model $q(z|x,\varphi)$.
To understand what is accomplished with this optimization, it is useful
to write the ELBO over a data set $\text{\ensuremath{\mathcal{D}}=\ensuremath{\{x_{n}\}_{n=1}^{N}}}$
using (\ref{eq:ELBO4}) as 
\begin{equation}
\sum_{n=1}^{N}\ln p(x_{n}|\theta)-\text{KL}\left(q(z|x_{n},\varphi)||p(z|x_{n},\theta)\right).\label{eq:ELBO Ch8}
\end{equation}
As such, for any fixed $\varphi$, optimizing the ELBO over $\theta$
maximizes the likelihood function in the first term under a variational
regularization term that penalizes posteriors $p(z|x,\theta)$ that
are significantly different from the selected variational posteriors
$q(z|x,\varphi)$. The choice of a given model for the variational
posterior hence drives learning, and should be treated as model or
hyperparameter selection \cite{bloginference1}.

The maximization of the ELBO over both model and variational parameters
can be carried out in different ways. As a first approach, one can
use EM by performing the E step via approximate inference to evaluate
the posterior of the latent variables. When VI is used for this purpose,
the resulting scheme is known as \emph{variational EM}. Alternatively,
one can use SGD with respect to both parameter vectors $\theta$ and
$\varphi$ by leveraging the REINFORCE method or the reparametrization
trick. The reparametrization trick approach is for instance used in
the\emph{ }VAE\emph{ }method for generative modelling \cite{VAE},
also known as Deep Gaussian Latent Models \cite{rezende2014stochastic},
as well as in the black-box learning approach of \cite{ranganath2015deep}
(see also \cite{mnih2014neural}). The KL divergence can also be substituted
by an adversarially learned divergence as in the GAN approach \cite{dumoulin2016adversarially}
(see Sec. \ref{subsec:GAN}).

When the variational parameters are updated using an M-projection,
rather than the I-projection that results from the maximization of
the ELBO, the approach of jointly optimizing model and variational
parameters yields the \emph{wake-sleep algorithm} \cite{hinton1995wake}.

\section{Summary}

Having observed in previous chapters that learning in probabilistic
models is often held back by the complexity of exact Bayesian inference
for hidden variables, this chapter has provided an overview of approximate,
lower-complexity, inference techniques. We have focused on MC and
VI methods, which are most commonly used. The treatment stressed the
impact of design choices in the selection of different types of approximation
criteria, such as M- and I-projection. It also covered the use of
approximate inference in learning problems. Techniques that improve
over the state of the art discussed in this chapter are being actively
investigated. Some additional topics for future research are covered
in the next chapter.

\section*{Appendix: M-Projection with the Exponential Family}

In this appendix, we consider the problem of obtaining the M-projection
of a distribution $p(z)$ into a model $q(z|\varphi)=Z(\varphi)^{-1}\text{exp}(\varphi^{T}u(z))$
from the exponential family  with sufficient statistics $u(z).$ We
will prove that, if there exists a value of $\varphi^{*}$ of the
natural parameter vector that satisfies the moment matching condition
(\ref{eq:M proj optimality}), then $q(z|\varphi^{*})$ is the M-projection.

We first write the KL divergence as
\begin{equation}
\begin{aligned}\text{KL}(p(z)||q(z|\varphi)) & =\text{ln}Z(\varphi)-\varphi^{T}\textrm{E}_{\mathrm{z\sim}p(z)}[u(\mathrm{z})]-H(p).\end{aligned}
\end{equation}
The difference between the KL divergence for a generic parameter vector
$\varphi$ and for the vector $\varphi^{*}$ satisfying (\ref{eq:M proj optimality})
can be written as
\begin{align}
 & \text{KL}(p(z)||q(z|\varphi))-\text{KL}(p(z)||q(z|\varphi^{*}))\nonumber \\
= & \text{ln\ensuremath{\left(\frac{Z(\varphi)}{Z(\varphi^{*})}\right)}}-(\varphi-\varphi^{*})^{T}\textrm{E}_{\mathrm{z\sim}p(z)}[u(\mathrm{z})]\nonumber \\
= & \textrm{E}_{\mathrm{z\sim}q(z|\varphi^{*})}\Bigg[\ln\ensuremath{\left(\frac{q(\mathrm{z}|\varphi^{*})}{q(\mathrm{z}|\varphi)}\right)}\Bigg]\nonumber \\
= & \text{KL}(q(z|\varphi^{*})||q(z|\varphi)).\label{eq:proof appendix Ch8}
\end{align}
Since the latter inequality is non-negative and equal to zero when
$\varphi=\varphi^{*},$ this choice of the natural parameters minimizes
the KL divergence.

\part{Conclusions}

\chapter{Concluding Remarks}

This monograph has provided a brief introduction to machine learning
by focusing on parametric probabilistic models for supervised and
unsupervised learning problems. It has endeavored to describe fundamental
concepts within a unified treatment starting from first principles.
Throughout the text, we have also provided pointers to advanced topics
that we have only been able only to mention or shortly touch upon.
Here, we offer a brief list of additional important aspects and open
problems that have not been covered in the preceding chapters.

\indent $\bullet$ \emph{Privacy}: In many applications, data sets
used to train machine learning algorithms contain sensitive private
information, such as personal preferences for recommendation systems.
It is hence important to ensure that the learned model does not reveal
any information about the individual entries of the training set.
This constraint can be formulated using the concept of differential
privacy. Typical techniques to guarantee privacy of individual data
points include adding noise to gradients when training via SGD and
relying on mixture of experts trained with different subsets of the
data \cite{Protection}.

\indent $\bullet$ \emph{Robustness}: It has been reported that various
machine learning models, including neural networks, are sensitive
to small variations in the data, producing the wrong response upon
minor, properly chosen, changes in the explanatory variables. Note
that such adversarially chosen examples, which cause a specific machine
to fail, are conceptually different from the randomly selected examples
that are assumed when defining the generalization properties of a
network. There is evidence that finding such examples is possible
even without knowing the internal structure of a machine, but solely
based on black-box observations \cite{2016arXiv160202697P}. Modifying
the training procedure in order to ensure robustness to adversarial
examples is an active area of research with important practical implications
\cite{goodfellow2014explaining}.

\indent $\bullet$ \emph{Computing platforms and programming frameworks}:
In order to scale up machine learning applications, it is necessary
to leverage distributed computing architectures and related standard
programming frameworks \cite{bekkerman2011scaling,angelino2016patterns}.
As a complementary and more futuristic approach, recent work has even
proposed to leverage the capabilities of annealing-based quantum computers
as samplers \cite{koren2016b} or discrete optimizers \cite{mott2017solving}.

\indent $\bullet$ \emph{Transfer learning}: Machines trained for
a certain task currently need to be re-trained in order to be re-purposed
for a different task. For instance, a machine that learned how to
drive a car would need to be retrained in order to learn how to drive
a truck. The field of transfer learning covers scenarios in which
one wishes to transfer the expertise acquired from some tasks to others.
Transfer learning includes different related paradigms, such as multitask
learning, lifelong learning, zero-shot learning, and domain adaptation
\cite{venkateswara}. In \emph{multitask learning}, several tasks
are learned simultaneously. Typical solutions for multitask learning
based on neural networks prescribe the presence of common hidden layers
among neural networks trained for different tasks \cite{bengio2012transfer}.
\emph{Lifelong learning} updates a machine trained on a number of
tasks to carry out a new task by leveraging the knowledge accumulated
during the previous training phases \cite{thrun1996learning}. \emph{Zero-shot
learning} refers to models capable of recognizing unseen classes with
training examples available only for related, but different, classes.
This often entails the task of learning representation of classes,
such as prototype vectors, that generate data in the class through
a fixed probabilistic mechanism \cite{fu2017recent}. Domain adaptation
will be discussed separately in the next point. 

\indent $\bullet$ \emph{Domain adaptation}: In many learning problems,
the available data has a different distribution from the data on which
the algorithm will be tested. For instance, in speech recognition,
one has available data for a given user at learning time, but it may
be desirable, after learning, to use the same machine for another
user. For supervised learning, this is typically modeled by assuming
that the distribution of the covariates $\mathrm{x}$ is different
during training and test, while the discriminative conditional distribution
$p(t|x)$ is the same for both phases \cite{venkateswara}. A generalization
of PAC theory analyses domain adaptation, obtaining bounds on the
generalization error under the desired test distribution as a function
of the difference between the training and test distributions \cite{blitzer2008learning}. 

\indent $\bullet$ \emph{Communication-efficient learning}: In distributed
computing platforms, data is typically partitioned among the processors
and communication among the processor entails latency and energy consumption.
An important research problem is that of characterizing the best trade-off
between learning performance and communication overhead \cite{zhang2013info}. 

\indent $\bullet$ \emph{Reinforcement learning}: Reinforcement learning
is at the heart of recent successes of machine learning methods in
acquiring the skills necessary to play video games or games against
human opponents (see, e.g., \cite{mnih2013playing}). In reinforcement
learning, one wishes to learn the optimal mapping, say $q(t|x,\theta)$,
between the observed state $x$ of the world and an action $t$. Unlike
supervised learning, the optimal action $t$ is not known, but the
machine observes a reward/ punishment signal depending on the effect
of the action. A popular approach, referred to as deep reinforcement
learning, models the mapping $q(t|x,\theta)$ using a deep neural
network. This is trained to maximize the average reward via SGD by
using the REINFORCE method (Chapter 8) to estimate the gradient \cite{Silverrl,Levinerl,blogRL,arulkumaran2017brief}.

\appendix

\chapter{Appendix A: Information Measures} 

In this appendix, we describe a principled and intuitive introduction
to information measures that builds on inference, namely estimation
and hypothesis testing. We focus on entropy, mutual information and
divergence measures. We also concentrate on discrete rvs. In the monograph,
we have taken the pragmatic approach of extending the definitions
to continuous variables by substituting sums with integrals. It is
worth noting that this approach does not come with any practical complications
when dealing with mutual information and divergence. Instead, the
continuous version of the entropy, known as differential entropy,
should be treated with care, as it does not satisfy some key properties
of the entropy such as non-negativity. 

\section{Entropy}

As proposed by Claude Shannon, the amount of information received
from the observation of a discrete random variable $\mathrm{x}\sim p(x)$
defined over a finite alphabet $\mathcal{X}$ should be measured by
the amount of uncertainty about its value prior to its measurement
\cite{Shannon}. To this end, we consider the problem of estimating
the value of $\mathrm{x}$ when one only knows the probabilistic model
$p(x)$. The key idea is that the observation of a random variable
$\mathrm{x}$ is more informative if its value is more difficult to
predict a priori, that is, based only on the knowledge of $p(x)$.

To formalize this notion, we need to specify: (\emph{i}) the type
of estimates that one is allowed to make on the value of $\mathrm{x}$;
and (\emph{ii}) the loss function $\ell$ that is used to measure
the accuracy of the estimate. We will proceed by considering two types
of estimates, namely \emph{point estimates}, whereby one needs to
commit to a specific value $\hat{x}$ as the estimate of $\mathrm{x}$;
and \emph{distributional estimates}, in which instead we are allowed
to produce a pmf $\hat{p}(x)$ over alphabet $\mathcal{X}$, hence
defining a profile of \char`\"{}beliefs\char`\"{} over the possible
values of rv $\mathrm{x}$. We will see below that the second approach
yields Shannon entropy, first encountered in this monograph in (\ref{eq:KL entropy Ch2}).

\textbf{Point Estimates.} Given a point estimate $\hat{x}$ and an
observed value $x\in\mathcal{\mathrm{\mathcal{X}}}$, as we have seen,
the estimation error can be measured by a non-negative loss function
$\ell(x,\hat{x})$, such the quadratic loss function and the 0-1 loss
function. For any given loss function $\ell$, based on the discussion
above, we can measure the information accrued by the observation of
$\mathrm{x}\sim p_{\mathrm{x}}$ by evaluating the average loss that
is incurred by the best possible a priori estimate of $\mathrm{x}$.
This leads to the definition of generalized entropy \cite{Grunwaldpaper}
\begin{equation}
H_{\ell}(p_{\mathrm{x}})=\min_{\hat{x}}\textrm{E\ensuremath{_{\mathrm{x}\sim p_{\mathrm{x}}}}[}\ell(\mathrm{x},\hat{x})],\label{eq:generalized entropy definition}
\end{equation}
where the estimate $\hat{x}$ is not necessarily constrained to lie
in the alphabet $\mathcal{X}$. As highlighted by the notation $H_{\ell}(p_{\mathrm{x}})$,
the generalized entropy depends on the pmf $p_{\mathrm{x}}$ and on
the loss function $\ell$. The notion of generalized entropy (\ref{eq:generalized entropy definition})
coincides with that of minimum Bayes risk for the given loss function
$\ell$.

For the quadratic loss function, the generalized entropy is the variance
of the distribution $H_{\ell_{2}}(p_{\mathrm{x}})=\textrm{var}(p_{\mathrm{x}})$.
To see this, impose the optimality condition $\textrm{\ensuremath{d}E[}(\mathrm{x}-\hat{x})^{2}]/\textrm{\ensuremath{d}}\hat{x}=0$
to conclude that the optimal point estimate is the mean $\hat{x}=\textrm{E}_{\mathrm{x}\sim p_{\mathrm{x}}}[\mathrm{x}].$
As for the 0-1 loss, the generalized entropy equals the minimum probability
of error for the detection of $\mathrm{x}$, that is,
\begin{equation}
H_{\ell_{0}}(p_{\mathrm{x}})=\min_{\hat{x}}\sum_{x\neq\hat{x}}p(x)=1-\max_{\hat{x}}p(\hat{x}).\label{eq:gen entropy 0}
\end{equation}
This is because the optimal estimate is the mode, i.e., the value
$\hat{x}$ with the largest probability $p(\hat{x})$. 

\textbf{Distributional Estimate.}\emph{ }We now consider a different
type of estimation problem in which we are permitted to choose a pmf
$\hat{p}(x)$ on the alphabet $\mathcal{X}$ as the estimate for the
outcome of variable $\mathrm{x}$. To facilitate intuition, we can
imagine $\hat{p}(x)$ to represent the fraction of one's wager that
is invested on the outcome of $\mathrm{x}$ being a specific value
$x$. Note that it may not be necessarily optimal to put all of one's
money on one value $x$! In fact, this depends on how we measure the
reward, or conversely the cost, obtained when a value $\mathrm{x}=x$
is realized.

To this end, we define a non-negative loss function $\ell(x,\hat{p}_{\mathrm{x}})$
representing the loss, or the ``negative gain'', suffered when the
value $\mathrm{x}=x$ is observed. This loss should sensibly be a
decreasing function of $\hat{p}(x)$ \textendash{} we register a smaller
loss, or conversely a larger gain, when we have wagered more on the
actual outcome $x$. As a fairly general class of loss functions,
we can hence define 
\begin{equation}
\ell(x,\hat{p}_{\mathrm{x}})=f(\hat{p}(x)),\label{eq:loss distributional}
\end{equation}
where $f$ is a decreasing function. More general classes of loss
functions are considered in \cite{DeGroot}.

Denote as $\Delta(\mathcal{\mathcal{X}})$ the simplex of pmfs defined
over alphabet $\mathcal{X}$. The generalized entropy can now be defined
in a way that is formally equivalent to (\ref{eq:generalized entropy definition}),
with the only difference being the optimization over pmf $\hat{p}_{\mathrm{x}}$
rather than over point estimate $\hat{x}$: 
\begin{equation}
H_{\ell}(p_{\mathrm{x}})=\min_{\hat{p}_{\mathrm{x}}\in\Delta(\mathcal{\mathcal{X}})}\textrm{E\ensuremath{_{\mathrm{x}\sim p_{\mathrm{x}}}}[}\ell(\mathrm{x},\hat{p}_{\mathrm{x}})].\label{eq:generalized entropy definition-1}
\end{equation}

A key e$\mathrm{x}$ample of loss function $\ell(x,\hat{p}_{\mathrm{x}})$
in class (\ref{eq:loss distributional}) is the \emph{log-loss} $\ell(x,\hat{p}_{\mathrm{x}})=-\log\hat{p}(x)$.
The log-loss has a strong motivation in terms of lossless compression.
In fact, as discussed in Sec. \ref{sec:Minimum-Description-Length},
by Kraft's inequality, it is possible to design a prefix-free \textendash{}
and hence decodable without delay \textendash{} lossless compression
scheme that uses $\left\lceil -\log\hat{p}(x)\right\rceil $ bits
to represent value $x$. As a result, the choice of a pmf $\hat{p}_{\mathrm{x}}$
is akin to the selection of a prefix-free lossless compression scheme
that requires a description of around $-\ln\hat{p}(x)$ bits to represent
value $x$. The expectation in (\ref{eq:generalized entropy definition-1})
measures the corresponding average number of bits required for lossless
compression by the given scheme.

Using the log-loss in (\ref{eq:generalized entropy definition}),
we obtain the Shannon entropy
\begin{align}
 & \min_{\hat{p}_{\mathrm{x}}\in\Delta(\mathcal{\mathcal{X}})}\textrm{E\ensuremath{_{\mathrm{x}\sim p_{\mathrm{x}}}}\ensuremath{[-}\ensuremath{\ln}\ensuremath{\ensuremath{\hat{p}}(\mathrm{x})}}],\label{eq:Shannon as log loss}\\
 & =\textrm{E\ensuremath{_{\mathrm{x}\sim p_{\mathrm{x}}}}\ensuremath{[-}\ensuremath{\ln}\ensuremath{\ensuremath{p}(\mathrm{x})}}]=H(p_{\mathrm{x}}).
\end{align}
In fact, imposing the optimality condition yields the optimal pmf
$\hat{p}(x)$ as $\hat{p}(x)=p(x)$. Equation (\ref{eq:Shannon as log loss})
reveals that the entropy $H(p_{\mathrm{x}})$ is the minimum average
log-loss when optimizing over all possible pmfs $\hat{p}_{\mathrm{x}}$.

It may seem at first glance that the choice $\hat{p}(x)=p(x)$ should
be optimal for most reasonable loss functions in class (\ref{eq:loss distributional}),
but this is not the case. In fact, when the alphabet $\mathcal{X}$
has more than two elements, it can be proved that the log-loss \textendash{}
more generally defined as $\ell(x,\hat{p}_{\mathrm{x}})=b\log\hat{p}(x)+c$
with $b\leq0$ and any $c$ \textendash{} is the only loss function
of the form (\ref{eq:loss distributional}) for which $\hat{p}(x)=p(x)$
is optimal \cite{Weissmanarxiv}.

As a final note, the generalized entropy $H_{\ell}(p_{\mathrm{x}})$
is a concave function of $p_{\mathrm{x}}$, which means that we have
the inequality $H_{\ell}(\lambda p_{\mathrm{x}}+(1-\lambda)q_{\mathrm{x}})\geq\lambda H_{\ell}(p_{\mathrm{x}})+(1-\lambda)H_{\ell}(q_{\mathrm{x}})$
for any two distributions $p_{\mathrm{x}}$ and $q_{\mathrm{x}}$
and any $0\leq\lambda\leq1$. This follows from the fact that the
entropy is the minimum over a family of linear functionals of $p_{\mathrm{x}}$
\cite{Boyd}. The concavity of $H_{\ell}(p_{\mathrm{x}})$ implies
that a variable $\mathrm{x}\sim\lambda p_{\mathrm{x}}+(1-\lambda)q_{\mathrm{x}}$
distributed according to the mixture of two distributions is more
``random'', i.e., it is more difficult to estimate, than both variables
$\mathrm{x}\sim p_{\mathrm{x}}$ and $\mathrm{x}\sim q_{\mathrm{x}}$.

\section{Conditional Entropy and Mutual Information}

Given two random variables $\mathrm{x}$ and $\mathrm{y}$ jointly
distributed according to a known probabilistic model $p(x,y)$, i.e.,
$(\mathrm{x},\mathrm{y})\sim p_{\mathrm{xy}}$, we now discuss how
to quantify the information that the observation of one variable,
say $\mathrm{y}$, brings about the other, namely $\mathrm{x}$. Following
the same approach adopted above, we can distinguish two inferential
scenarios for this purpose: in the first, a point estimate $\hat{x}(y)$
of $\mathrm{x}$ needs to be produced based on the observation of
a value $\mathrm{y}=y$ and the knowledge of the joint pmf $p_{\mathrm{x}\mathrm{y}}$;
while, in the second, we are allowed to choose a pmf $\hat{p}_{\mathrm{x}|\mathrm{y}=y}$
as the estimate of $\mathrm{x}$ given the observation $\mathrm{y}=y.$

\emph{Point Estimate}: Assuming point estimates and given a loss function
$\ell(x,\hat{x})$, the generalized conditional entropy for an observation
$\mathrm{y}=y$ is defined as the minimum average loss 
\begin{equation}
H_{\ell}(p_{\mathrm{x}|\mathrm{y}=y})=\min_{\hat{x}(y)}\textrm{E\ensuremath{_{\mathrm{x}\sim p_{\mathrm{x}|\mathrm{y}=y}}}[}\ell(\mathrm{x},\hat{x}(y))|\mathrm{y}=y].\label{eq:generalized conditional entropy}
\end{equation}
Note that this definition is consistent with (\ref{eq:generalized entropy definition})
as applied to the conditional pmf $p_{\mathrm{x}|\mathrm{y}=y}$.
Averaging over the distribution of the observation $\mathrm{y}$ yields
the generalized conditional entropy 
\begin{equation}
H_{\ell}(\mathrm{x}|\mathrm{y})=\textrm{E}_{\mathrm{y}\sim p_{\mathrm{y}}}[H_{\ell}(p_{\mathrm{x}|\mathrm{y}})].\label{eqgeneratlized conditional entropy 1}
\end{equation}
It is emphasized that the generalized conditional entropy depends
on the joint distribution $p_{\mathrm{x}\mathrm{y}}$, while (\ref{eq:generalized conditional entropy})
depends only on the conditional pmf $p_{\mathrm{x}|\mathrm{y}=y}$.

For the squared error, the generalized conditional entropy can be
easily seen to be the average conditional variance $H_{\ell_{2}}(\mathrm{x}|\mathrm{y})=\textrm{E}_{\mathrm{y}\sim p_{\mathrm{y}}}[\textrm{var}$$(p_{\mathrm{x}|\mathrm{y}})${]},
since the a posteriori mean $\hat{x}(y)=\textrm{E}_{\mathrm{x}\sim p_{\mathrm{x}|\mathrm{y}=y}}[\mathrm{x}|\mathrm{y}=y]$
is the optimal estimate. For the 0-1 loss, the generalized conditional
entropy $H_{\ell_{0}}(\mathrm{x}|\mathrm{y})$ is instead equal to
the minimum probability of error for the detection of $\mathrm{x}$
given $\mathrm{y}$ and the MAP estimate $\hat{x}(y)=\textrm{argmax}_{\hat{x}\in\mathcal{\mathrm{x}}}p(\hat{x}|y)$
is optimal.

\emph{Distributional Estimate}: Assume now that we are allowed to
choose a pmf $\hat{p}_{\mathrm{x}|\mathrm{y}=y}$ as the estimate
of $\mathrm{x}$ given the observation $\mathrm{y}=y$, and that we
measure the estimation loss via a function $\ell(x,\hat{p}_{\mathrm{x}})$
as in (\ref{eq:loss distributional}). The definition of generalized
conditional entropy for a given value of $\mathrm{y}=y$ follows directly
from the arguments above and is given as $H_{\ell}(p_{\mathrm{x}|\mathrm{y}=y})$,
while the generalized conditional entropy is (\ref{eqgeneratlized conditional entropy 1}).
With the log-loss function, the definition above can be again seen
to coincide with Shannon conditional entropy $H(\mathrm{x}|\mathrm{y})=\textrm{E\ensuremath{_{\mathrm{x},\mathrm{y}\sim p_{\mathrm{x},\mathrm{y}}}}[}-\ln p(\mathrm{x}|\mathrm{y})]$.

If $\mathrm{x}$ and $\mathrm{y}$ are independent, we have the equality
$H_{\ell}(\mathrm{x}|\mathrm{y})=H_{\ell}(\mathrm{x})$. Furthermore,
since in (\ref{eq:generalized conditional entropy}) we can always
choose estimates that are independent of $\mathrm{y}$, we generally
have the inequality $H_{\ell}(\mathrm{x}|\mathrm{y})\leq H_{\ell}(\mathrm{x})$:
observing $\mathrm{y}$, on average, can only decrease the entropy.
Note, however, that it is not true that $H_{\ell}(p_{\mathrm{x}|\mathrm{y}=y})$
is necessarily smaller than $H_{\ell}(\mathrm{x})$ \cite{Cover}.

\emph{Mutual Information}: The inequality $H_{\ell}(\mathrm{x}|\mathrm{y})\leq H_{\ell}(\mathrm{x})$
justifies the definition of generalized mutual information with respect
to the given loss function $\ell$ as 
\begin{equation}
I_{\ell}(\mathrm{x};\mathrm{y})=H_{\ell}(\mathrm{x})-H_{\ell}(\mathrm{x}|\mathrm{y}).\label{eq:mutual information}
\end{equation}
The mutual information measures the decrease in average loss that
is obtained by observing $\mathrm{y}$ as compared to having only
prior information about $p_{\mathrm{x}}$. This notion of mutual information
is in line with the concept of statistical information proposed by
DeGroot (see \cite{DeGroot} for a recent treatment). With the log-loss,
the generalized mutual information (\ref{eq:mutual information})
reduces to Shannon's mutual information. As shown in \cite{jiao2015justification},
the log-loss is in fact the only loss function, up to multiplicative
factors, under which the generalized mutual information (\ref{eq:mutual information})
satisfies the data processing inequality, as long as the alphabet
of $\mathrm{x}$ has more than two elements. 

\section{Divergence Measures}

We now discuss a way to quantify the ``difference'' between two
given probabilistic models $p_{\mathrm{x}}$ and $q_{\mathrm{x}}$
defined over the same alphabet $\mathcal{X}.$ We consider here the
viewpoint of binary hypothesis testing as a theoretical framework
in which to tackle the issue. Other related approaches have also found
applications in machine learning, including optimal transport theory
and kernel methods \cite{Wasserstein}.

We consider the following standard binary hypothesis testing problem.
Given an observation $\mathrm{x}$, decide whether $\mathrm{x}$ was
generated from pmf $p_{\mathrm{x}}$ or from pmf $q_{\mathrm{x}}.$
To proceed, we define a decision rule $T(x)$, which should have the
property that it is increasing with the certainty that a value $\mathrm{x}=x$
is generated from $p_{\mathrm{x}}$ rather than $q_{\mathrm{x}}$.
For e$\mathrm{x}$ample, in practice, one may impose a threshold on
the rule $T(x)$ so that, when $T(x)$ is larger than the threshold,
a decision is made that $\mathrm{x}=x$ was generated from $p_{\mathrm{x}}$.

In order to design the decision rule $T(x)$, we again minimize a
loss function or, equivalently, maximize a merit function. For convenience,
here we take the latter approach, and define the problem of maximizing
the merit function
\begin{equation}
\textrm{E}_{\mathrm{x}\sim p_{\mathrm{x}}}[T(\mathrm{x})]-\textrm{E}_{\mathrm{x}\sim q_{\mathrm{x}}}[g(T(\mathrm{x}))]\label{eq:merit function}
\end{equation}
over the rule $T(x)$, where $g$ is a concave increasing function.
This criterion can be motivated as follows: (\emph{i}) It increases
if $T(\mathrm{x})$ is large, on average, for values of $\mathrm{x}$
generated from $p_{\mathrm{x}}$; and (\emph{ii}) it decreases if,
upon expectation, $T(\mathrm{x})$ is large for values of $\mathrm{x}$
generated from $q_{\mathrm{x}}.$ The function $g$ can be used to
define the relative importance of errors made in favor of one distribution
or the other. From this discussion, the optimal value of (\ref{eq:merit function})
can be taken to be a measure of the distance between the two pmfs.
This yields the following definition of divergence between two pmfs
\begin{equation}
D_{f}(p_{\mathrm{x}}||q_{\mathrm{x}})=\max_{T(x)}\textrm{ E}_{\mathrm{x}\sim p_{\mathrm{x}}}[T(\mathrm{x})]-\textrm{E}_{\mathrm{x}\sim q_{\mathrm{x}}}[g(T(\mathrm{x}))],\label{eq:Df}
\end{equation}
where the subscript $f$ will be justified below.

Under suitable differentiability assumptions on function $g$ (see
\cite{fGAN} for generalizations), taking the derivative with respect
to $T(x)$ for all $x\in\mathcal{\mathrm{\mathcal{\mathrm{x}}}}$
yields the optimality condition $g'(T(x))=p(x)/q(x)$. This relationship
reveals the connection between the optimal detector $T(x)$ and the
LLR $p(x)/q(x)$. Plugging this result into (\ref{eq:Df}), it can
be directly checked that the following equality holds \cite{Nguyen}
\begin{equation}
D_{f}(p_{\mathrm{x}}||q_{\mathrm{x}})=\textrm{E}_{\mathrm{x}\sim q_{\mathrm{x}}}\left[f\left(\frac{p_{\mathrm{x}}(\mathrm{x})}{q_{\mathrm{x}}(\mathrm{x})}\right)\right],\label{eq:f divergence}
\end{equation}
where the function $f(x)=g^{*}(x)$ is the convex dual function of
$g(t)$, which is defined as $g^{*}(x)=\sup_{t}\left(xt-g(t)\right)$.
Note that dual function $f$ is always convex \cite{Boyd}.

Under the additional constraint $f(1)=0$, definition (\ref{eq:f divergence})
describes a large class of divergence measures parametrized by the
convex function $f$, which are known as $f$-divergences or Ali-Silvey
distance measures \cite{Duchicourse}. Note that the constraint $f(1)=0$
ensures that the divergence is zero when the pmfs $p_{\mathrm{x}}$
and $q_{\mathrm{x}}$ are identical. Among their key properties, $f$-divergences
satisfy the data processing inequality \cite{Duchicourse}.

For e$\mathrm{x}$ample, the choice $g(t)=\exp(t-1)$, which gives
the dual convex $f(x)=x\ln x$, yields the optimal detector $T(x)=1+\ln(p(x)/q(x))$
and the corresponding divergence measure (\ref{eq:f divergence})
is the standard KL divergence KL$(p_{\mathrm{x}}||q_{\mathrm{x}})$.
As another instance of $f$-divergence, with $g(t)=-\ln(2-\exp(t))$
we obtain the optimal detector $T(x)=\ln(2p_{\mathrm{x}}(x)/p_{\mathrm{x}}(x)+q_{\mathrm{x}}(x))$,
and $D_{f}(p_{\mathrm{x}}||q_{\mathrm{x}})$ becomes the Jensen-Shannon
divergence\footnote{The Jensen-Shannon divergence can also be interpreted as the mutual
information $I(s;\mathrm{x})$}. For reference, the latter can be written as 
\begin{equation}
\textrm{JS}(p_{\mathrm{x}}||q_{\mathrm{x}})=\textrm{ KL\ensuremath{\left(p_{\mathrm{x}}\parallel m_{\mathrm{x}}\right)}+KL\ensuremath{\left(q_{\mathrm{x}}\parallel m_{\mathrm{x}}\right)}},
\end{equation}
where $m_{\mathrm{x}}(x)=(p_{\mathrm{x}}(x)+q_{\mathrm{x}}(x))/2.$\footnote{The Jensen-Shannon divergence, as defined above, is proportional to
the mutual information $I(\mathrm{s};\mathrm{x})$ for the joint distribution
$p_{\mathrm{s},\mathrm{x}}(s,x)=1/2\cdot p_{\mathrm{x}|\mathrm{s}}(x|s)$
with binary ${\rm s},$ and conditional pmf defined as $p_{\mathrm{x}|\mathrm{s}}(x|0)=p_{\mathrm{x}}(x)$
and $p_{\mathrm{x}|\mathrm{s}}(x|s)(x|1)=q_{\mathrm{x}}(x)$.} Another special case, which generalizes the KL divergence and other
metrics, is the $\alpha$-divergence discussed in Chapter 8 (see (\ref{eq:alpha divergence Ch8})),
which is obtained with $f(x)=(\alpha(x-1)-(x^{\alpha}-1))/(\alpha(1-\alpha))$
for some real-valued parameter $\alpha$. We refer to \cite{fGAN,Duchicourse}
for other examples.

The discussion above justified the adoption of the loss function (\ref{eq:Df})
in a heuristic fashion. It is, however, possible to derive formal
relationships between the error probability of binary hypothesis testing
and \emph{f}-divergences \cite{Berisha}. We also refer to the classical
Sanov lemma and Stein lemma as fundamental applications of KL divergence
to large deviation and hypothesis testing \cite{Cover}.



\chapter{Appendix B: KL Divergence and Exponential Family}

In this appendix, we provide a general expression for the KL divergence
between two distributions $p(x|\eta_{1})$ and $p(x|\eta_{2})$ from
the same regular exponential family with log-partition function $A(\cdot)$,
sufficient statistics $u(x)$, and moment parameters $\mu_{1}$ and
$\mu_{2}$, respectively. We recall from Chapter 3 that the log-partition
function is convex and that we have the identity $\nabla A(\eta)=\mu.$ 

The KL divergence between the two distributions can be translated
into a divergence on the space of natural parameters. In particular,
the following relationship holds \cite{Amari}
\begin{equation}
\textrm{KL}(p(x|\eta_{1})||p(x|\eta_{2}))=D_{A}(\eta_{2},\eta_{1}),\label{eq:Appendix B1}
\end{equation}
where $D_{A}(\eta_{2},\eta_{1})$ represents the Bregman divergence
with generator function given by the log-partition function $A(\cdot)$,
that is
\begin{align}
D_{A}(\eta_{2},\eta_{1}) & =A(\eta_{2})-A(\eta_{1})-(\eta_{2}-\eta_{1})^{T}\nabla A(\eta_{1})\nonumber \\
 & =A(\eta_{2})-A(\eta_{1})-(\eta_{2}-\eta_{1})^{T}\mu_{1}.\label{eq:Appendix B2}
\end{align}
The first line of (\ref{eq:Appendix B2}) is the general definition
of the Bregman divergence $D_{A}(\cdot,\cdot)$ with a generator function
$A(\cdot),$ while the second follows from the relationship (\ref{eq:gradient log part-1}).
Note that the Bregman divergence can be proved to be non-negative,
and convex in its first argument, but not necessarily in the second
argument. The equality (\ref{eq:Appendix B1})-(\ref{eq:Appendix B2})
can be proved by using the definition of exponential family via the
following equality
\begin{align}
\textrm{KL}(p(x|\eta_{1})||p(x|\eta_{2})) & =\textrm{E}_{\mathrm{x}\sim p(x|\eta_{1})}\left[\log\frac{p(\mathrm{x}|\eta_{1})}{p(\mathrm{x}|\eta_{2})}\right]\nonumber \\
 & =\textrm{E}_{\mathrm{x}\sim p(x|\eta_{1})}[(\eta_{1}-\eta_{2})^{T}u(\mathrm{x})]-A(\eta_{1})+A(\eta_{2}).
\end{align}

Recalling again that we have the equality $\nabla A(\eta_{1})=\mu_{1}$,
the relationship (\ref{eq:Appendix B1})-(\ref{eq:Appendix B2}) can
be approximated as 
\begin{equation}
\begin{split}\textrm{KL}(p(x|\eta_{1})||p(x|\eta_{2})) & =\frac{1}{N}(\eta_{1}-\eta_{2})^{T}J_{\eta_{1}}(\eta_{1}-\eta_{2})+O(||\eta_{1}-\eta_{2}||^{3})\end{split}
,
\end{equation}
where $J_{\eta}=-\text{E}\left[\nabla_{\eta}^{2}\text{ln}p(\mathrm{x}|\eta)\right]$
is the Fisher information matrix. This expansion holds given the relationship
$\nabla_{\eta}^{2}A(\eta)=J_{\eta}$ \cite{Duchicourse}.

It is also possible to write a relationship analogous to (\ref{eq:Appendix B1})-(\ref{eq:Appendix B2})
in terms of mean parameters. This is done by using the convex conjugate
function
\begin{equation}
A^{*}(\mu)=\sup_{\eta}\eta^{T}\mu-A(\eta),
\end{equation}
where the maximization is over the feasible set of natural parameters.
In fact, the optimization over $\eta$ yields the natural parameter
$\eta$ corresponding to the mean parameter $\mu$, i.e., $\nabla A(\eta)=\mu$.
It hence follows from (\ref{eq:Appendix B2}) that we have 
\begin{align}
\textrm{KL}(p(x|\mu_{1})||p(x|\mu_{2})) & =A^{*}(\mu_{1})-A^{*}(\mu_{2})-(\mu_{1}-\mu_{2})^{T}\eta_{2}\nonumber \\
 & =D_{A^{*}}(\mu_{1},\mu_{2}),
\end{align}
where in the second line we have used the inverse mapping $\nabla A*(\mu)=\eta$
between mean and natural parameters (which holds for minimal families).

Chernoff divergence measures, including the Bhattacharyya distance,
can also be written in closed form for exponential families \cite{Nielsen}.

\begin{acknowledgements}
	Osvaldo Simeone has received funding from the European Research Council (ERC) under the European Union’s Horizon 2020 research and innovation programme (grant agreement No 725731).
\end{acknowledgements}

\backmatter  

\printbibliography

\end{document}